\documentclass[acmtog,nonacm,screen]{acmart}

\usepackage{booktabs} 

\setcitestyle{nosort,square} 

\title{\methodname{}: Glossy Object Capture under Non-Distant Lighting}  

\newcommand{\affuci}[0]{%
    \affiliation{%
        \institution{University of California, Irvine}
        \city{San Jose}
        \state{California}
        \country{USA}
    }
}
\newcommand{\affadobe}[0]{%
    \affiliation{%
        \institution{Adobe Research}
        \city{San Jose}
        \state{California}
        \country{USA}
    }
}

\author{Guangyan Cai}
\affuci
\affadobe
\email{gcai3@uci.edu}

\author{Fujun Luan}
\affadobe
\email{fluan@adobe.com}

\author{Milo\v{s} Ha\v{s}an}
\affadobe
\email{mihasan@adobe.com}

\author{Kai Zhang}
\affadobe
\email{kaiz@adobe.com}

\author{Sai Bi}
\affadobe
\email{sbi@adobe.com}

\author{Zexiang Xu}
\affadobe
\email{zexu@adobe.com}

\author{Iliyan Georgiev}
\affiliation{%
    \institution{Adobe Research}
    \country{UK}
}
\email{igeorgiev@adobe.com}

\author{Shuang Zhao}
\affuci
\email{shz@ics.uci.edu}

\authorsaddresses{} 

\renewcommand\footnotetextcopyrightpermission[1]{} 

\usepackage{mathtools,stmaryrd,nicefrac}
\usepackage[many]{tcolorbox}
\usepackage{dsfont}
\usepackage{bm}
\usepackage{ifthen,comment,xspace}
\usepackage{overpic}
\usepackage{subcaption}
\usepackage{cleveref}
\usepackage[outline]{contour}
\contourlength{0.075em}

\usepackage[linesnumbered,ruled,vlined]{algorithm2e} 

\SetAlFnt{\small}
\SetAlCapFnt{\small}
\SetAlCapNameFnt{\small}
\SetAlCapHSkip{0pt}
\let\oldnl\nl
\newcommand{\nonl}{\renewcommand{\nl}{\let\nl\oldnl}}
\definecolor{cmtClr}{HTML}{028A0F}

\usepackage{tikz}
\usetikzlibrary{calc}

\usepackage{enumitem}
\setlist[itemize]{parsep=0pt,partopsep=0pt,leftmargin=*,itemsep=5pt}
\setlist[enumerate]{parsep=0pt,partopsep=0pt,leftmargin=*,itemsep=5pt}

\definecolor{orange_cubic}{rgb}{.9765, .5887, .3569}
\definecolor{purple_cubic}{rgb}{.4706, 0, .5216}
\definecolor{green_cubic}{rgb}{.28603, .81178, .5008}

\definecolor{grayLL}{rgb}{.98, .98, .98}
\definecolor{grayL}{rgb}{.9, .9, .9}
\definecolor{purpleL}{rgb}{.9735, .95, .9761}
\definecolor{purpleD}{rgb}{.8941, .8, .9043}
\definecolor{greenL}{rgb}{.9643, .9906, .9750}
\definecolor{greenD}{rgb}{.7145, .9249, .7999}
\definecolor{greenDD}{rgb}{.3145, .6249, .3999}
\definecolor{orangeLL}{rgb}{0.9991, 0.9846, 0.9759}
\definecolor{orangeL}{rgb}{.9982, .9692, .9518}
\definecolor{orangeD}{rgb}{.9929, .8766, .8071}

\definecolor{redL}{rgb}{1.0, 0.95, 0.95}
\definecolor{redD}{rgb}{1.0, 0.8, 0.8}
\definecolor{redDD}{rgb}{1.0, 0.4, 0.4}
\definecolor{yellowL}{rgb}{1.0, 1.0, 0.95}
\definecolor{yellowD}{rgb}{0.95, 0.95, 0.6}
\definecolor{yellowDD}{rgb}{0.8, 0.8, 0.2}
\definecolor{blueLL}{rgb}{0.98, 0.98, 1.0}
\definecolor{blueL}{rgb}{0.95, 0.95, 1.0}
\definecolor{blueD}{rgb}{0.8, 0.8, 1.0}
\definecolor{blueDD}{rgb}{0.6, 0.6, 1.0}

\definecolor{uciBlue}{RGB}{0,100,164}
\definecolor{uciBlueL}{RGB}{127.5, 177.5, 209.5}
\definecolor{uciOrange}{RGB}{247,141,45}
\definecolor{uciOrangeL}{RGB}{251,198,150}

\newtcolorbox{myBox}{%
	colback=grayLL,colframe=grayL,top=1mm,bottom=1mm,left=1mm,right=1mm%
}
\newtcolorbox{myTitledBox}[2][]{%
	colback=grayLL,colframe=grayL,top=1mm,bottom=1mm,left=1mm,right=1mm,enlarge top by=0.5em,title={#2},fonttitle=\bfseries\small\color{gray},#1%
}

\tcolorboxenvironment{myitemizeP}{blanker,before skip=6pt,after skip=6pt,borderline west={3mm}{0pt}{purpleD}}

\tcolorboxenvironment{myitemizeG}{blanker,before skip=6pt,after skip=6pt,borderline west={3mm}{0pt}{greenD}}

\tcolorboxenvironment{myitemizeO}{blanker,before skip=6pt,after skip=6pt,borderline west={3mm}{0pt}{orangeD}}
\newtcbtheorem{thm}{Theorem}%
{colback=white,colframe=orangeD,top=1mm,bottom=1mm,left=1mm,right=1mm,enlarge top by=1mm,fonttitle=\bfseries\color{black}}{thm}
\makeatletter
\newcommand{\dotr}[1]{%
	\mathpalette\@dotr{#1}%
}
\newcommand*{\@dotr}[2]{%
	\sbox0{$\m@th#1#2$}%
	\usebox{0}%
	\raisebox{\dimexpr\ht0-\height}{$\m@th#1\@smallbullet#1\bullet$}%
	\kern\scriptspace
}
\newcommand*{\@smallbullet}[2]{%
	\scalebox{.4}{$\m@th#1#2$}%
}
\makeatother

\newcommand{\bx}{\boldsymbol{x}}

\newcommand{\calL}{\mathcal{L}}

\newcommand{\D}{\mathrm{d}}

\newcommand{\lightpath}{\bar{\bx}}

\newcommand{\loss}{\calL}
\newcommand{\Llap}{\loss_{\mathrm{lap}}}

\newcommand{\wh}{\omega_{\mathrm{h}}}

\newlength{\tikzLen}

\definecolor{darkgreen}{rgb}{0.05,0.6,0.05}

\newcommand{\mesh}{\mathcal{M}}

\newcommand{\params}{\boldsymbol{\xi}}

\definecolor{blueL}{RGB}{127.5, 177.5, 209.5}
\definecolor{orangeL}{RGB}{251,198,150}

\newcommand{\emesh}{\mesh_{\mathrm{env}}}

\newcommand{\methodname}{\textsc{PBIR-NIE}\xspace}
\newcommand{\envmapname}{\textsc{Envmap++}\xspace}

\newcommand{\best}[1]{\textcolor{blueL}{#1}}
\newcommand{\secondbest}[1]{\textcolor{orangeL}{#1}}

\begin{document}
\begin{abstract}
Glossy objects present a significant challenge for 3D reconstruction from multi-view input images under natural lighting. In this paper, we introduce \methodname{}, an inverse rendering framework designed to holistically capture the geometry, material attributes, and surrounding illumination of such objects. We propose a novel parallax-aware non-distant environment map as a lightweight and efficient lighting representation, accurately modeling the near-field background of the scene, which is commonly encountered in real-world capture setups. This feature allows our framework to accommodate complex parallax effects beyond the capabilities of standard infinite-distance environment maps. Our method optimizes an underlying signed distance field (SDF) through physics-based differentiable rendering, seamlessly connecting surface gradients between a triangle mesh and the SDF via neural implicit evolution (NIE). To address the intricacies of highly glossy BRDFs in differentiable rendering, we integrate the antithetic sampling algorithm to mitigate variance in the Monte Carlo gradient estimator. Consequently, our framework exhibits robust capabilities in handling glossy object reconstruction, showcasing superior quality in geometry, relighting, and material estimation.
\end{abstract}


\begin{teaserfigure}
    \centering
    \includegraphics[width=\linewidth]{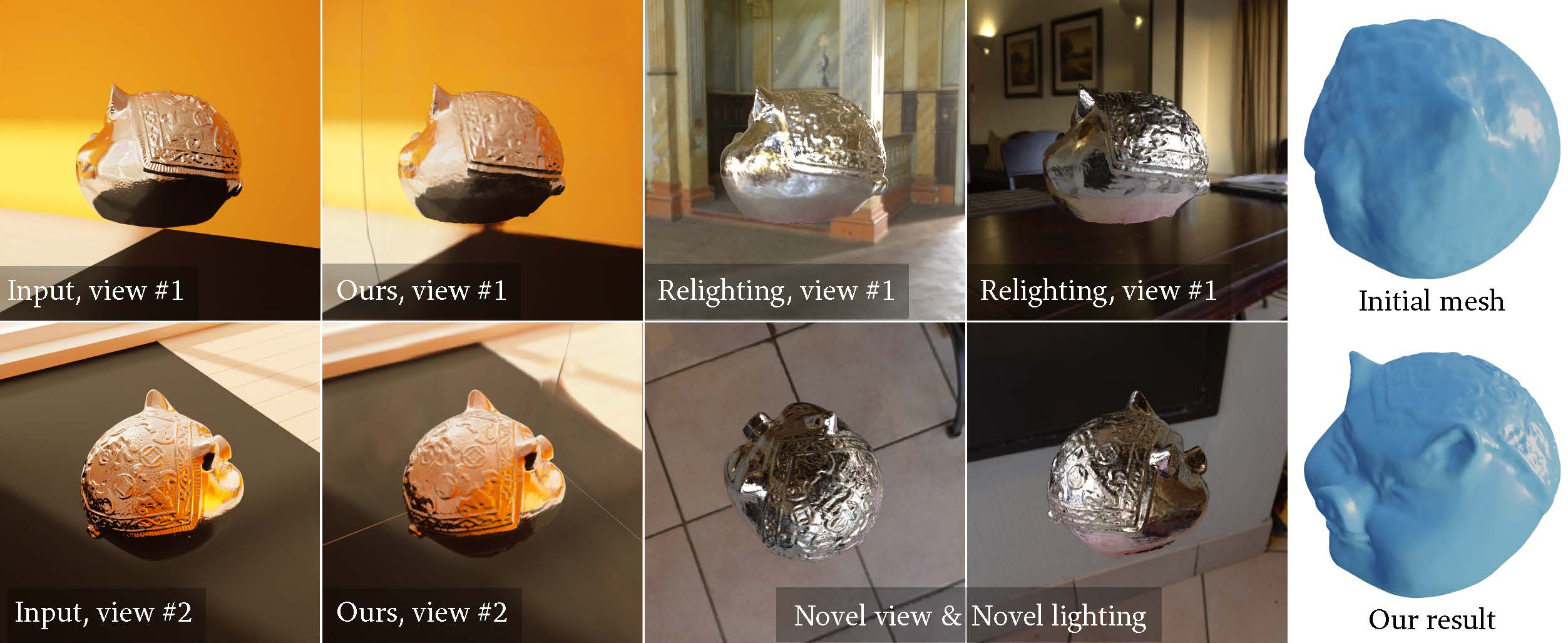}
    \vspace{-6mm}
    \caption{\textbf{We propose \methodname{}, a physics-based inverse rendering pipeline} that optimizes an object's shape, glossy surface reflectance, and non-distant lighting representation. Our method faithfully recovers the shiny and specular appearance, produces relighting results with high fidelity, and accurately captures geometric details from a rough visual hull initialization.}
    \label{fig:teaser}
\end{teaserfigure}

\maketitle
%

\section{Introduction}
\label{sec:intro}
The joint reconstruction of an object's surface geometry, material reflectance, and its surrounding illumination, commonly referred to as inverse rendering, stands as a foundational task in computer vision and graphics for 3D content creation, with applications across various fields: film making, game production, product design, and AR/VR. Despite its broad applicability, the reconstruction of non-diffuse objects remains particularly challenging due to the presence of specular reflections. These reflective surfaces introduce inconsistencies in color when captured from multiple viewpoints, creating difficulties for multi-view reconstruction methods (such as NeRF~\cite{mildenhall2020nerf} and 3D Gaussian Splatting~\cite{kerbl20233d}) that heavily rely on view consistency and feature matching.

Recovering the appearance of glossy objects presents challenges due to the high-frequency reflections of complex surrounding environments. Previous inverse rendering methods often rely on an \emph{infinitely distant} 2D environment map~\cite{munkberg2022extracting}, either in a discretized form or parameterized using mixtures of spherical Gaussians~\cite{zhang2021physg, jin2023tensoir} for representing the background. However, the infinite-distance assumption is easily violated in real-world capture setups. One solution for capturing complex background parallax effects is to represent the lighting as a neural radiance field (NeRF)~\cite{mildenhall2020nerf}, as demonstrated by prior works~\cite{ling2024nerf, wang2024inverse, zhuangNeAIPreconvolutedRepresentation2023}. Utilizing NeRF as a background can however be computationally expensive, and the absence of an efficient importance sampling method further complicates its practical use in inverse rendering applications. To address this, we propose a lightweight background representation, \envmapname{}, designed to efficiently model both near-field and far-field background illumination with a parallax-aware environment map. This representation strikes a balance between quality and performance, as demonstrated in our glossy object reconstruction task. 

Prior inverse rendering works~\cite{luan2021unified,caiPhysicsBasedInverse2022, sunNeuralPBIRReconstructionShape2023} rely on good initialization of the surface topology (e.g. via neural rendering methods such as NeuS~\cite{wang2021neus}), and further refine the geometric details with a differentiable renderer. On the other hand, neural implicit evolution (NIE)~\cite{mehtaLevelSetTheory2022} formulates a level-set evolution method for parametrically defined implicit surfaces that does not require mesh extraction to be differentiable. We show that this method can be combined with a physics-based differentiable renderer and allows topology change during optimization. We integrate NIE into our framework and optimize an underlying neural signed distance field (SDF). This integration allows our method to handle objects with complex topology, including thin structures, holes, and other subtle geometric features; thus, it is robust to poor initialization. Our framework, dubbed \methodname{}, not only captures the geometry, material properties, and surrounding illumination of glossy objects but also handles topology changes seamlessly.

Lastly, when dealing with highly specular BRDFs (i.e. with very low roughness) in physics-based differentiable rendering, the naively estimated geometric gradients can be very noisy. \citet{zhang2021antithetic} demonstrate that antithetic sampling can significantly reduce the variance in such cases. However, despite enabling faster convergence, this method can be tedious to implement, especially for indirect illumination where the number of light paths grows exponentially. We propose a simple yet efficient modified variant of antithetic sampling, incorporating Russian roulette and roughness regularization beyond secondary bounces of each light path. This variant is easy to implement and we show it is robust for glossy-object reconstruction.

In summary, our contributions include:
\begin{itemize}
    \item \envmapname, a lightweight non-distant lighting representation that efficiently models both the near- and far-field background illumination with a parallax-aware environment map, overcoming limitations of infinitely distant and NeRF emitters.
    \item Integration of neural implicit evolution into PBIR, allowing for topological changes and enhancing the framework's ability to handle complex object geometries.
    \item An efficient antithetic sampling variant that improves the handling of highly glossy BRDFs in differentiable rendering with gradient variance reduction for fast and accurate reconstruction.
    \item An end-to-end pipeline achieving state-of-the-art geometry, material, and lighting estimation for glossy-object reconstruction, enabling realistic view synthesis and relighting.
\end{itemize}


\section{Related Work}
\label{sec:related}
\subsection{Neural Surface Reconstruction}
Neural rendering, particularly neural implicit representation, has seen significant advancements in 3D reconstruction in recent years. Neural radiance fields (NeRF)~\cite{mildenhall2020nerf} and its variants~\cite{muller2022instant, chen2022tensorf} utilize volume rendering on scene representations with neural density fields and view-dependent color fields, resulting in impressive photorealism in novel view synthesis. However, geometry extracted from volumetric density fields often exhibits flawed surfaces. 

Alternatively, representing the underlying geometry with a signed distance field (SDF) has shown promise for improved surface reconstruction in recent neural SDF-based approaches such as NeuS~\cite{wang2021neus}, VolSDF~\cite{yariv2021volume}, and  PermutoSDF~\cite{rosu2023permutosdf} Unfortunately, when capturing glossy objects with shiny and metallic materials (such as a soda can, a stainless kettle, or a polished silver spoon), both density-based and SDF-based approaches struggle to faithfully reconstruct the geometry. 

\citet{caiPhysicsBasedInverse2022} leverage MeshSDF~\cite{remelli2020meshsdf}, a differentiable version of Marching Cubes, to implicitly optimize an SDF by rendering the extracted mesh. However, this method is computationally expensive. Conversely, \citet{viciniDifferentiableSignedDistance2022} and \citet{bangaru2022differentiable} reparameterize the discontinuities in direct SDF rendering to avoid meshes entirely, but this approach complicates extending the method to multiple bounces and implementing variance reduction techniques. 

\subsection{Glossy Surface Reconstruction}
Recently, glossy surface reconstruction has received increased attention in the neural and inverse rendering community. Ref-NeRF~\cite{verbin2022ref} introduced integrated directional encoding, replacing NeRF's view-dependent color field with a representation of reflected radiance based on surface normals, which improved the reconstructed surface quality. This approach was extended to SDF-based frameworks in Ref-NeuS~\cite{ge2023ref}. SpecNeRF~\cite{ma2023specnerf} proposed 3D Gaussian-based encoding to enhance NeRF's reflection modeling capabilities. Neural directional encoding~\cite{wuneural2024} transfers feature-grid-based spatial encoding into the angular domain and considers near-field specular interreflections with cone tracing, further improving the modeling of complex reflections. Neural plenoptic function~\cite{wang2024inverse} proposed to represent global illumination via a 5D representation based on NeRFs and raytracing. Our most relevant baseline is NeRO~\cite{liuNeRONeuralGeometry2023}, which introduced a two-stage, NeuS-based pipeline that explicitly incorporates the rendering equation into the neural reconstruction framework, demonstrating superior geometry quality on reflective objects. However, their pipeline relies on approximation to ensure the computation is tractable, for instance, using neural networks to predict occlusion and indirect illumination. While our work shares a similar goal, we primarily approach the problem by solving the rendering equation without approximation.


\subsection{Material and Lighting Estimation}
Beyond surface geometry, inverse rendering~\cite{marschner1998inverse, ramamoorthi2001signal} typically also involves estimating the material properties (e.g., SVBRDF) and, in some cases, the surrounding illumination (depending on the capture setup). Traditional data-driven acquisition methods~\cite{xia2016recovering, dong2014appearance, dong2010manifold, aittala2013practical, nam2018practical, nam2016simultaneous, zhou2016sparse} often assume sparsity in spatially-varying surface reflectance and frame the capture as a complex non-linear optimization problem. Similarly, recent neural reconstruction methods address this analysis-by-synthesis problem through differentiable rendering for intrinsic decomposition. They employ various differentiable rendering techniques, including neural renderers (PhySG~\cite{zhang2021physg}, NeRFactor~\cite{zhang2021nerfactor}, TensoIR~\cite{jin2023tensoir} and others~\cite{boss2021nerd, boss2021neural, zhang2022iron, zhang2022modeling, kuang2022neroic, srinivasan2021nerv}), fast differentiable rasterizers~\cite{munkberg2022extracting} or differentiable Monte Carlo raytracers~\cite{luan2021unified, caiPhysicsBasedInverse2022, sunNeuralPBIRReconstructionShape2023, hasselgren2022shape}. 
Our approach operates within the differentiable path tracing framework, parameterizing spatially-varying surface reflectance with an analytic microfacet BRDF model using the GGX distribution~\cite{walter2007microfacet}. However, unlike previous methods that typically represent lighting with a 2D distant environment map, we also consider near-field background illumination, crucial for glossy object reconstruction. Similar to \citet{ling2024nerf} and NeRO~\cite{liuNeRONeuralGeometry2023} that represent the background illumination with non-distant environment emitters (either a NeRF or two direct-indirect separable lighting MLPs), we introduce a lightweight, parallax-aware environment map representation. This approach demonstrates robustness to near-field and far-field lighting conditions while remaining efficient in inverse rendering optimization. 

\begin{figure*}[t]
\centering
\includegraphics[width=0.86\textwidth]{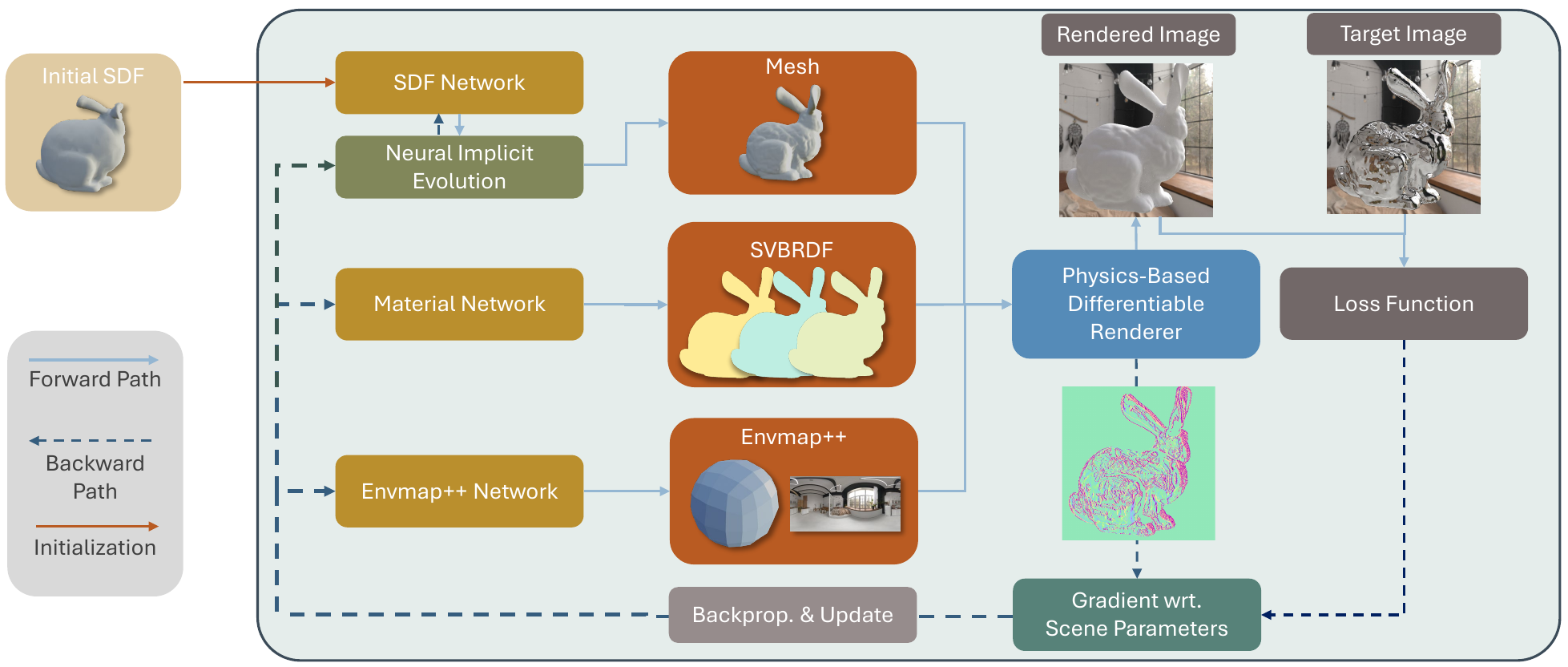}
\vspace{-0.7\baselineskip}
\caption{\textbf{Overview of our \methodname pipeline.} Our pipeline takes a set of multi-view images capturing a glossy object and an initial shape as input. It then reconstructs the scene's geometry, material properties, and lighting using a physics-based inverse rendering (PBIR) approach. The iterative refinement process includes: 
1) \textbf{Forward Pass}: Rendering an image by employing physics-based differentiable rendering. This involves using an explicit mesh extracted with a non-differentiable Marching Cubes algorithm to represent the neural implicit surface for shape, and material networks for surface properties, while leveraging information from input training views. Additionally, \envmapname{} is utilized for enhanced lighting representation, replacing the standard infinite-distance environment map to handle non-distant background illumination.
2) \textbf{Backward Pass}: Comparing the rendered image to the ground truth and computing gradients with respect to scene parameters. We use neural implicit evolution (NIE)~\cite{mehtaLevelSetTheory2022} to facilitate the backpropagation of gradients from the extracted mesh to the neural implicit surface, bypassing the non-differentiable extraction step. 
3) \textbf{Update}: Adjusting scene parameters (geometry, material, lighting) via backpropagation to minimize the difference between the rendered and ground truth image.
}
\label{fig:our_pipeline}
\end{figure*}

\section{Preliminaries}
\label{sec:prelim}


In this section, we briefly revisit mathematical and algorithmic preliminaries related to physics-based differentiable rendering~\cite{zhao2020physics}, and discuss the advantages and disadvantages of the differentiable renderer we use and how it affects our pipeline.

Given a virtual object described by parameters $\params$, we render the image with Monte Carlo rendering based on the path integral formulation introduced by \citet{veach1997robust}:
\begin{equation}
    \label{eq:veach_path_integral}
    I = \int_{\Omega} f(\lightpath) \, \D \mu(\lightpath),
\end{equation}
where $\Omega := \cup_{N\geq 1}\mathcal{M}^{N+1}$ is the path space consisting of light transport paths $\lightpath = (\bx_0, \bx_1, ..., \bx_N)$ with $\mathcal{M}$ being the union of all object surfaces, $f$ is the measurement contribution function, and $\mu$ is the corresponding area-product measure. 

Computing image gradients involves differentiating pixel intensities in Eq.~\eqref{eq:veach_path_integral} with respect to $\params$, which is not a trivial process due to the existence of discontinuities in the integrand. \citet{Zhang:2020:PSDR} and \citet{bangaru2020warpedsampling} presented two different paradigms for tackling this problem: one directly track the discontinuities and one eliminates the discontinuities via reparameterization. Both can accurately differentiate Eq.~\eqref{eq:veach_path_integral}.


Given estimation of the gradient $\frac{\D I}{\D\params}$, we are able to reconstruct scene parameters from images using an optimization approach: render images using the initial scene parameters, compute the loss with regard to the target images, obtain the gradient of the loss with respect to the scene parameters, update the scene parameters using gradient descent and repeat until convergence.

\section{Our Method}
\label{sec:ours}

Our \textbf{\methodname{}} pipeline reconstructs the shape, material, and background lighting of opaque objects from multi-view images with known camera poses. While it can handle diffuse objects, it is specifically designed to perform well on glossy ones, which are more sensitive to light reconstruction quality due to reflected background details. Previous methods often use an environment map to represent background illumination, assuming light originates from infinitely far away. However, this results in blurry reconstructions under indoor lighting, often compensated by increased roughness. To address this, we introduce a new model \textbf{\envmapname{}} (Sec. \ref{sec:Envmap++}), a lightweight non-distant lighting representation that is suitable for both near-field and far-field lighting conditions. Additionally, optimizing the shape and material of glossy objects requires special considerations; for shape optimization, we employ neural implicit evolution with careful initialization to ensure robust geometry handling (Sec. \ref{sec:shape}); we introduce a modified version of antithetic sampling~\cite{zhang2021antithetic} for variance reduction in the Monte Carlo gradient estimation (Sec. \ref{sec:material}).

The differentiable renderer is a crucial component of our pipeline. We have selected Mitsuba 3 \cite{jakob2022mitsuba3} due to its implementation of numerous state-of-the-art techniques pertinent to our application. 
However, one limitation constrains our choice of scene primitives: although neural networks serve as effective representations for primitives, evaluating them within the rendering loop is suboptimal. Mitsuba 3 supports two evaluation modes—megakernel and wavefront—enabled by its auto-differentiation engine, Dr. Jit \cite{Jakob2020DrJit}. While the megakernel mode is significantly more efficient than the wavefront mode for Monte Carlo rendering, it currently does not support the evaluation of neural networks within the kernel. Conversely, the wavefront mode allows for such evaluations but incurs substantial performance and memory costs. To facilitate practical applications, we opt for the megakernel mode, trading some flexibility for efficiency by evaluating all neural networks prior to the rendering step. This trade-off plays an important role in our pipeline design. 

\subsection{\envmapname{}: Non-Distant Environment Map}
\label{sec:Envmap++}


We propose representing the background using a deformed spherical emitter, optimizing its vertices to simulate the parallax effect. Assume (without loss of generality) the object and cameras are bounded by a unit sphere $\mathcal{S}$. The emitter consists of a mesh $\emesh$ deformed from the starting sphere $S$. Each vertex $x^{\mathcal{M}}$ of $\emesh$ is represented as $x - n(x) d(x)$, where $x$ and $n(x)$ are the corresponding vertex and vertex normal on $\mathcal{S}$, and $d(x) > 0$ is the displacement amount to be optimized. Furthermore, $L_{\mathrm{e}}(x)$ is the optimized uniformly emitted radiance of each vertex. The normals point inside the sphere to ensure visibility. To handle the wide range of $d$, we adopt the \textit{inverted sphere parameterization} from NeRF++ \cite{zhangNeRFAnalyzingImproving2020} and optimize $r(x) \in (0,1)$ such that $d(x)=\frac{1}{r(x)}-1$. This adaptation leads to our light representation termed \envmapname{}.

More generally, for any bounding sphere, $\emesh$ can be represented as:
\begin{equation}
\emesh = \{s \left(x - n(x) (\frac{1}{r(x)} - 1) \right) + c \; |  \; x \in \mathcal{S}  \}
\end{equation}
where $s\in \mathbb{R}_{>0}$ is a scaling factor (bounding sphere radius) and $c \in \mathbb{R}^3$ is the center of the bounding sphere.

In our implementation, we employ a deformed cube for the mesh structure of $\emesh$ to ensure even triangle distribution, analogous to cube mapping techniques. Direct optimization of $r(x)$ and radiance values $L_{\mathrm{e}}(x)$ can be unstable. Hence, we utilize small MLPs with \textit{permutohedral encoding} \cite{rosuPermutoSDFFastMultiView2023} to predict these values for a given $x \in \mathcal{S}$, where $x$ is represented as a 3D unit vector when querying the encoding, to avoid poles.
Laplacian loss $\Llap(\emesh)$ is further applied to smooth the background emitter mesh. 

While the actual background rarely resembles a sphere, this approximation suffices in practice as a non-distant light representation. When vertices are positioned infinitely far away ($r \rightarrow 0$ in the limit), this representation approaches the traditional environment map, accommodating a mixture of near-field and far-field lighting scenarios. In addition, since this is simply a textured area light, we can importance sample it without additional effort.

\subsection{Shape Reconstruction}
\label{sec:shape}

\subsubsection{Shape Initialization}
Our pipeline uses an initialization stage to start with potentially imperfect predictions of object shape, reflectance, and illumination, similar to Neural-PBIR~\cite{sunNeuralPBIRReconstructionShape2023}. In a later stage, initialized by these predictions, we refine the initial results to obtain the final high-quality reconstruction. We found two initialization to perform the best in different cases. First, our pipeline utilizes the initial SDF reconstructed by PermutoSDF~\cite{rosuPermutoSDFFastMultiView2023} due to its robustness in handling diffuse and rough glossy objects.

Second, for highly glossy objects where PermutoSDF can fail and produce holes, we implement a voxel-based visual hull algorithm~\cite{laurentini1994visual} to obtain the initial shape. We first estimate a set of masks based of the input images using a salient object segmentation model such as U$^{2}$-Net~\cite{qin2020u2}. Then we back-project the voxels within the scene bounding box to determine if they are within the foreground mask for a given view. We keep a voxel if the fraction of input views where it falls into the foreground mask is above a given threshold. This approach tends to produce fewer holes and broken geometry features, despite having less accurate boundaries than the result from PermutoSDF. The resulting occupancy grid can be transformed into an SDF using a Euclidean distance transform and serve as our initialization. 

\subsubsection{Neural Implicit Evolution}  
\label{sec:nie}

\begin{figure}[tb]
\centering
\includegraphics[width=1.\linewidth]{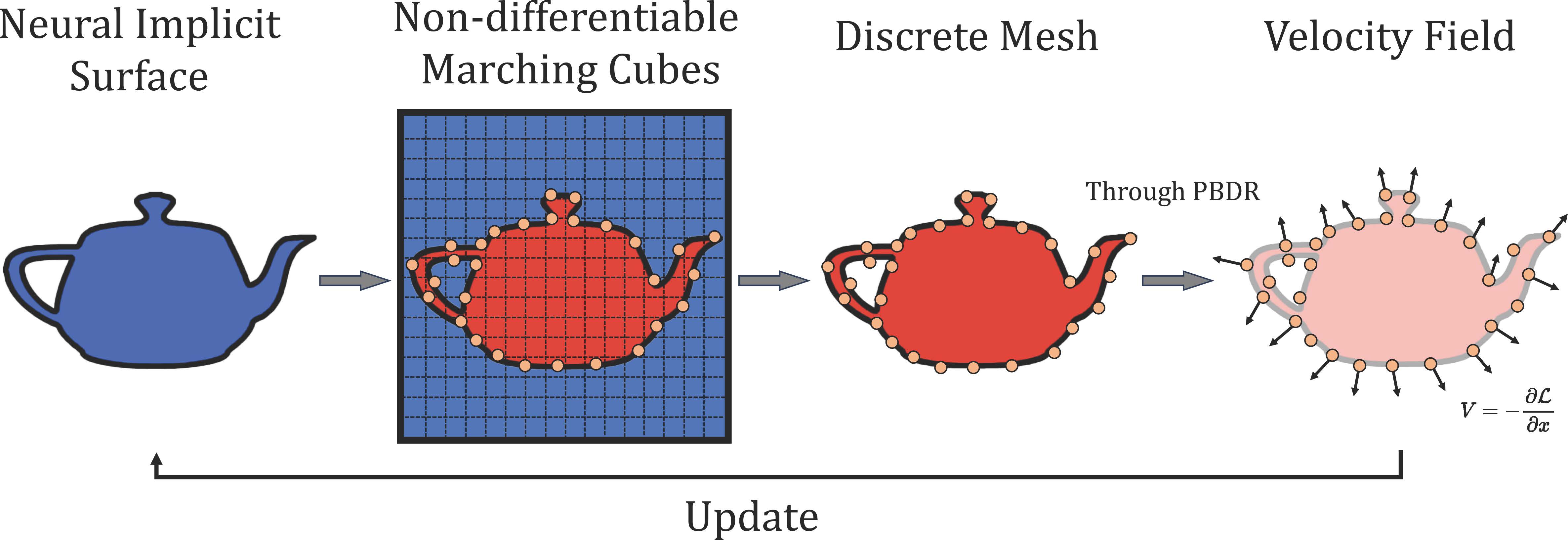}
\vspace{-1.5\baselineskip}
\caption{\textbf{Neural Implicit Evolution (NIE).} Here we illustrate our \methodname{} pipeline for optimizing the underlying geometry using neural implicit evolution (NIE)~\cite{mehtaLevelSetTheory2022}. We represent the geometry with a neural signed distance field (SDF) and employ a surface extraction algorithm, such as non-differentiable marching cubes, to obtain a discretized surface mesh. Next, we compute mesh vertex gradients using physics-based differentiable rendering (PBDR) and update the neural SDF through NIE with the obtained velocity field.}
\end{figure}

Specular highlights or inaccurate segmentation can create holes in our initial mesh, so our shape optimization routine needs to handle topological changes to correct these issues. Further, many objects have thin features such as cup handles. While representing the shape using a neural implicit function would be an ideal solution, only a few works \cite{caiPhysicsBasedInverse2022, bangaru2022differentiable, viciniDifferentiableSignedDistance2022} have successfully combined this approach with a physically based differentiable renderer despite its popularity in neural surface reconstruction. The root of this problem is that accurately estimating the derivative of Equation \ref{eq:veach_path_integral} requires handling a boundary term \cite{Zhang:2020:PSDR}, which requires tracking discontinuities at object boundaries, but it is particularly challenging for implicit surfaces.


Our pipeline \textbf{\methodname{}} incorporates neural implicit evolution (NIE)~\cite{mehtaLevelSetTheory2022} as the backbone for propagating surface gradients from mesh-based differentiable rendering to the underlying neural SDF. Given a neural implicit function $\Phi(x; \theta)$, where $\theta$ is the set of parameters of a neural network, we use its isosurface $\mesh=\{x \,|\, \Phi(x; \theta) = 0\}$ to represent the object we are reconstructing. NIE optimizes this surface by first extracting a mesh using non-differentiable Marching Cubes and obtains the vertex gradients w.r.t the training loss, $\frac{\partial L}{\partial x}$, through a differentiable renderer. It then constructs a flow field $V=-\frac{\partial L}{\partial x}$ and creates the target implicit function $\phi(x)=\Phi(x; \theta)-\Delta t (\nabla \Phi(x;\theta) \cdot V)$, where $\Delta t$ is the time step. Finally, the implicit function $\Phi(x; \theta)$ is optimized by minimizing its difference with the target implicit function using gradient descent: 
\begin{equation}
    \min _\theta J(\theta)=\frac{1}{|\mesh|} \sum_{x \in \mesh}\|\phi(x)-\Phi(x; \theta)\|^2.
\end{equation}

NIE is provably better than MeshSDF (further reading in \cite{mehtaLevelSetTheory2022}) and it can be integrated into an existing differentiable renderer without requiring differentiable marching cubes; however, only minimal examples of differentiable rendering are shown in the original NIE work; our method is the first to use it in a full inverse rendering pipeline for geometry, materials and lighting.

We introduce two adjustments to NIE for reconstructing glossy objects. First, we clamp the magnitude of the flow field. When the roughness of the object is very low, $\frac{\partial L}{\partial x}$ becomes extremely high and disturbs the training stability. With the formulation of NIE, as long as the implicit surface is evolving towards the correct direction, the magnitude of the flow field is not important. Thus, we simply rescale $\frac{\partial L}{\partial x}$ so that its magnitude is at most $\epsilon_v$.

Second, although NIE is designed to work with level set functions (where the eikonal constraint $||\nabla_x \Phi(x;\theta)|| = 1$ is not enforced), we found that the absence of this enforcement leads to degraded results at high learning rates. To speed up training, we add the eikonal constraint and enforce the implicit function to behave as a signed distance function. The commonly used eikonal loss $\mathcal{L}_{\text{Eik}}=\frac{1}{2}||\nabla_x \Phi(x;\theta) - 1||^2$ is in fact unstable, as demonstrated in \cite{liDistanceRegularizedLevel2010a, yangStEikStabilizingOptimization2023}. As a result, we use the DRLSE loss by \citet{liDistanceRegularizedLevel2010a}: 
\begin{equation}
    \mathcal{L}_{\text{DRLSE}}(s)= 
    \begin{cases}
        \frac{1}{(2 \pi)^2}(1-\cos (2 \pi s)), & \text { if } s \leq 1 \\ 
        \frac{1}{2}(s-1)^2, & \text { if } s > 1
    \end{cases},
\end{equation}
where $s=||\nabla_x \Phi(x;\theta)||$. We usually use a weight of 1e-2 in our experiments.



\begin{figure}
     \centering
     \begin{subfigure}[b]{0.116\textwidth}
         \centering
         \includegraphics[trim={140 140 140 140}, clip, width=\linewidth]{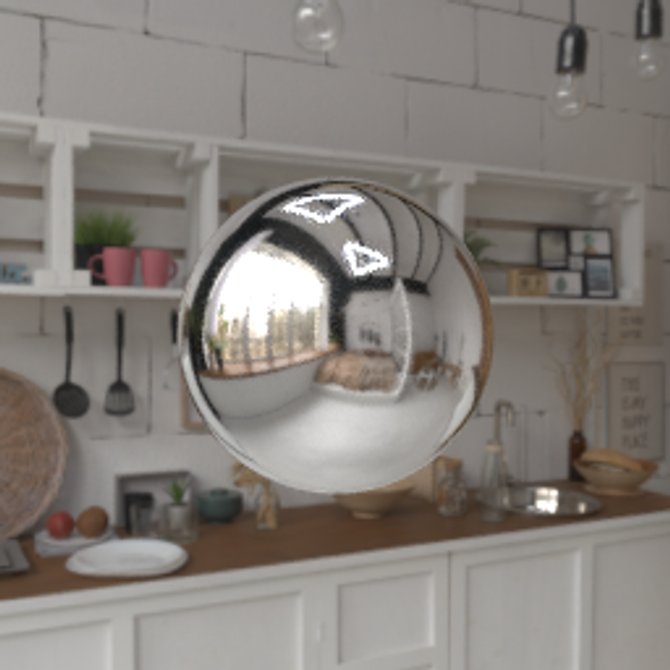}
         \caption{Rendering}
         \label{fig:rgb_as}
     \end{subfigure}
     \hfill
     \begin{subfigure}[b]{0.116\textwidth}
         \centering
         \includegraphics[trim={105 105 105 105}, clip, width=\linewidth]{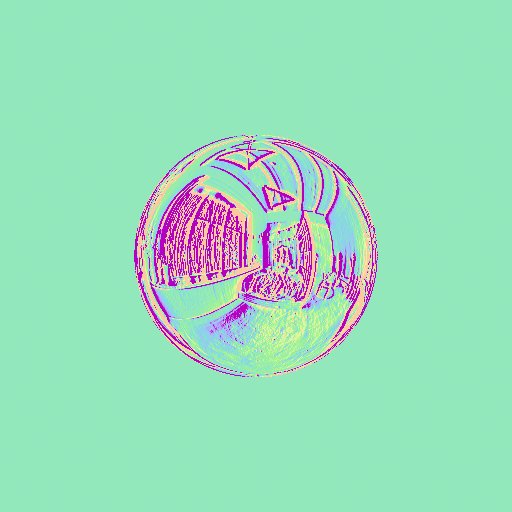}
         \caption{Reference (FD)}
         \label{fig:as_fd}
     \end{subfigure}
     \hfill
     \begin{subfigure}[b]{0.116\textwidth}
         \centering
         \includegraphics[trim={105 105 105 105}, clip, width=\linewidth]{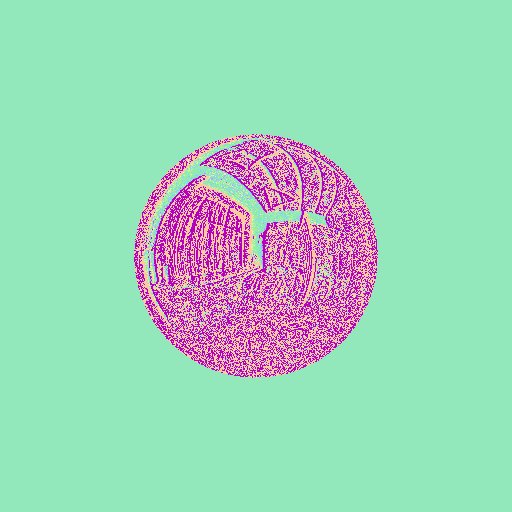}
         \caption{AD w/o AS}
         \label{fig:as_ad}
     \end{subfigure}
     \hfill
     \begin{subfigure}[b]{0.116\textwidth}
         \centering
         \includegraphics[trim={105 105 105 105}, clip, width=\linewidth]{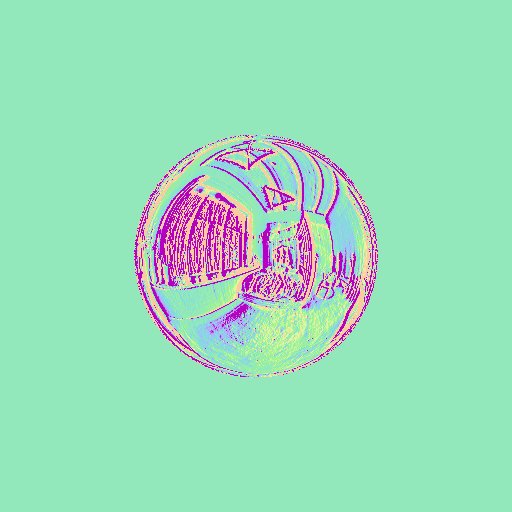}
         \caption{Ours}
         \label{fig:as_ad_anti}
     \end{subfigure}
    \vspace{-1.5\baselineskip}
    \caption{\textbf{Antithetic Sampling.} When dealing with highly glossy objects (such as a chrome ball in Fig.~\ref{fig:rgb_as}), traditional BSDF sampling techniques may result in high variance gradients. In Fig.~\ref{fig:as_fd}, we compute finite differences and display the ground-truth gradient image corresponding to the shape translation of a glossy object with a low roughness value of 0.05. Without antithetic sampling (Fig.~\ref{fig:as_ad}), the gradient image appears noisy, leading to unstable training. However, by applying antithetic sampling (Fig.~\ref{fig:as_ad_anti}), we achieve a significantly more reliable Monte Carlo gradient estimation with the same number of samples.}
    \label{fig:as}
\end{figure}

\subsection{Material}
\label{sec:material}
\subsubsection{BRDF Model} 
We model the material using a simplified version of the Disney BRDF~\cite{karis2013real}. It consists of a diffuse lobe and an anisotropic microfacet specular lobe modeled with the GGX normal distribution function \cite{ggx}. It is also adopted by other works for inverse rendering~\cite{sunNeuralPBIRReconstructionShape2023, luan2021unified, bi2020deep}. Although we use an explicit mesh for rendering, the mesh's connectivity and topology change constantly as the underlying implicit surface evolves. Thus, applying 2D textures to it is impractical. While we can encode the material properties using an MLP, querying them at each intersection point within the rendering loop requires turning off the megakernel mode in Mitsuba 3, resulting in significant performance regression. Instead, we store the predicted material properties at the mesh vertices and then interpolate them to estimate the material properties at the intersection points. The meshes extracted with marching cubes are very high-resolution, so the interpolation errors remain small. If higher-resolution textures are desired, we can run an additional stage with explicit UV mapping after the topology has stabilized. 

\subsubsection{Antithetic Sampling} 
\citet{zhang2021antithetic} found that using glossy or near-specular BRDFs with traditional sampling techniques results in high variance in gradient estimation (Fig. \ref{fig:as}). This excessive noise negatively impacts shape optimization. The root of this issue is that the derivative of the normal distribution function $\frac{\mathrm{d} D}{\mathrm{d} \wh}$ grows rapidly as the roughness decreases. The remedy they proposed is to create an antithetic sample $\wh'$ by mirroring the sampled half vector $\wh$ along the normal direction, which has the same $D$-value. Since $\frac{\mathrm{d} D}{\mathrm{d} \wh}=-\frac{\mathrm{d} D}{\mathrm{d} \wh'}$, the derivatives cancel out and the variance of the gradient estimation is reduced. 

Since our goal is to reconstruct glossy objects, employing this technique is essential to ensure robustness. The antithetic sample requires us to trace an additional path. To simplifies our implementation, instead of tracing two paths at each intersection, we render the image twice, once with normal BRDF sampling and once with antithetic sampling, and average them at the end. For multi-bounces, we only do antithetic sampling at the first bounce. The same random seed is used in the two passes to ensure correlation. This simplification is effective as demonstrated in Fig.~\ref{fig:as_ad_anti}.

\section{Implementation details}
\label{sec:impl}
Our pipeline makes heavy use of neural networks for scene primitives. All of our neural networks share the same architecture: a small MLP with permutohedral encoding~\cite{rosu2023permutosdf}. This ensures memory efficiency while maintaining expressiveness. We use 4 hidden layers of 32 neurons for the SDF network, and 2 hidden layers of 32 neurons for the material network. The emitted radiance $L_{\mathrm{e}}(x)$ and inverted distance $r(x)$ in \envmapname are predicted by two separate MLPs with 2 hidden layers of 32 neurons.

We use the Adam optimizer~\cite{kingma2014adam}, with learning rates ranging from 1e-3 to 1e-4 depending on the scenes. We optimize most scenes with 2000 iterations with a batch size of 1. The resolution of the non-differentiable marching cubes during the NIE step is $256^3$. As mentioned in Sec. \ref{sec:Envmap++}, we use a sphere mesh constructed as a deformed cube as the starting mesh. The resolution of each face of the cube is $32^2$.

We use a modified version of the \texttt{prb\_projective} integrator from Mitsuba 3 with antithetic sampling support. By default, it supports paths of arbitrary depth using the Russian roulette stopping criterion.

\section{Results}
\label{sec:results}
To demonstrate the effectiveness of our method, we present reconstructions on synthetic input images and the real-world capture dataset Stanford-ORB~\cite{kuang2024stanford}. We compare the reconstructions obtained with our pipeline against three state-of-the-art baselines: NeRO~\cite{liuNeRONeuralGeometry2023}, Neural-PBIR~\cite{sunNeuralPBIRReconstructionShape2023}, and NeRF Emitter~\cite{ling2024nerf}. We demonstrate superior reconstruction quality in terms of geometry and lighting (Sec.~\ref{sec:comparison_with_baselines}). Additionally, we conduct ablation studies to evaluate several components of our pipeline (Sec.~\ref{sec:evaluation_and_ablation}). Please refer to the supplement for more results.

\subsection{Comparison with Baselines}
\label{sec:comparison_with_baselines}

\paragraph*{Comparison with NeRO~\cite{liuNeRONeuralGeometry2023}.} 

\begin{figure*}[htbp]
\centering
\resizebox{0.93\textwidth}{!}{%
\setlength{\tabcolsep}{1.0pt}
\renewcommand{\arraystretch}{0.5}
\begin{tabular}{cccccccccc}
& \multicolumn{3}{c}{Reference} & \multicolumn{3}{c}{Ours} & \multicolumn{3}{c}{NeRO~\cite{liuNeRONeuralGeometry2023} } \\ 
\rotatebox{90}{\parbox{0.11\textwidth}{\centering \textsc{Bell} }} &
\includegraphics[trim={0 0 0 0}, clip, width=0.11\textwidth]{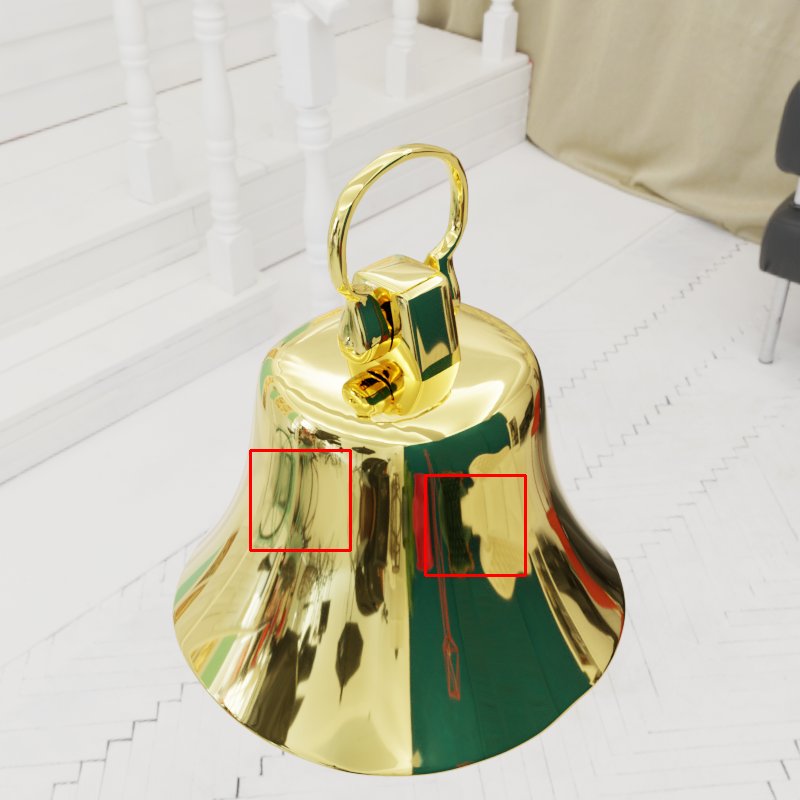} &
\includegraphics[trim={0 0 0 0}, clip, width=0.11\textwidth]{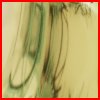} &
\includegraphics[trim={0 0 0 0}, clip, width=0.11\textwidth]{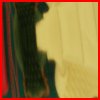} &
\includegraphics[trim={0 0 0 0}, clip, width=0.11\textwidth]{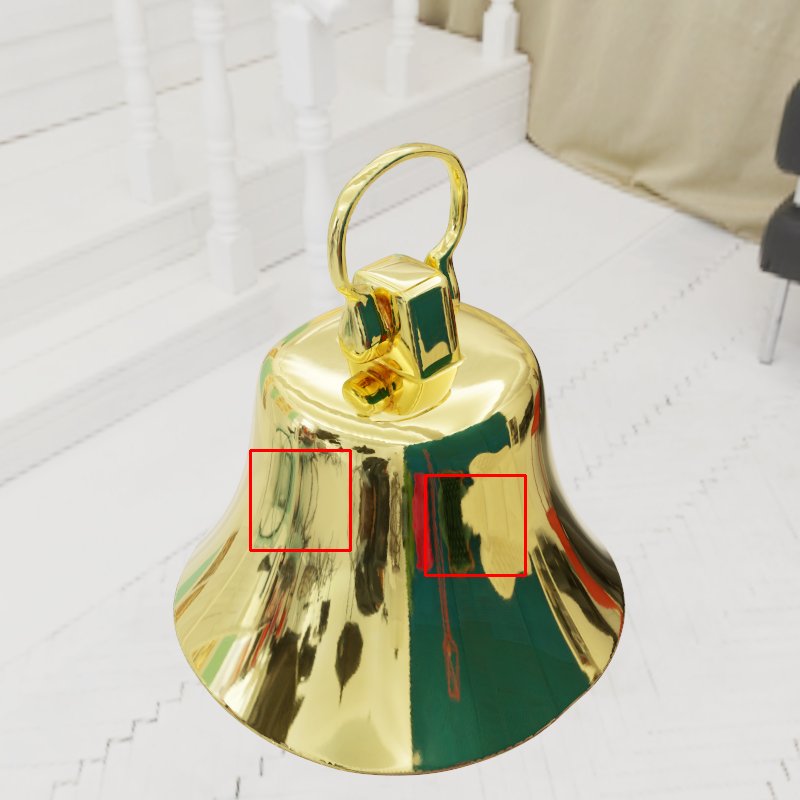} &
\includegraphics[trim={0 0 0 0}, clip, width=0.11\textwidth]{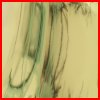} &
\includegraphics[trim={0 0 0 0}, clip, width=0.11\textwidth]{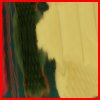} &
\includegraphics[trim={0 0 0 0}, clip, width=0.11\textwidth]{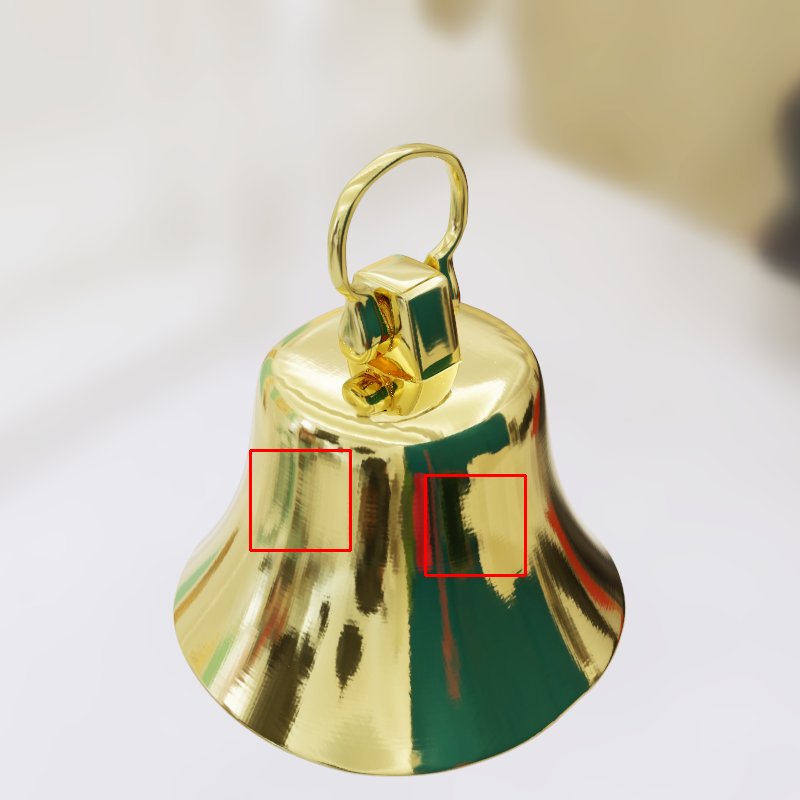} &
\includegraphics[trim={0 0 0 0}, clip, width=0.11\textwidth]{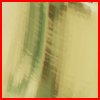} &
\includegraphics[trim={0 0 0 0}, clip, width=0.11\textwidth]{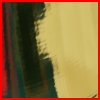} \\ 
\rotatebox{90}{\parbox{0.11\textwidth}{\centering \textsc{Teapot} }} & 
\includegraphics[trim={0 0 0 0}, clip, width=0.11\textwidth]{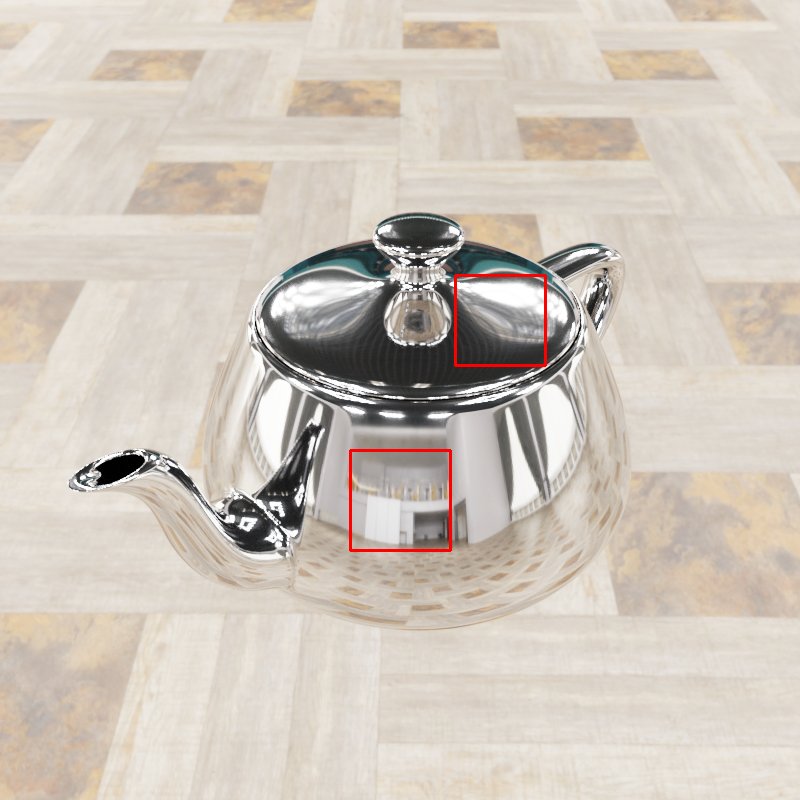} &
\includegraphics[trim={0 0 0 0}, clip, width=0.11\textwidth]{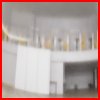} &
\includegraphics[trim={0 0 0 0}, clip, width=0.11\textwidth]{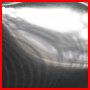} &
\includegraphics[trim={0 0 0 0}, clip, width=0.11\textwidth]{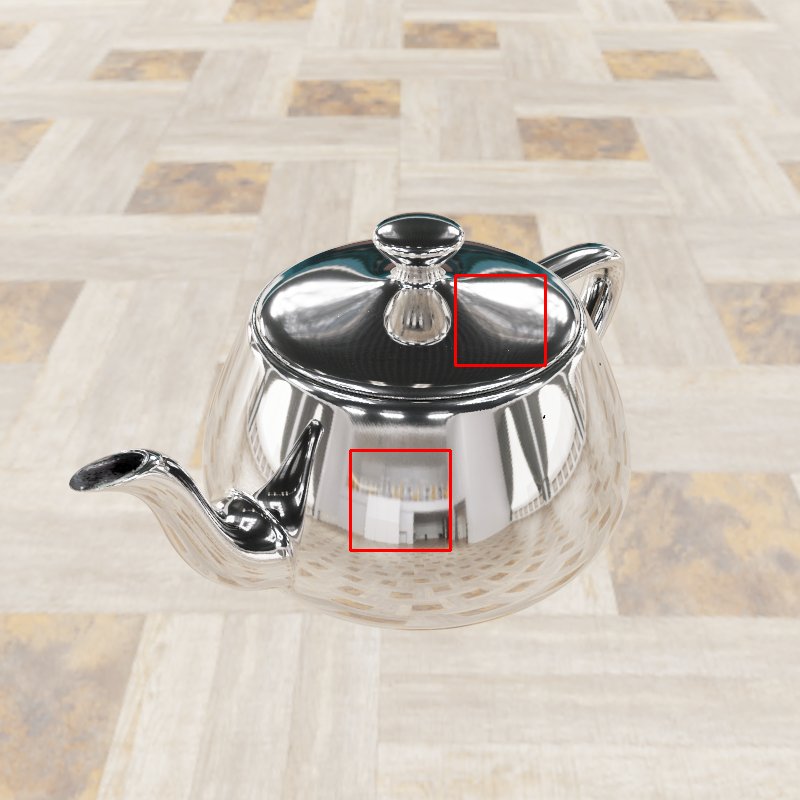} &
\includegraphics[trim={0 0 0 0}, clip, width=0.11\textwidth]{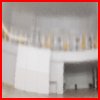} &
\includegraphics[trim={0 0 0 0}, clip, width=0.11\textwidth]{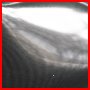} &
\includegraphics[trim={0 0 0 0}, clip, width=0.11\textwidth]{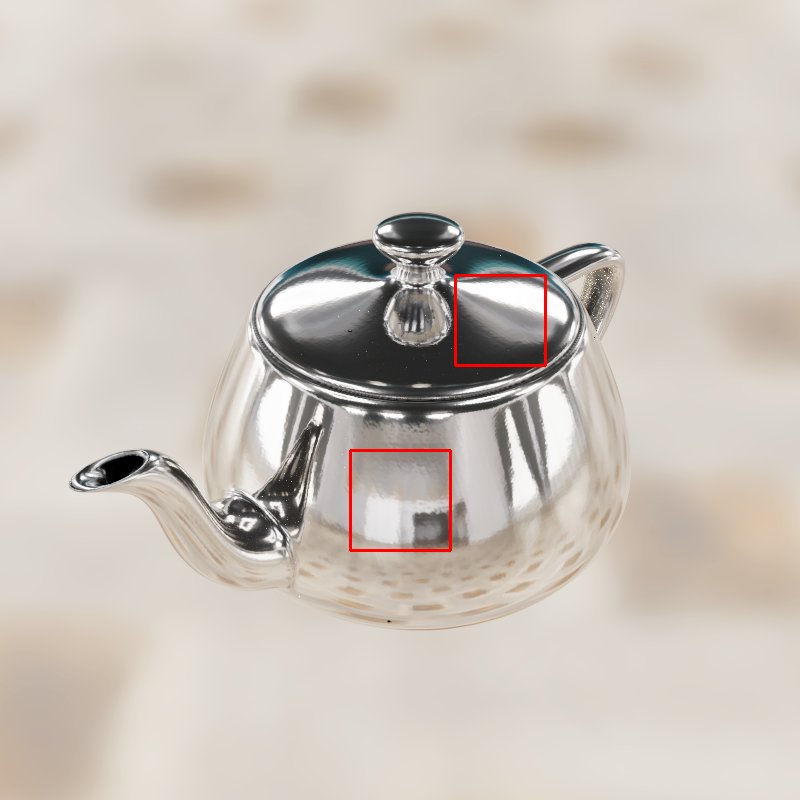} &
\includegraphics[trim={0 0 0 0}, clip, width=0.11\textwidth]{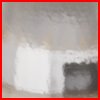} &
\includegraphics[trim={0 0 0 0}, clip, width=0.11\textwidth]{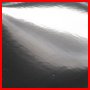} \\ 
\rotatebox{90}{\parbox{0.11\textwidth}{\centering \textsc{Potion} }} & \includegraphics[trim={0 0 0 0}, clip, width=0.11\textwidth]{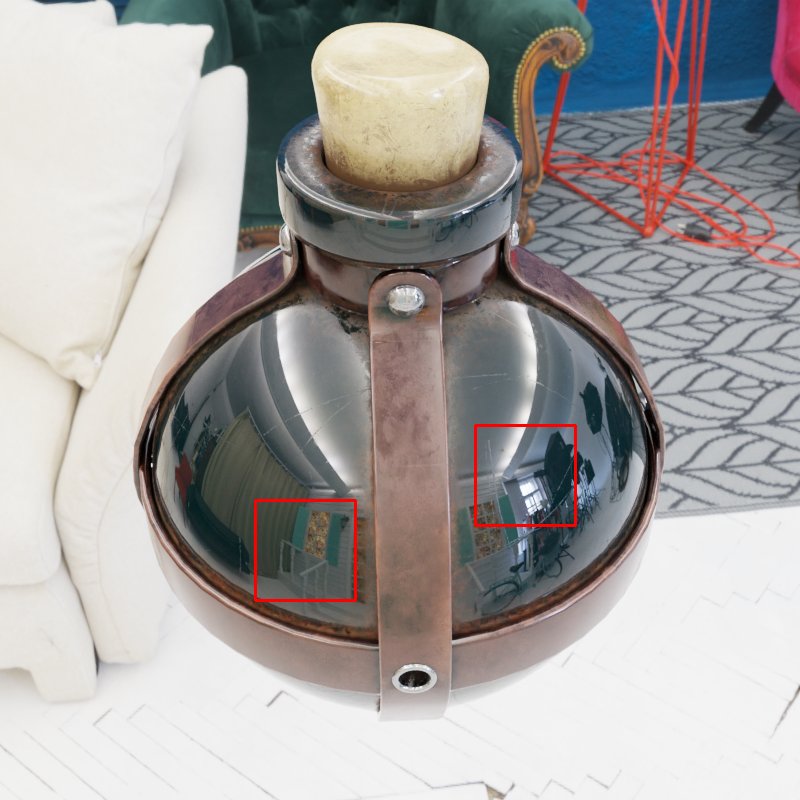} &
\includegraphics[trim={0 0 0 0}, clip, width=0.11\textwidth]{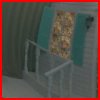} &
\includegraphics[trim={0 0 0 0}, clip, width=0.11\textwidth]{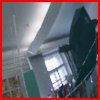} &
\includegraphics[trim={0 0 0 0}, clip, width=0.11\textwidth]{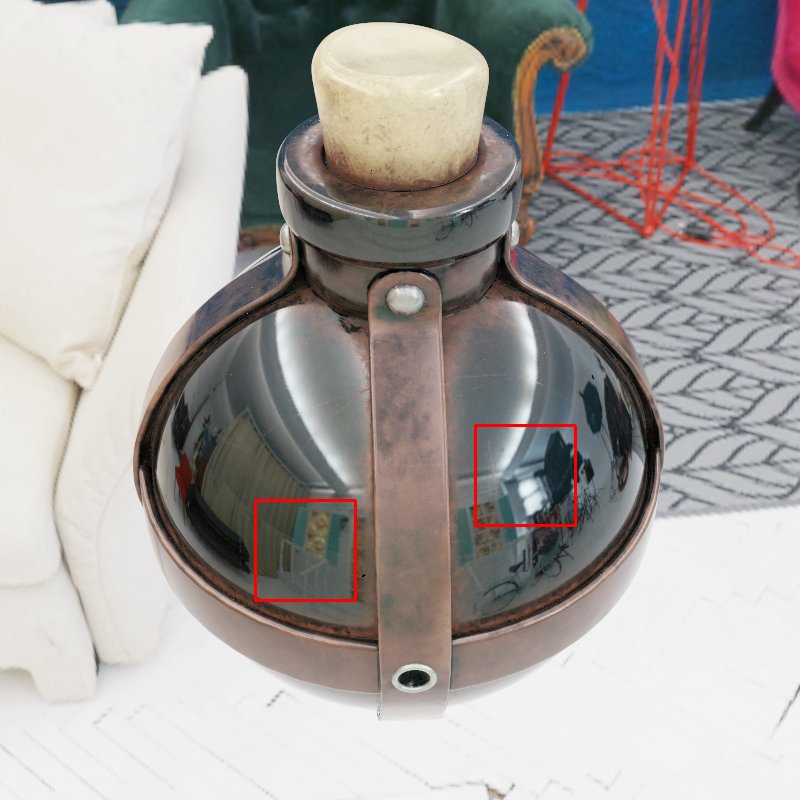} &
\includegraphics[trim={0 0 0 0}, clip, width=0.11\textwidth]{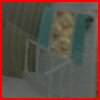} &
\includegraphics[trim={0 0 0 0}, clip, width=0.11\textwidth]{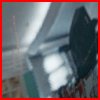} &
\includegraphics[trim={0 0 0 0}, clip, width=0.11\textwidth]{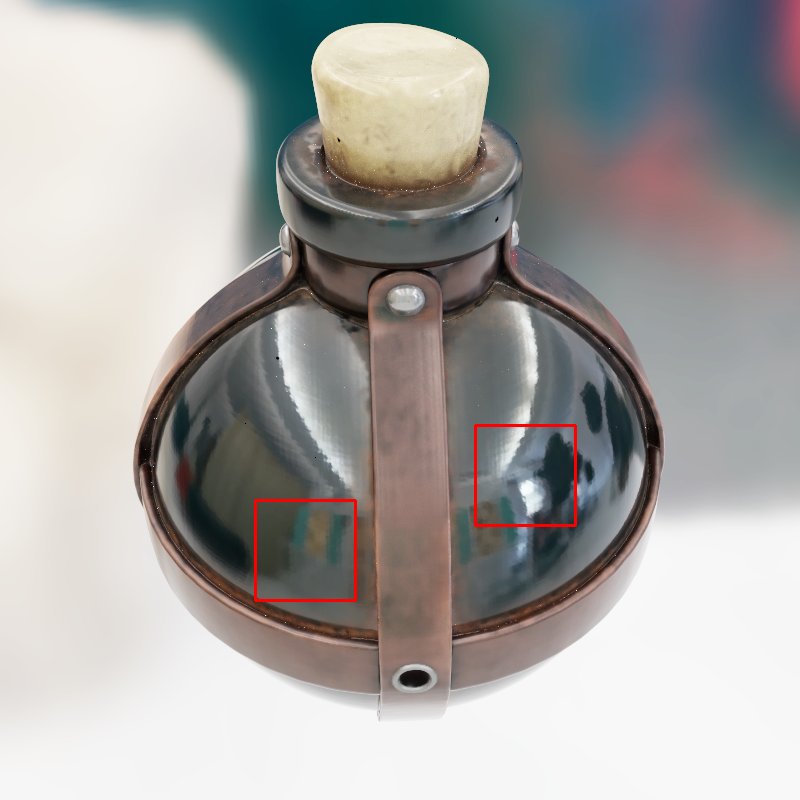} &
\includegraphics[trim={0 0 0 0}, clip, width=0.11\textwidth]{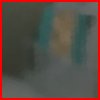} &
\includegraphics[trim={0 0 0 0}, clip, width=0.11\textwidth]{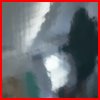} \\ 
\end{tabular}
}
\vspace{-1.0\baselineskip}
\caption{\textbf{Comparison against NeRO~\cite{liuNeRONeuralGeometry2023} on material \& lighting reconstruction.} We evaluate the quality of material and lighting reconstruction using NeRO's glossy synthetic dataset. In this experiment, we use the same geometry as NeRO and compare NeRO's stage II results with ours. Our \methodname{} demonstrates superior reconstruction quality on the appearance of glossy objects, capturing detailed highlights from the environment more effectively.}
\label{fig:nero_comparison1_results}
\end{figure*}

\begin{figure*}[htbp]
\centering
\resizebox{0.93\textwidth}{!}{%
\setlength{\tabcolsep}{1.0pt}
\renewcommand{\arraystretch}{0.5}
\begin{tabular}{cccccccccc}
& \multicolumn{3}{c}{Reference} & \multicolumn{3}{c}{Ours} & \multicolumn{3}{c}{NeRO~\cite{liuNeRONeuralGeometry2023} } \\ 
\rotatebox{90}{\parbox{0.11\textwidth}{\centering \textsc{Bell} }} &
\includegraphics[trim={50 0 50 100}, clip, width=0.11\textwidth]{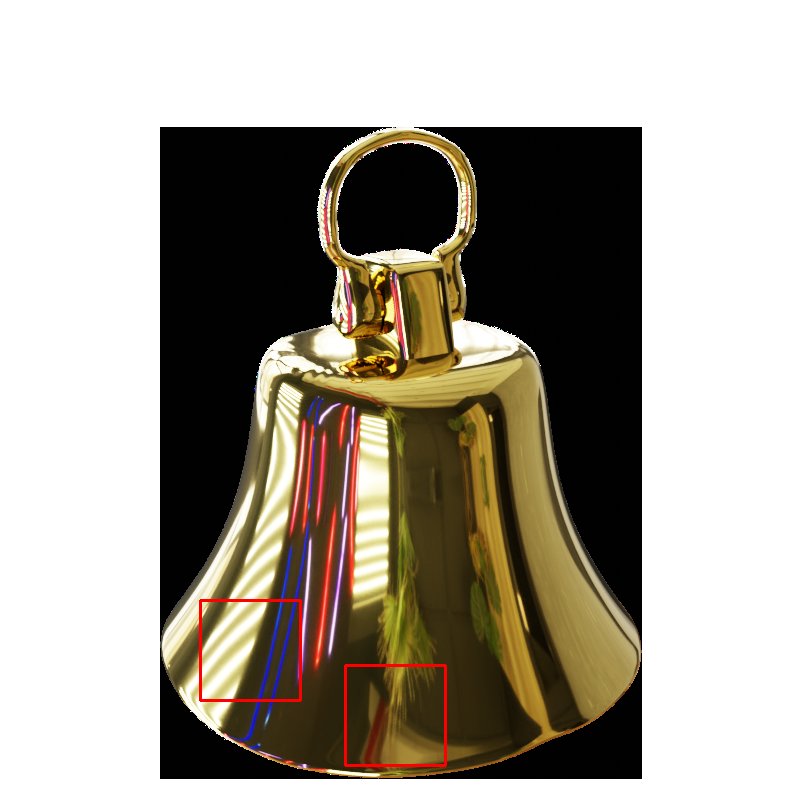} &
\includegraphics[trim={0 0 0 0}, clip, width=0.11\textwidth]{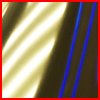} &
\includegraphics[trim={0 0 0 0}, clip, width=0.11\textwidth]{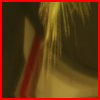} &
\includegraphics[trim={50 0 50 100}, clip, width=0.11\textwidth]{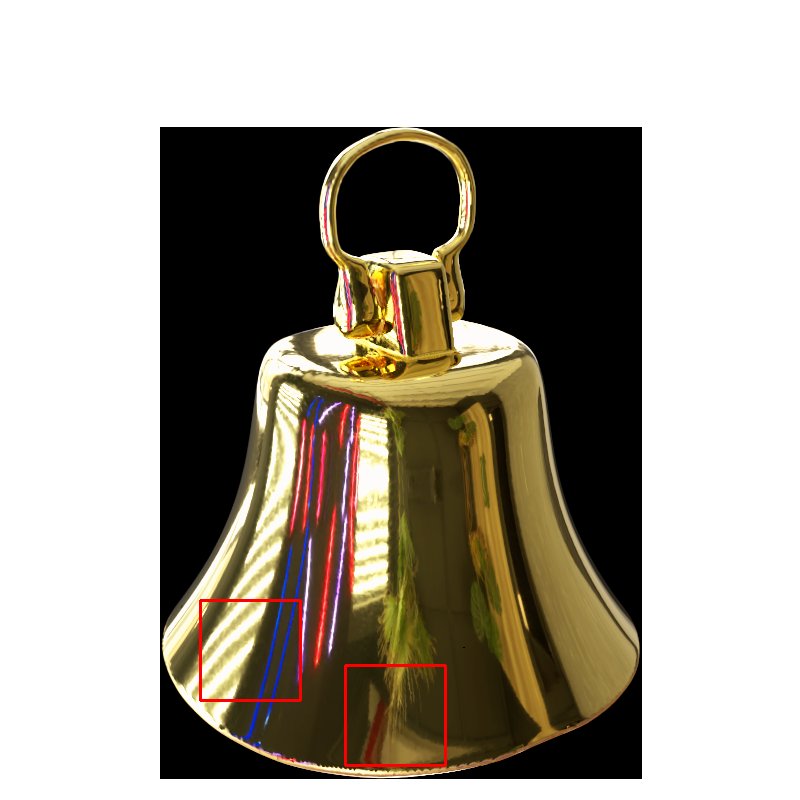} &
\includegraphics[trim={0 0 0 0}, clip, width=0.11\textwidth]{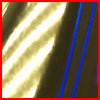} &
\includegraphics[trim={0 0 0 0}, clip, width=0.11\textwidth]{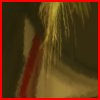} &
\includegraphics[trim={50 0 50 100}, clip, width=0.11\textwidth]{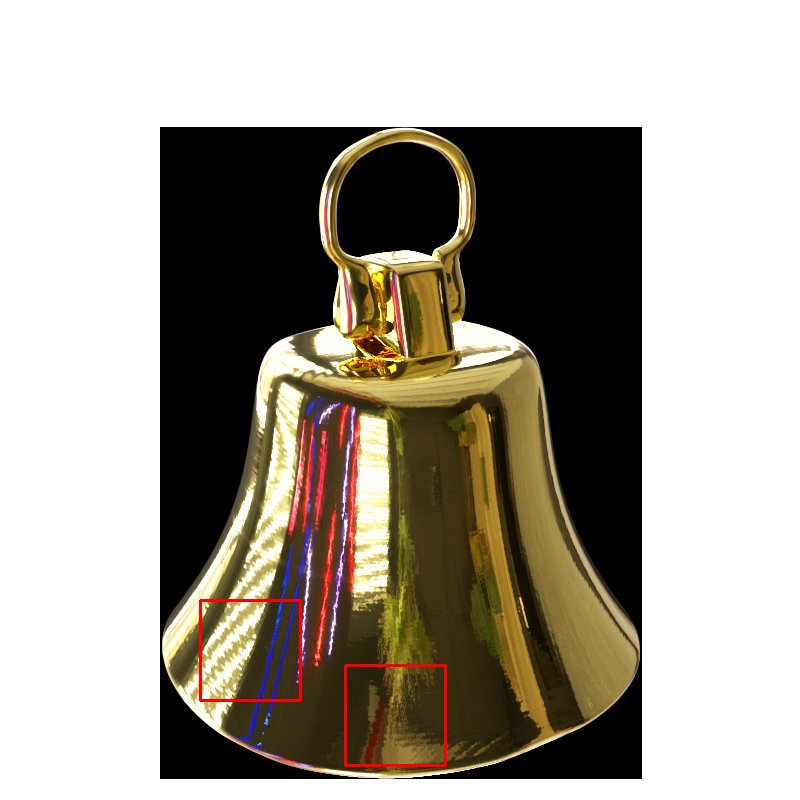} &
\includegraphics[trim={0 0 0 0}, clip, width=0.11\textwidth]{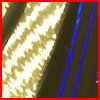} &
\includegraphics[trim={0 0 0 0}, clip, width=0.11\textwidth]{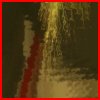} \\ 
\rotatebox{90}{\parbox{0.11\textwidth}{\centering \textsc{Cat} }} &
\includegraphics[trim={50 0 50 100}, clip, width=0.11\textwidth]{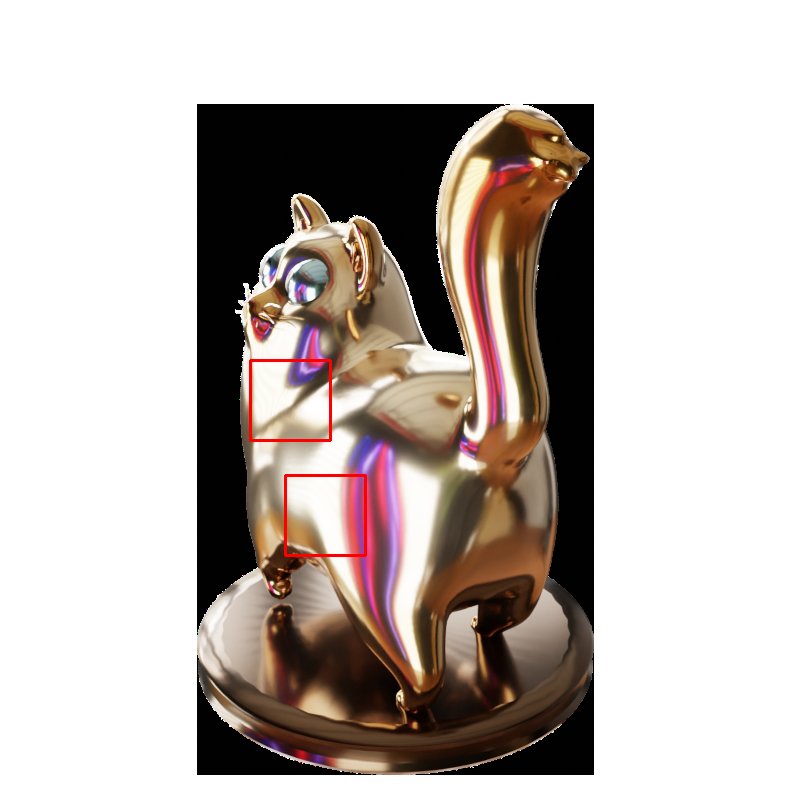} &
\includegraphics[trim={0 0 0 0}, clip, width=0.11\textwidth]{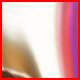} &
\includegraphics[trim={0 0 0 0}, clip, width=0.11\textwidth]{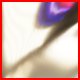} &
\includegraphics[trim={50 0 50 100}, clip, width=0.11\textwidth]{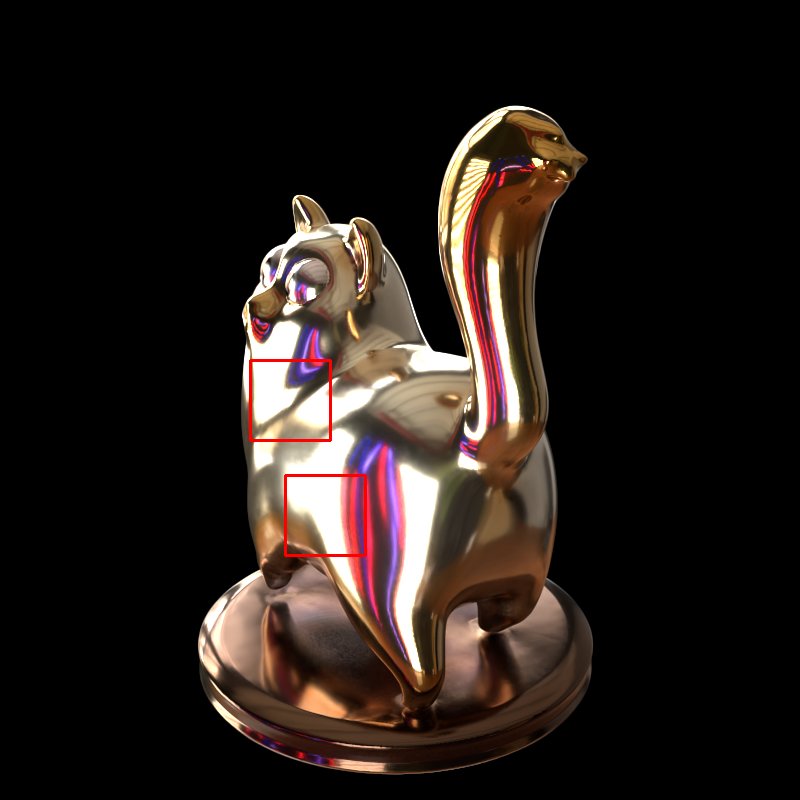} &
\includegraphics[trim={0 0 0 0}, clip, width=0.11\textwidth]{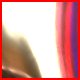} &
\includegraphics[trim={0 0 0 0}, clip, width=0.11\textwidth]{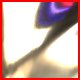} &
\includegraphics[trim={50 0 50 100}, clip, width=0.11\textwidth]{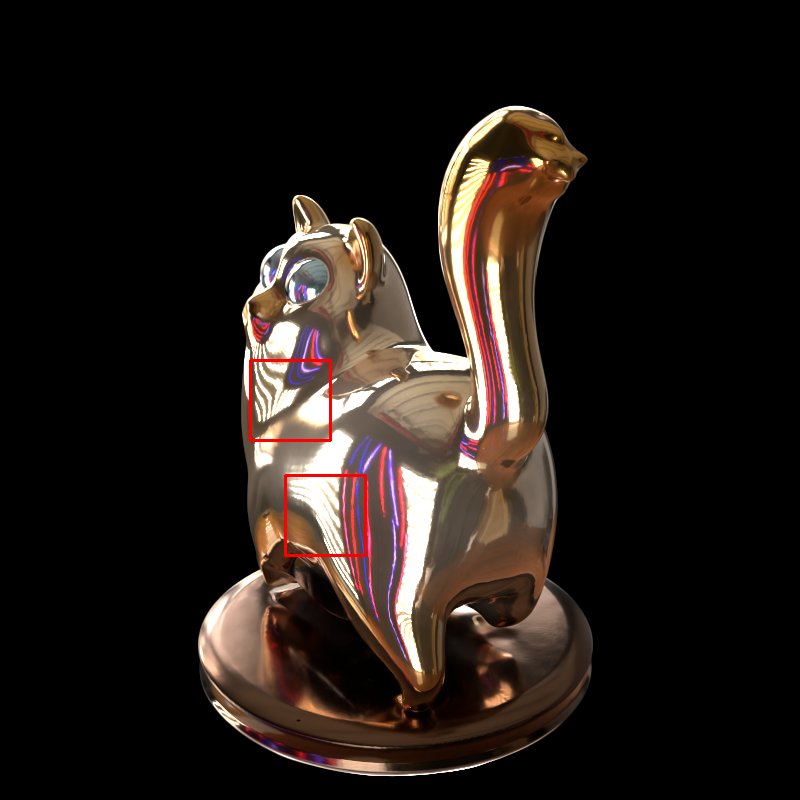} &
\includegraphics[trim={0 0 0 0}, clip, width=0.11\textwidth]{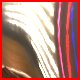} &
\includegraphics[trim={0 0 0 0}, clip, width=0.11\textwidth]{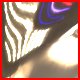} \\ 
\rotatebox{90}{\parbox{0.11\textwidth}{\centering \textsc{Teapot} }} &
\includegraphics[trim={0 0 0 0}, clip, width=0.11\textwidth]{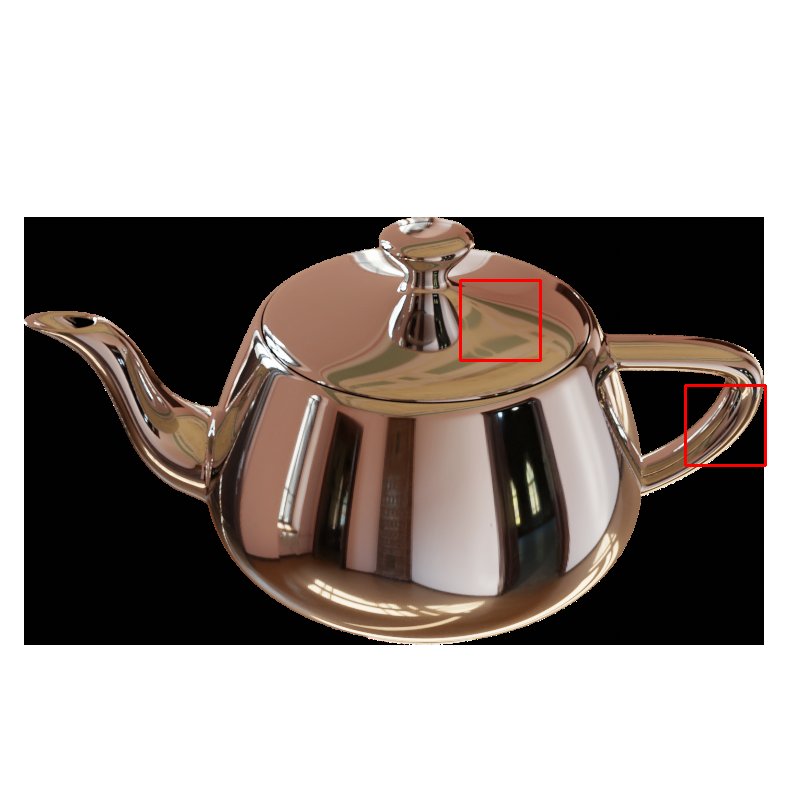} &
\includegraphics[trim={0 0 0 0}, clip, width=0.11\textwidth]{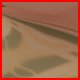} &
\includegraphics[trim={0 0 0 0}, clip, width=0.11\textwidth]{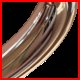} &
\includegraphics[trim={0 0 0 0}, clip, width=0.11\textwidth]{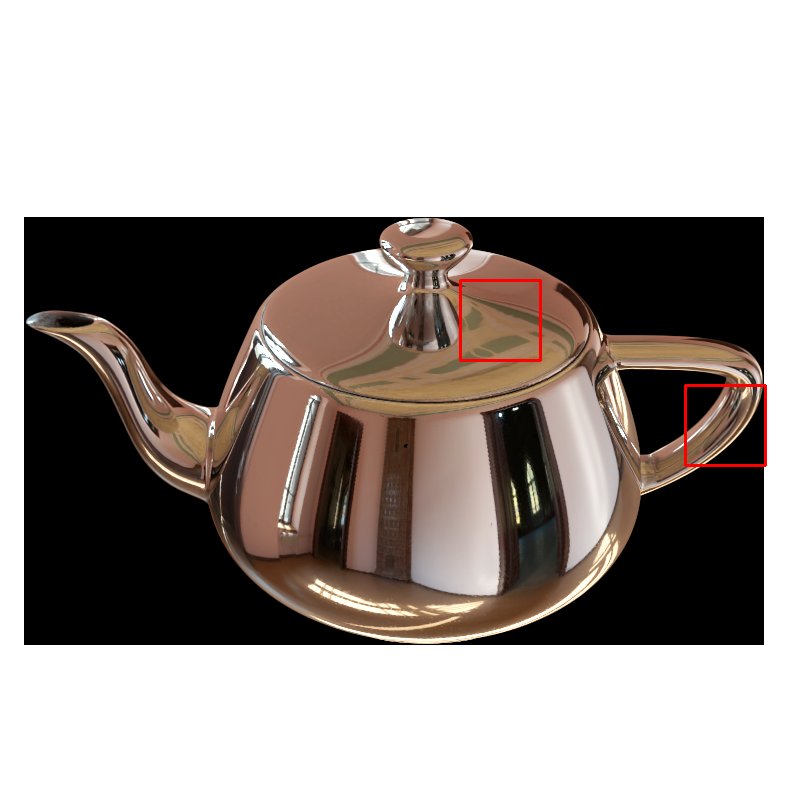} &
\includegraphics[trim={0 0 0 0}, clip, width=0.11\textwidth]{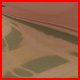} &
\includegraphics[trim={0 0 0 0}, clip, width=0.11\textwidth]{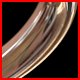} &
\includegraphics[trim={0 0 0 0}, clip, width=0.11\textwidth]{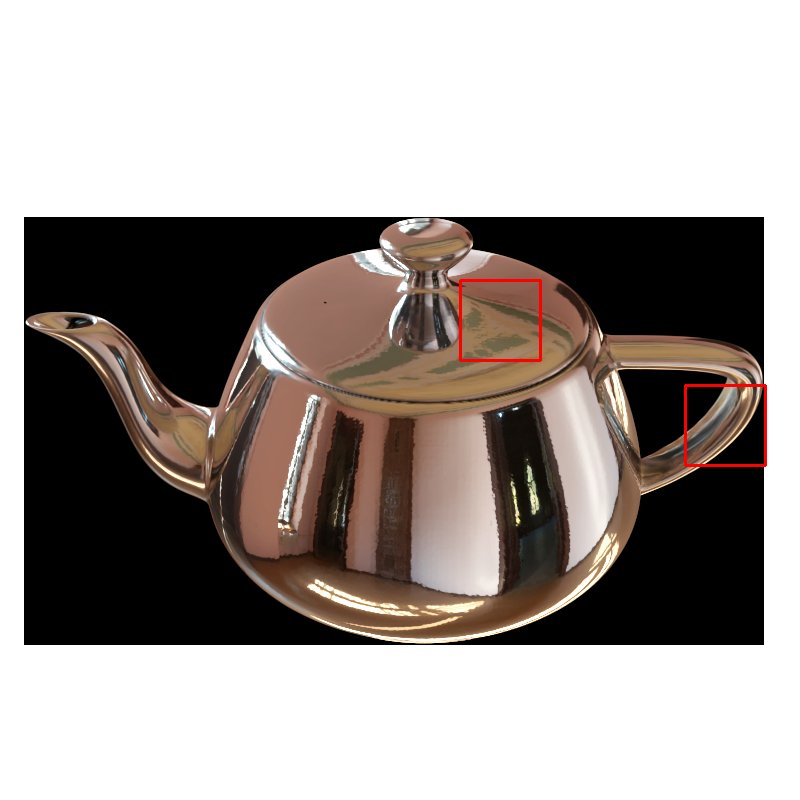} &
\includegraphics[trim={0 0 0 0}, clip, width=0.11\textwidth]{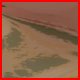} &
\includegraphics[trim={0 0 0 0}, clip, width=0.11\textwidth]{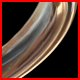} \\ 
\end{tabular}
}
\vspace{-1.0\baselineskip}
\caption{\textbf{Relighting quality comparison with NeRO~\cite{liuNeRONeuralGeometry2023}.} Similar to the above figure, but here we render under a novel view and lighting. Using the same geometry as NeRO, we compare the stage II results of NeRO with those of our \methodname{}. Oure relighting captures highlights on glossy objects in novel environments more accurately.}
\label{fig:nero_comparison2_results}
\end{figure*}

\begin{figure*}[htbp]
\centering
\resizebox{0.9\textwidth}{!}{%
\setlength{\tabcolsep}{1.0pt}
\renewcommand{\arraystretch}{0.5}
\begin{tabular}{cccccc}
Reference & Ours & NeRO & Reference & Ours & NeRO \\
\includegraphics[trim={70 73 70 67}, clip, width=0.166\textwidth]{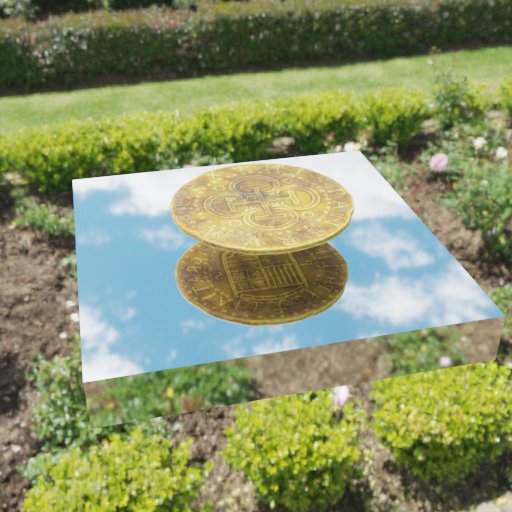} &
\includegraphics[trim={70 73 70 67}, clip, width=0.166\textwidth]{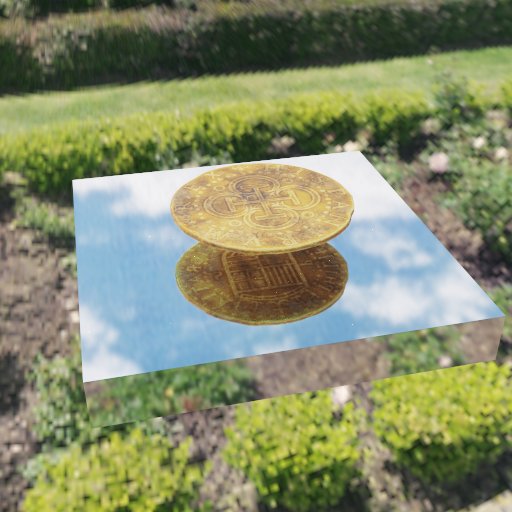} &
\includegraphics[trim={70 73 70 67}, clip, width=0.166\textwidth]{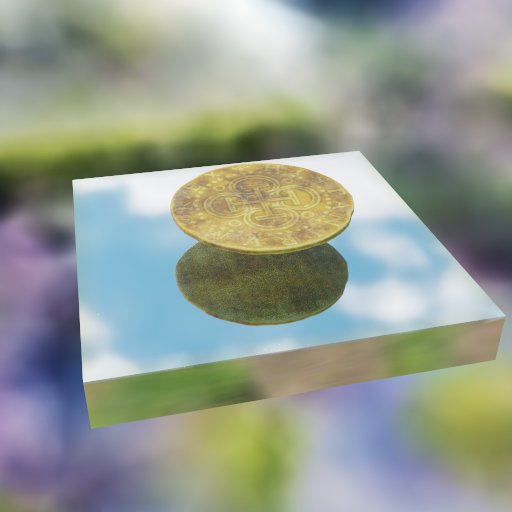} &
\includegraphics[trim={70 73 70 67}, clip, width=0.166\textwidth]{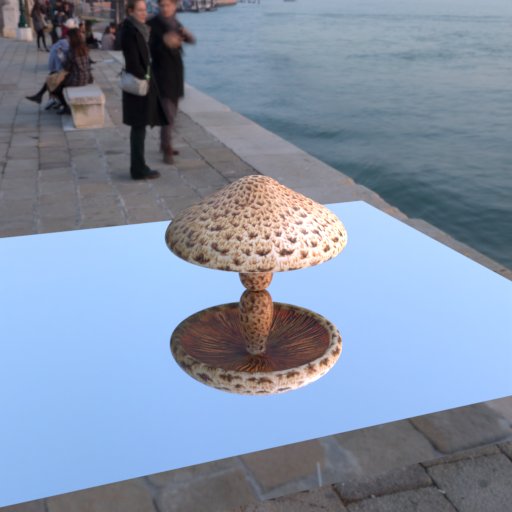} &
\includegraphics[trim={70 73 70 67}, clip, width=0.166\textwidth]{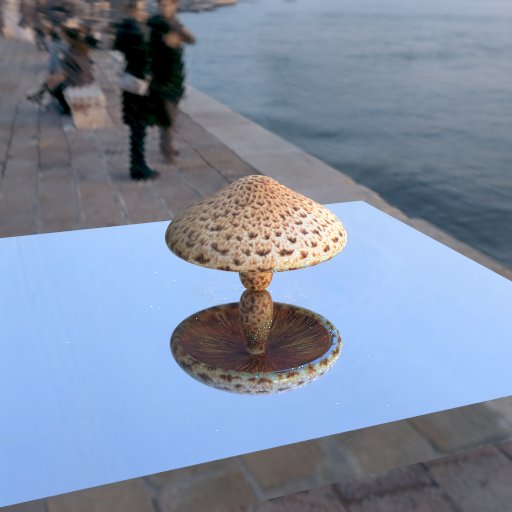} &
\includegraphics[trim={70 73 70 67}, clip, width=0.166\textwidth]{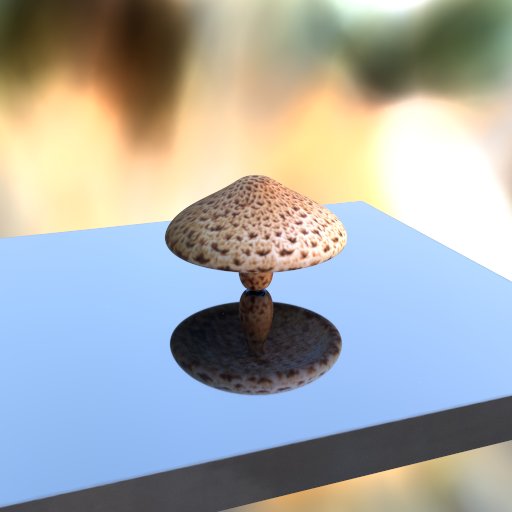} \\
\multicolumn{6}{c}{\rule{1.\textwidth}{0.4pt}} \\
\multicolumn{6}{c}{{\small \textbf{Recovered albedo of the objects' bottoms through specular reflections.}}} \\
\includegraphics[trim={70 73 70 67}, clip, width=0.166\textwidth]{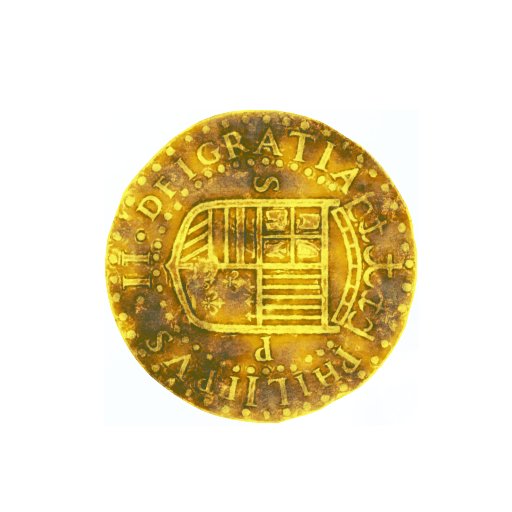} &
\includegraphics[trim={70 73 70 67}, clip, width=0.166\textwidth]{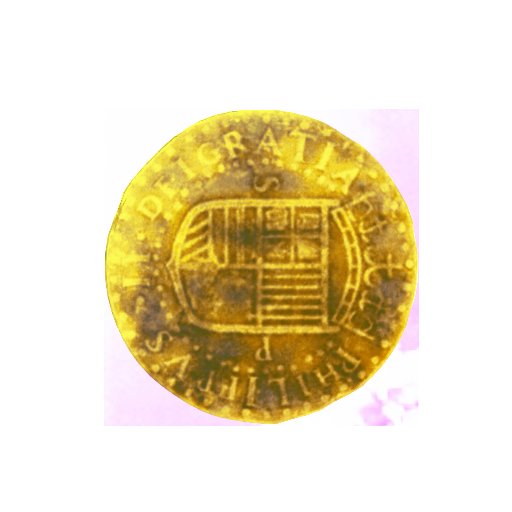} &
\includegraphics[trim={70 73 70 67}, clip, width=0.166\textwidth]{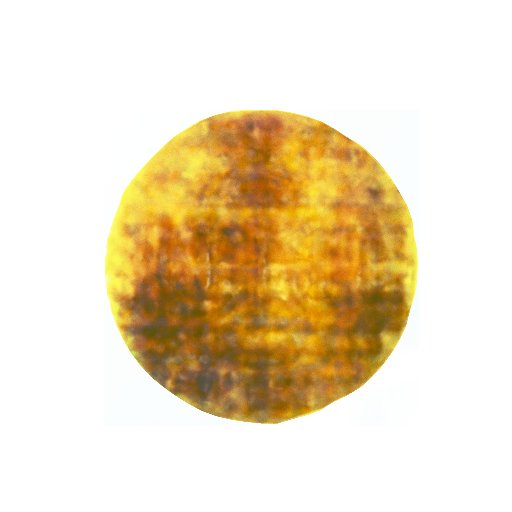} &
\includegraphics[trim={50 50 50 50}, clip, width=0.166\textwidth]{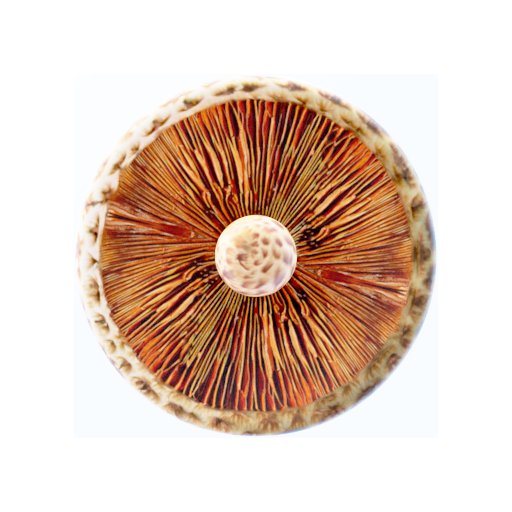} &
\includegraphics[trim={50 50 50 50}, clip, width=0.166\textwidth]{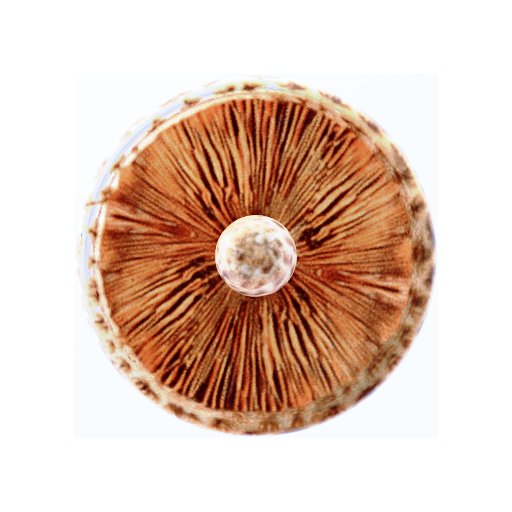} &
\includegraphics[trim={50 50 50 50}, clip, width=0.166\textwidth]{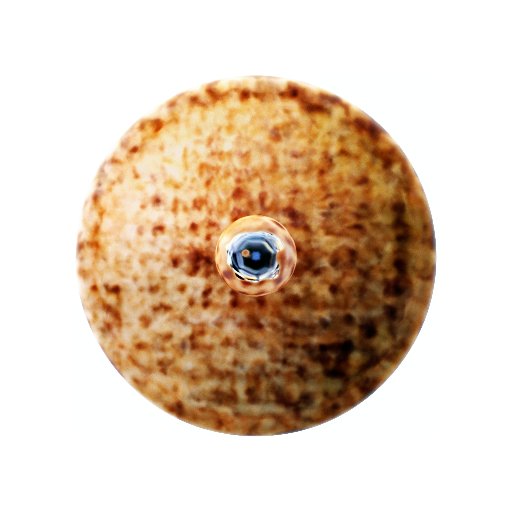} \\
Reference & Ours & NeRO & Reference & Ours & NeRO \\
\end{tabular}
}
\vspace{-1.0\baselineskip}
\caption{\textbf{Comparison against NeRO~\cite{liuNeRONeuralGeometry2023} on glossy interreflections.} In these two scenes,  an object (a coin or a mushroom) is placed above a specular table, where the input views can observe the bottom of the object through specular reflections on the table. Our pipeline successfully recovers the texture details on the object, while NeRO fails to and produces blurry reconstructions.}
\label{fig:nero_comparison3_results}
\end{figure*}

We conducted extensive evaluations comparing our method with NeRO~\cite{liuNeRONeuralGeometry2023}, focusing on the recovery of texture details and relighting quality in scenes with glossy interreflections. Figs.~\ref{fig:nero_comparison1_results}, \ref{fig:nero_comparison2_results}, and \ref{fig:nero_comparison3_results} illustrate these comparisons in various settings.
Fig.~\ref{fig:nero_comparison3_results} demonstrates our ability to reconstruct detailed material using interreflection, thanks to the correct simulation of global illumination. NeRO fails at this task because it predicts indirect lighting using a neural network. Fig.~\ref{fig:nero_comparison2_results} highlights the relighting quality on NeRO's glossy synthetic objects \textsc{bell}, \textsc{cat}, and \textsc{teapot}. Using the same geometry as NeRO, our method captures detailed highlights and reflections more effectively, demonstrating superior relighting performance.
Lastly, Fig.~\ref{fig:nero_comparison1_results}, we present a detailed comparison of texture recovery through specular reflections. Two scenes involving a coin and a mushroom placed above a specular table demonstrate that our method successfully reconstructs fine texture details observed through specular reflections, whereas NeRO produces blurrier reconstructions.

\begin{figure*}[htbp]
\centering
\resizebox{0.93\textwidth}{!}{%
\setlength{\tabcolsep}{1.0pt}
\renewcommand{\arraystretch}{0.5}
\begin{tabular}{cccccccccc}
& \multicolumn{3}{c}{Reference} & \multicolumn{3}{c}{Ours} & \multicolumn{3}{c}{Neural-PBIR~\cite{sunNeuralPBIRReconstructionShape2023} } \\ 
\rotatebox{90}{\parbox{0.11\textwidth}{\centering \textsc{Spot} }} &
\includegraphics[trim={0 0 0 0}, clip, width=0.11\textwidth]{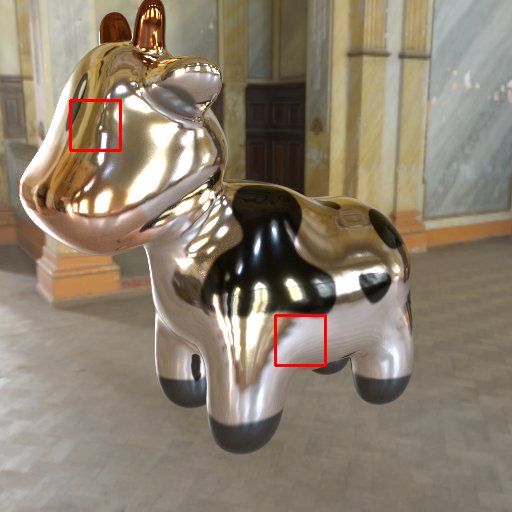} &
\includegraphics[trim={0 0 0 0}, clip, width=0.11\textwidth]{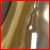} &
\includegraphics[trim={0 0 0 0}, clip, width=0.11\textwidth]{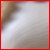} &
\includegraphics[trim={0 0 0 0}, clip, width=0.11\textwidth]{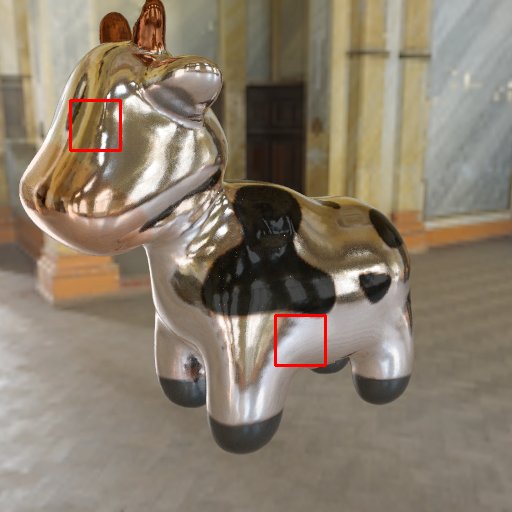} &
\includegraphics[trim={0 0 0 0}, clip, width=0.11\textwidth]{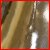} &
\includegraphics[trim={0 0 0 0}, clip, width=0.11\textwidth]{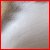} &
\includegraphics[trim={0 0 0 0}, clip, width=0.11\textwidth]{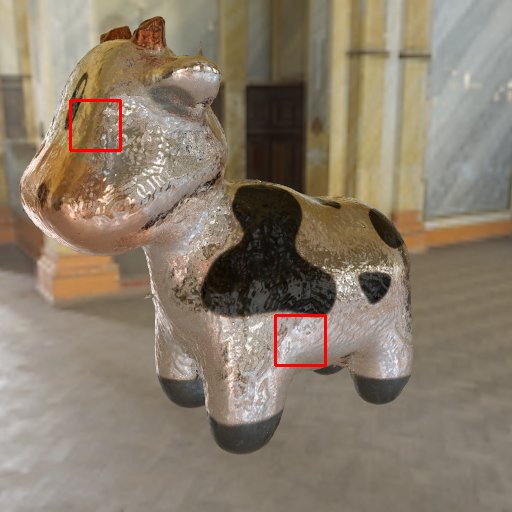} &
\includegraphics[trim={0 0 0 0}, clip, width=0.11\textwidth]{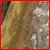} &
\includegraphics[trim={0 0 0 0}, clip, width=0.11\textwidth]{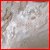} \\ 
\rotatebox{90}{\parbox{0.11\textwidth}{\centering \textsc{Knot} }} &
\includegraphics[trim={0 0 0 0}, clip, width=0.11\textwidth]{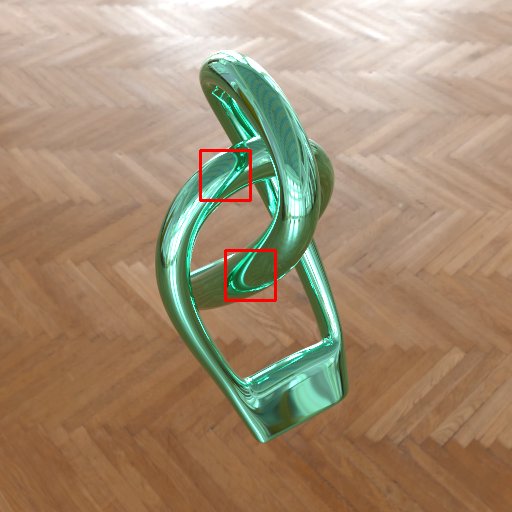} &
\includegraphics[trim={0 0 0 0}, clip, width=0.11\textwidth]{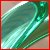} &
\includegraphics[trim={0 0 0 0}, clip, width=0.11\textwidth]{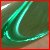} &
\includegraphics[trim={0 0 0 0}, clip, width=0.11\textwidth]{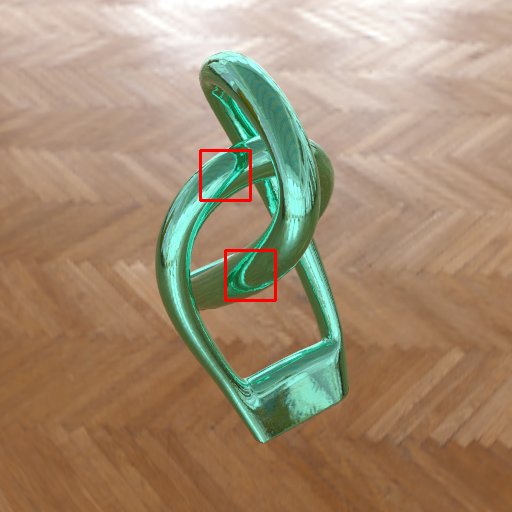} &
\includegraphics[trim={0 0 0 0}, clip, width=0.11\textwidth]{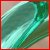} &
\includegraphics[trim={0 0 0 0}, clip, width=0.11\textwidth]{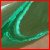} &
\includegraphics[trim={0 0 0 0}, clip, width=0.11\textwidth]{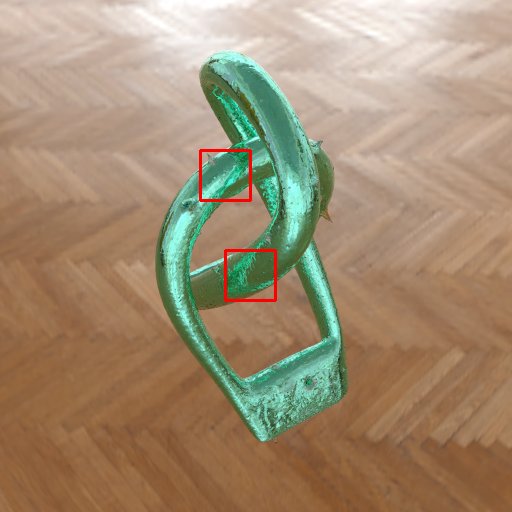} &
\includegraphics[trim={0 0 0 0}, clip, width=0.11\textwidth]{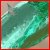} &
\includegraphics[trim={0 0 0 0}, clip, width=0.11\textwidth]{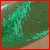} \\ 
\rotatebox{90}{\parbox{0.11\textwidth}{\centering \textsc{Cross} }} &
\includegraphics[trim={0 0 0 0}, clip, width=0.11\textwidth]{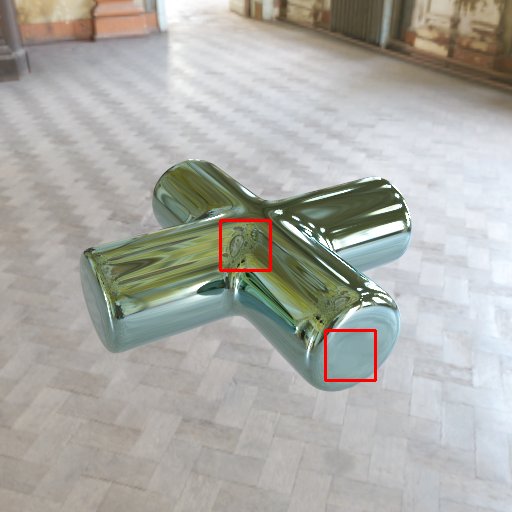} &
\includegraphics[trim={0 0 0 0}, clip, width=0.11\textwidth]{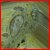} &
\includegraphics[trim={0 0 0 0}, clip, width=0.11\textwidth]{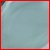} &
\includegraphics[trim={0 0 0 0}, clip, width=0.11\textwidth]{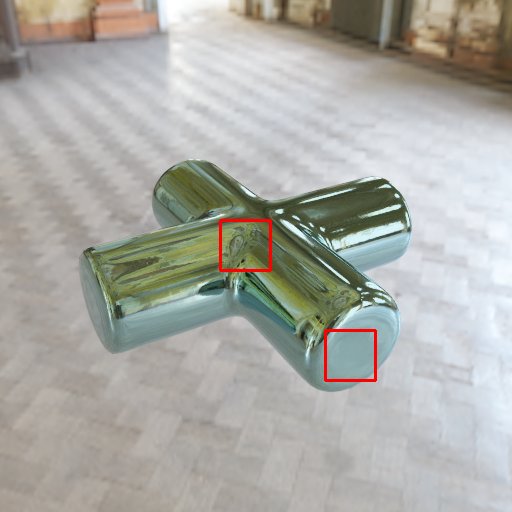} &
\includegraphics[trim={0 0 0 0}, clip, width=0.11\textwidth]{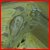} &
\includegraphics[trim={0 0 0 0}, clip, width=0.11\textwidth]{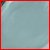} &
\includegraphics[trim={0 0 0 0}, clip, width=0.11\textwidth]{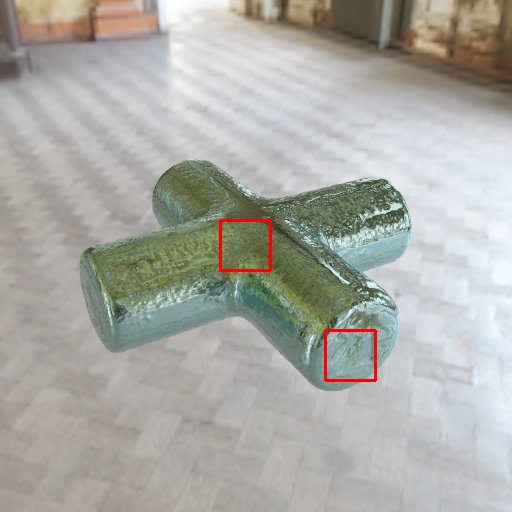} &
\includegraphics[trim={0 0 0 0}, clip, width=0.11\textwidth]{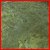} &
\includegraphics[trim={0 0 0 0}, clip, width=0.11\textwidth]{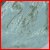} 
\end{tabular}
}
\vspace{-1.0\baselineskip}
\caption{\textbf{Comparison against Neural-PBIR~\cite{sunNeuralPBIRReconstructionShape2023} on glossy object reconstruction.} We evaluate the quality of glossy object reconstruction using Neural-PBIR and our method. Due to the fact that Neural-PBIR's initialization stage often fails for glossy objects, we specifically evaluated the mesh refinement stage. In this experiment, we start with the same initial geometry obtained from a visual hull and compare the physics-based inverse rendering (PBIR) mesh refinement. Our \methodname{} demonstrates superior reconstruction quality, effectively capturing detailed highlights from the environment. In contrast, Neural-PBIR struggles to reproduce glossy materials and geometric details.}
\label{fig:neu-pbir_comparison1_results}
\end{figure*}

\paragraph*{Comparison with Neural-PBIR~\cite{sunNeuralPBIRReconstructionShape2023}.}
We compared our method with Neural-PBIR~\cite{sunNeuralPBIRReconstructionShape2023}, focusing on the reconstruction quality of glossy objects. Due to the fact that Neural-PBIR's initialization stage often fails for glossy objects, we specifically evaluated the mesh refinement stage. Fig.~\ref{fig:neu-pbir_comparison1_results} shows the results for three different objects: \textsc{spot},  \textsc{knot}, and  \textsc{cross}. Starting from the same initial geometry obtained from a visual hull, we refined the meshes using our physics-based inverse rendering (PBIR) approach. Our method demonstrates superior reconstruction quality, effectively capturing detailed highlights and reflections from the environment. This is evident in the sharper and more detailed insets of our results compared to those of Neural-PBIR. In contrast, Neural-PBIR struggles to accurately reproduce the glossy materials and fine geometric details, resulting in less realistic reconstructions.

\begin{figure}[tb]
\centering 
\includegraphics[width=1\linewidth]{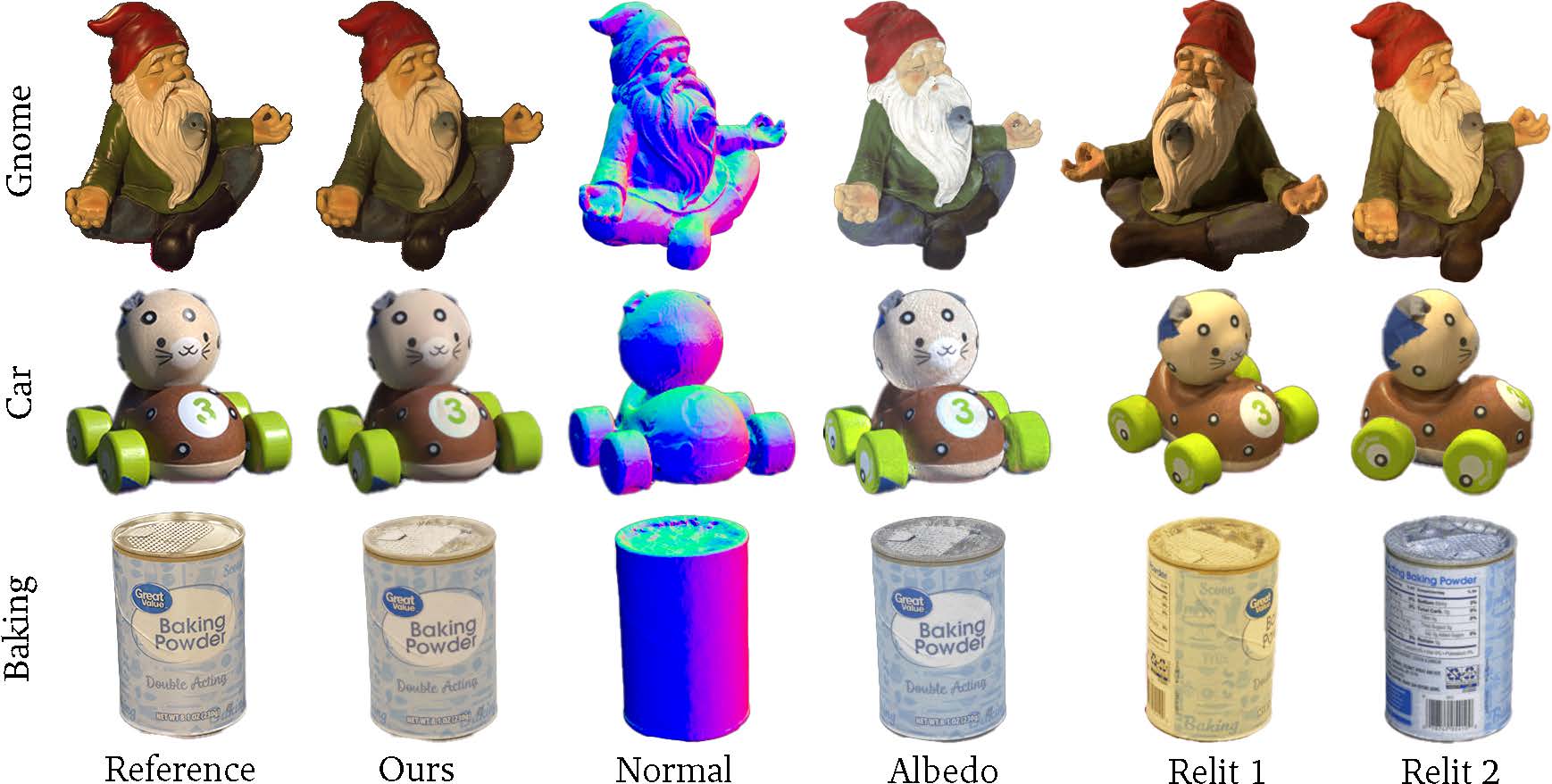}
\vspace{-1.0\baselineskip}
\caption{\textbf{Our results on Stanford-ORB~\cite{kuang2024stanford} dataset.} }
\label{fig:sorb_quality}
\end{figure}

\begin{table*}
\centering
\caption{
{\bf Geometry, relighting, and view-interpolation quality on Stanford-ORB dataset~\cite{kuang2024stanford}. }
}
{
\resizebox{0.8\textwidth}{!}{
\begin{tabular}{ @{}l@{\hskip 5pt} @{}c@{} @{}c@{\hskip 5pt} c@{\hskip 5pt} c@{} c@{\hskip 5pt} @{}c@{\hskip 5pt} @{}c@{\hskip 5pt} c@{\hskip 5pt} c@{} c@{\hskip 5pt} @{}c@{\hskip 5pt} @{}c@{\hskip 5pt} c@{\hskip 5pt} c@{} @{}c@{} }
    \toprule
    && \multicolumn{3}{c}{Geometry} && \multicolumn{4}{c}{Novel Scene Relighting} && \multicolumn{4}{c}{Novel View Synthesis} \\
    \cmidrule{3-5}
    \cmidrule{7-10}
    \cmidrule{12-15}
    Method && Depth$\downarrow$ & Normal$\downarrow$ & Shape$\downarrow$ && PSNR-H$\uparrow$ & PSNR-L$\uparrow$ & SSIM$\uparrow$ & LPIPS$\downarrow$ & & PSNR-H$\uparrow$ & PSNR-L$\uparrow$ & SSIM$\uparrow$ & LPIPS$\downarrow$ \\
    \midrule
    PhySG~\cite{zhang2021physg} &&
    1.90 & 0.17 & 9.28 && 21.81 & 28.11 & 0.960 & 0.055 && 24.24 & 32.15 & 0.974 & 0.047
    \\
    NVDiffRec~\cite{munkberg2022extracting} &&
    \underline{0.31} & 0.06 & 0.62 && 22.91 & 29.72 & 0.963 & 0.039 && 21.94 & 28.44 & 0.969 & 0.030
    \\
    NeRD~\cite{boss2021nerd} &&
    1.39 & 0.28 & 13.7 && 23.29 & 29.65 & 0.957 & 0.059 && 25.83 & 32.61 & 0.963 & 0.054
    \\
    NeRFactor~\cite{zhang2021nerfactor} &&
    0.87 & 0.29 & 9.53 && 23.54 & 30.38 & 0.969 & 0.048 && 26.06 & 33.47 & 0.973 & 0.046
    \\
    InvRender~\cite{zhang2022modeling} &&
    0.59 & 0.06 & \underline{0.44} && 23.76 & 30.83 & 0.970 & 0.046 && 25.91 & 34.01 & 0.977 & 0.042
    \\
    NVDiffRecMC~\cite{hasselgren2022shape} &&
    0.32 & \textbf{0.04} & 0.51 && \underline{24.43} & \underline{31.60} & 0.972 & 0.036 && \underline{28.03} & \underline{36.40} & 0.982 & 0.028
    \\
    Neural-PBIR~\cite{sunNeuralPBIRReconstructionShape2023} &&
    \textbf{0.30} & 0.06 & \textbf{0.43} && 26.01 & 33.26 & \textbf{0.979} & \textbf{0.023} && \textbf{28.83} & \textbf{36.80} & \textbf{0.986} & \textbf{0.019}
    \\

    PBIR-NIE (ours) &&
    0.50 & \underline{0.05} & 0.64 && \textbf{26.26} & \textbf{33.46} & \underline{0.977} & \underline{0.028} && 27.06 & 35.09 & \underline{0.983} & \underline{0.023}
    \\
    \bottomrule
\end{tabular}
}
}
\label{tab:stanfordorb}
\end{table*}

\begin{figure*}[htbp]
\centering
\resizebox{0.92\textwidth}{!}{%
\setlength{\tabcolsep}{1.0pt}
\renewcommand{\arraystretch}{0.5}
\begin{tabular}{ccccccc}
 & Rendering (view \# 1)  & Rendering (view \# 2) & Geometry (front \& back) & Rendering (view \# 1)  & Rendering (view \# 2) & Geometry \\
\rotatebox{90}{\parbox{0.166\textwidth}{\centering Reference}} &
\includegraphics[trim={0 0 0 0}, clip, height=0.166\textwidth]{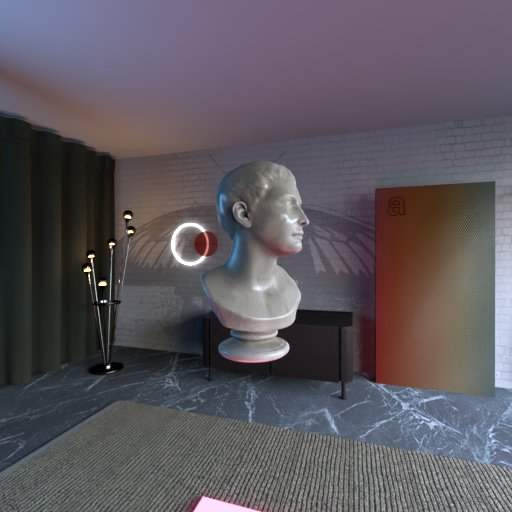} &
\includegraphics[trim={0 0 0 0}, clip, height=0.166\textwidth]{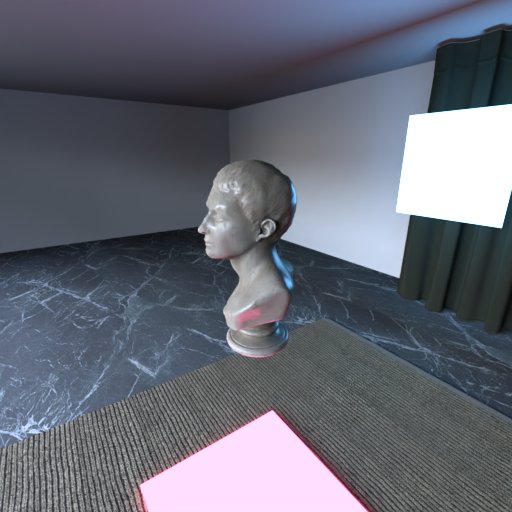} &
\includegraphics[trim={0 0 0 0}, clip, height=0.166\textwidth]{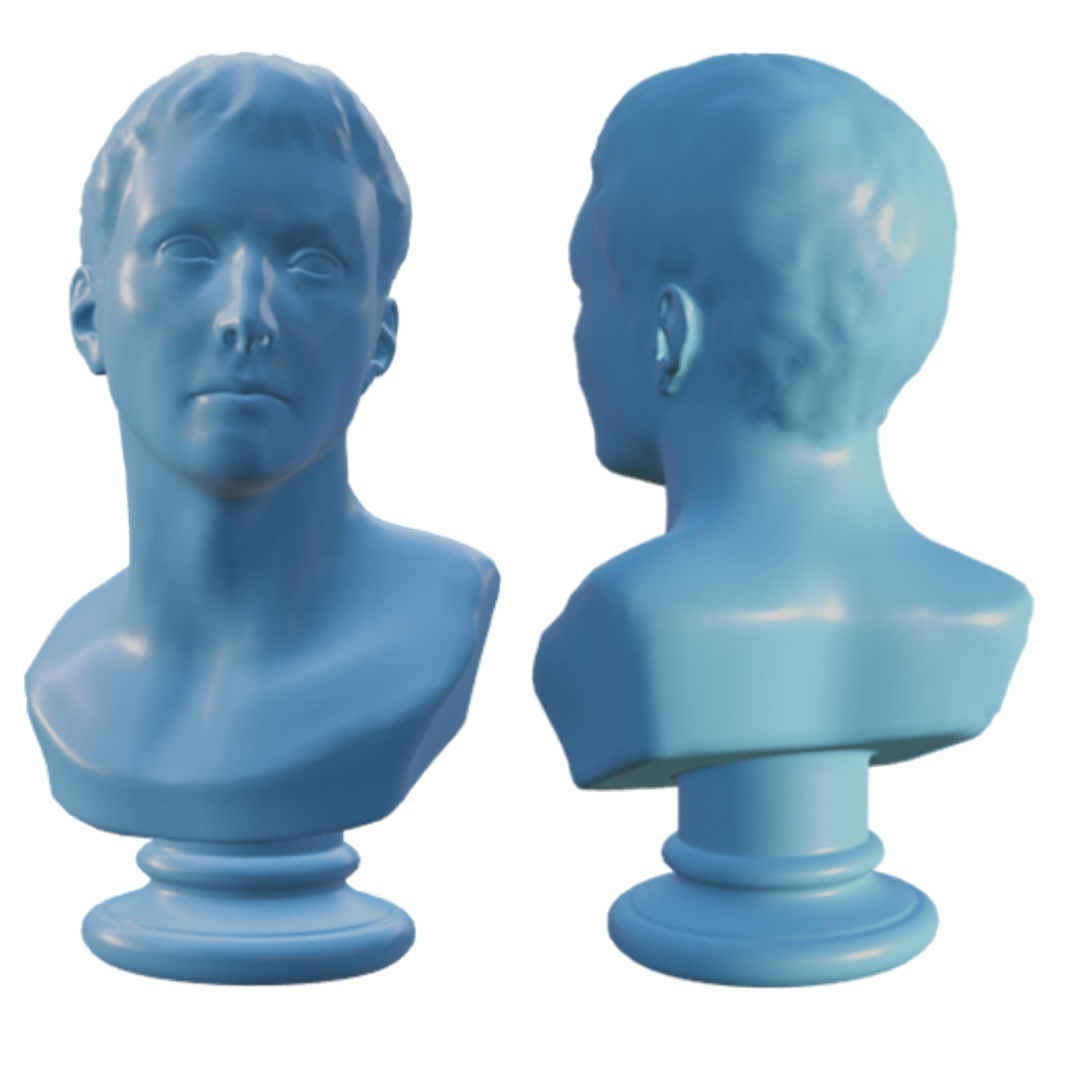} &
\includegraphics[trim={0 0 0 0}, clip, height=0.166\textwidth]{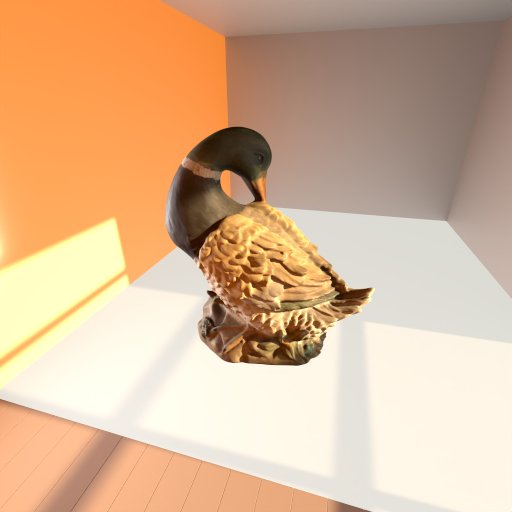} &
\includegraphics[trim={0 0 0 0}, clip, height=0.166\textwidth]{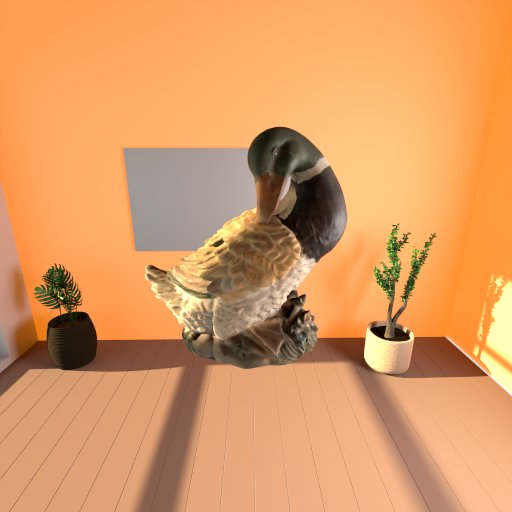} &
\includegraphics[trim={0 0 0 0}, clip, height=0.166\textwidth]{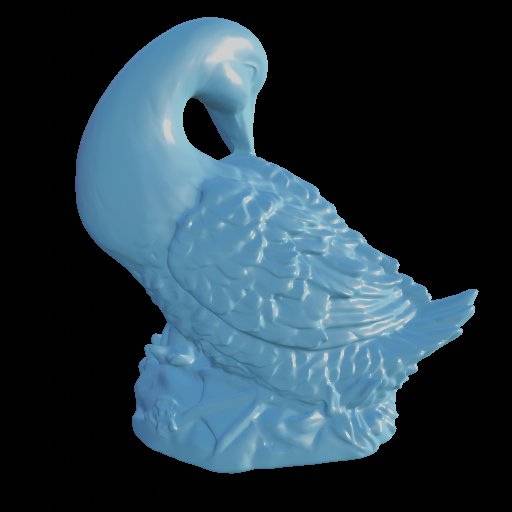} \\
& PSNR: \best{$\mathbf{27.24}$} / $31.43$ dB  & PSNR: \best{$\mathbf{26.72}$} / $31.86$ dB & Chamfer: \best{$\mathbf{9.26 \times 10^{-5}}$} & PSNR: \best{$\mathbf{26.99}$} / $31.95$ dB & PSNR: \secondbest{$\mathbf{26.57}$} / $31.72$ dB & Chamfer: $3.08 \times 10^{-4}$ \\
\rotatebox{90}{\parbox{0.166\textwidth}{\centering NeRF Emitter}} &
\includegraphics[trim={0 0 0 0}, clip, height=0.166\textwidth]{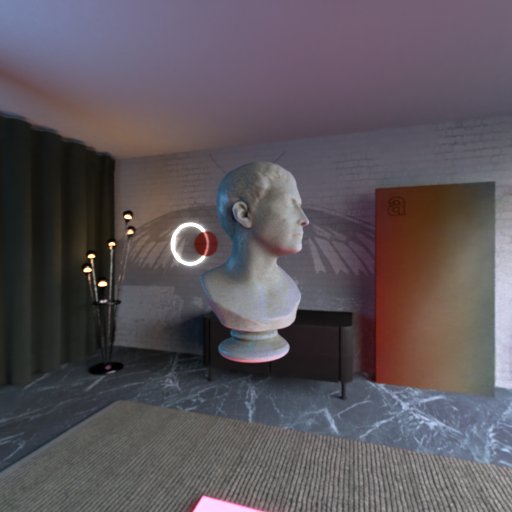} &
\includegraphics[trim={0 0 0 0}, clip, height=0.166\textwidth]{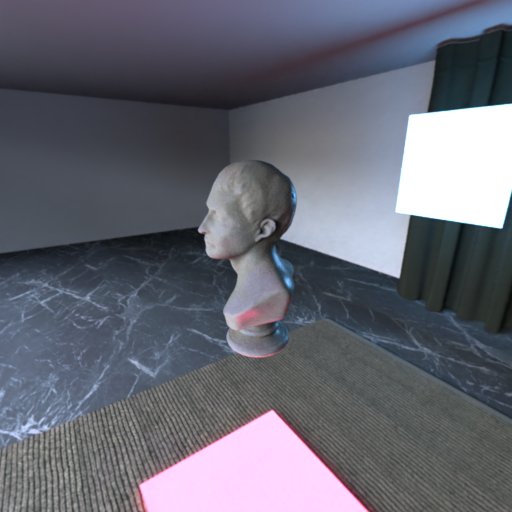} &
\includegraphics[trim={0 0 0 0}, clip, height=0.166\textwidth]{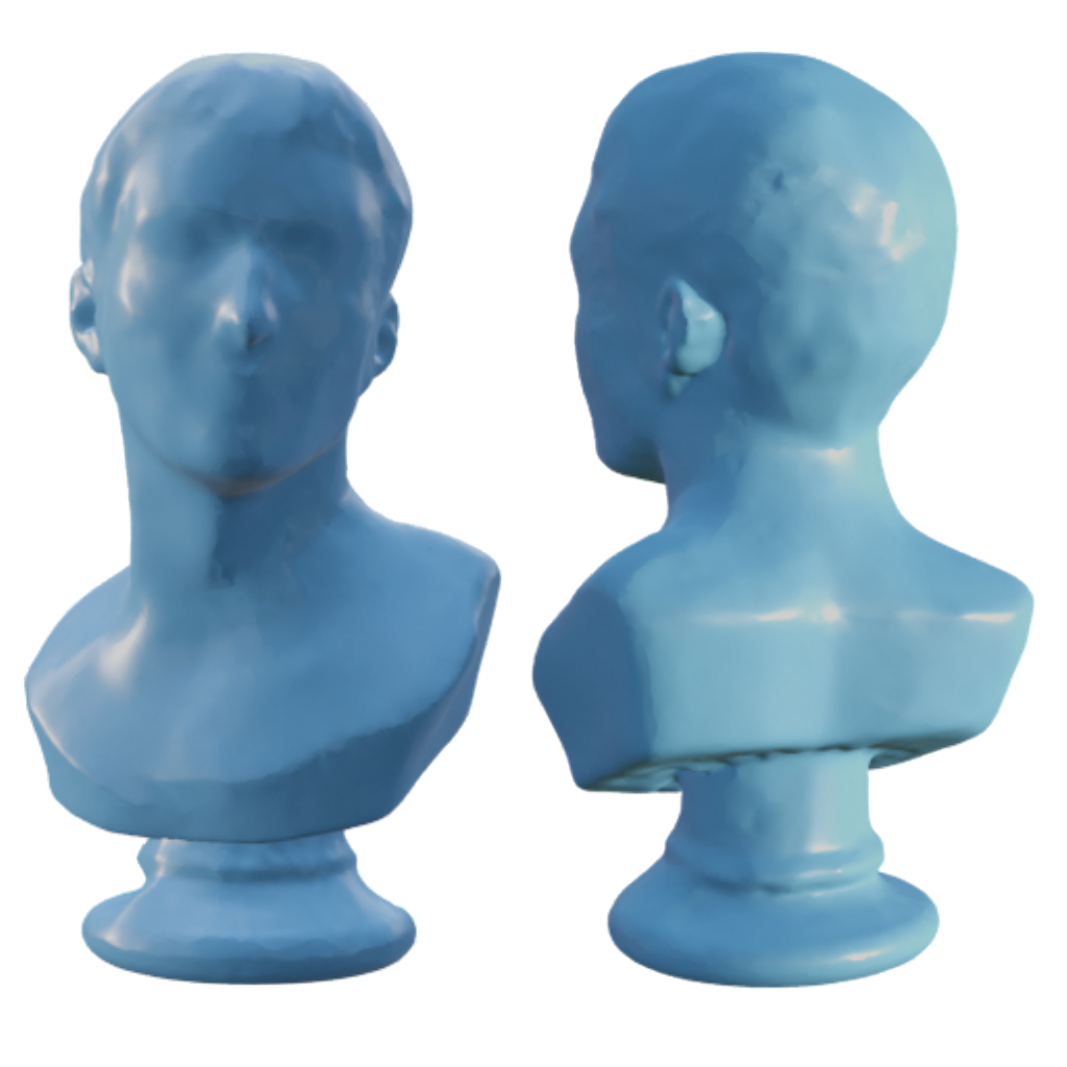} &
\includegraphics[trim={0 0 0 0}, clip, height=0.166\textwidth]{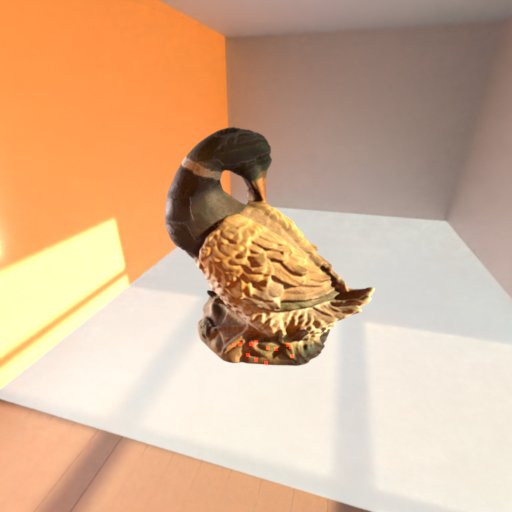} &
\includegraphics[trim={0 0 0 0}, clip, height=0.166\textwidth]{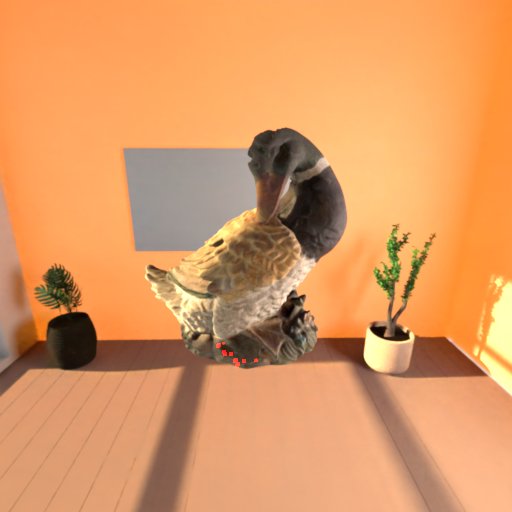} &
\includegraphics[trim={0 0 0 0}, clip, height=0.166\textwidth]{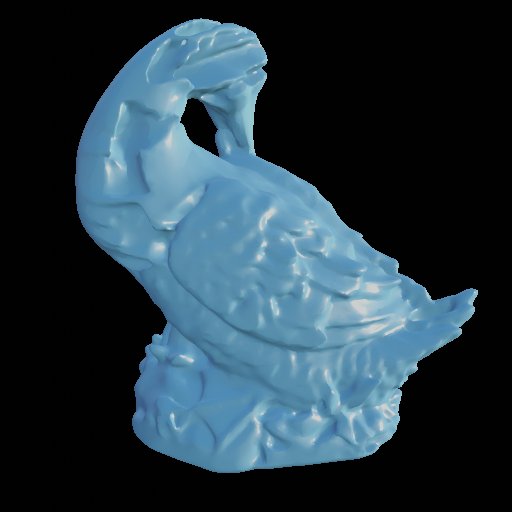} \\
& PSNR: \secondbest{$\mathbf{25.75}$} / $24.32$ dB  & PSNR: \secondbest{$\mathbf{25.14}$} / $19.47$ dB & Chamfer: \secondbest{$\mathbf{2.81 \times 10^{-4}}$} & PSNR: \secondbest{$\mathbf{26.35}$} / $24.92$ dB & PSNR: \best{$\mathbf{26.96}$} / $25.69$ dB & Chamfer: \best{$\mathbf{5.49 \times 10^{-5}}$} \\
\rotatebox{90}{\parbox{0.166\textwidth}{\centering \envmapname{} (\textbf{ours})}} &
\includegraphics[trim={0 0 0 0}, clip, height=0.166\textwidth]{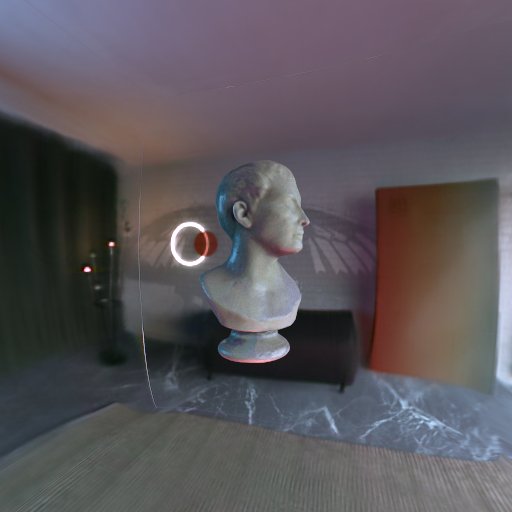} &
\includegraphics[trim={0 0 0 0}, clip, height=0.166\textwidth]{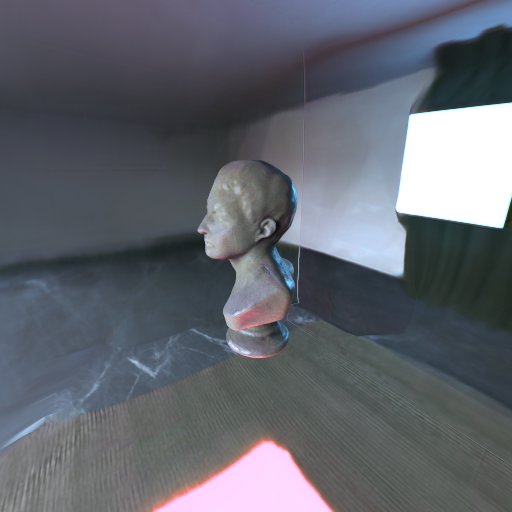} &
\includegraphics[trim={0 0 0 0}, clip, height=0.166\textwidth]{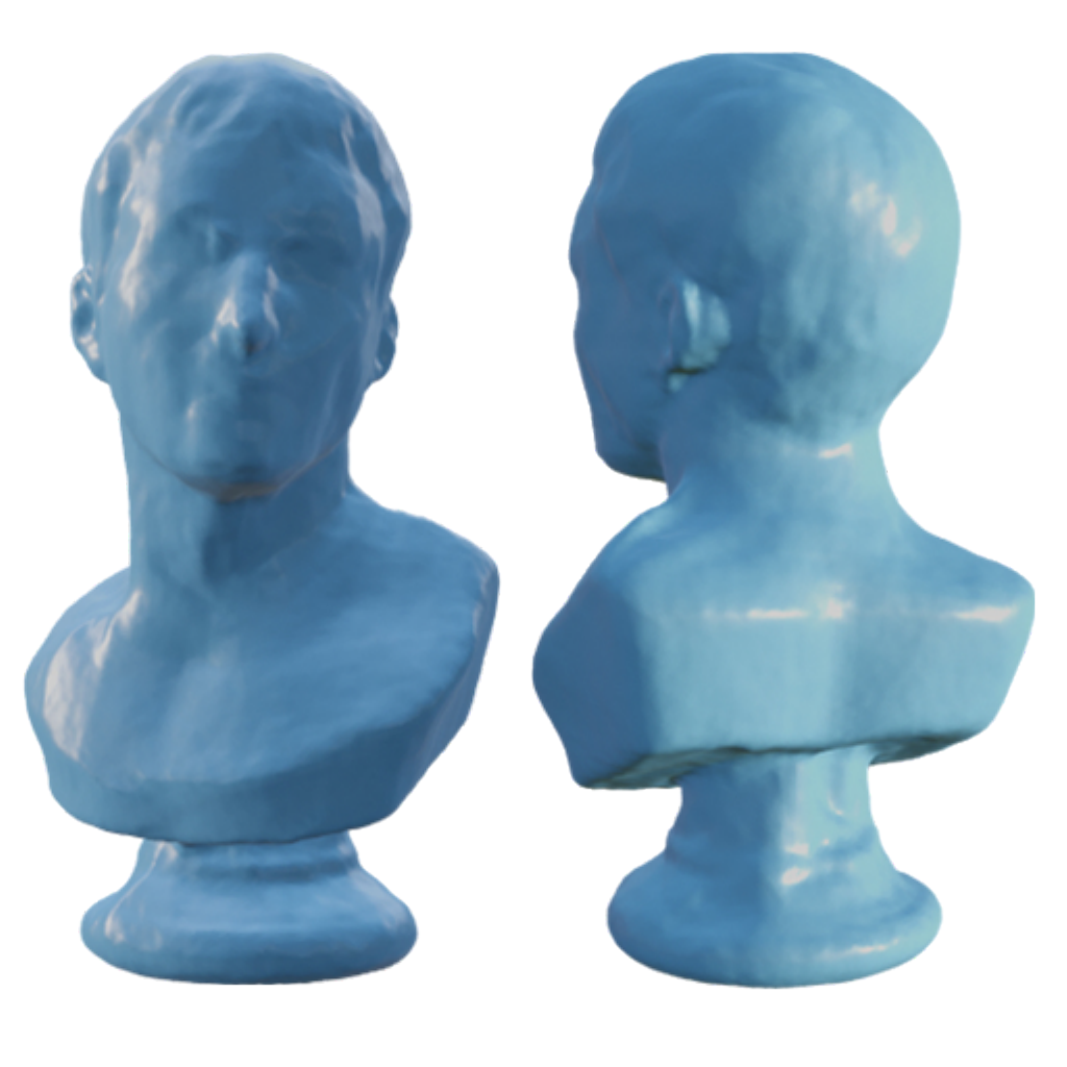} &
\includegraphics[trim={0 0 0 0}, clip, height=0.166\textwidth]{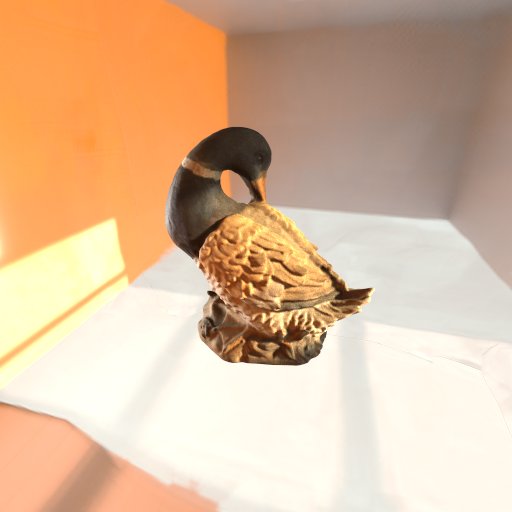} &
\includegraphics[trim={0 0 0 0}, clip, height=0.166\textwidth]{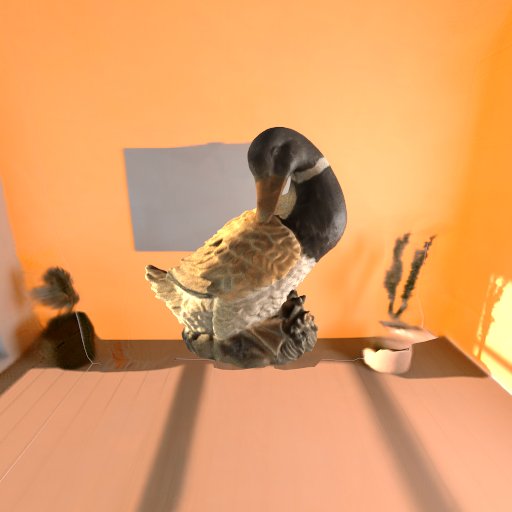} &
\includegraphics[trim={0 0 0 0}, clip, height=0.166\textwidth]{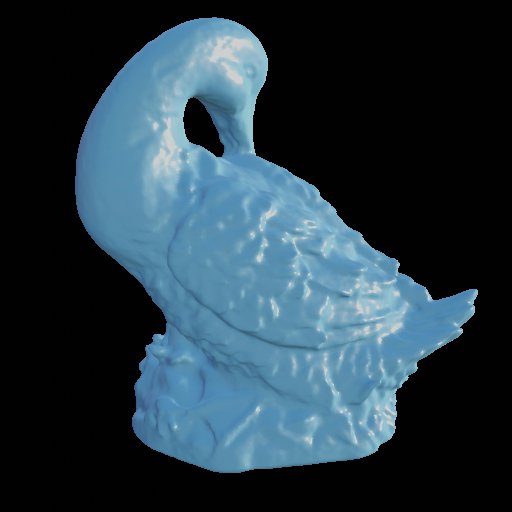} \\
& PSNR: $24.48$ / $16.10$ dB & PSNR: $24.24$ / $13.86$ dB & Chamfer: $4.41 \times 10^{-4}$ & PSNR: $25.05$ / $20.22$ dB & PSNR: $25.31$ / $15.76$ dB & Chamfer: \secondbest{$\mathbf{9.73 \times 10^{-5}}$}  \\
\rotatebox{90}{\parbox{0.166\textwidth}{\centering Envmap}} &
\includegraphics[trim={0 0 0 0}, clip, height=0.166\textwidth]{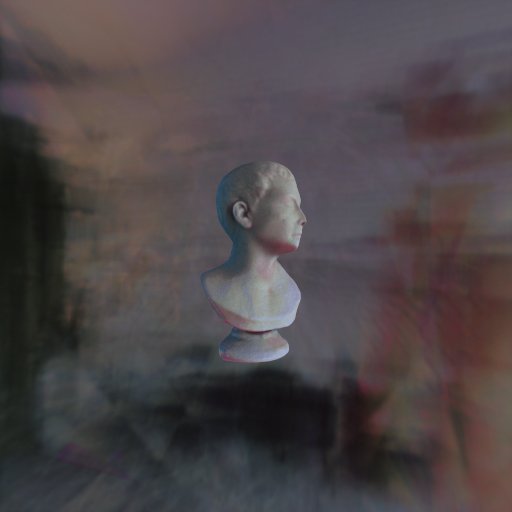} &
\includegraphics[trim={0 0 0 0}, clip, height=0.166\textwidth]{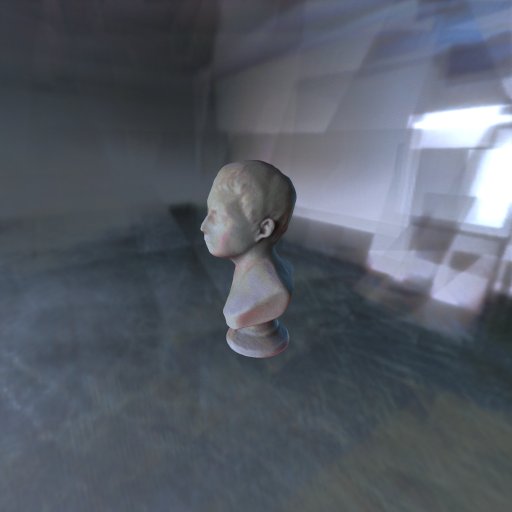} &
\includegraphics[trim={0 0 0 0}, clip, height=0.166\textwidth]{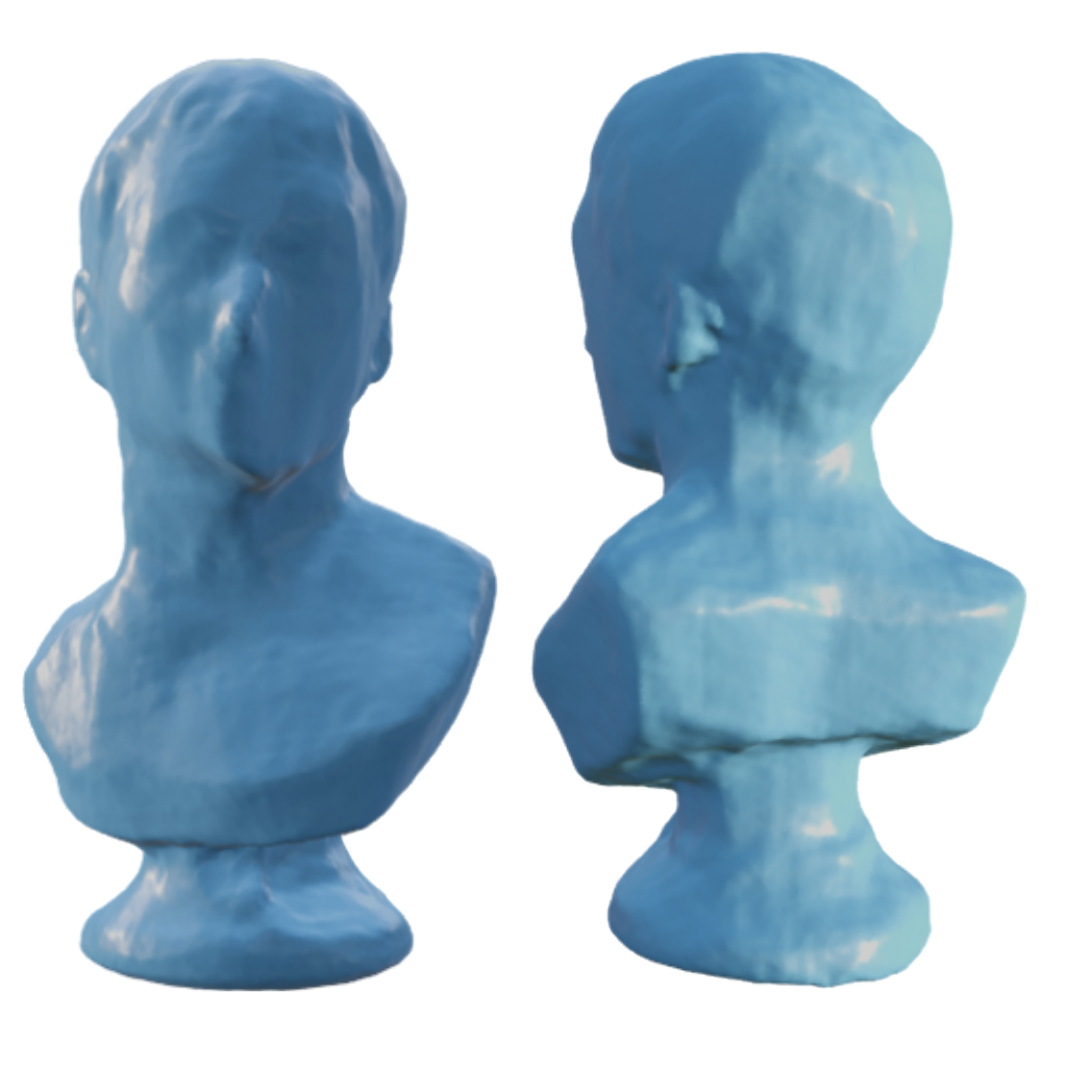} &
\includegraphics[trim={0 0 0 0}, clip, height=0.166\textwidth]{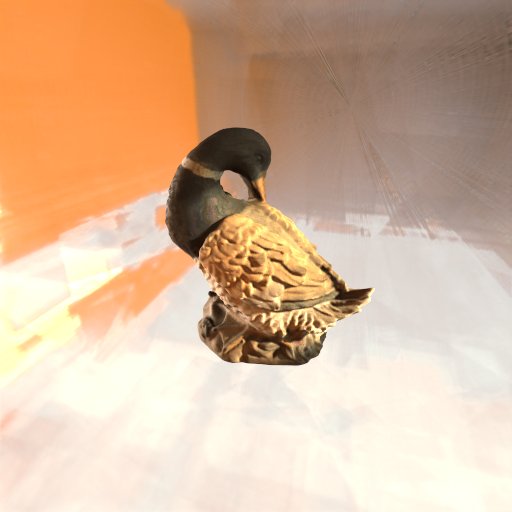} &
\includegraphics[trim={0 0 0 0}, clip, height=0.166\textwidth]{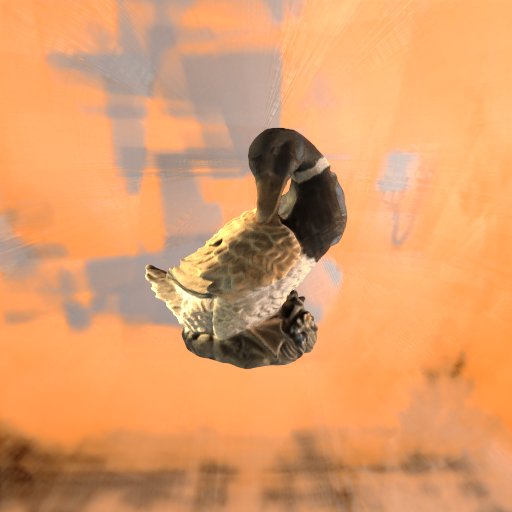} &
\includegraphics[trim={0 0 0 0}, clip, height=0.166\textwidth]{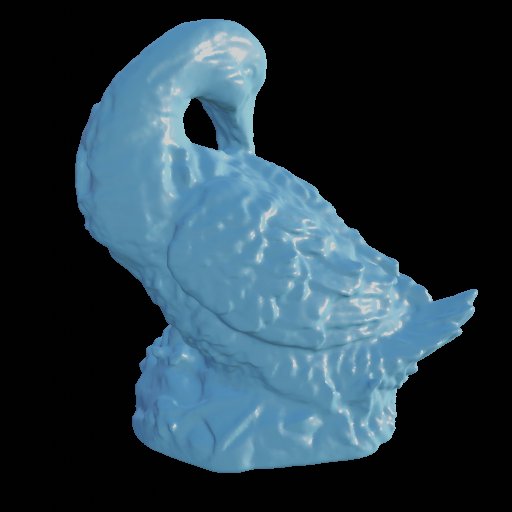} \\
\end{tabular}
}
\vspace{-1.0\baselineskip}
\caption{\textbf{Comparison of lighting representations: NeRF Emitter~\cite{ling2024nerf}, our \envmapname{}, and standard environment map.} In these two scenes, we jointly optimize the object shape, materials, and surrounding lighting with each lighting representation. We report the quantitative results — PSNR for the foreground-only region and the whole image (in the format 'PSNR: (foreground / whole) dB'), as well as Chamfer distance — above the corresponding image, and color-code the \best{\textbf{best}} and \secondbest{\textbf{second best}} method accordingly. Our pipeline successfully recovers the object's geometry, while the standard environment map fails in these scenes due to the violation of the infinite-distance assumption and background parallax. NeRF Emitter uses a NeRF for background lighting, which is computationally expensive. In contrast, our \envmapname{} is a lightweight representation, providing more efficient inverse rendering optimization.}
\label{fig:nerf_emitter_comparison_results}
\end{figure*}

\paragraph*{Comparison with NeRF Emitter~\cite{ling2024nerf} and standard environment map.} We evaluate our \envmapname{} lighting representation against two baselines: NeRF Emitter~\cite{ling2024nerf}, which uses a NeRF for background illumination, and the standard environment map commonly used in prior inverse rendering frameworks. Since NeRF emitter can only work under direct illumination and has no support for antithetic sampling as of writing, we compare these methods using diffuse objects under direct illumination. As shown in Fig.~\ref{fig:nerf_emitter_comparison_results}, we assess the joint optimization of object shape, material, and lighting for two scenes, \textsc{Sculpture} and \textsc{Duck}. Both objects are placed inside an indoor room where the surrounding environment violates the infinite-distance assumption of the standard environment map. Our lightweight \envmapname{} lighting representation enables superior inverse rendering quality and geometry reconstruction, comparable to the computationally expensive NeRF Emitter results.

\paragraph*{Stanford-ORB~\cite{kuang2024stanford} results.} We validate our pipeline on the real-world inverse rendering dataset Stanford-ORB~\cite{kuang2024stanford}. The results demonstrate the effectiveness of our method in terms of geometry reconstruction, relighting, and view interpolation. Our method achieves the best performance on novel scene relighting, while being comparable to Neural-PBIR~\cite{sunNeuralPBIRReconstructionShape2023} on geometry reconstruction and novel view synthesis. Qualitative results are shown in Fig.~\ref{fig:sorb_quality}, and quantitative results are summarized in Table~\ref{tab:stanfordorb}.

\subsection{Evaluations and Ablations}
\label{sec:evaluation_and_ablation}

\begin{figure}[htp]
\centering
\resizebox{0.93\linewidth}{!}{%
\setlength{\tabcolsep}{1.0pt}
\renewcommand{\arraystretch}{0.5}
\begin{tabular}{ccc}
 Reference & Ours (\envmapname) & Envmap \\ 
\includegraphics[trim={0 0 0 0}, clip, width=0.32\linewidth]{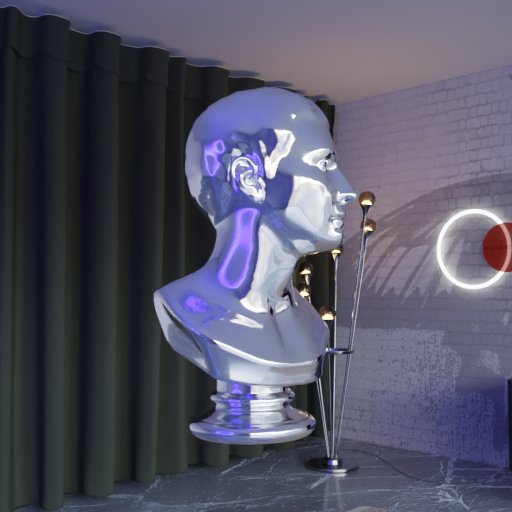} &
\includegraphics[trim={0 0 0 0}, clip, width=0.32\linewidth]{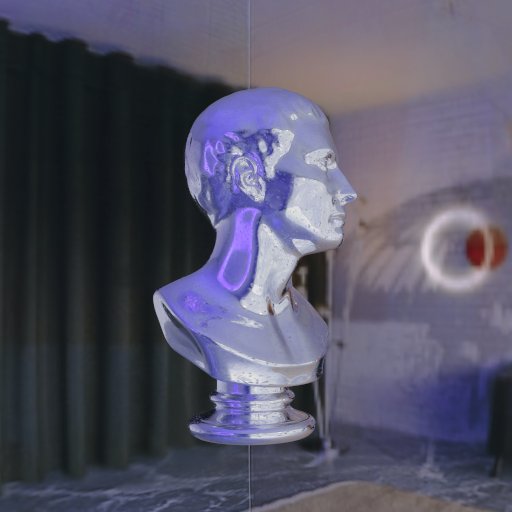}&
\includegraphics[trim={0 0 0 0}, clip, width=0.32\linewidth]{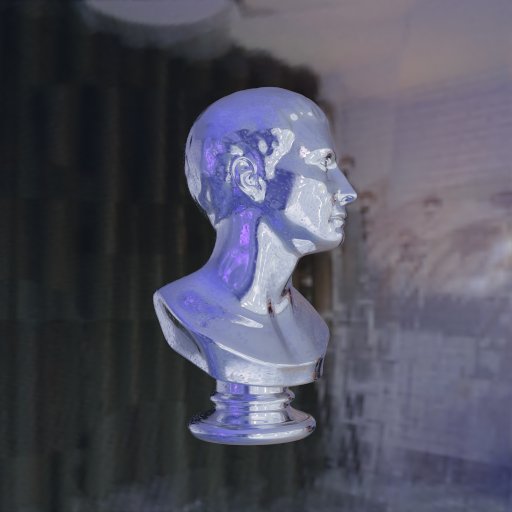}\\
\includegraphics[trim={175 175 200 200}, clip, width=0.32\linewidth]{images/ablation_envmap_pp/training_view_gt.png} &
\includegraphics[trim={175 175 200 200}, clip, width=0.32\linewidth]{images/ablation_envmap_pp/traing_view_envmap++.png}&
\includegraphics[trim={175 175 200 200}, clip, width=0.32\linewidth]{images/ablation_envmap_pp/traing_view_envmap.png}\\
\end{tabular}
}
\vspace{-1.0\baselineskip}
\caption{\textbf{Ablation on \envmapname{}.} We evaluate the quality of glossy object appearance acquisition under non-distant background illumination using our proposed \envmapname{} vs. standard environment map lighting.}
\label{fig:ablation1_env_pp}
\end{figure}

\paragraph*{Importance of \envmapname{}.} In this experiment, we ablate the importance of our proposed \envmapname{} for capturing glossy objects in a near-field environment. We use ground-truth geometry and compare the appearance of the glossy object with \envmapname{} versus a standard environment map. As shown in Fig.~\ref{fig:ablation1_env_pp}, a highly glossy sculpture is placed in an indoor room, presenting strong parallax in the background. Our \envmapname{} successfully captures this non-distant lighting, resulting in a more accurate appearance capture. In contrast, the baseline using standard environment map lighting representation fails to represent the near-field background illumination, leading to poorer quality.

\begin{figure}
     \centering
     \begin{subfigure}[b]{0.116\textwidth}
         \centering
         \includegraphics[trim={50 50 50 50}, clip, width=\linewidth]{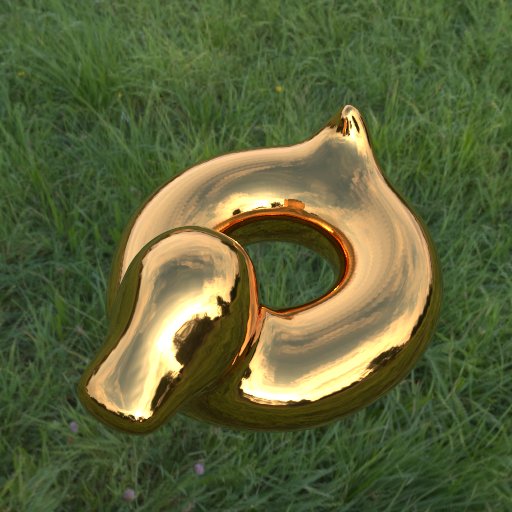}
         \caption{Reference}
         \label{fig:abla_nie_gt}
     \end{subfigure}
     \hfill
     \begin{subfigure}[b]{0.116\textwidth}
         \centering
         \includegraphics[trim={50 50 50 50}, clip, width=\linewidth]{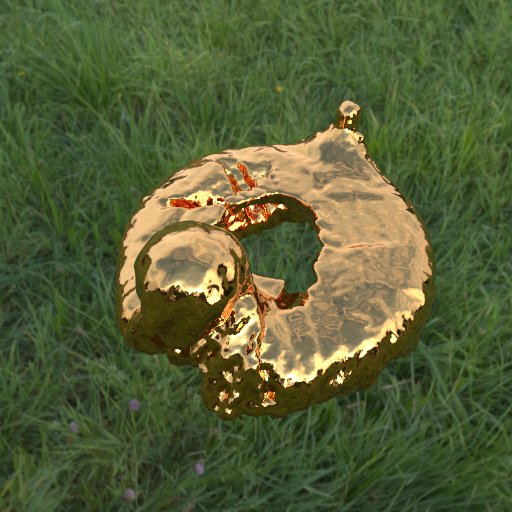}
         \caption{Initial}
         \label{fig:abla_nie_init}
     \end{subfigure}
     \hfill
     \begin{subfigure}[b]{0.116\textwidth}
         \centering
         \includegraphics[trim={50 50 50 50}, clip, width=\linewidth]{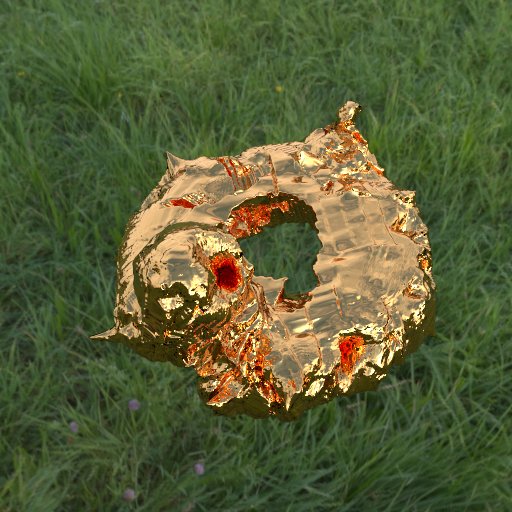}
         \caption{Large Steps}
         \label{fig:abla_nie_ls}
     \end{subfigure}
     \hfill
     \begin{subfigure}[b]{0.116\textwidth}
         \centering
         \includegraphics[trim={50 50 50 50}, clip, width=\linewidth]{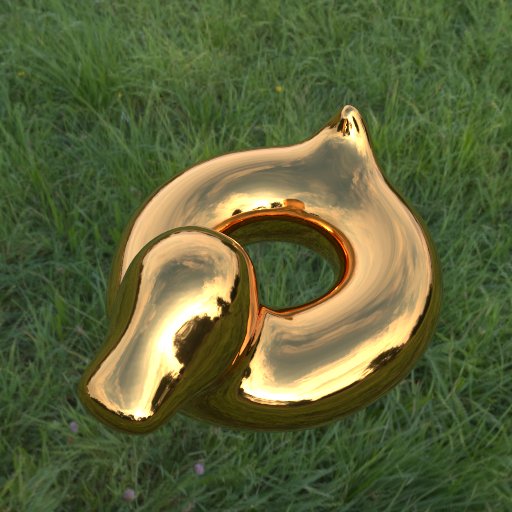}
         \caption{Ours (NIE)}
         \label{fig:abla_nie_nie}
     \end{subfigure}
    \vspace{-1.5\baselineskip}
    \caption{\textbf{Ablation on NIE \cite{mehtaLevelSetTheory2022}} We demonstrate the necessity of NIE when given poor initialization. }
    \label{fig:ablation_nie}
\end{figure}

\paragraph*{NIE vs. Large Steps~\cite{Nicolet2021Large}.} In this experiment, we ablate the NIE component of our pipeline by comparing the reconstructed geometry with fixed lighting and material against the reconstruction produced by the \emph{Large Steps} algorithm~\cite{Nicolet2021Large}. The \emph{Large Steps} algorithm operates on an explicit mesh with fixed topology, whereas \emph{NIE} uses the explicit mesh as a proxy to optimize an implicit surface capable of topology changes. This capability is crucial when dealing with poor initializations that contain holes, as demonstrated in Fig.~\ref{fig:abla_nie_ls} and Fig.~\ref{fig:ablation_nie}, where \emph{NIE} outperforms \emph{Large Steps}. Additionally, the gradient clamping step described in Sec.~\ref{sec:nie} helps stabilize the optimization process without creating spikes, as shown in Fig.~\ref{fig:abla_nie_ls}.


\section{Discussion and Conclusion}
\label{sec:conclusion}

\paragraph*{Limitations and future work}

Although \envmapname{} provides a lightweight solution to address the non-distant illumination problem, its representational power is greatly limited by the underlying geometry: a deformed cube cannot possibly model complex backgrounds, especially when the background is very close to the object or when occlusion is present. Additionally, it cannot accurately model strong directional light sources due to the lack of directional information in the radiance texture. Nevertheless, we found it satisfactory for common scenes. Developing lightweight emitters with greater representational power would be of future interest. 

Another issue we aim to further explore is the ambiguity between lighting and material. This ambiguity often manifests as "baking," for instance, reconstructed textures may incorrectly embed specular highlights. While humans can easily identify and resolve such artifacts, optimizers see these artifacts as just one of many possible solutions. To find the optimal solution, a strong prior in material and lighting is necessary.

\paragraph*{Conclusion}
In this work, we introduced an inverse rendering framework, \methodname{}, for reconstructing highly glossy objects' geometry, material, and surrounding illumination. By integrating neural implicit evolution, we achieved robust handling of complex object topology without the need for careful initialization. Our lightweight background representation, \envmapname{}, efficiently models both near-field and far-field background illumination, offering a more efficient solution for inverse rendering tasks. To tackle challenges with highly glossy BRDFs, we integrated an efficient variant of antithetic sampling, enabling faster convergence and more accurate reconstruction. Our pipeline delivers state-of-the-art results in geometry, material, and lighting estimation, enabling realistic view synthesis and relighting. 

%
\bibliographystyle{ACM-Reference-Format}
\bibliography{references}
\clearpage

\clearpage

\section{Supplemental Results}
\begin{figure*}[t]
    \centering
    \resizebox{1.\textwidth}{!}{%
    \setlength{\tabcolsep}{1.5pt}
    \renewcommand{\arraystretch}{0.5}
    \begin{tabular}{ccc|ccc|cc}
    &  \multicolumn{2}{c}{Novel view rendering} & \multicolumn{3}{c}{Relighting}  & Normal & Albedo\\
    \rotatebox{90}{\parbox{0.133\textwidth}{\centering \textsc{Ball} }} &
    \includegraphics[trim={50 50 50 50}, clip, height=0.133\textwidth]{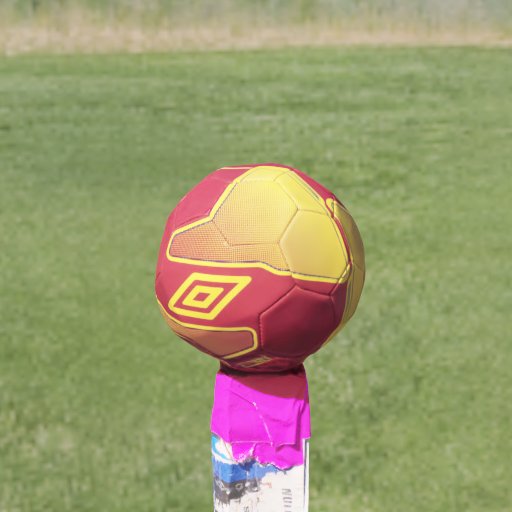} &
    \includegraphics[trim={50 50 50 50}, clip, height=0.133\textwidth]{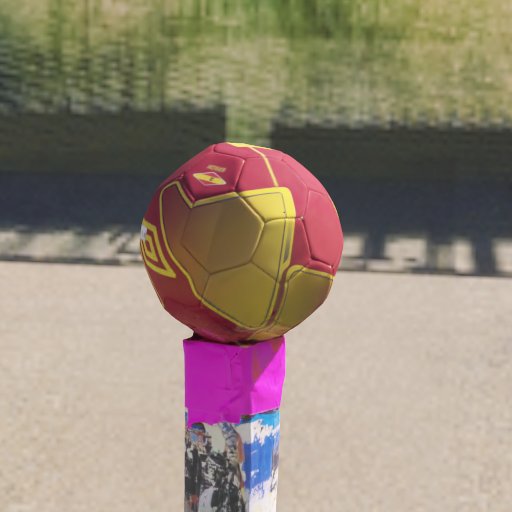} &
    \includegraphics[trim={50 50 50 50}, clip, height=0.133\textwidth]{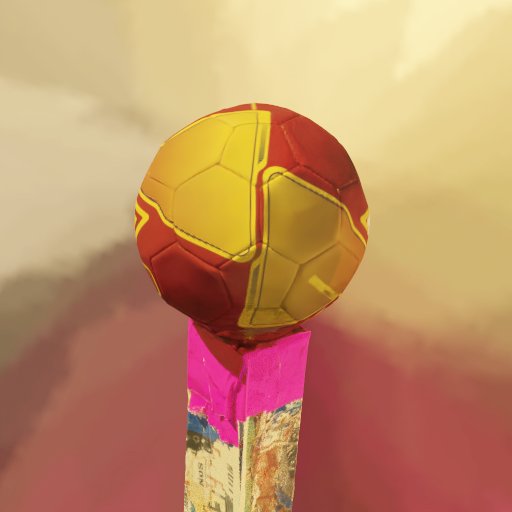} &
    \includegraphics[trim={50 50 50 50}, clip, height=0.133\textwidth]{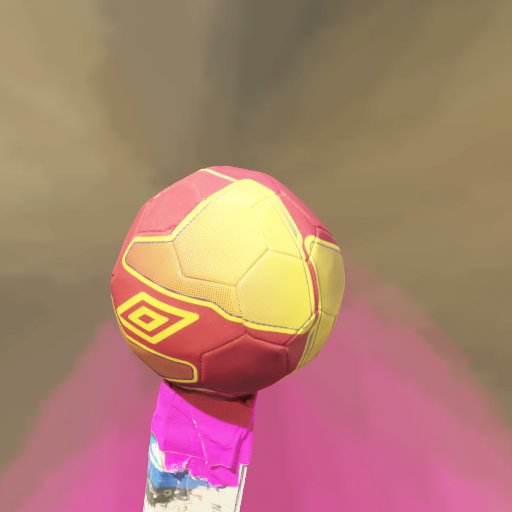} &
    \includegraphics[trim={50 50 50 50}, clip, height=0.133\textwidth]{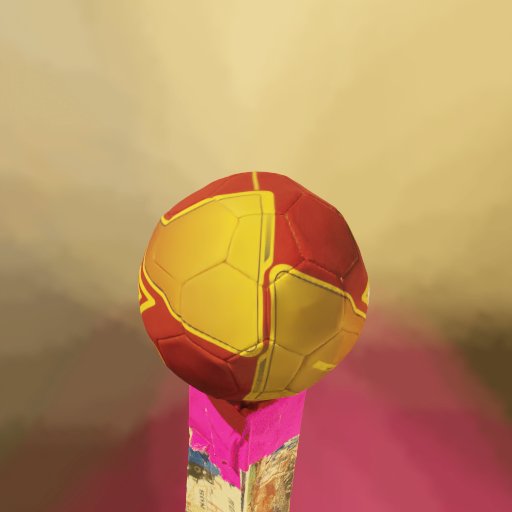} &
    \includegraphics[trim={50 50 50 50}, clip, height=0.133\textwidth]{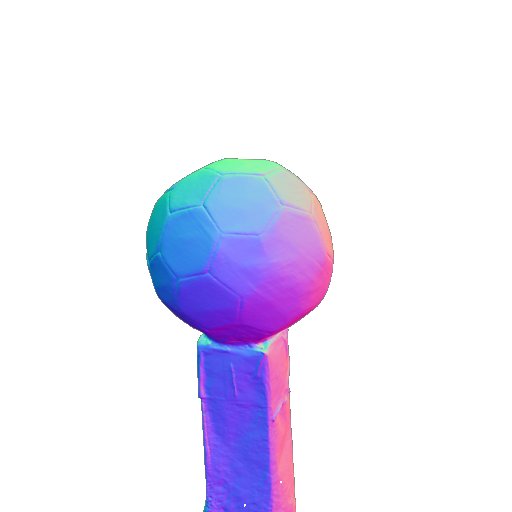} &
    \includegraphics[trim={50 50 50 50}, clip, height=0.133\textwidth]{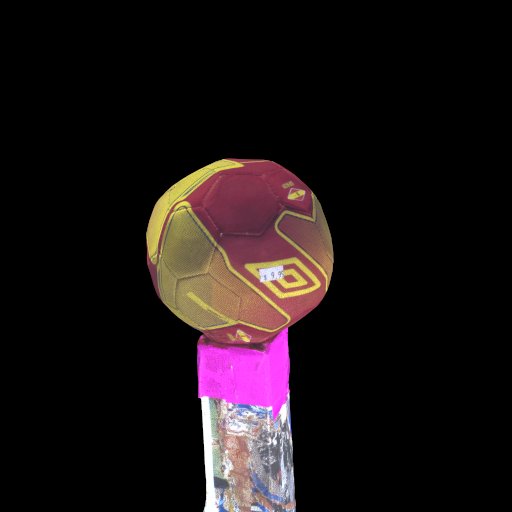} \\
    \rotatebox{90}{\parbox{0.133\textwidth}{\centering \textsc{Baking} }} &
    \includegraphics[trim={50 50 50 50}, clip, height=0.133\textwidth]{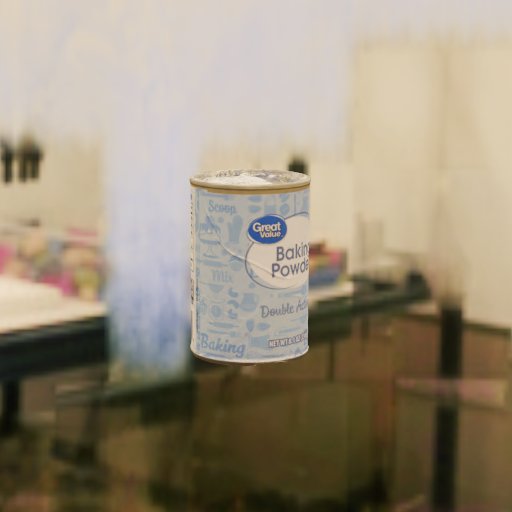} &
    \includegraphics[trim={50 50 50 50}, clip, height=0.133\textwidth]{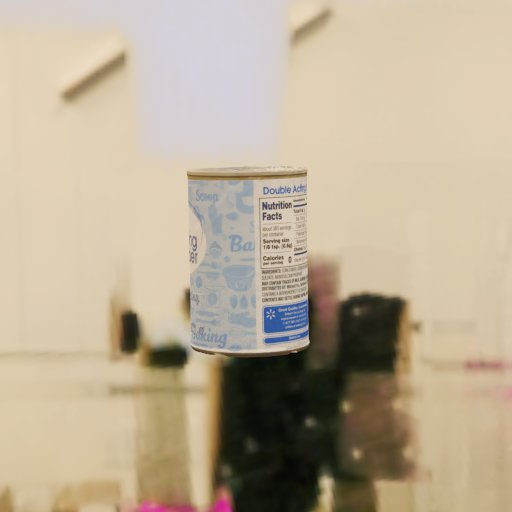} &
    \includegraphics[trim={50 50 50 50}, clip, height=0.133\textwidth]{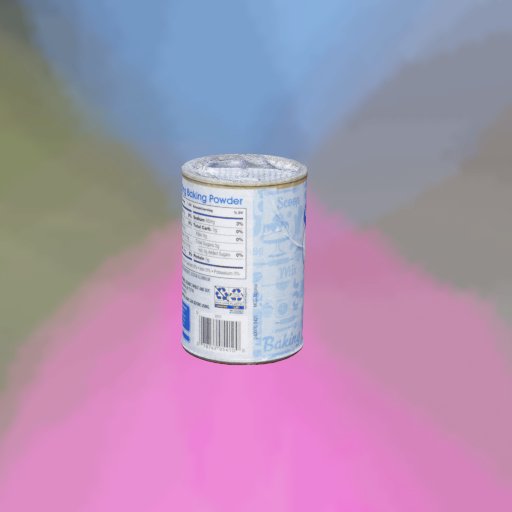} &
    \includegraphics[trim={50 50 50 50}, clip, height=0.133\textwidth]{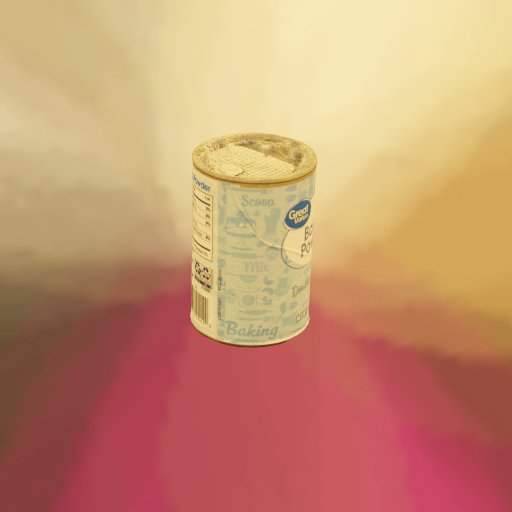} &
    \includegraphics[trim={50 50 50 50}, clip, height=0.133\textwidth]{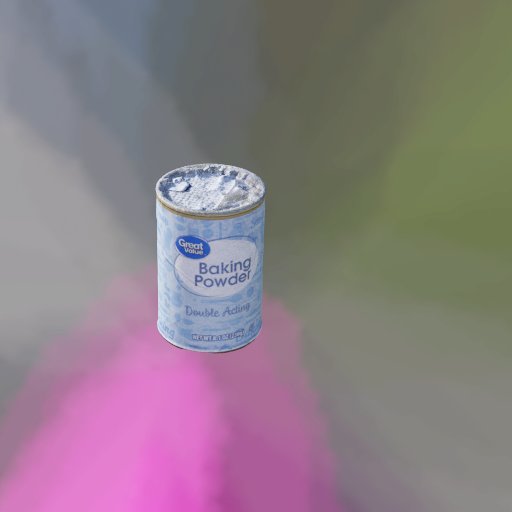} &
    \includegraphics[trim={50 50 50 50}, clip, height=0.133\textwidth]{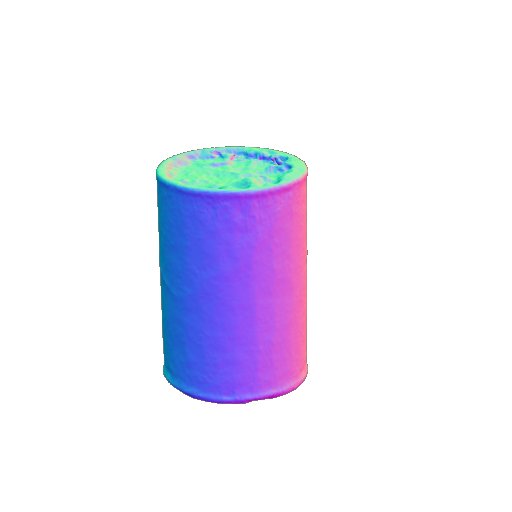} &
    \includegraphics[trim={50 50 50 50}, clip, height=0.133\textwidth]{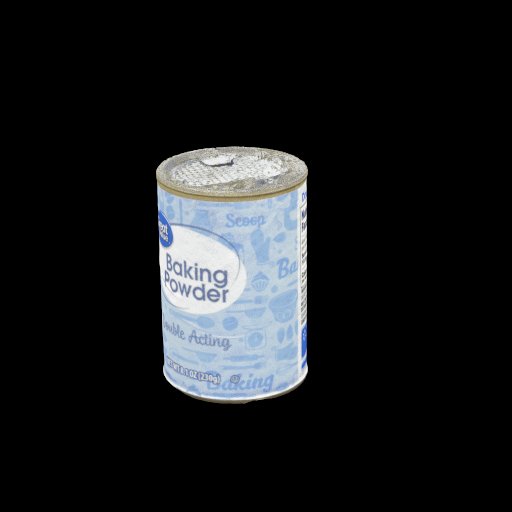} 
    \\
    \rotatebox{90}{\parbox{0.133\textwidth}{\centering \textsc{Blocks} }} &
    \includegraphics[trim={50 50 50 50}, clip, height=0.133\textwidth]{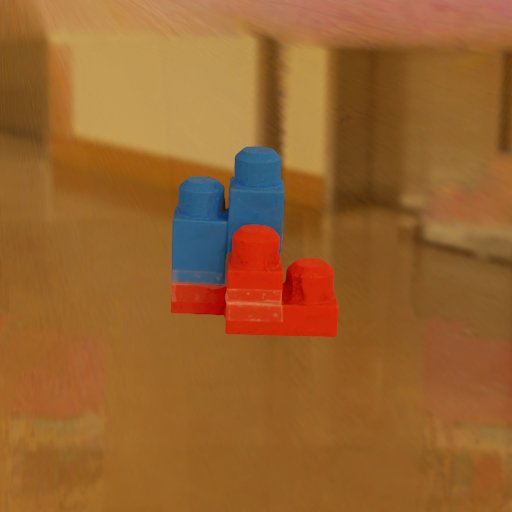} &
    \includegraphics[trim={50 50 50 50}, clip, height=0.133\textwidth]{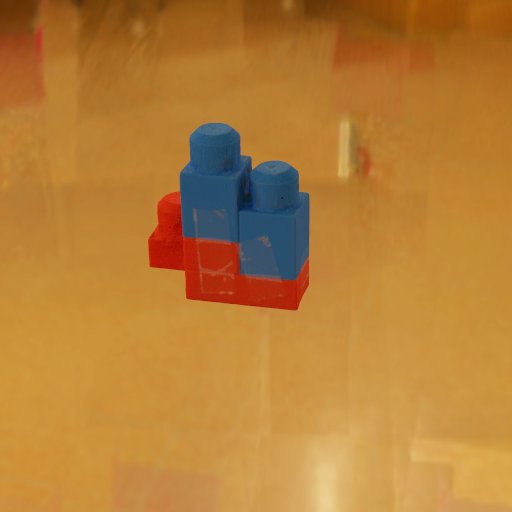} &
    \includegraphics[trim={50 50 50 50}, clip, height=0.133\textwidth]{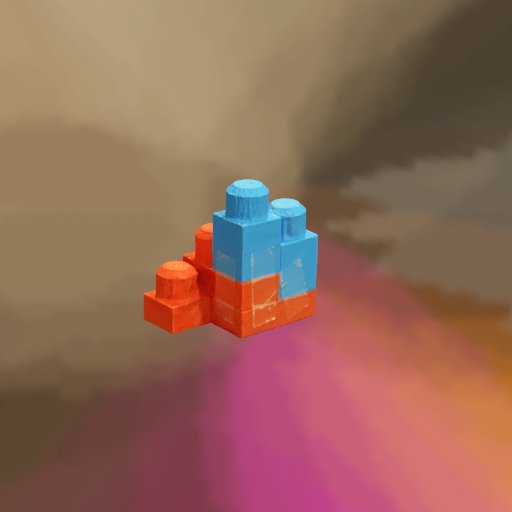} &
    \includegraphics[trim={50 50 50 50}, clip, height=0.133\textwidth]{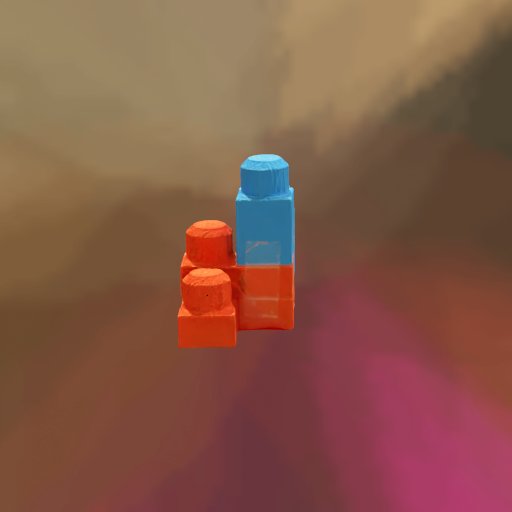} &
    \includegraphics[trim={50 50 50 50}, clip, height=0.133\textwidth]{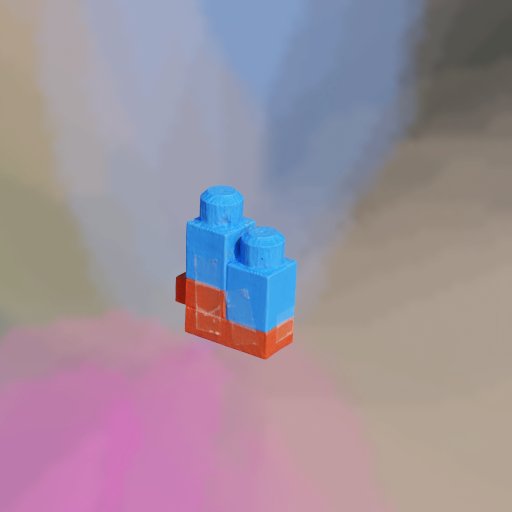} &
    \includegraphics[trim={50 50 50 50}, clip, height=0.133\textwidth]{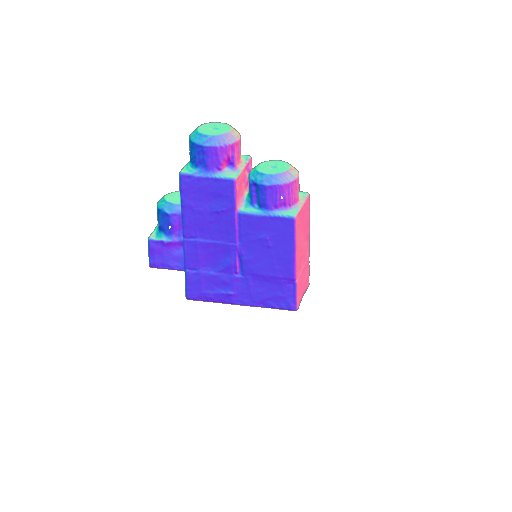} &
    \includegraphics[trim={50 50 50 50}, clip, height=0.133\textwidth]{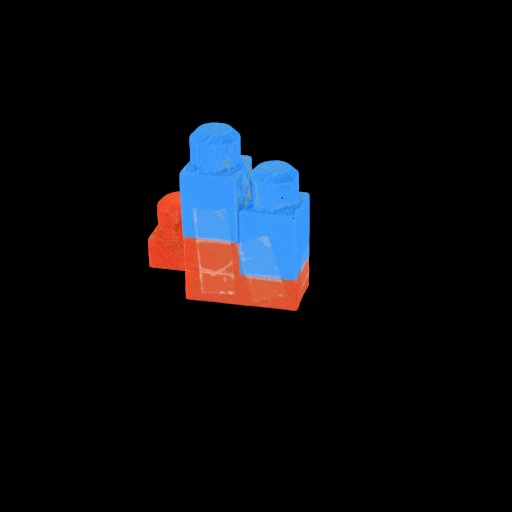} 
    \\
    \rotatebox{90}{\parbox{0.133\textwidth}{\centering \textsc{Cactus} }} &
    \includegraphics[trim={50 50 50 50}, clip, height=0.133\textwidth]{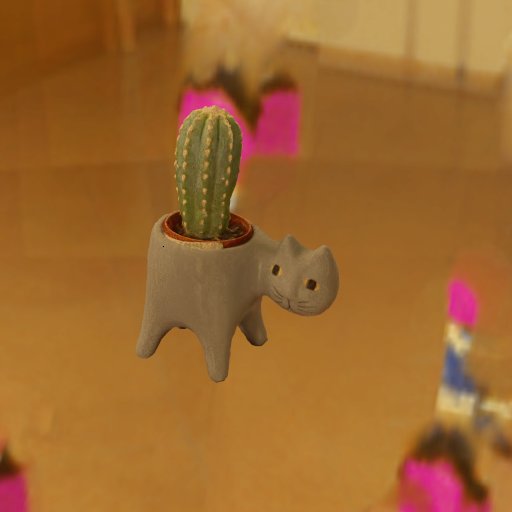} &
    \includegraphics[trim={50 50 50 50}, clip, height=0.133\textwidth]{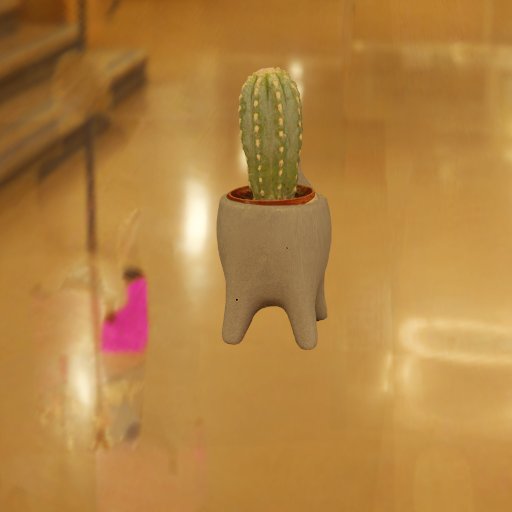} &
    \includegraphics[trim={50 50 50 50}, clip, height=0.133\textwidth]{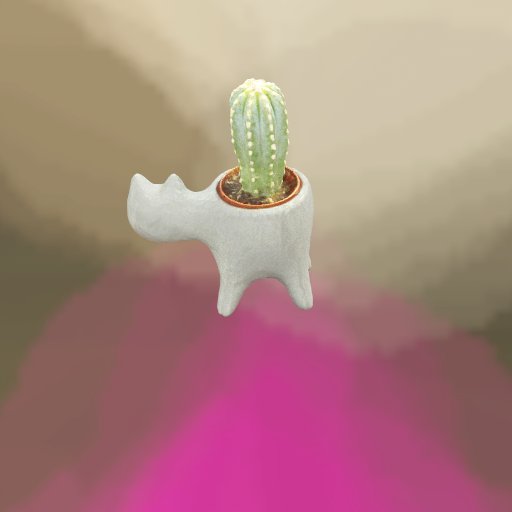} &
    \includegraphics[trim={50 50 50 50}, clip, height=0.133\textwidth]{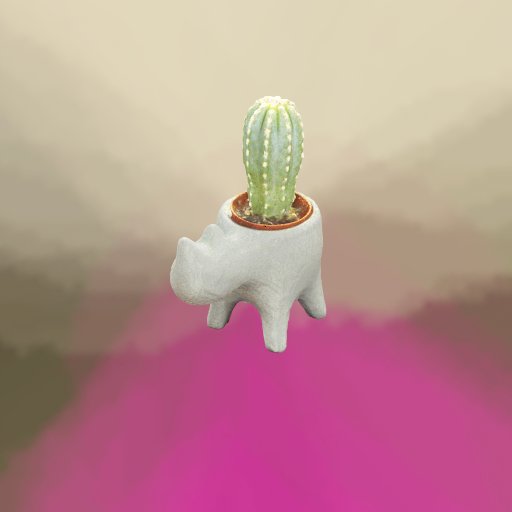} &
    \includegraphics[trim={50 50 50 50}, clip, height=0.133\textwidth]{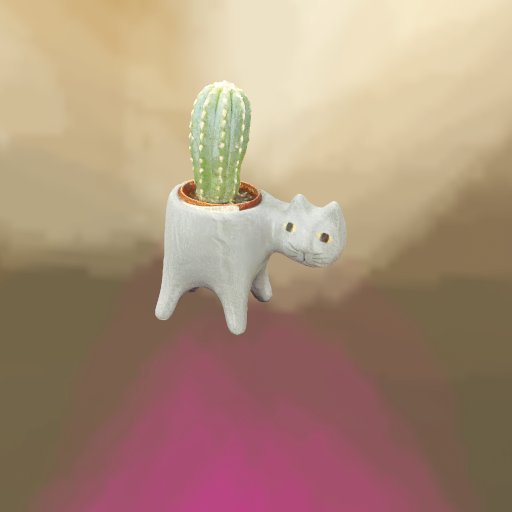} &
    \includegraphics[trim={50 50 50 50}, clip, height=0.133\textwidth]{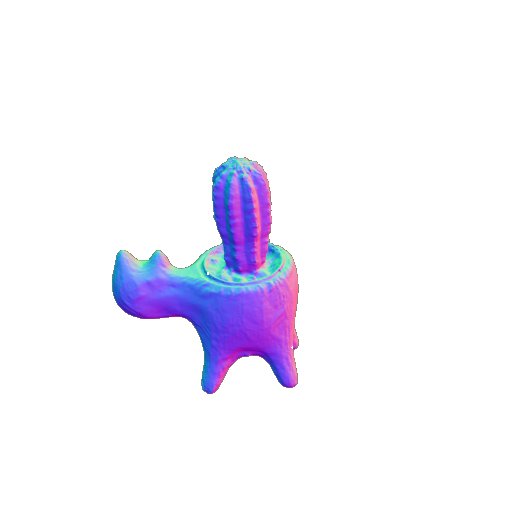} &
    \includegraphics[trim={50 50 50 50}, clip, height=0.133\textwidth]{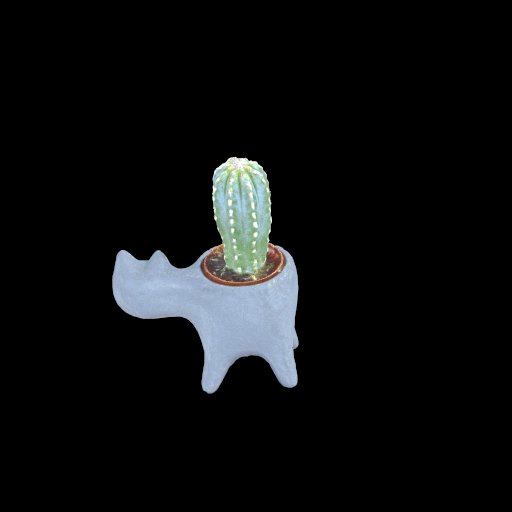} 
    \\
    \rotatebox{90}{\parbox{0.133\textwidth}{\centering \textsc{Car} }} &
    \includegraphics[trim={50 50 50 50}, clip, height=0.133\textwidth]{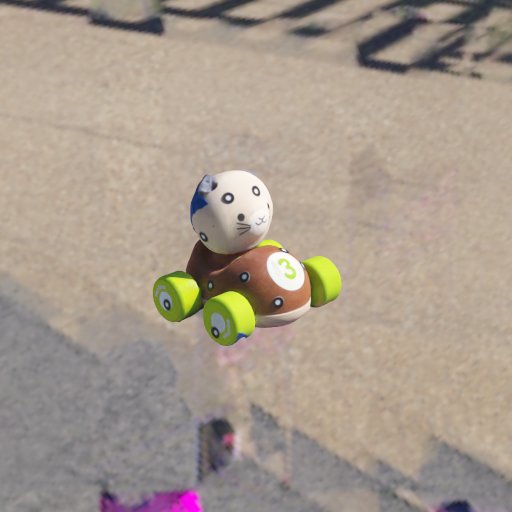} &
    \includegraphics[trim={50 50 50 50}, clip, height=0.133\textwidth]{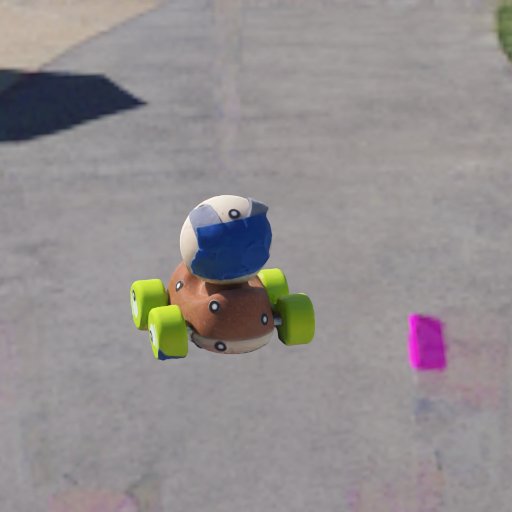} &
    \includegraphics[trim={50 50 50 50}, clip, height=0.133\textwidth]{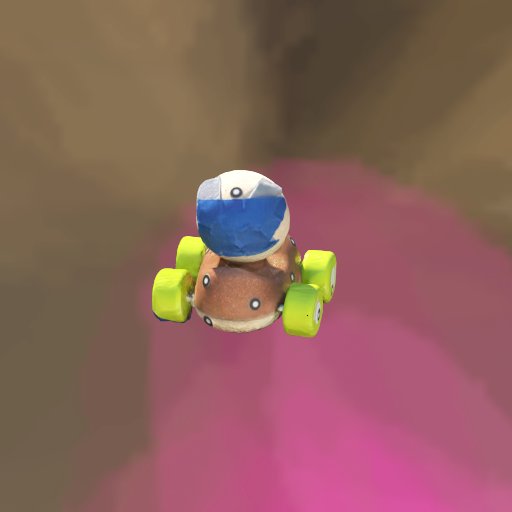} &
    \includegraphics[trim={50 50 50 50}, clip, height=0.133\textwidth]{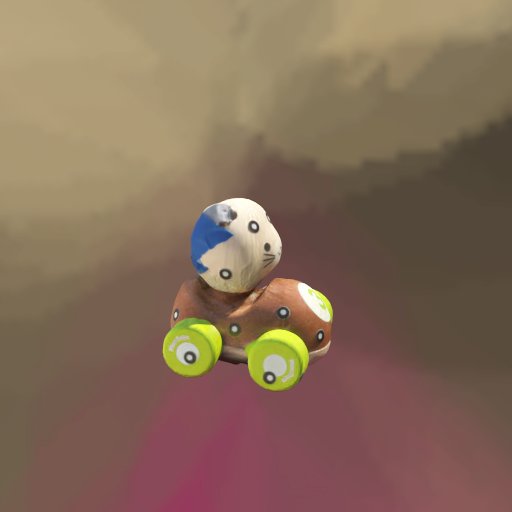} &
    \includegraphics[trim={50 50 50 50}, clip, height=0.133\textwidth]{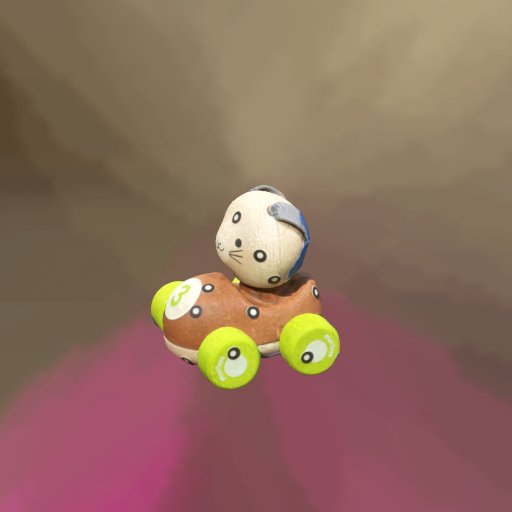} &
    \includegraphics[trim={50 50 50 50}, clip, height=0.133\textwidth]{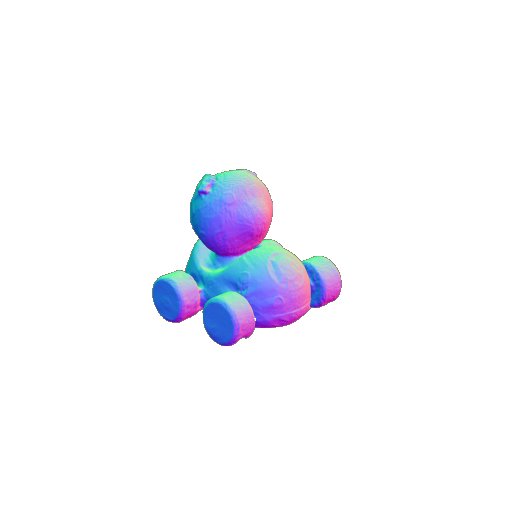} &
    \includegraphics[trim={50 50 50 50}, clip, height=0.133\textwidth]{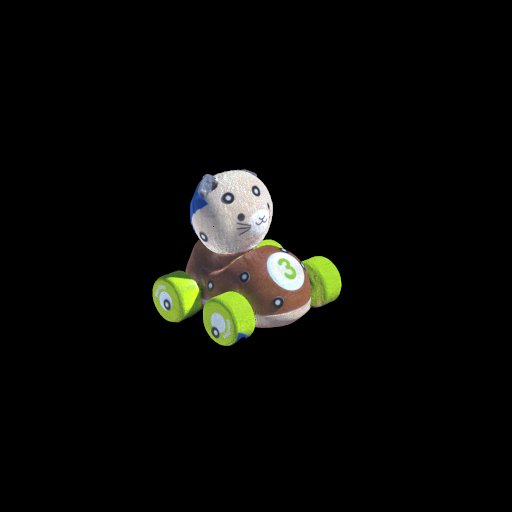} 
    \\
    \rotatebox{90}{\parbox{0.133\textwidth}{\centering \textsc{Gnome} }} &
    \includegraphics[trim={50 50 50 50}, clip, height=0.133\textwidth]{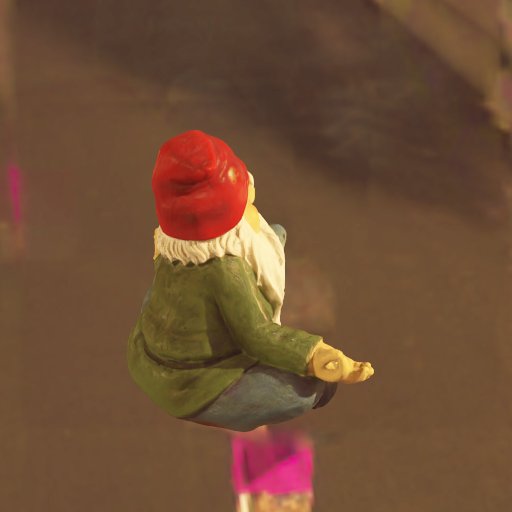} &
    \includegraphics[trim={50 50 50 50}, clip, height=0.133\textwidth]{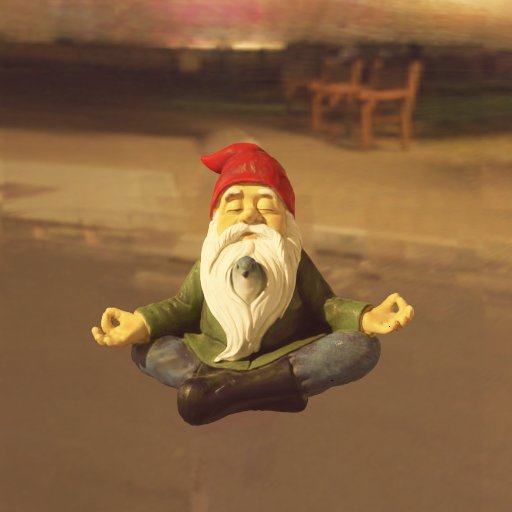} &
    \includegraphics[trim={20 20 20 20}, clip, height=0.133\textwidth]{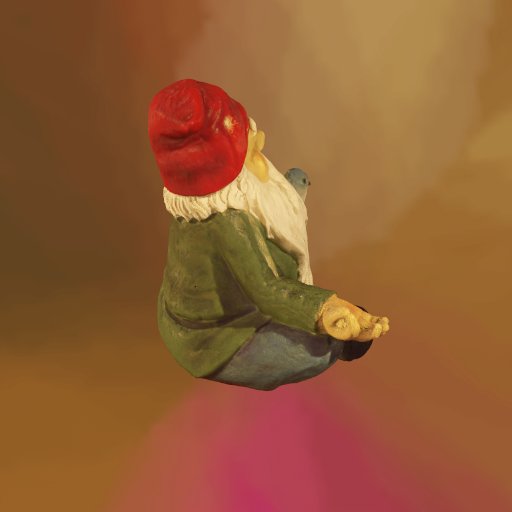} &
    \includegraphics[trim={20 20 20 20}, clip, height=0.133\textwidth]{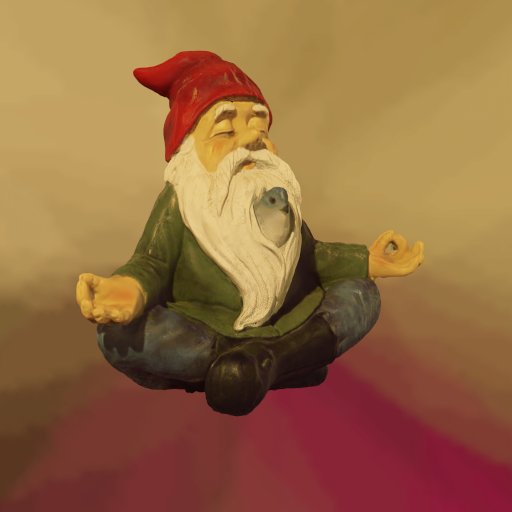} &
    \includegraphics[trim={20 20 20 20}, clip, height=0.133\textwidth]{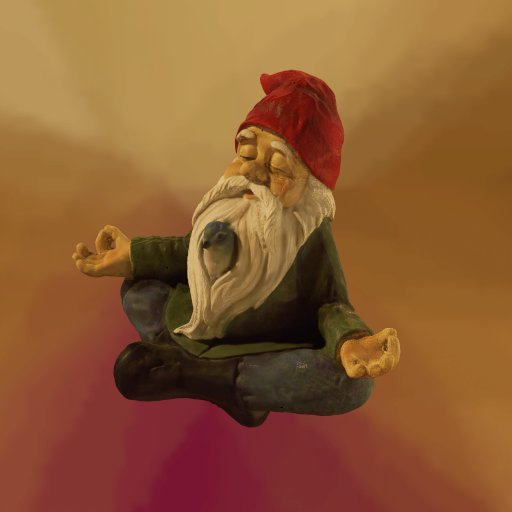} &
    \includegraphics[trim={50 50 50 50}, clip, height=0.133\textwidth]{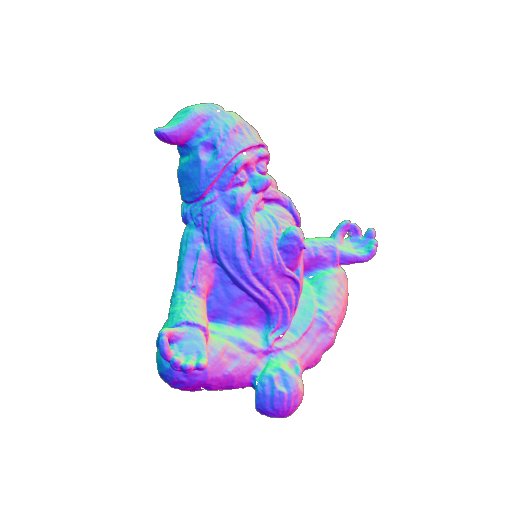} &
    \includegraphics[trim={50 50 50 50}, clip, height=0.133\textwidth]{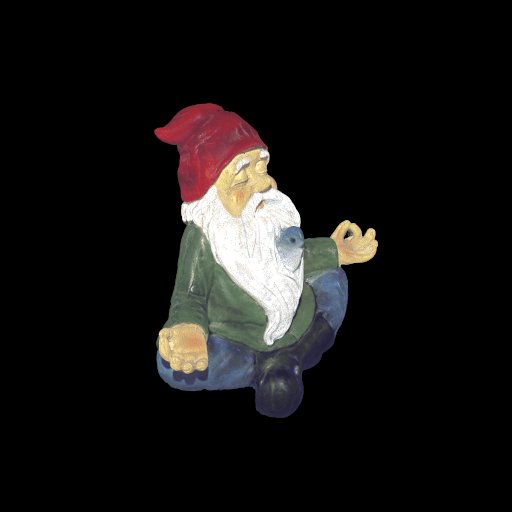} 
    \\
    \rotatebox{90}{\parbox{0.133\textwidth}{\centering \textsc{Pepsi} }} &
    \includegraphics[trim={50 50 50 50}, clip, height=0.133\textwidth]{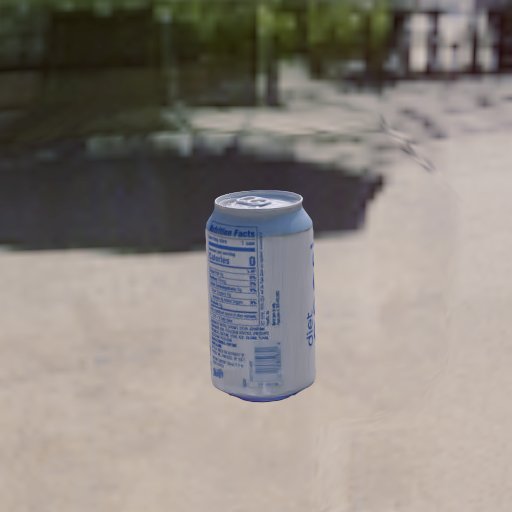} &
    \includegraphics[trim={50 50 50 50}, clip, height=0.133\textwidth]{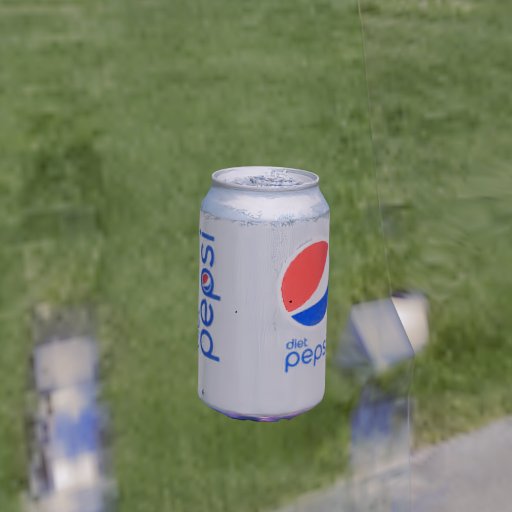} &
    \includegraphics[trim={0 50 100 50}, clip, height=0.133\textwidth]{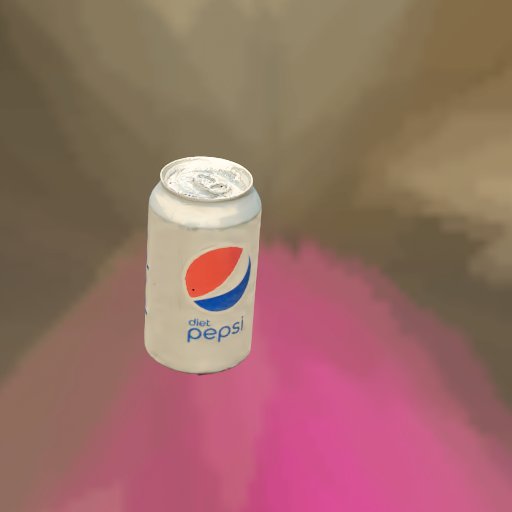} &
    \includegraphics[trim={50 50 50 50}, clip, height=0.133\textwidth]{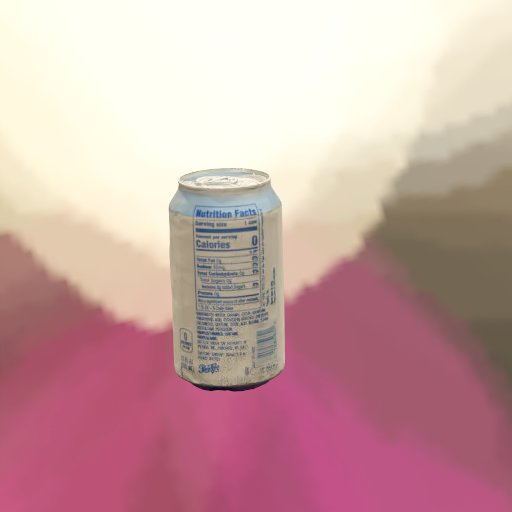} &
    \includegraphics[trim={50 50 50 50}, clip, height=0.133\textwidth]{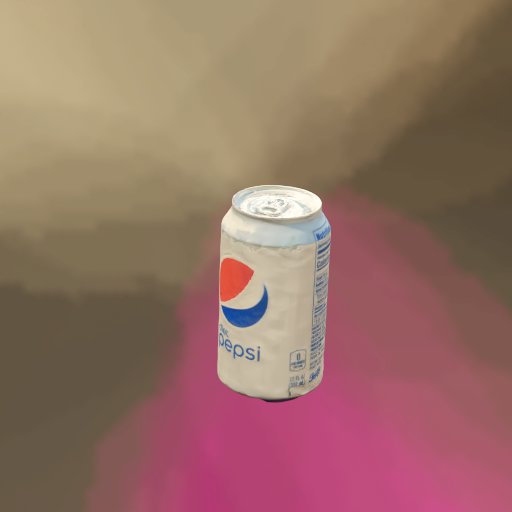} &
    \includegraphics[trim={50 50 50 50}, clip, height=0.133\textwidth]{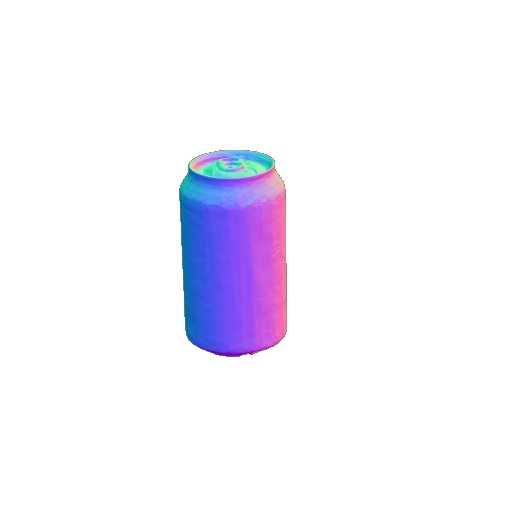} &
    \includegraphics[trim={50 50 50 50}, clip, height=0.133\textwidth]{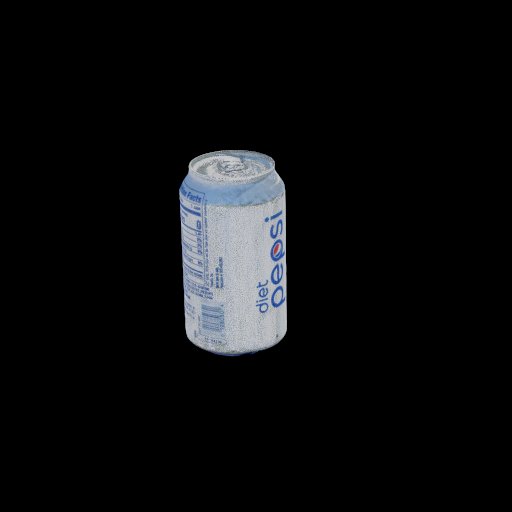} 
    \\
    \end{tabular}
    }
    \caption{\textbf{Additional results of our pipeline on Stanford-ORB~\cite{kuang2024stanford} data.} .}
    \label{fig:supp_sorb}
\end{figure*}

\begin{figure*}[t]
    \centering
    \resizebox{1.\textwidth}{!}{%
    \setlength{\tabcolsep}{1.5pt}
    \renewcommand{\arraystretch}{0.5}
    \begin{tabular}{ccc|cc|ccc}
    &  \multicolumn{2}{c}{Novel view rendering} & \multicolumn{2}{c}{Recovered lighting}  & \multicolumn{3}{c}{Relighting}\\
    \rotatebox{90}{\parbox{0.133\textwidth}{\centering Ours }} &
    \includegraphics[trim={0 0 0 0}, clip, height=0.133\textwidth]{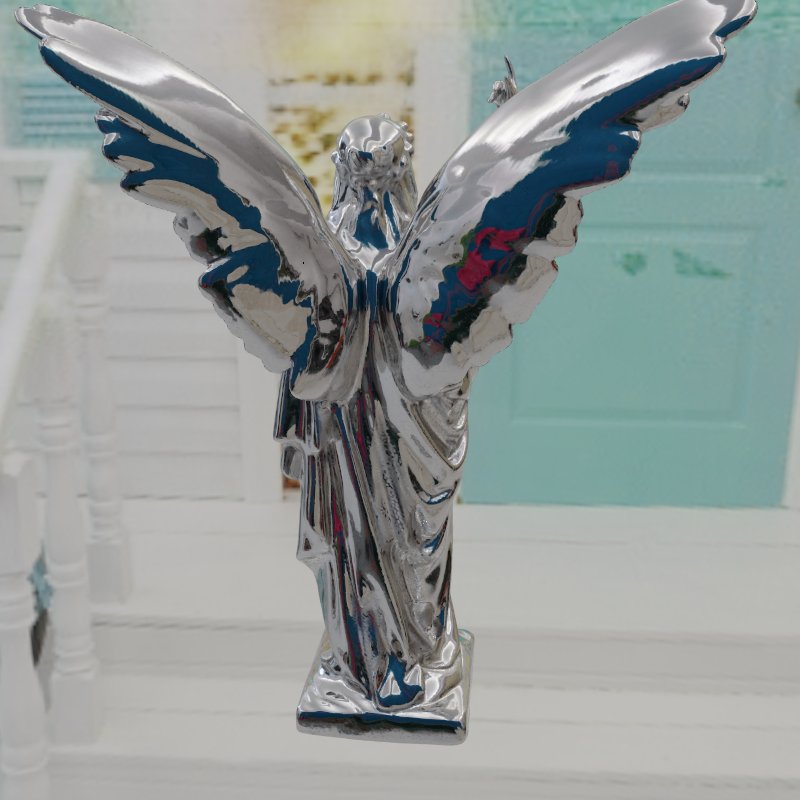} &
    \includegraphics[trim={0 0 0 0}, clip, height=0.133\textwidth]{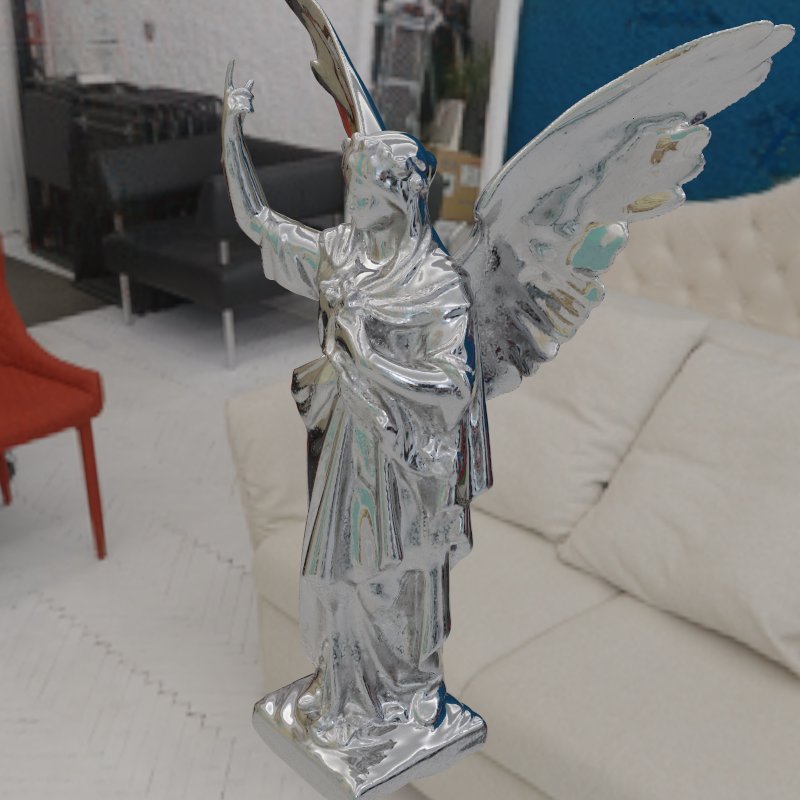} &
    \multicolumn{2}{c}{\includegraphics[trim={0 0 0 0}, clip, height=0.133\textwidth]{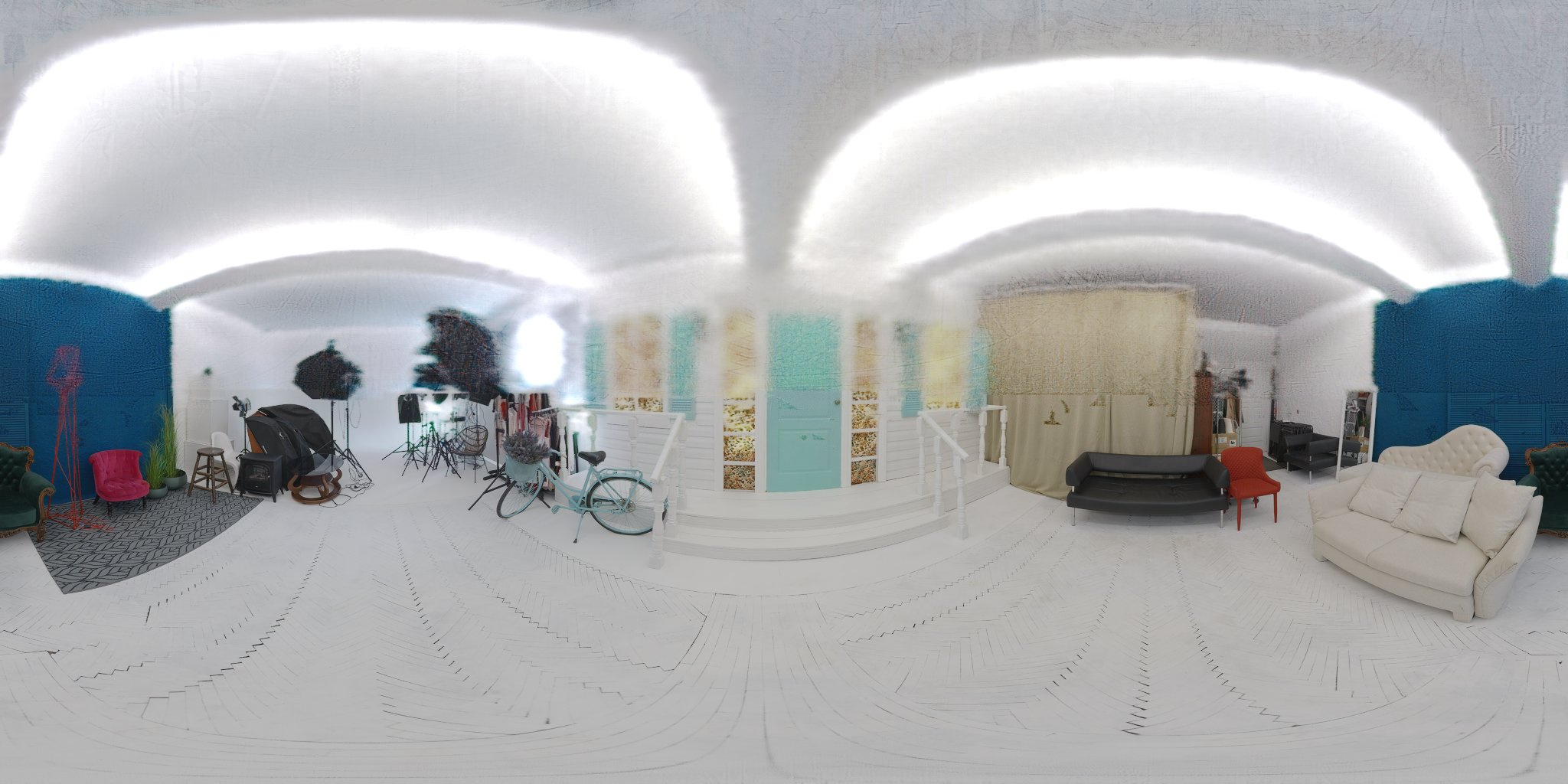}  } &
    \includegraphics[trim={0 0 0 0}, clip, height=0.133\textwidth]{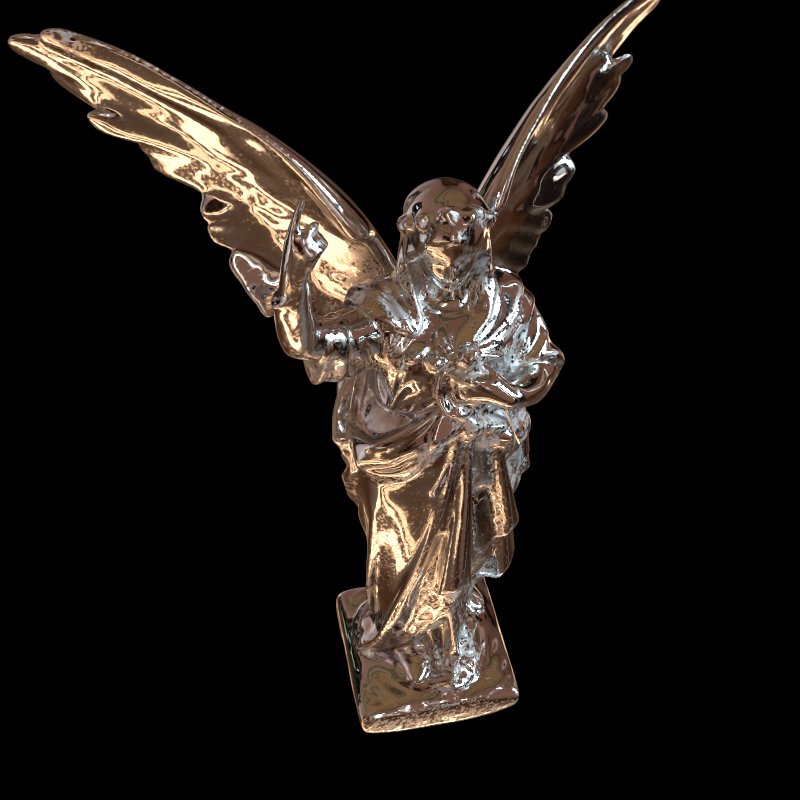}  &
    \includegraphics[trim={0 0 0 0}, clip, height=0.133\textwidth]{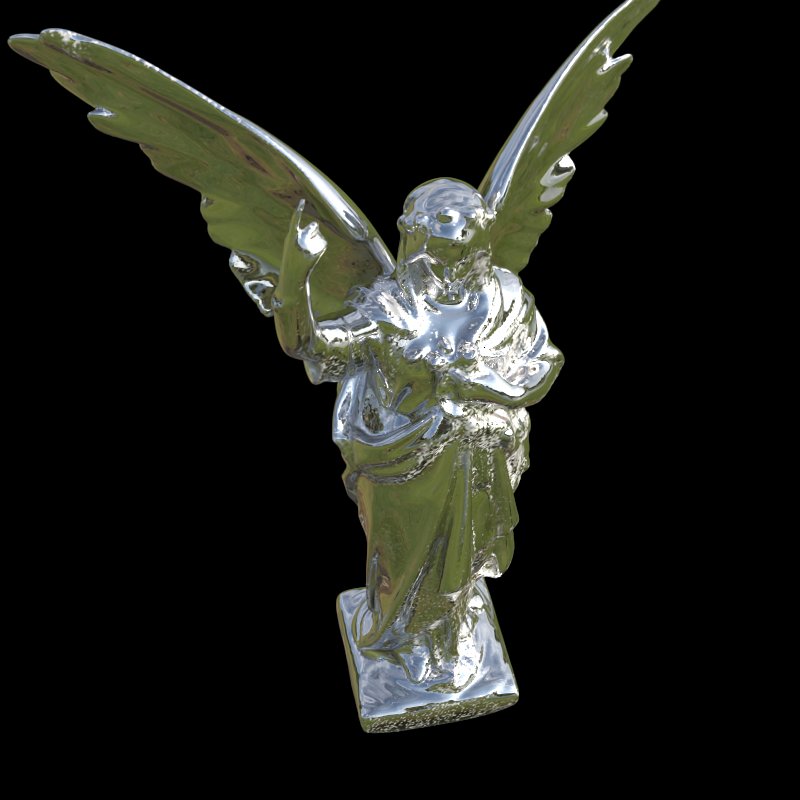}  &
    \includegraphics[trim={0 0 0 0}, clip, height=0.133\textwidth]{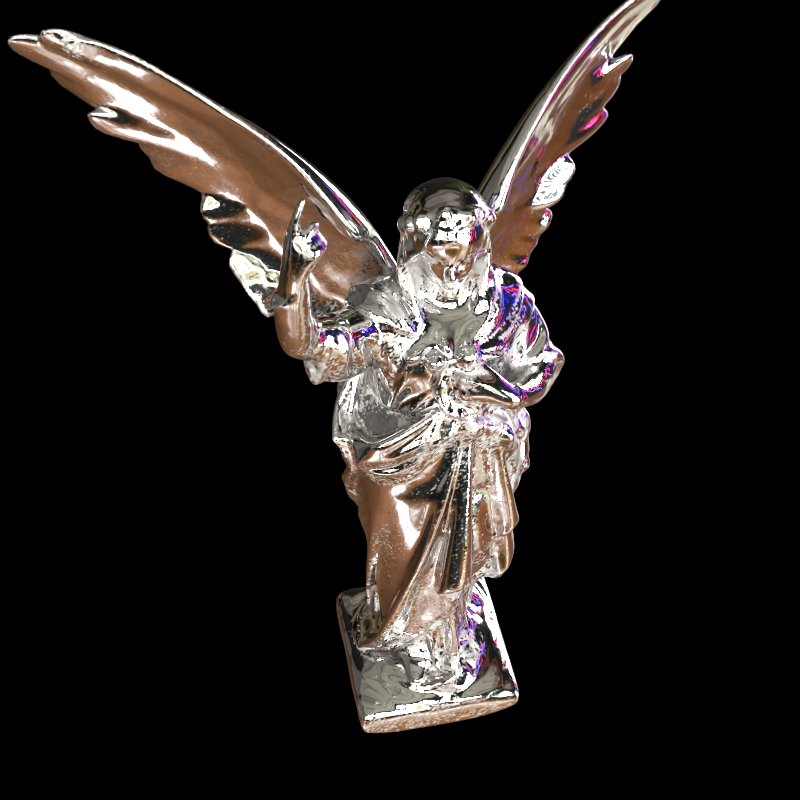} \\
    \rotatebox{90}{\parbox{0.133\textwidth}{\centering NeRO }} &
    \includegraphics[trim={0 0 0 0}, clip, height=0.133\textwidth]{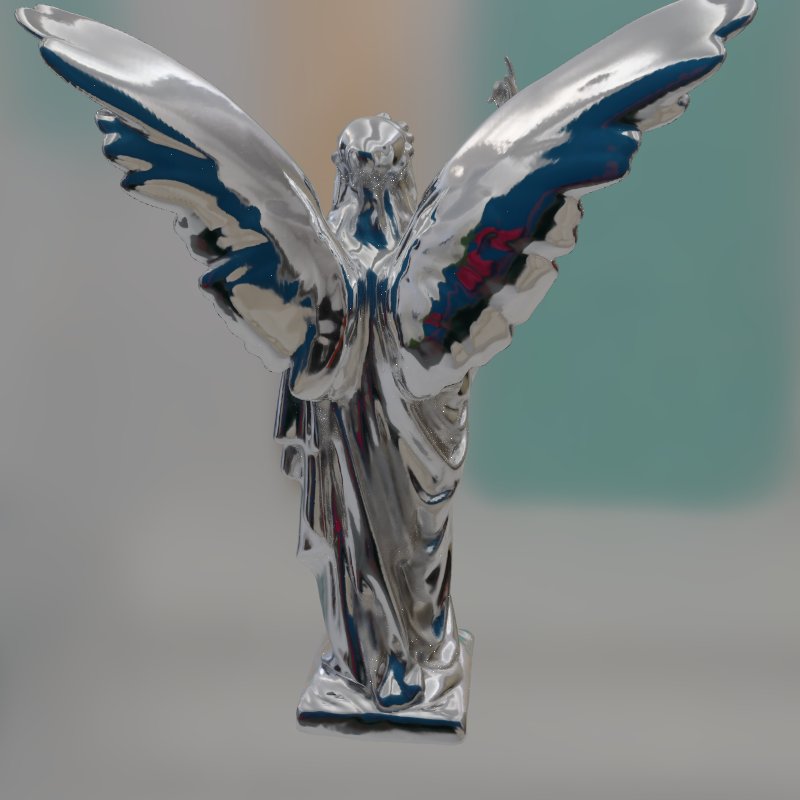} &
    \includegraphics[trim={0 0 0 0}, clip, height=0.133\textwidth]{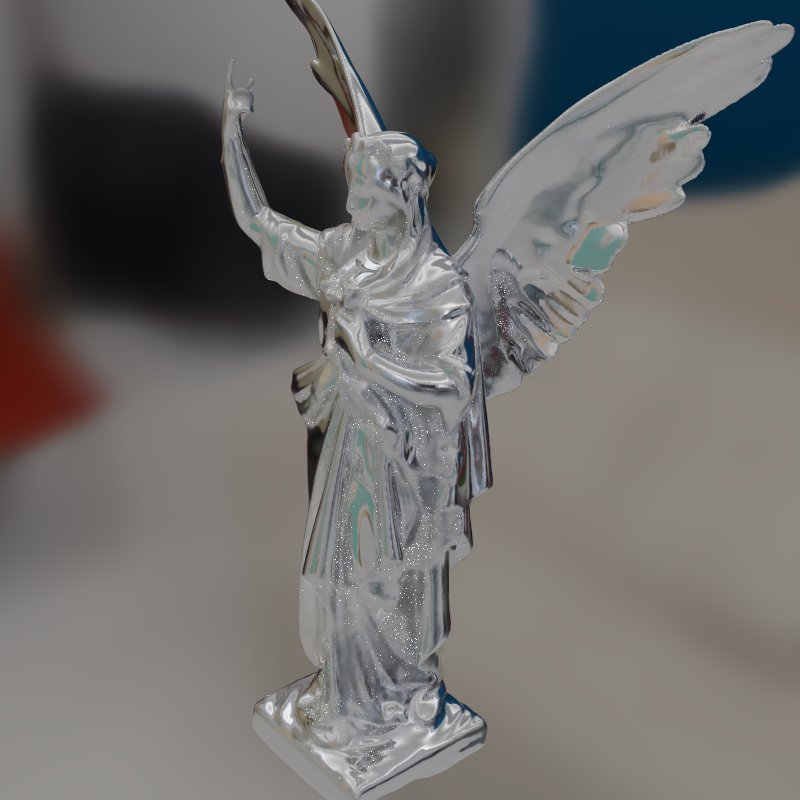} &
    \multicolumn{2}{c}{\includegraphics[trim={0 0 0 0}, clip, height=0.133\textwidth]{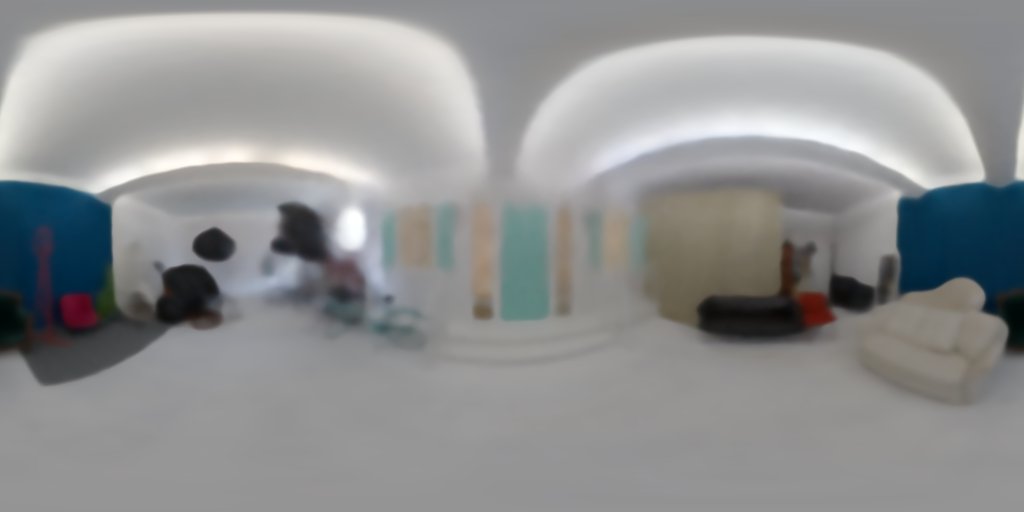}  } &
    \includegraphics[trim={0 0 0 0}, clip, height=0.133\textwidth]{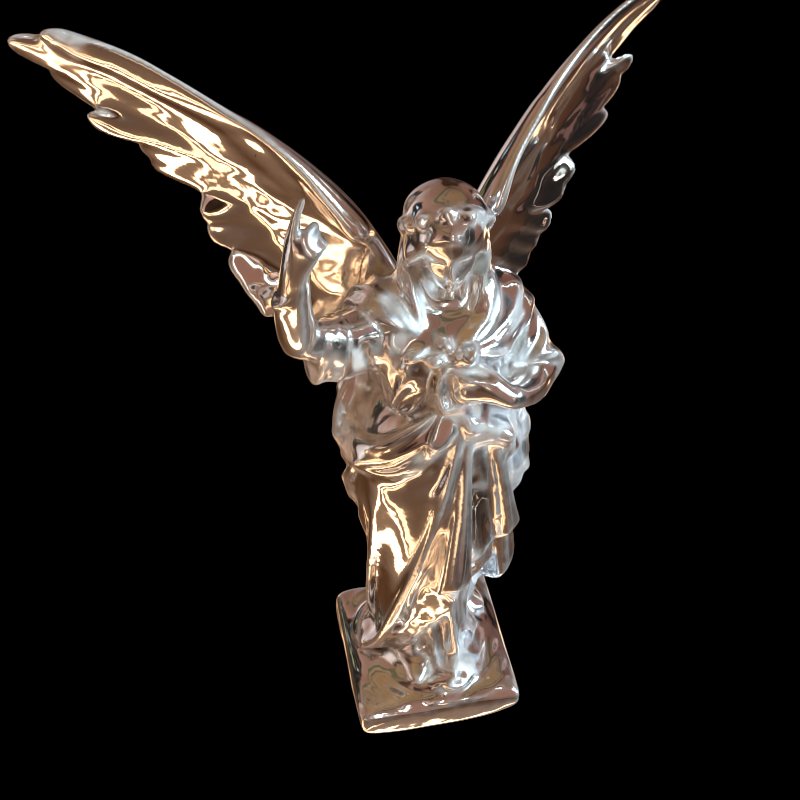}  &
    \includegraphics[trim={0 0 0 0}, clip, height=0.133\textwidth]{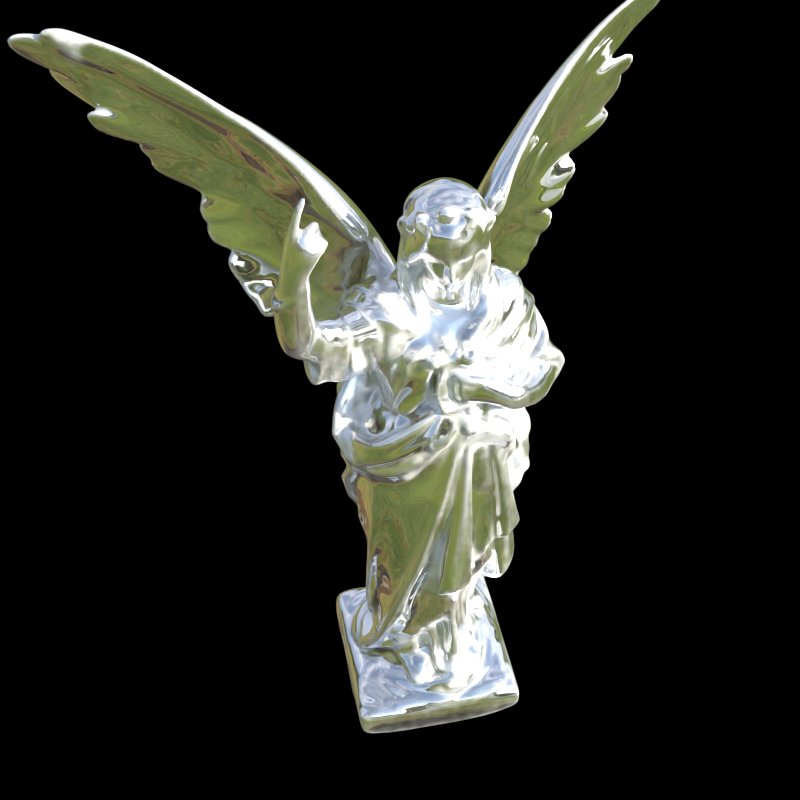}  &
    \includegraphics[trim={0 0 0 0}, clip, height=0.133\textwidth]{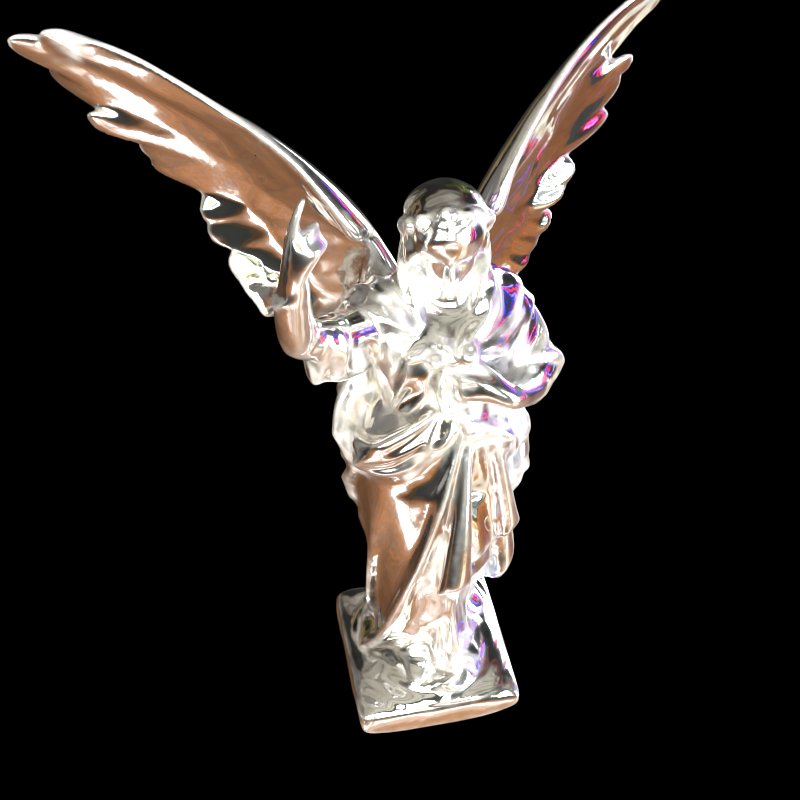} \\ \hline
    \rotatebox{90}{\parbox{0.133\textwidth}{\centering Ours }} &
    \includegraphics[trim={0 0 0 0}, clip, height=0.133\textwidth]{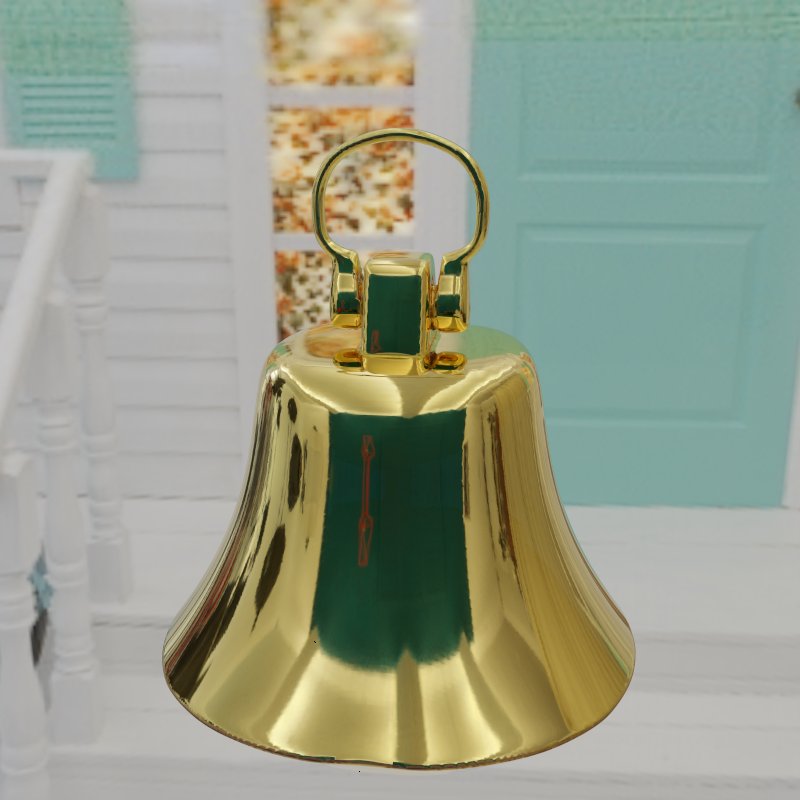} &
    \includegraphics[trim={0 0 0 0}, clip, height=0.133\textwidth]{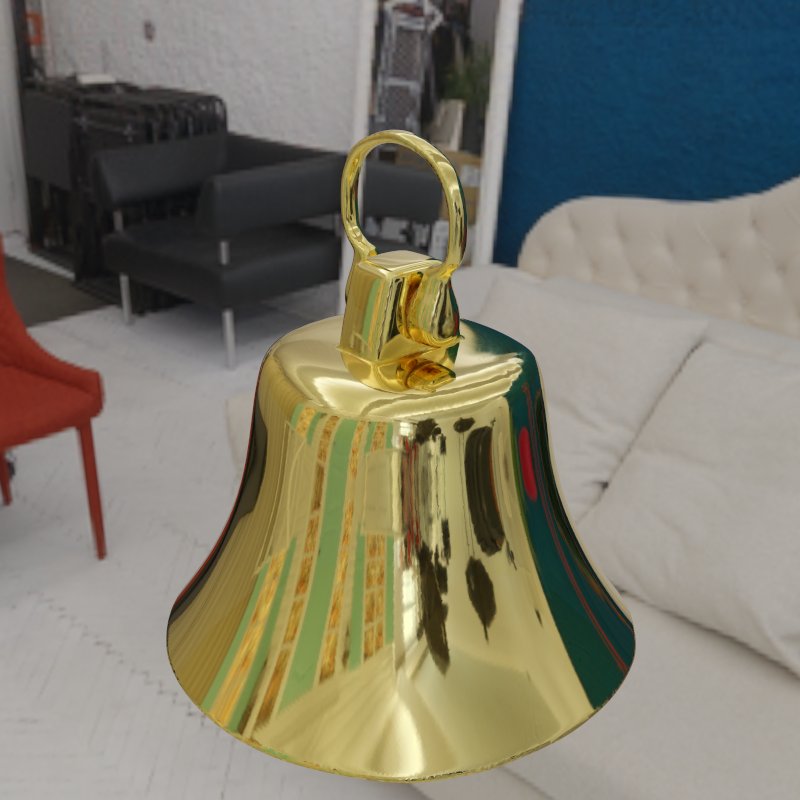} &
    \multicolumn{2}{c}{\includegraphics[trim={0 0 0 0}, clip, height=0.133\textwidth]{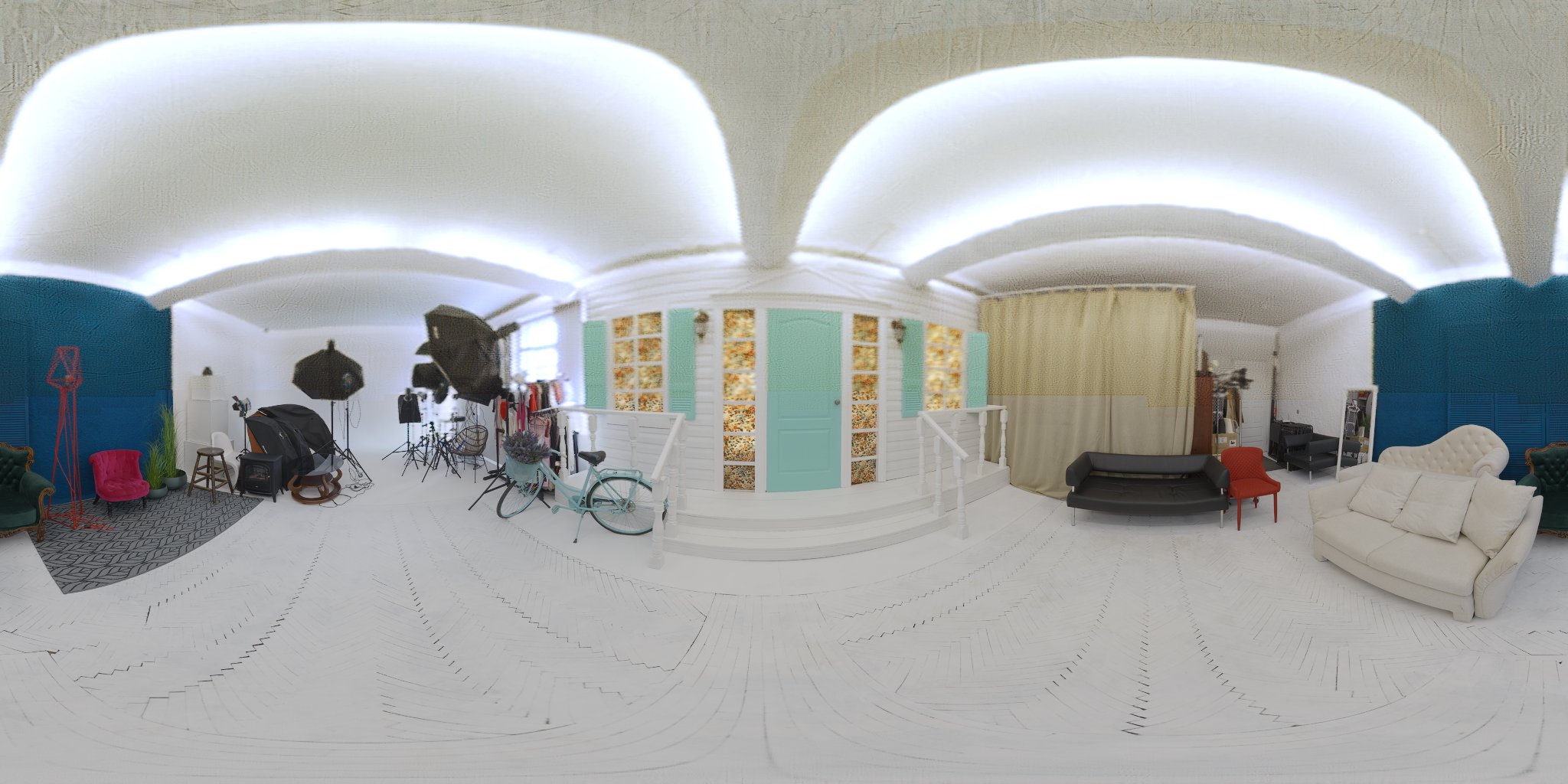}  } &
    \includegraphics[trim={0 0 0 0}, clip, height=0.133\textwidth]{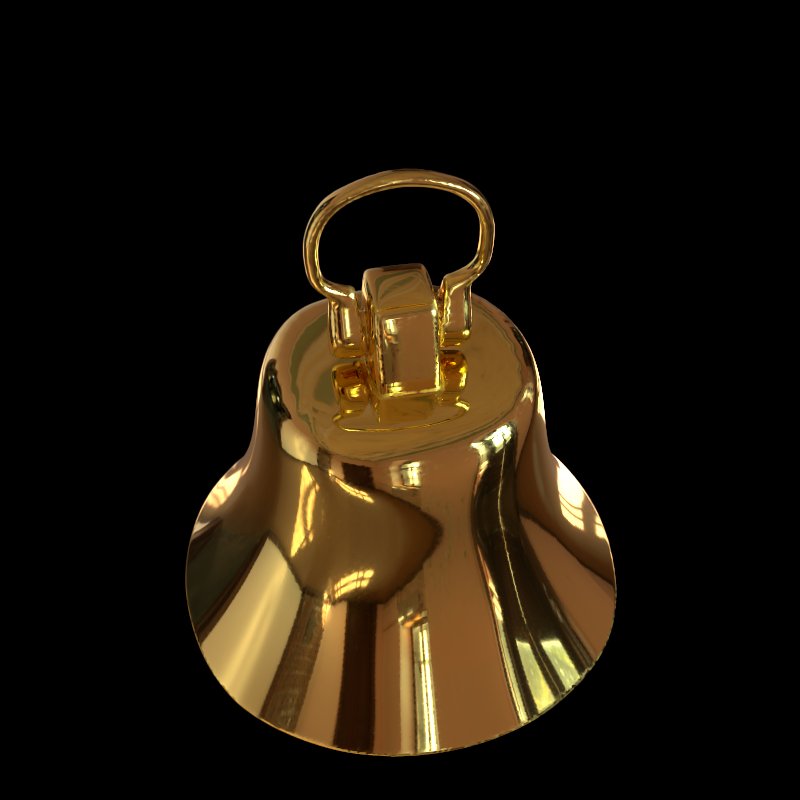}  &
    \includegraphics[trim={0 0 0 0}, clip, height=0.133\textwidth]{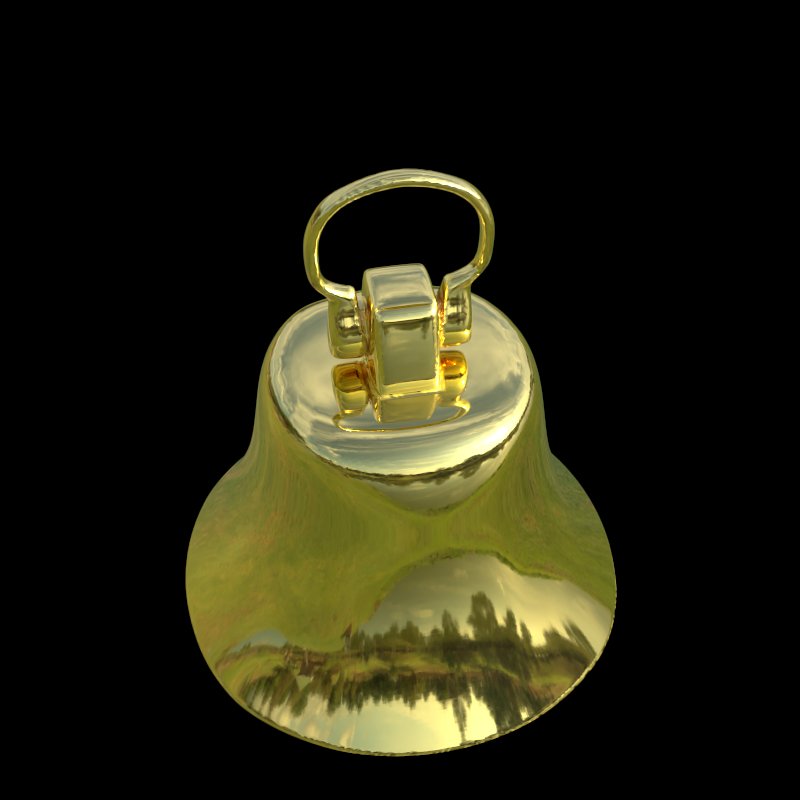}  &
    \includegraphics[trim={0 0 0 0}, clip, height=0.133\textwidth]{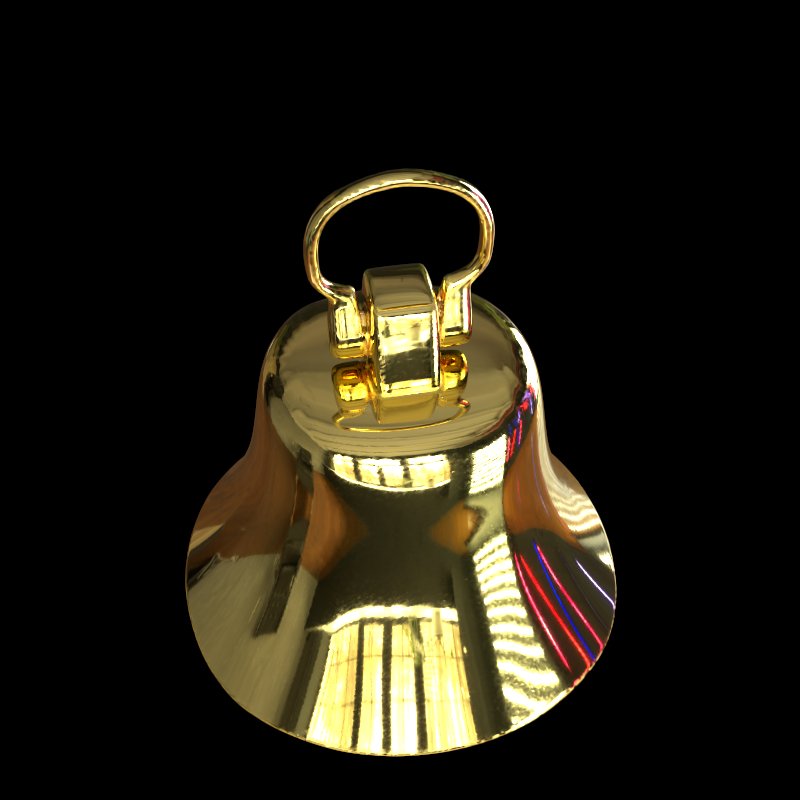} \\
    \rotatebox{90}{\parbox{0.133\textwidth}{\centering NeRO }} &
    \includegraphics[trim={0 0 0 0}, clip, height=0.133\textwidth]{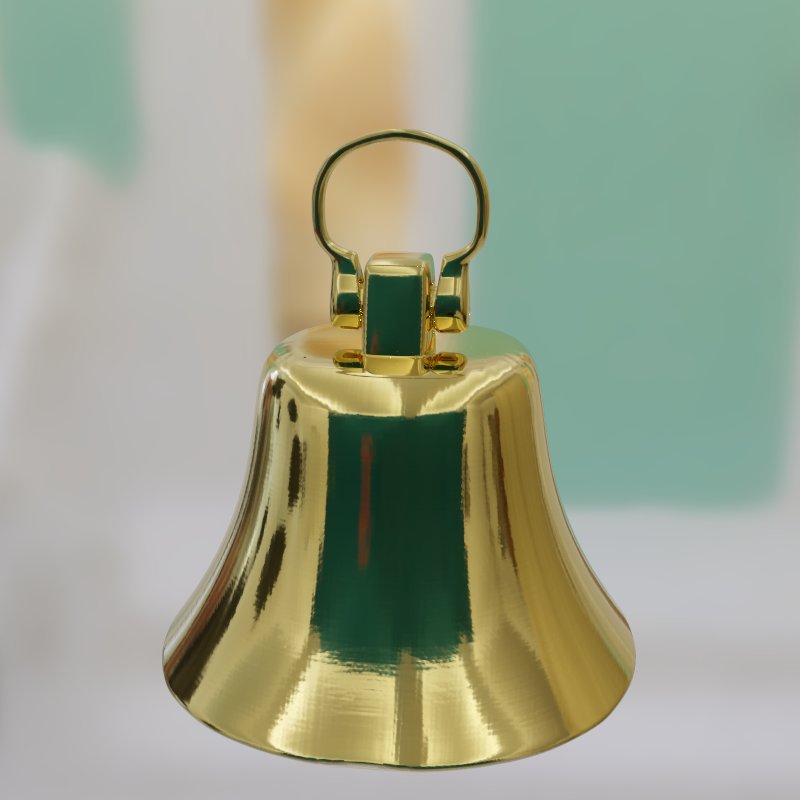} &
    \includegraphics[trim={0 0 0 0}, clip, height=0.133\textwidth]{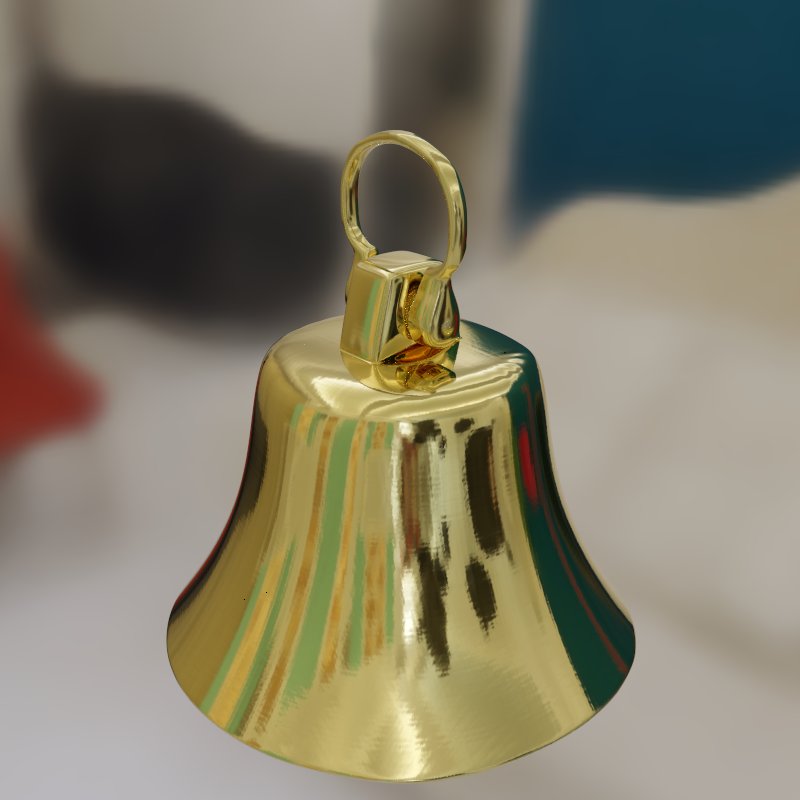} &
    \multicolumn{2}{c}{\includegraphics[trim={0 0 0 0}, clip, height=0.133\textwidth]{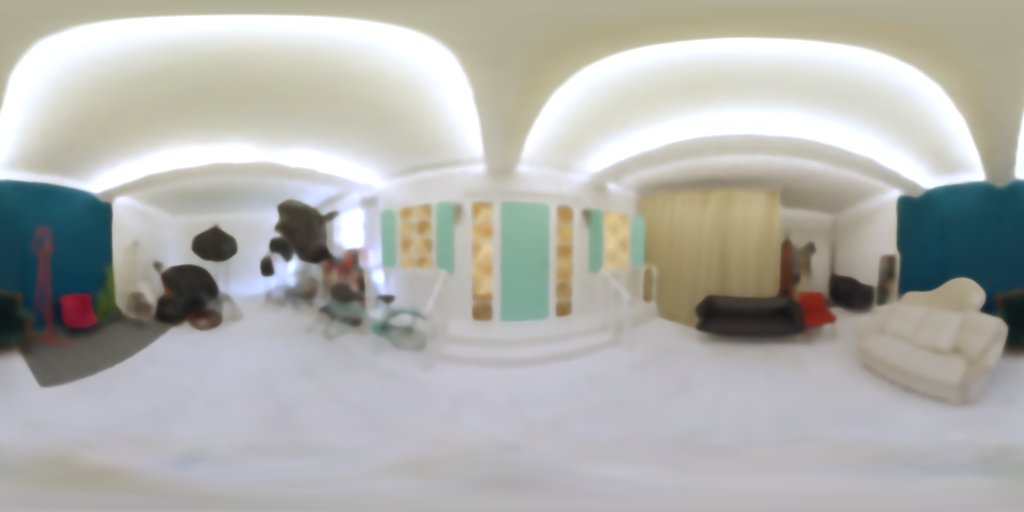}  } &
    \includegraphics[trim={0 0 0 0}, clip, height=0.133\textwidth]{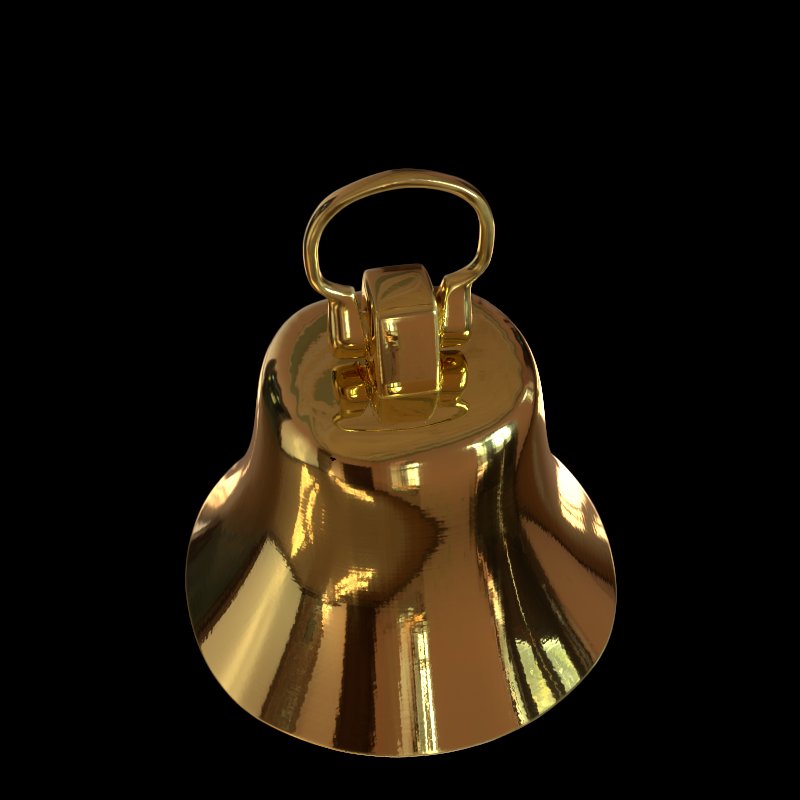}  &
    \includegraphics[trim={0 0 0 0}, clip, height=0.133\textwidth]{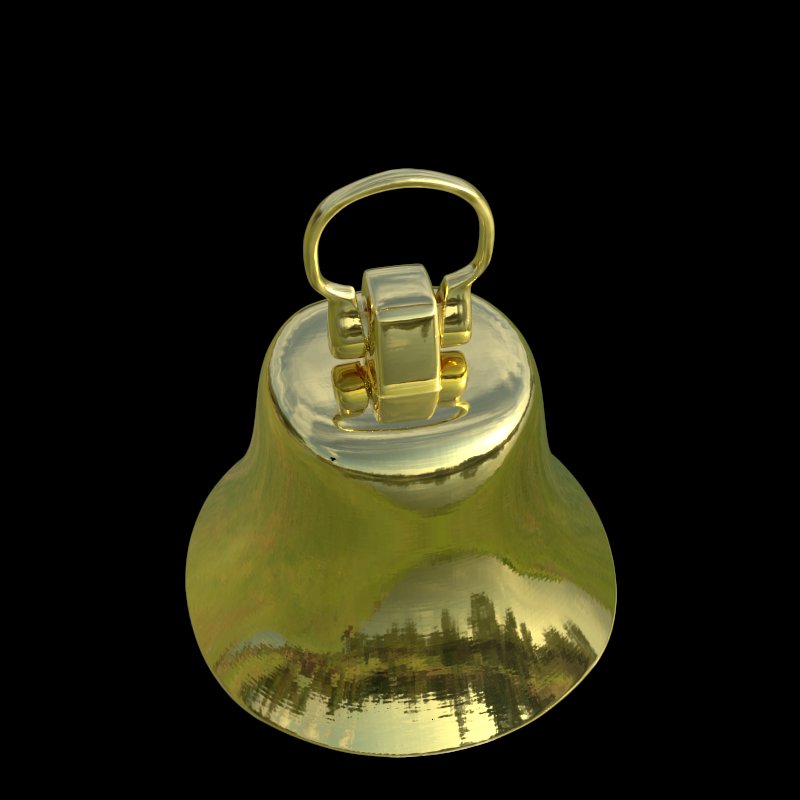}  &
    \includegraphics[trim={0 0 0 0}, clip, height=0.133\textwidth]{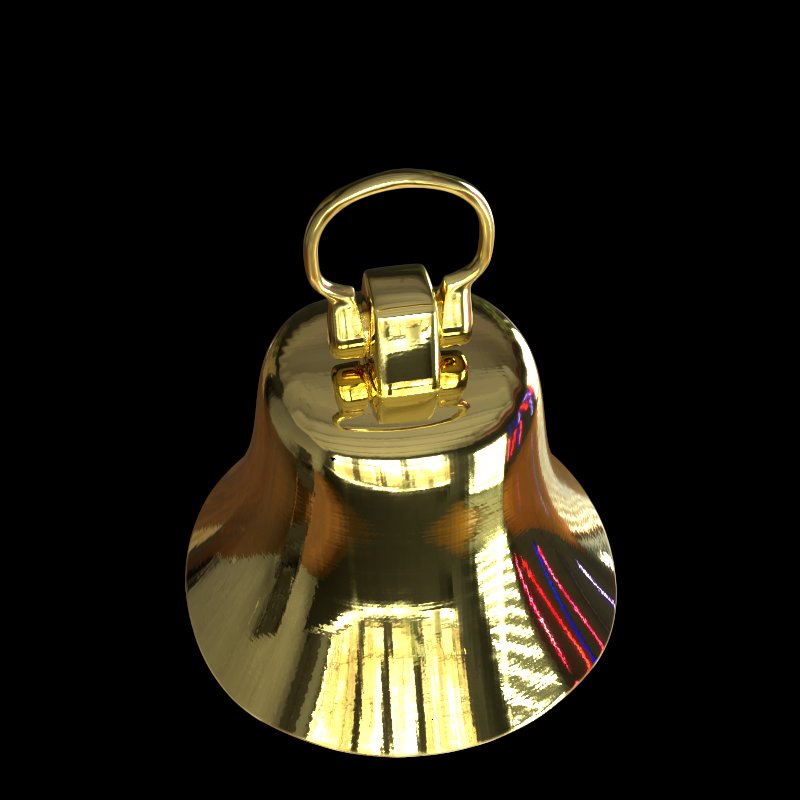} \\ \hline
    \rotatebox{90}{\parbox{0.133\textwidth}{\centering Ours }} &
    \includegraphics[trim={0 0 0 0}, clip, height=0.133\textwidth]{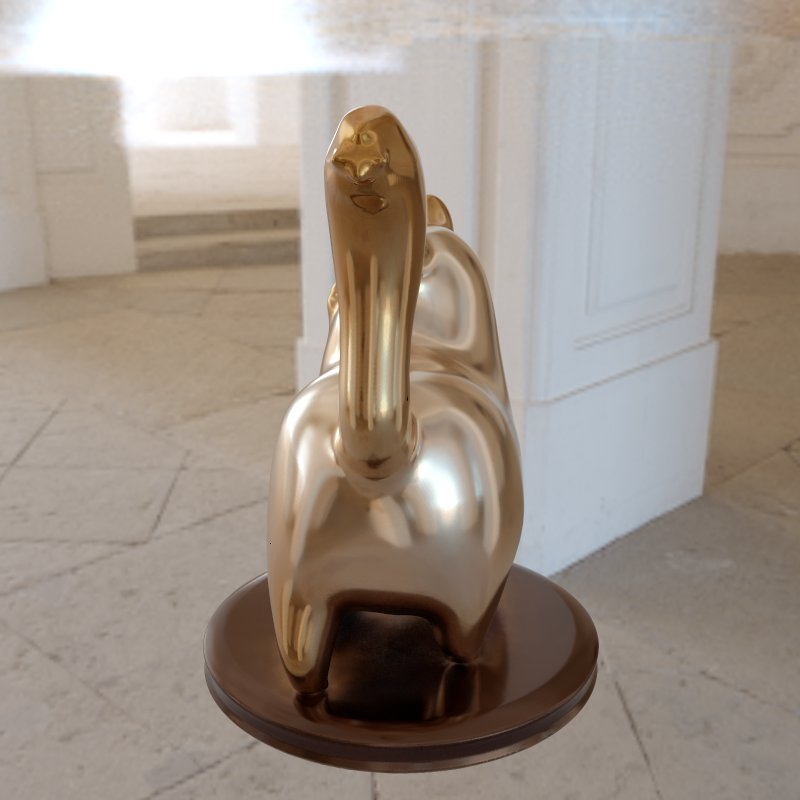} &
    \includegraphics[trim={0 0 0 0}, clip, height=0.133\textwidth]{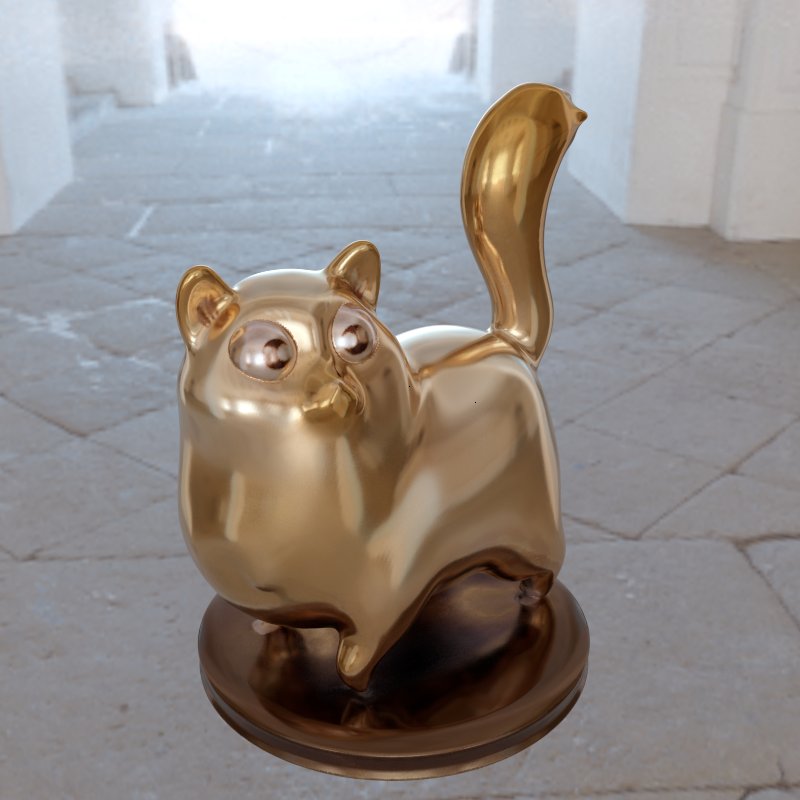} &
    \multicolumn{2}{c}{\includegraphics[trim={0 0 0 0}, clip, height=0.133\textwidth]{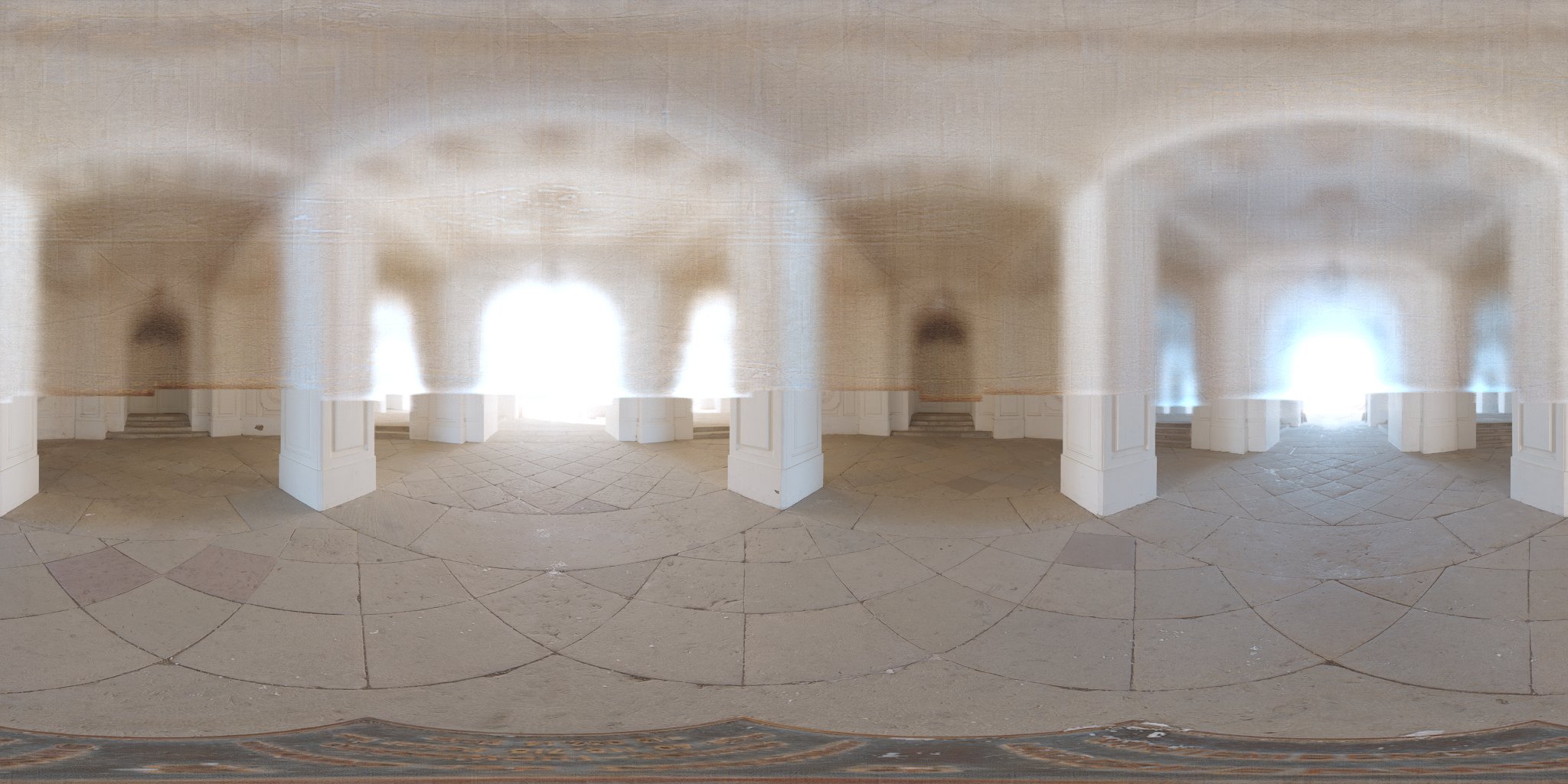}  } &
    \includegraphics[trim={0 0 0 0}, clip, height=0.133\textwidth]{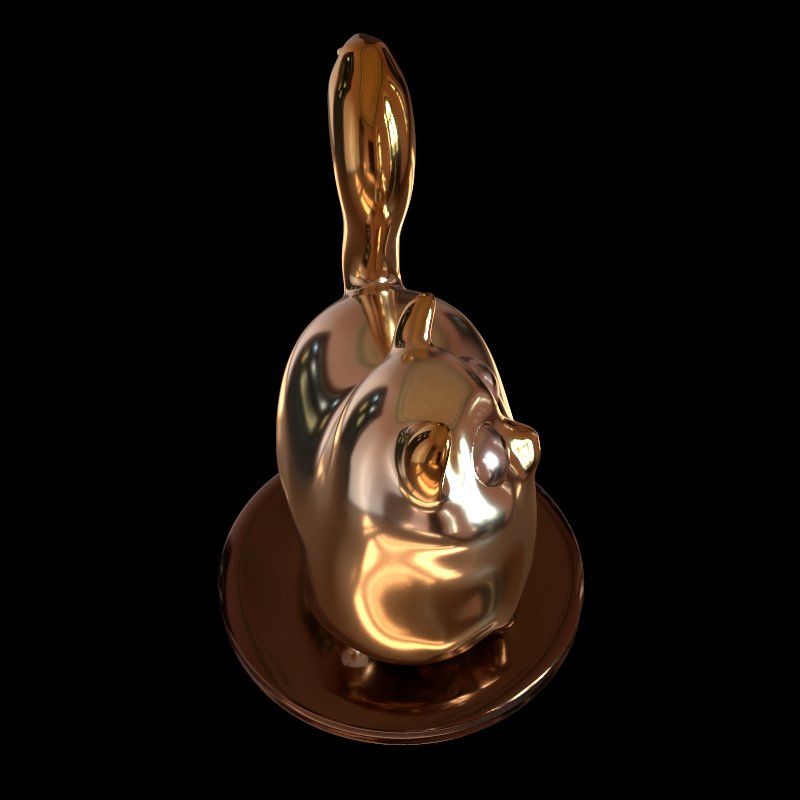}  &
    \includegraphics[trim={0 0 0 0}, clip, height=0.133\textwidth]{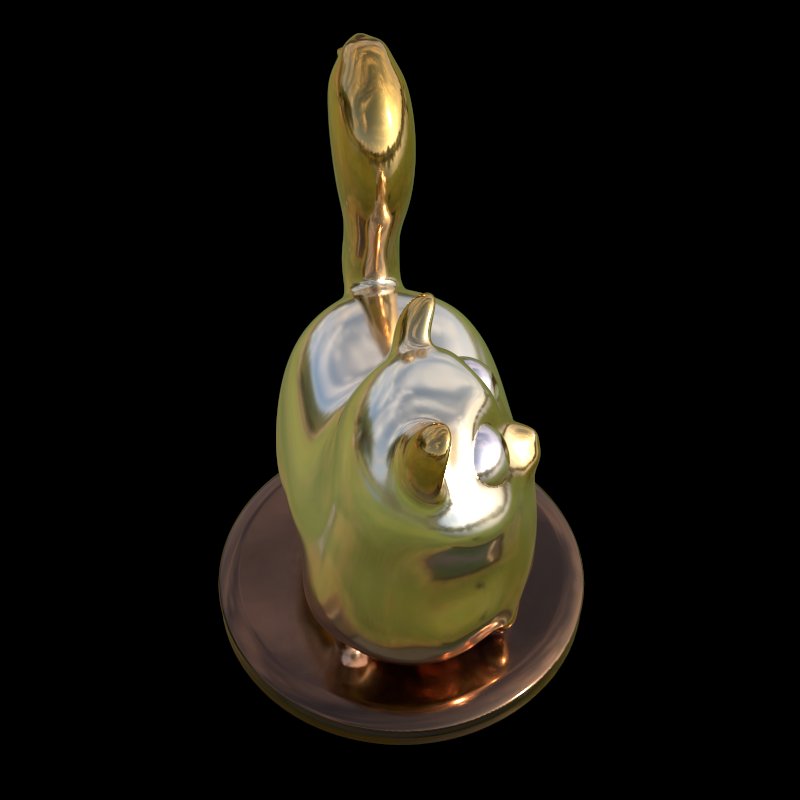}  &
    \includegraphics[trim={0 0 0 0}, clip, height=0.133\textwidth]{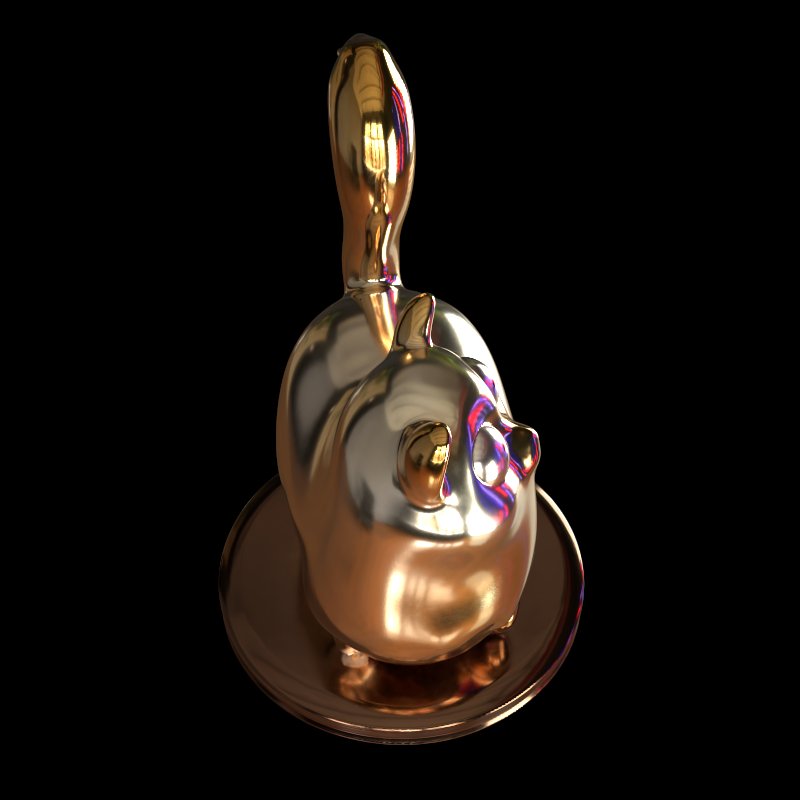} \\
    \rotatebox{90}{\parbox{0.133\textwidth}{\centering NeRO }} &
    \includegraphics[trim={0 0 0 0}, clip, height=0.133\textwidth]{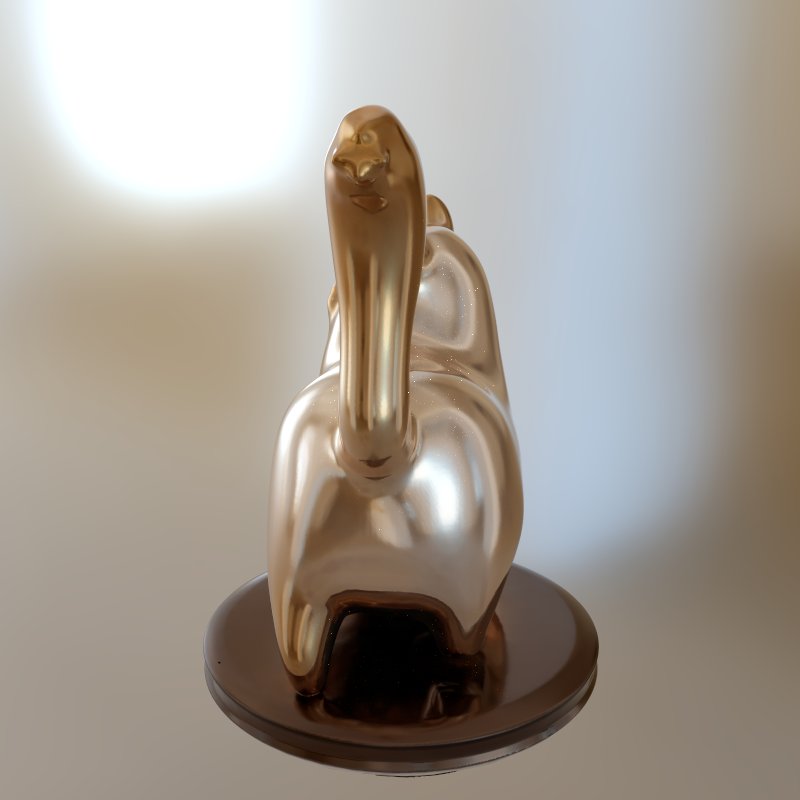} &
    \includegraphics[trim={0 0 0 0}, clip, height=0.133\textwidth]{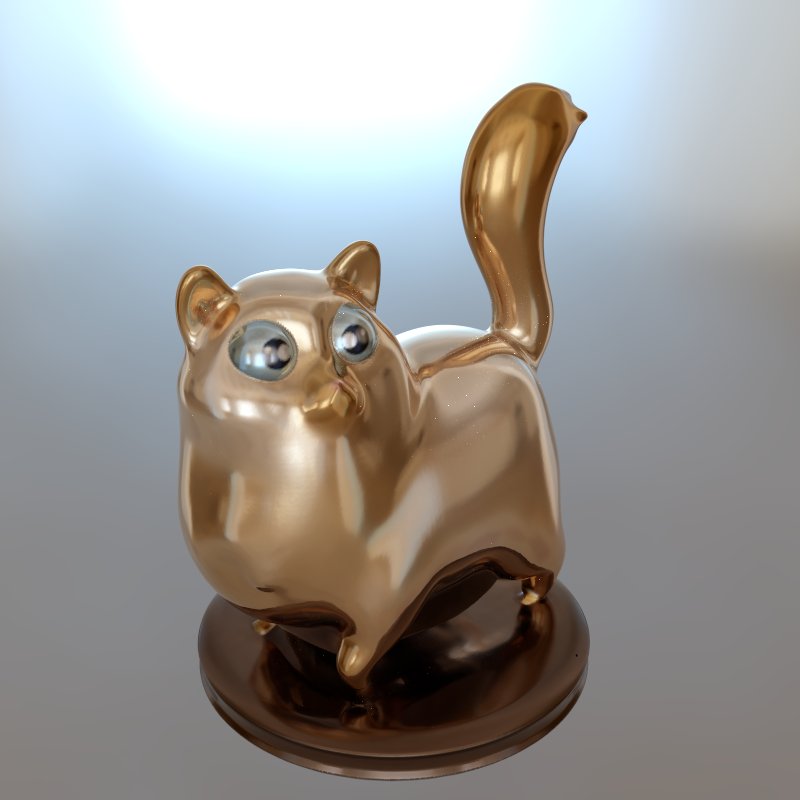} &
    \multicolumn{2}{c}{\includegraphics[trim={0 0 0 0}, clip, height=0.133\textwidth]{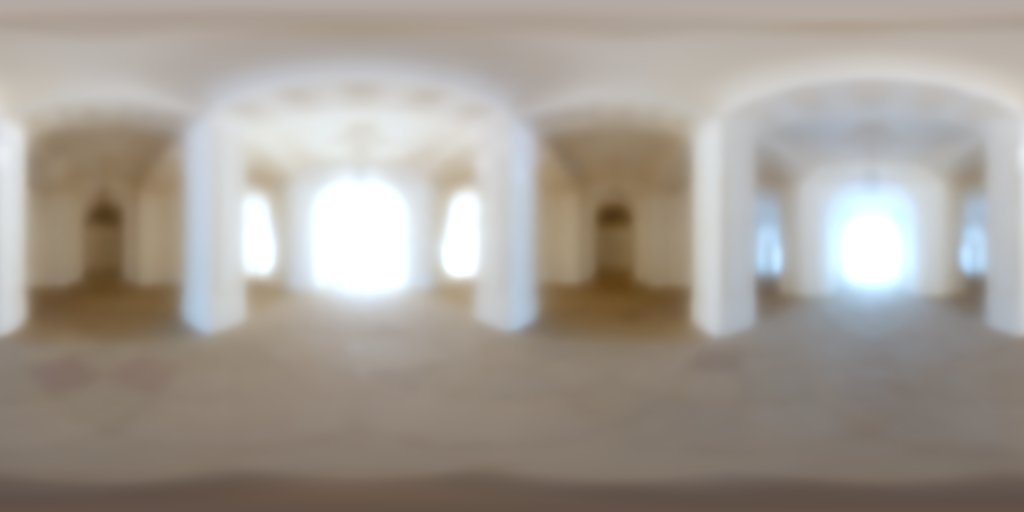}  } &
    \includegraphics[trim={0 0 0 0}, clip, height=0.133\textwidth]{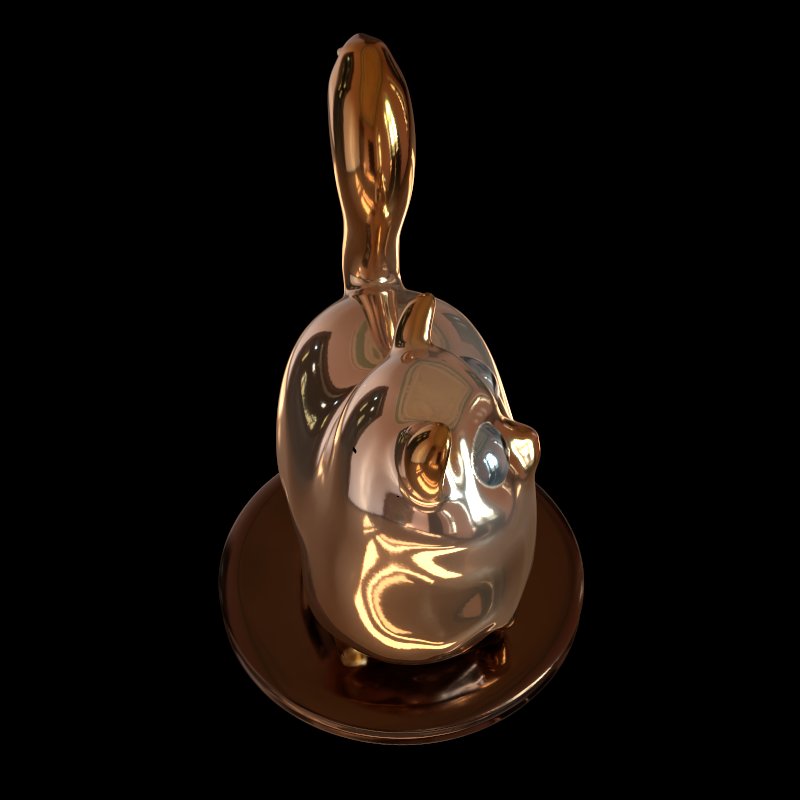}  &
    \includegraphics[trim={0 0 0 0}, clip, height=0.133\textwidth]{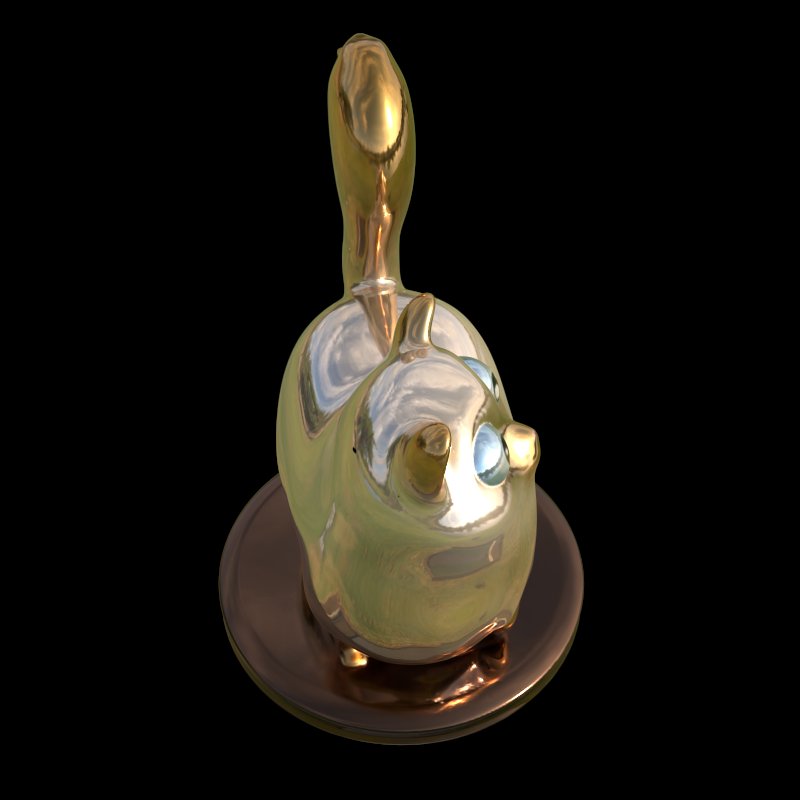}  &
    \includegraphics[trim={0 0 0 0}, clip, height=0.133\textwidth]{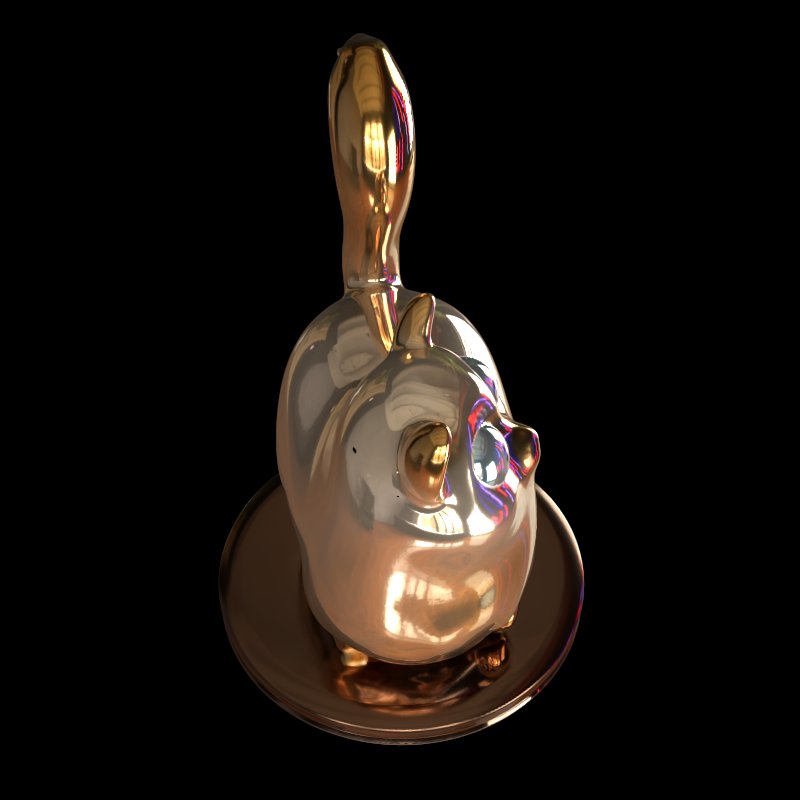} \\ \hline
    \rotatebox{90}{\parbox{0.133\textwidth}{\centering Ours }} &
    \includegraphics[trim={0 0 0 0}, clip, height=0.133\textwidth]{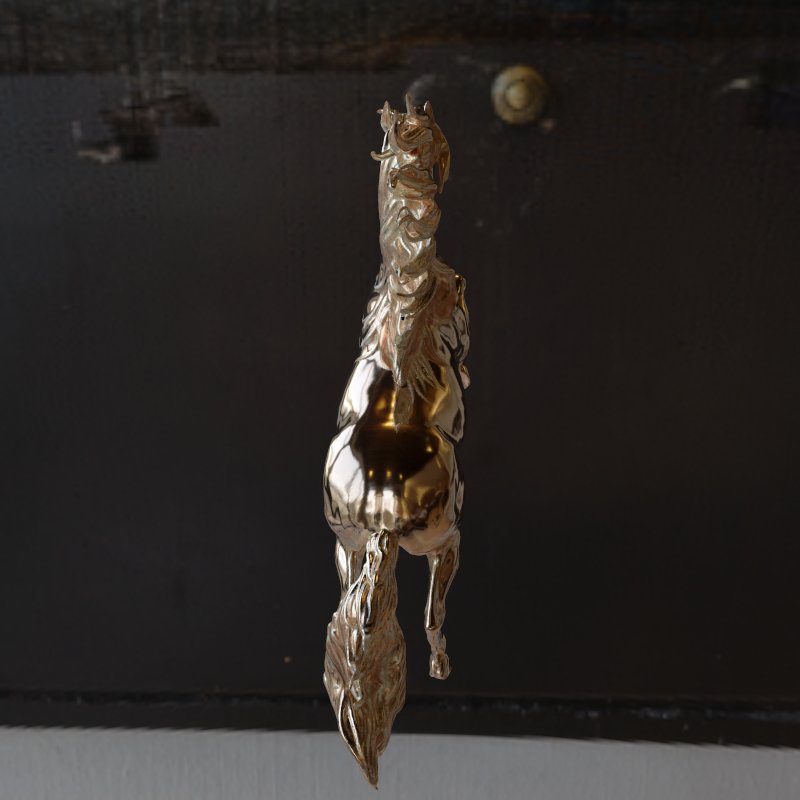} &
    \includegraphics[trim={0 0 0 0}, clip, height=0.133\textwidth]{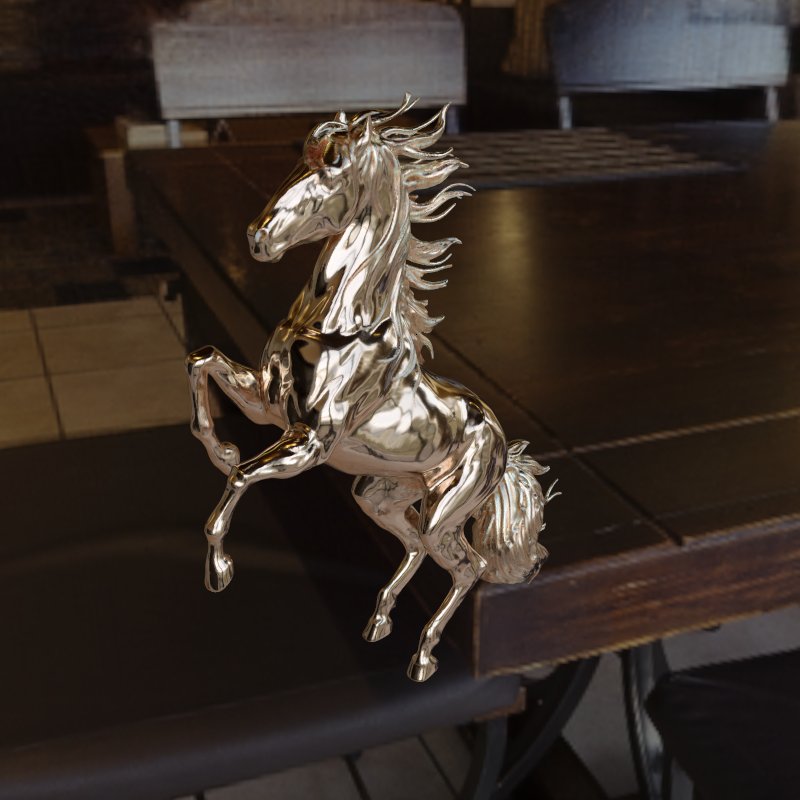} &
    \multicolumn{2}{c}{\includegraphics[trim={0 0 0 0}, clip, height=0.133\textwidth]{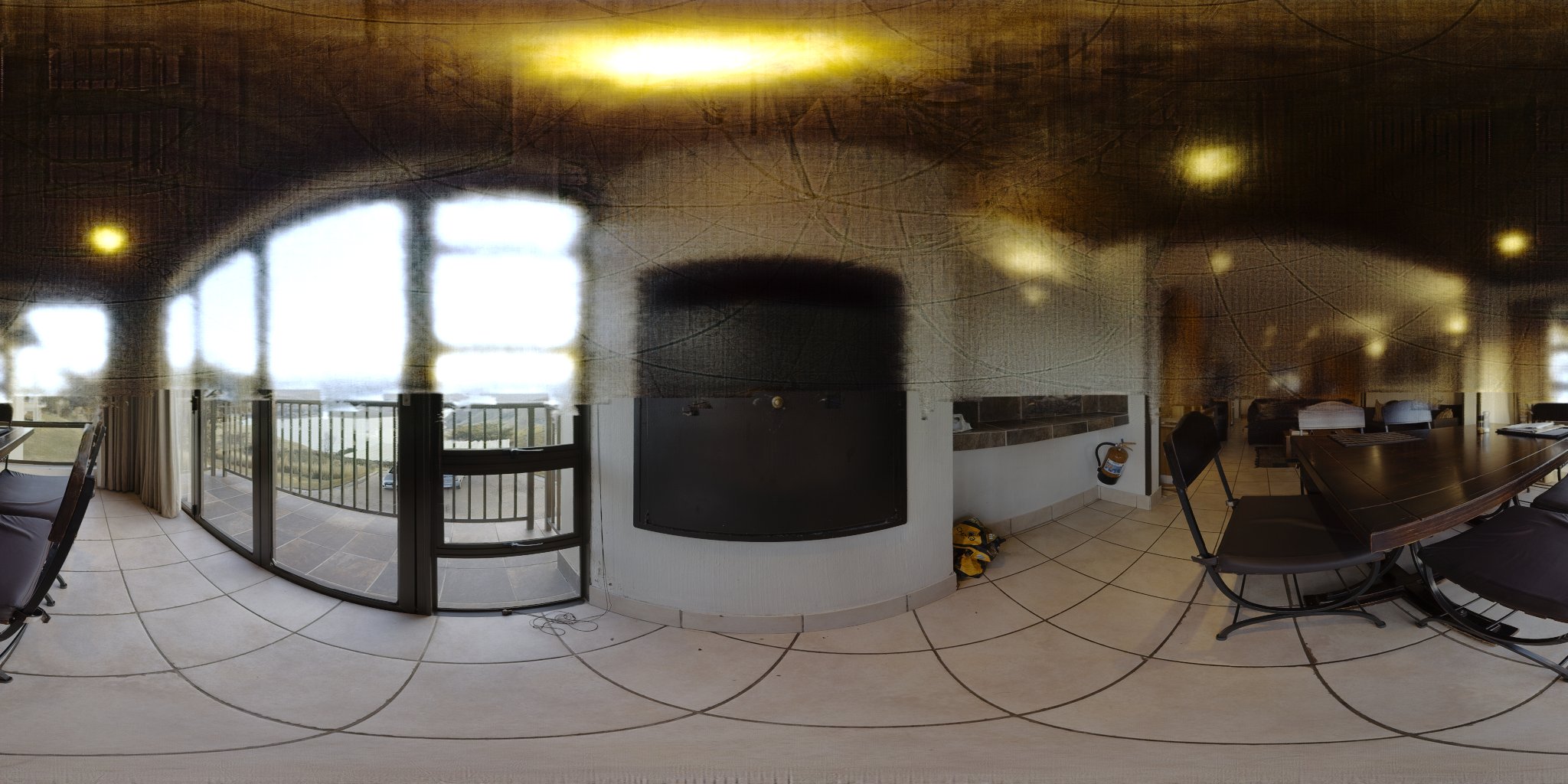}  } &
    \includegraphics[trim={0 0 0 0}, clip, height=0.133\textwidth]{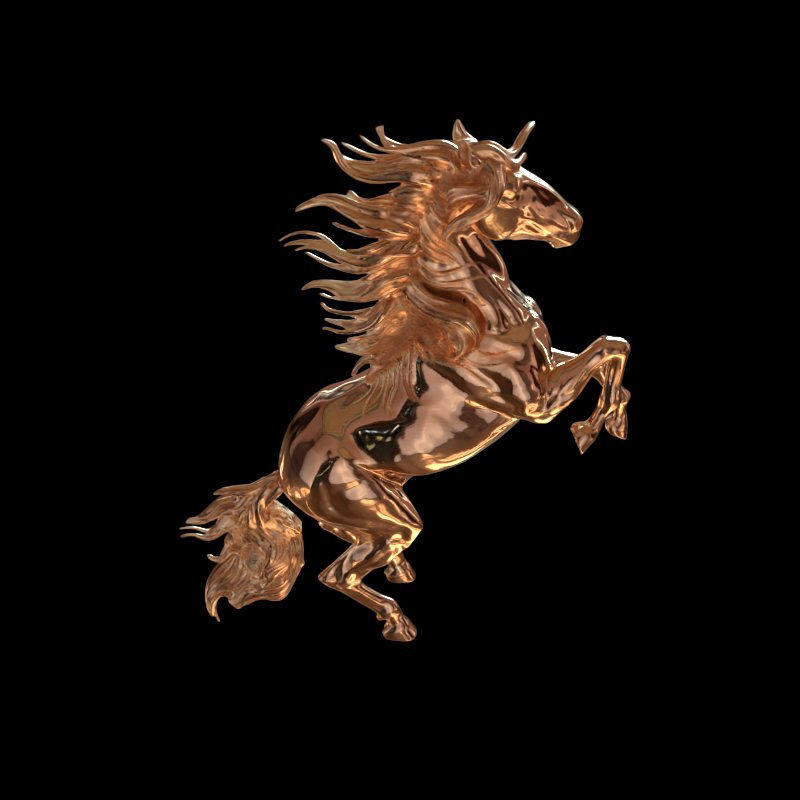}  &
    \includegraphics[trim={0 0 0 0}, clip, height=0.133\textwidth]{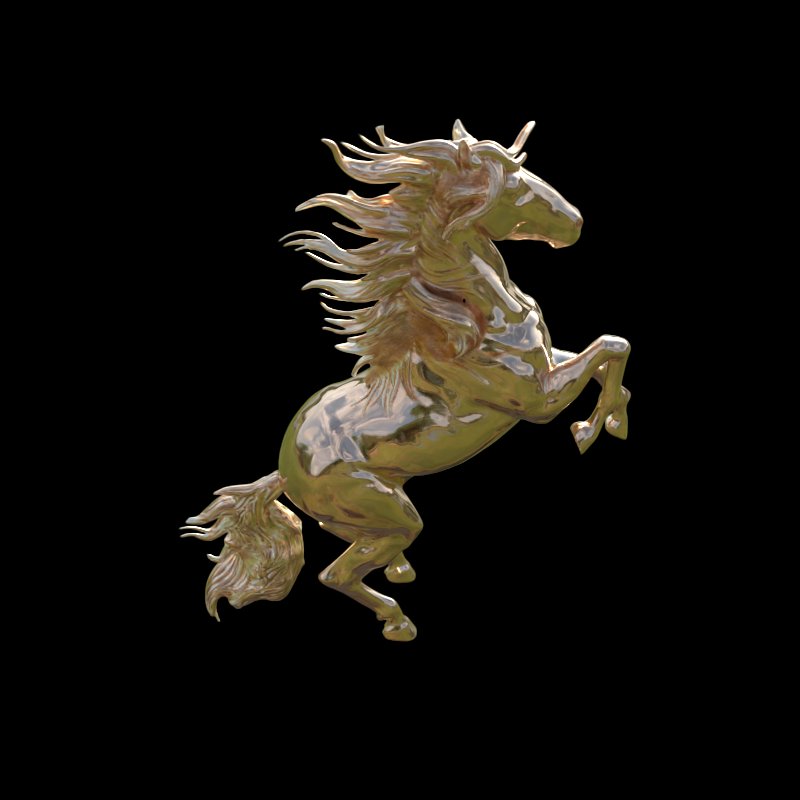}  &
    \includegraphics[trim={0 0 0 0}, clip, height=0.133\textwidth]{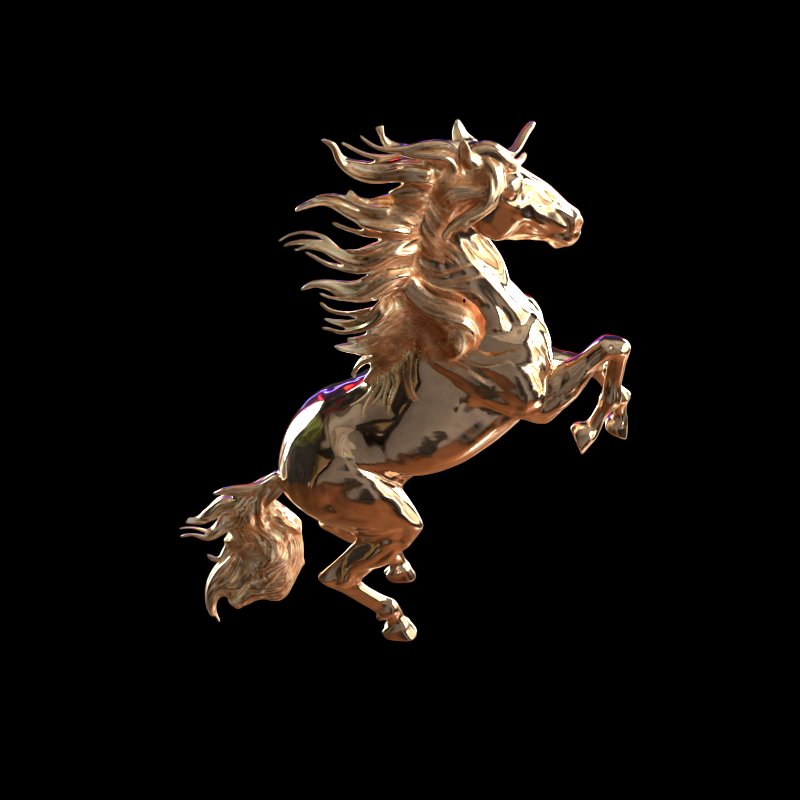} \\
    \rotatebox{90}{\parbox{0.133\textwidth}{\centering NeRO }} &
    \includegraphics[trim={0 0 0 0}, clip, height=0.133\textwidth]{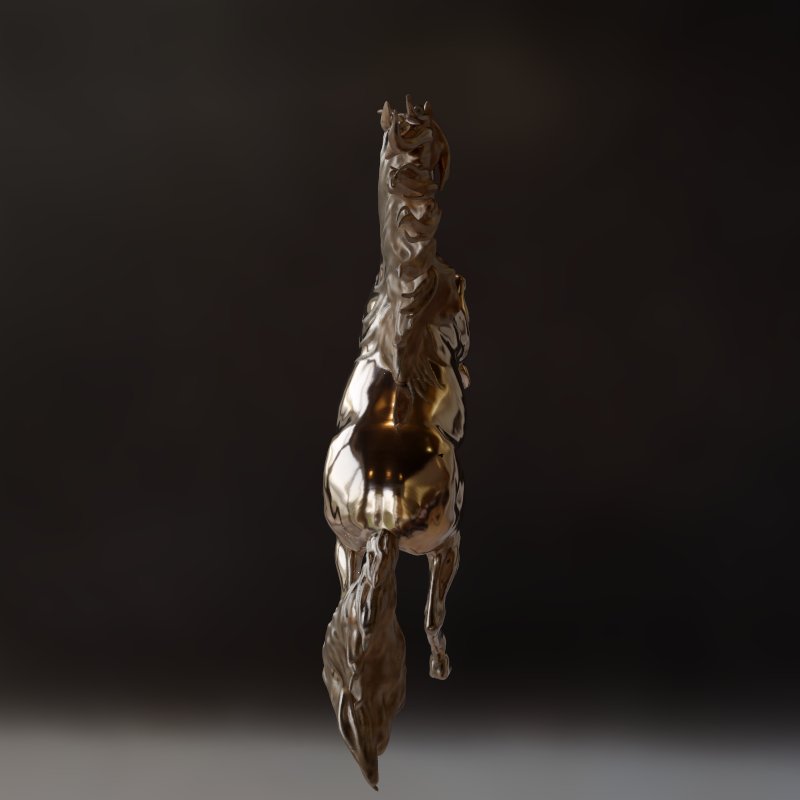} &
    \includegraphics[trim={0 0 0 0}, clip, height=0.133\textwidth]{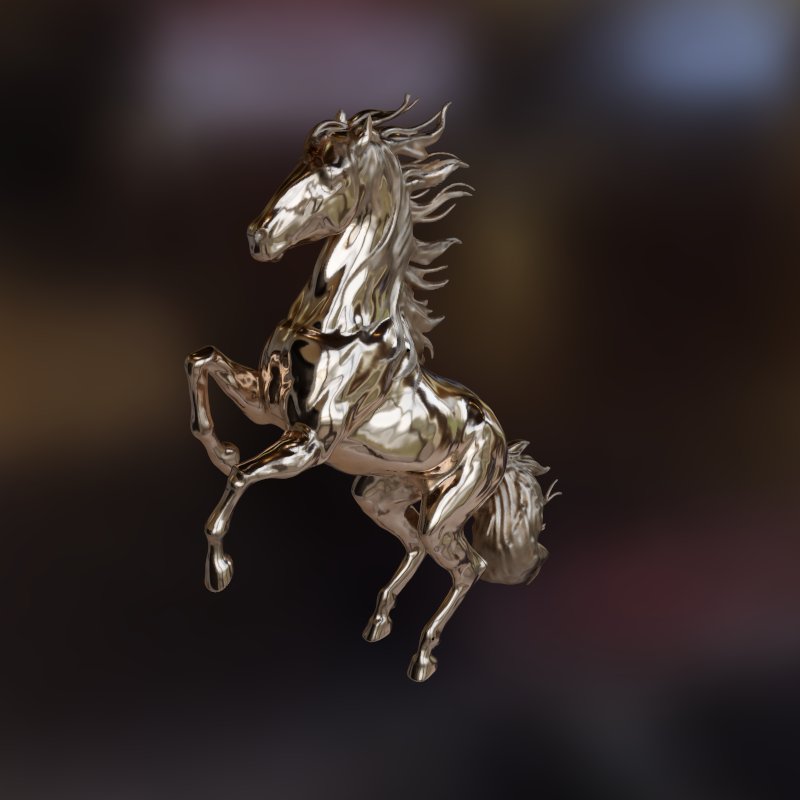} &
    \multicolumn{2}{c}{\includegraphics[trim={0 0 0 0}, clip, height=0.133\textwidth]{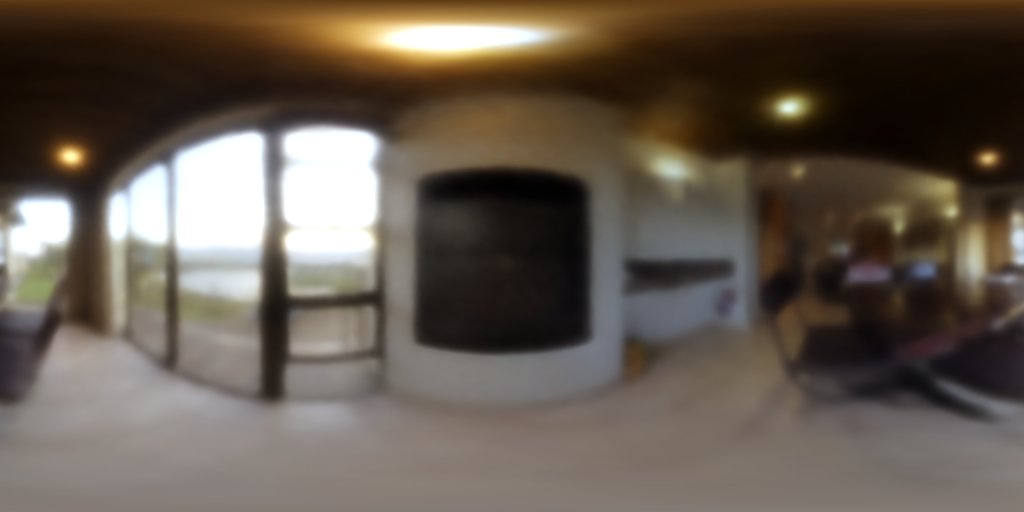}  } &
    \includegraphics[trim={0 0 0 0}, clip, height=0.133\textwidth]{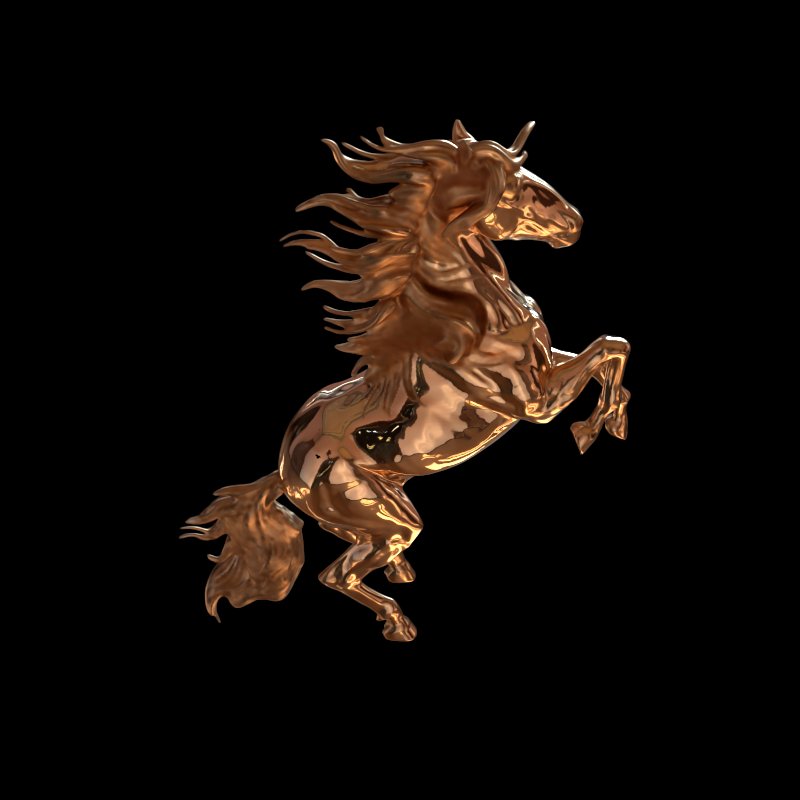}  &
    \includegraphics[trim={0 0 0 0}, clip, height=0.133\textwidth]{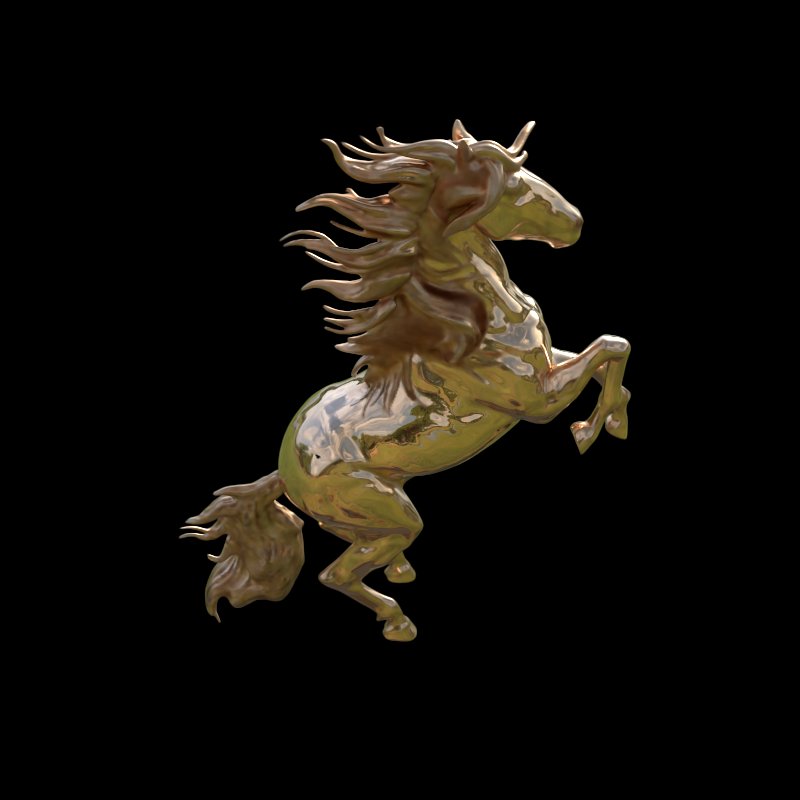}  &
    \includegraphics[trim={0 0 0 0}, clip, height=0.133\textwidth]{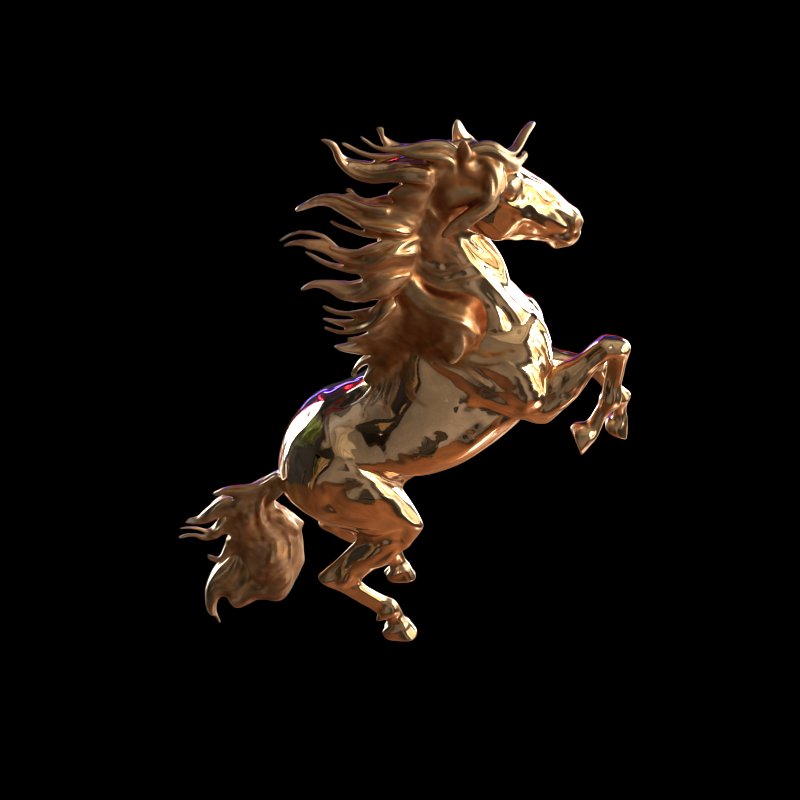} \\ \hline
    \end{tabular}
    }
    \caption{\textbf{Additional comparisons of our method against NeRO~\cite{liuNeRONeuralGeometry2023} on glossy synthetic data. } Please zoom in to better compare the results. }
    \label{fig:supp_nero1}
\end{figure*}

\begin{figure*}[t]
    \centering
    \resizebox{1.\textwidth}{!}{%
    \setlength{\tabcolsep}{1.5pt}
    \renewcommand{\arraystretch}{0.5}
    \begin{tabular}{ccc|cc|ccc}
    &  \multicolumn{2}{c}{Novel view rendering} & \multicolumn{2}{c}{Recovered lighting}  & \multicolumn{3}{c}{Relighting}\\
    \rotatebox{90}{\parbox{0.133\textwidth}{\centering Ours }} &
    \includegraphics[trim={0 0 0 0}, clip, height=0.133\textwidth]{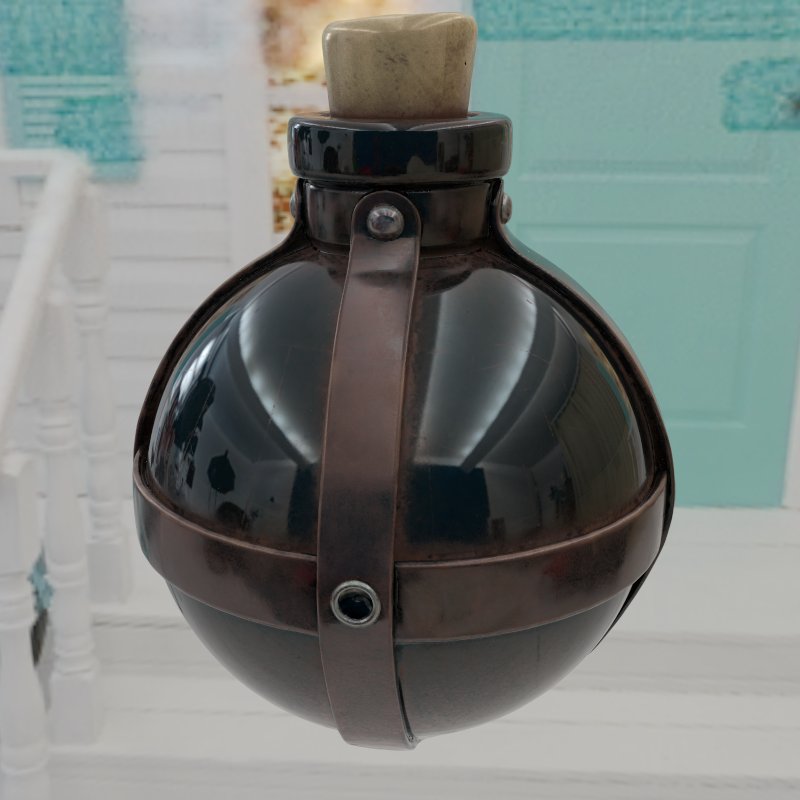} &
    \includegraphics[trim={0 0 0 0}, clip, height=0.133\textwidth]{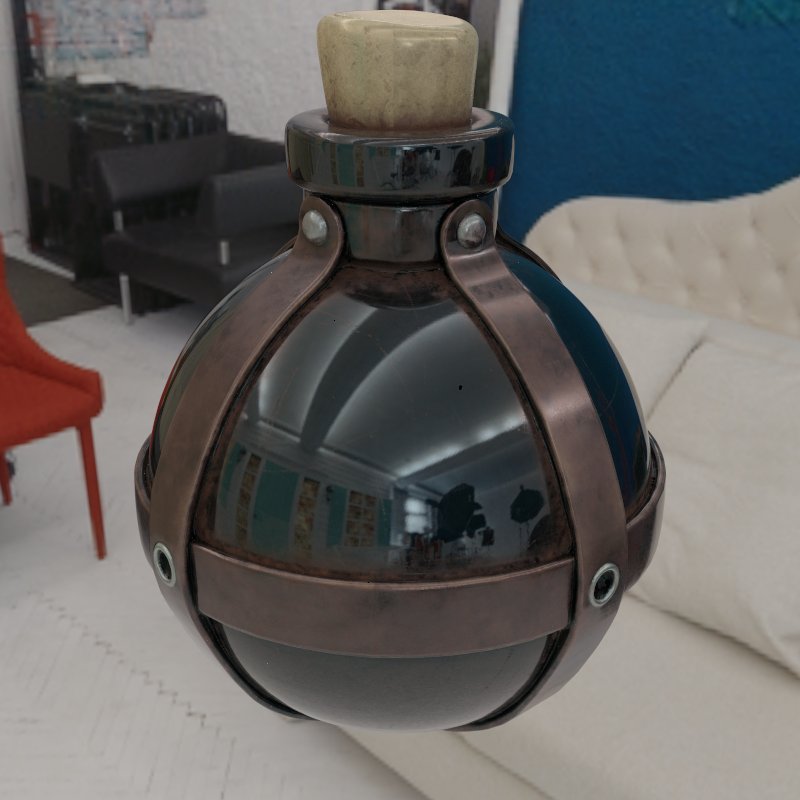} &
    \multicolumn{2}{c}{\includegraphics[trim={0 0 0 0}, clip, height=0.133\textwidth]{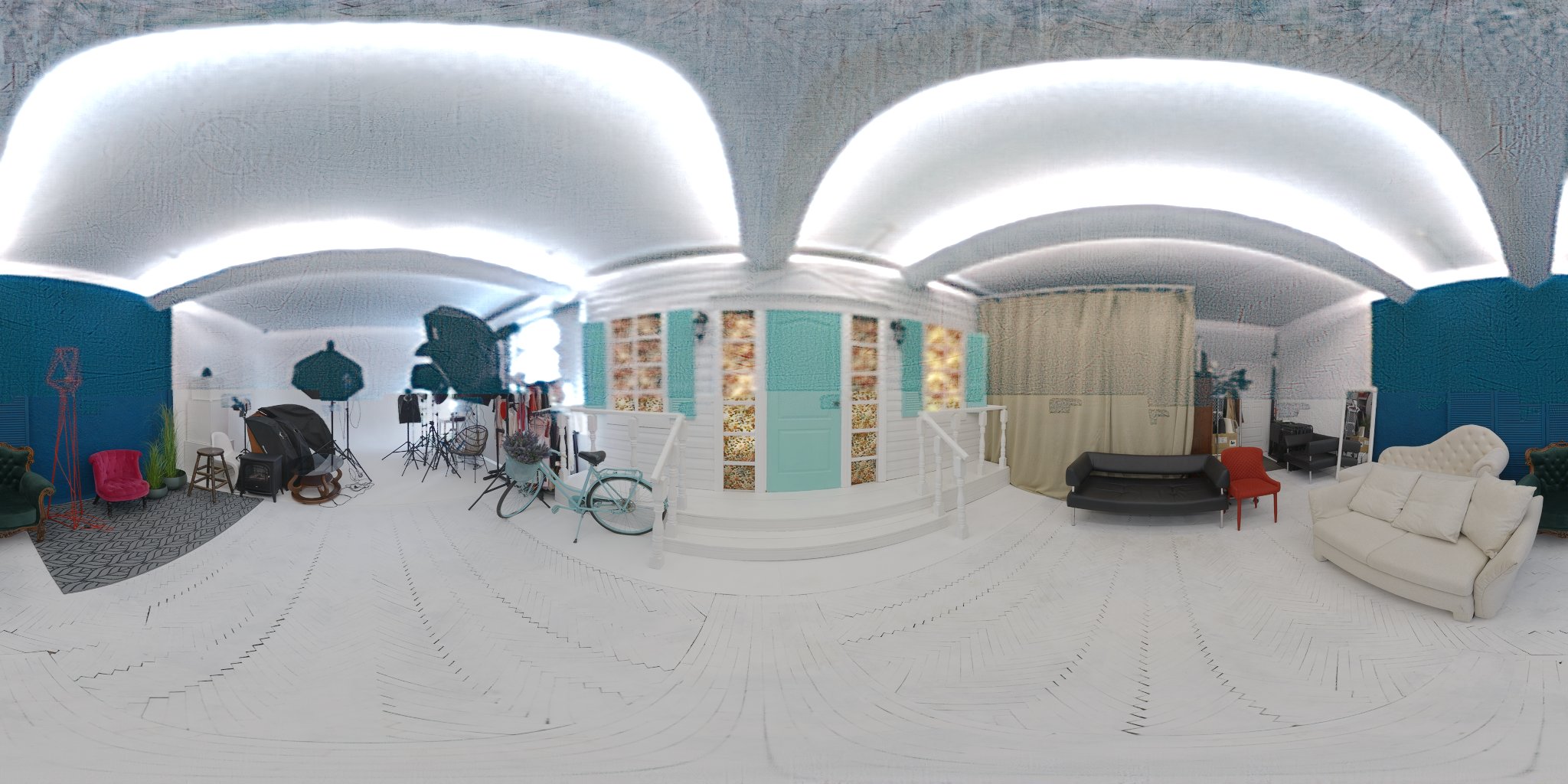}  } &
    \includegraphics[trim={0 0 0 0}, clip, height=0.133\textwidth]{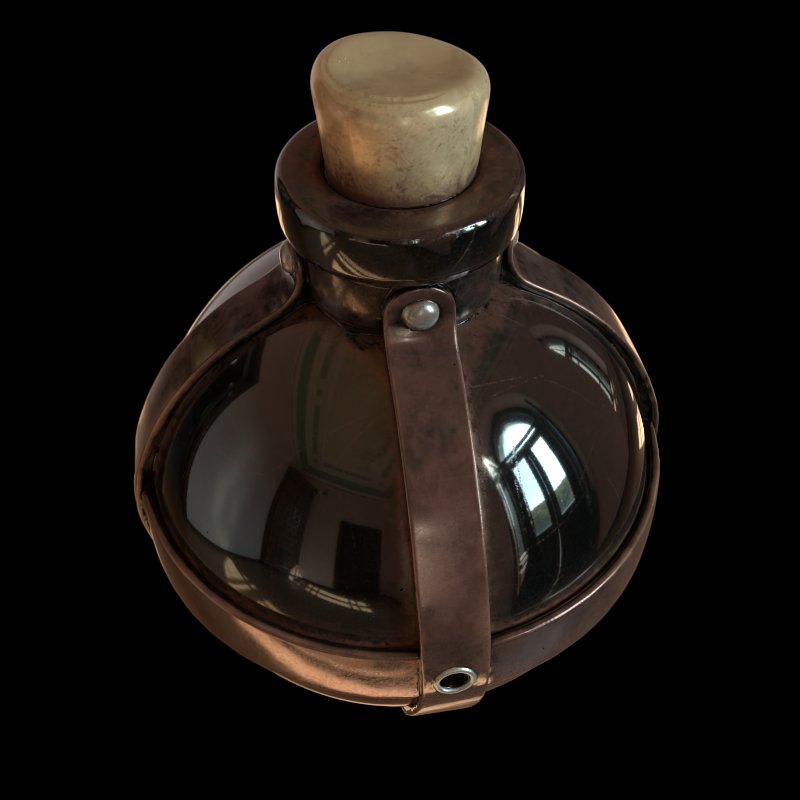}  &
    \includegraphics[trim={0 0 0 0}, clip, height=0.133\textwidth]{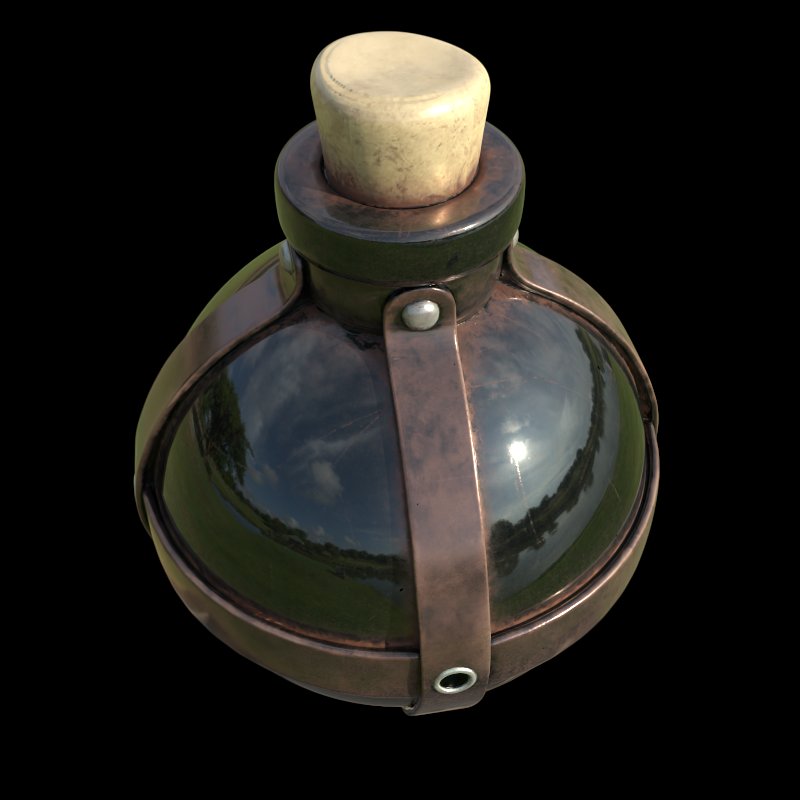}  &
    \includegraphics[trim={0 0 0 0}, clip, height=0.133\textwidth]{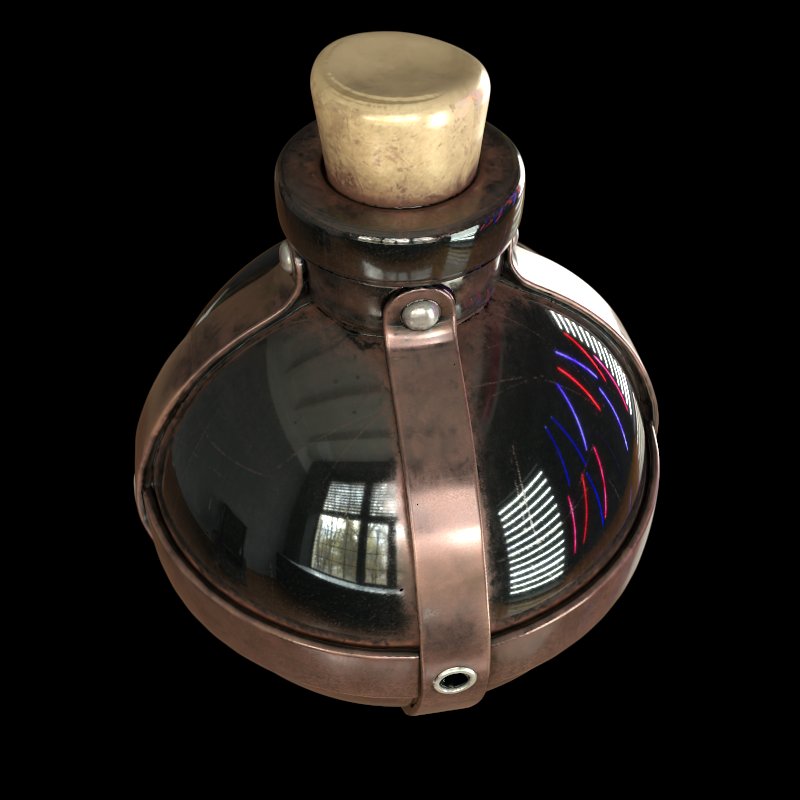} \\
    \rotatebox{90}{\parbox{0.133\textwidth}{\centering NeRO }} &
    \includegraphics[trim={0 0 0 0}, clip, height=0.133\textwidth]{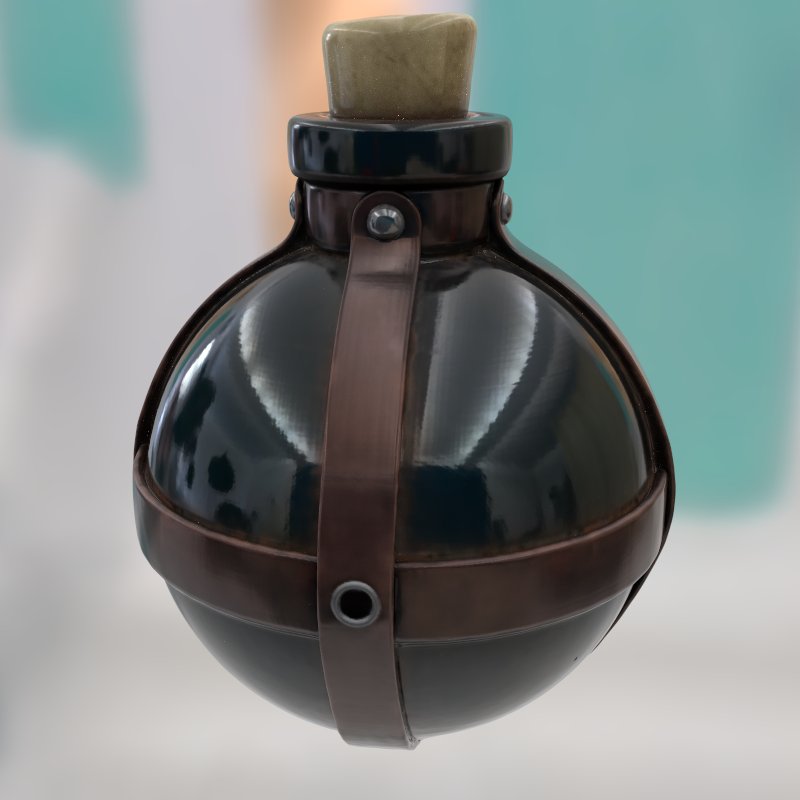} &
    \includegraphics[trim={0 0 0 0}, clip, height=0.133\textwidth]{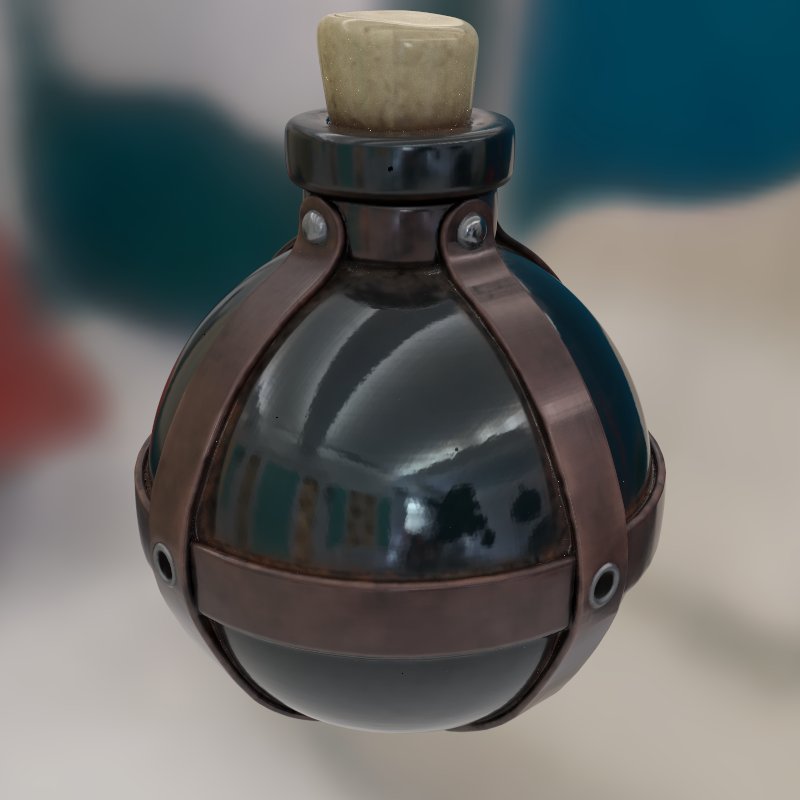} &
    \multicolumn{2}{c}{\includegraphics[trim={0 0 0 0}, clip, height=0.133\textwidth]{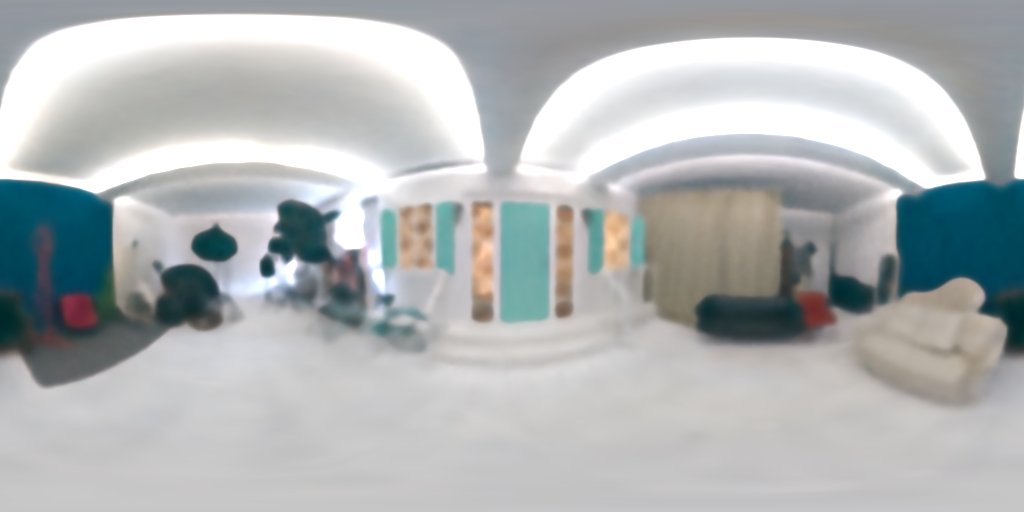}  } &
    \includegraphics[trim={0 0 0 0}, clip, height=0.133\textwidth]{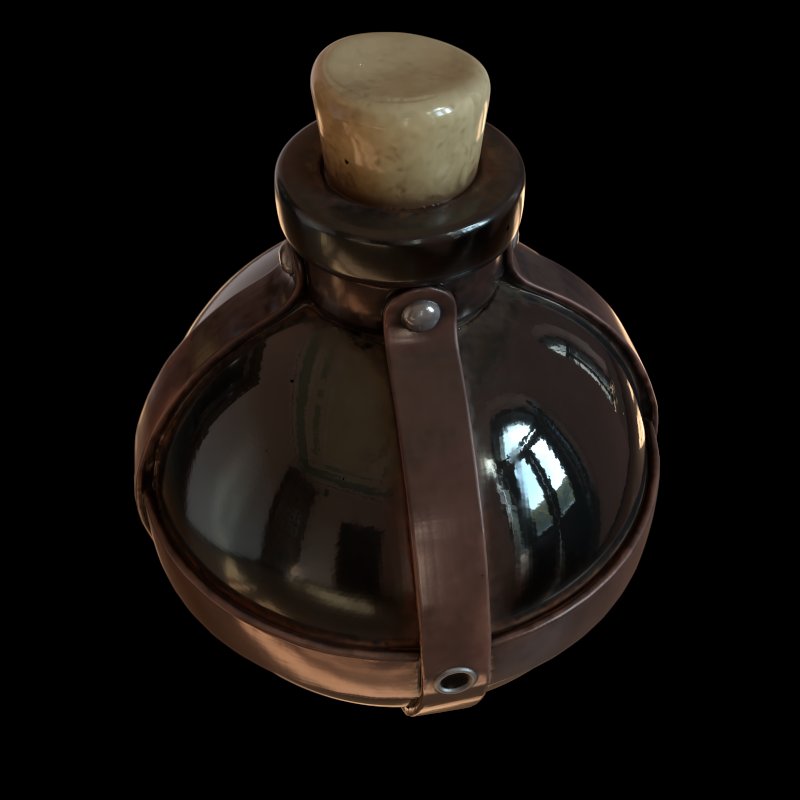}  &
    \includegraphics[trim={0 0 0 0}, clip, height=0.133\textwidth]{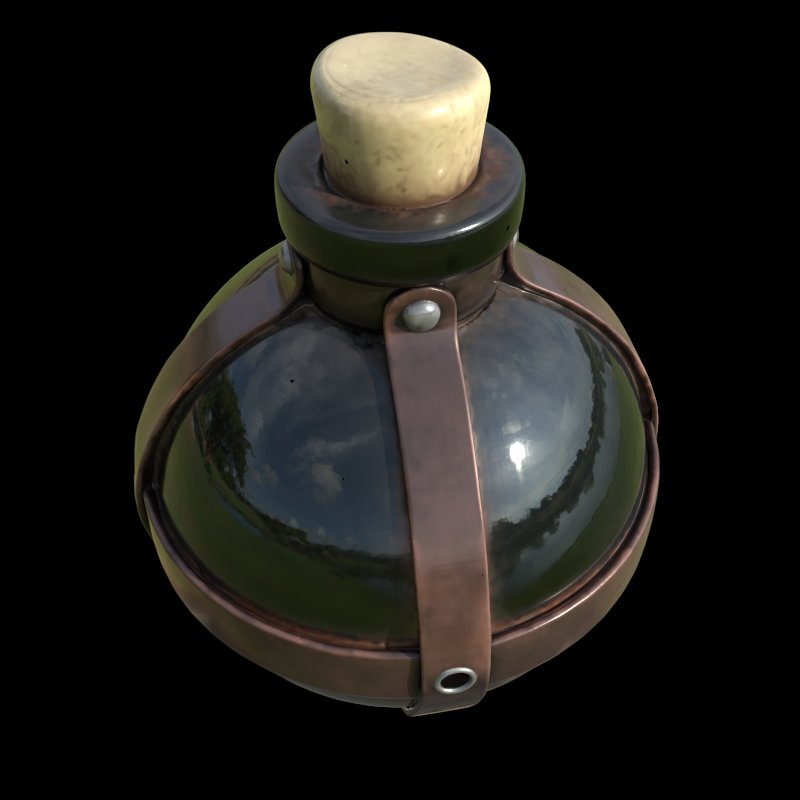}  &
    \includegraphics[trim={0 0 0 0}, clip, height=0.133\textwidth]{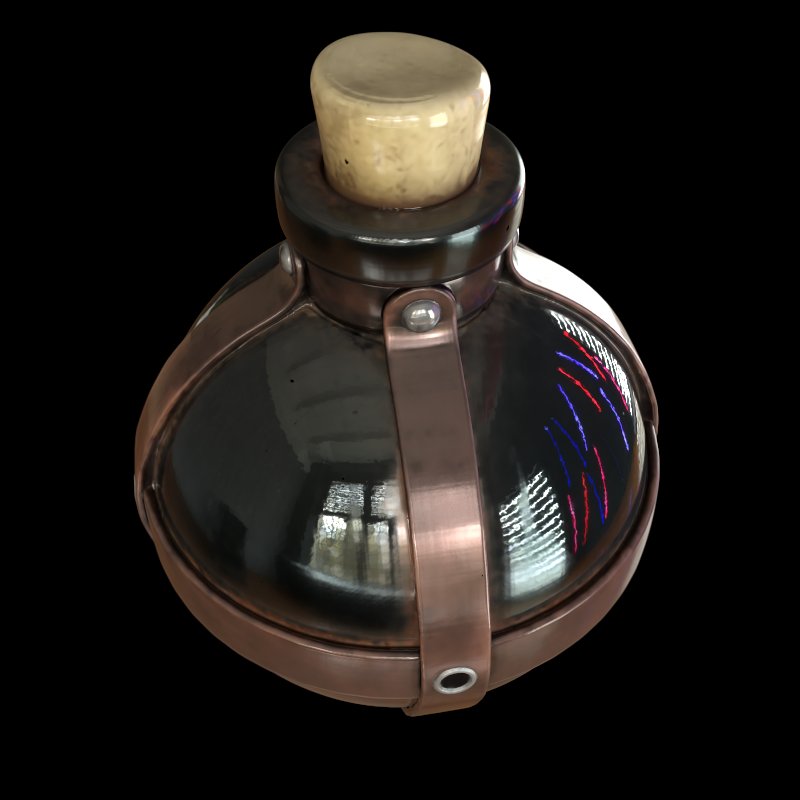} \\ \hline
    \rotatebox{90}{\parbox{0.133\textwidth}{\centering Ours }} &
    \includegraphics[trim={0 0 0 0}, clip, height=0.133\textwidth]{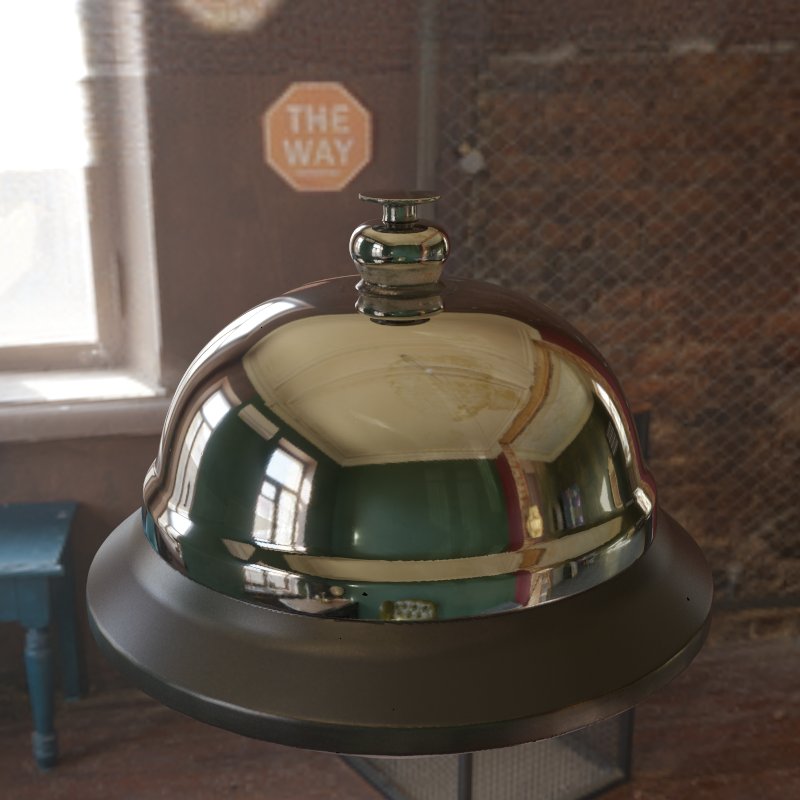} &
    \includegraphics[trim={0 0 0 0}, clip, height=0.133\textwidth]{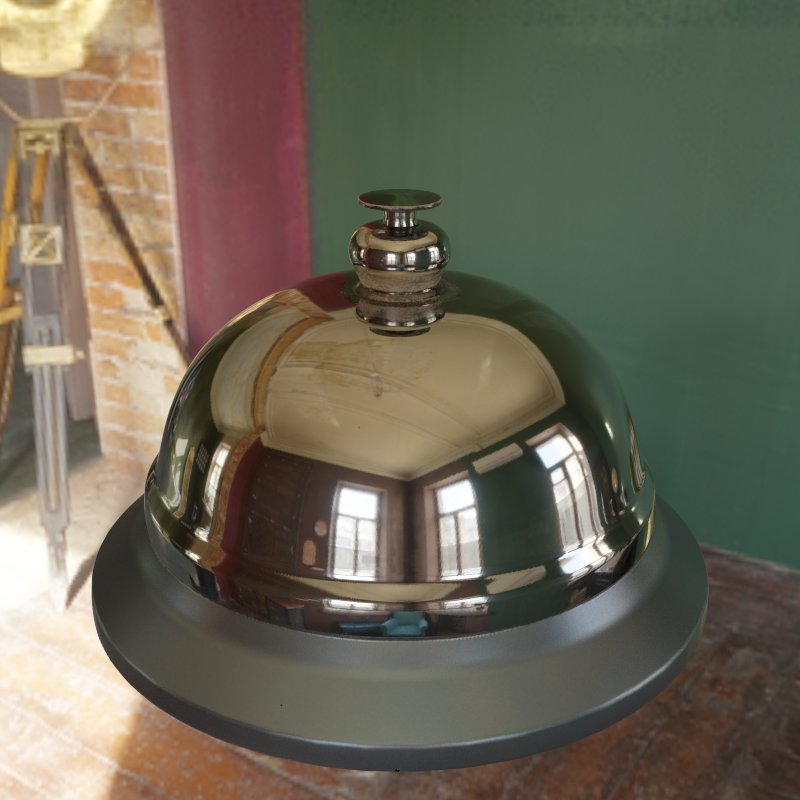} &
    \multicolumn{2}{c}{\includegraphics[trim={0 0 0 0}, clip, height=0.133\textwidth]{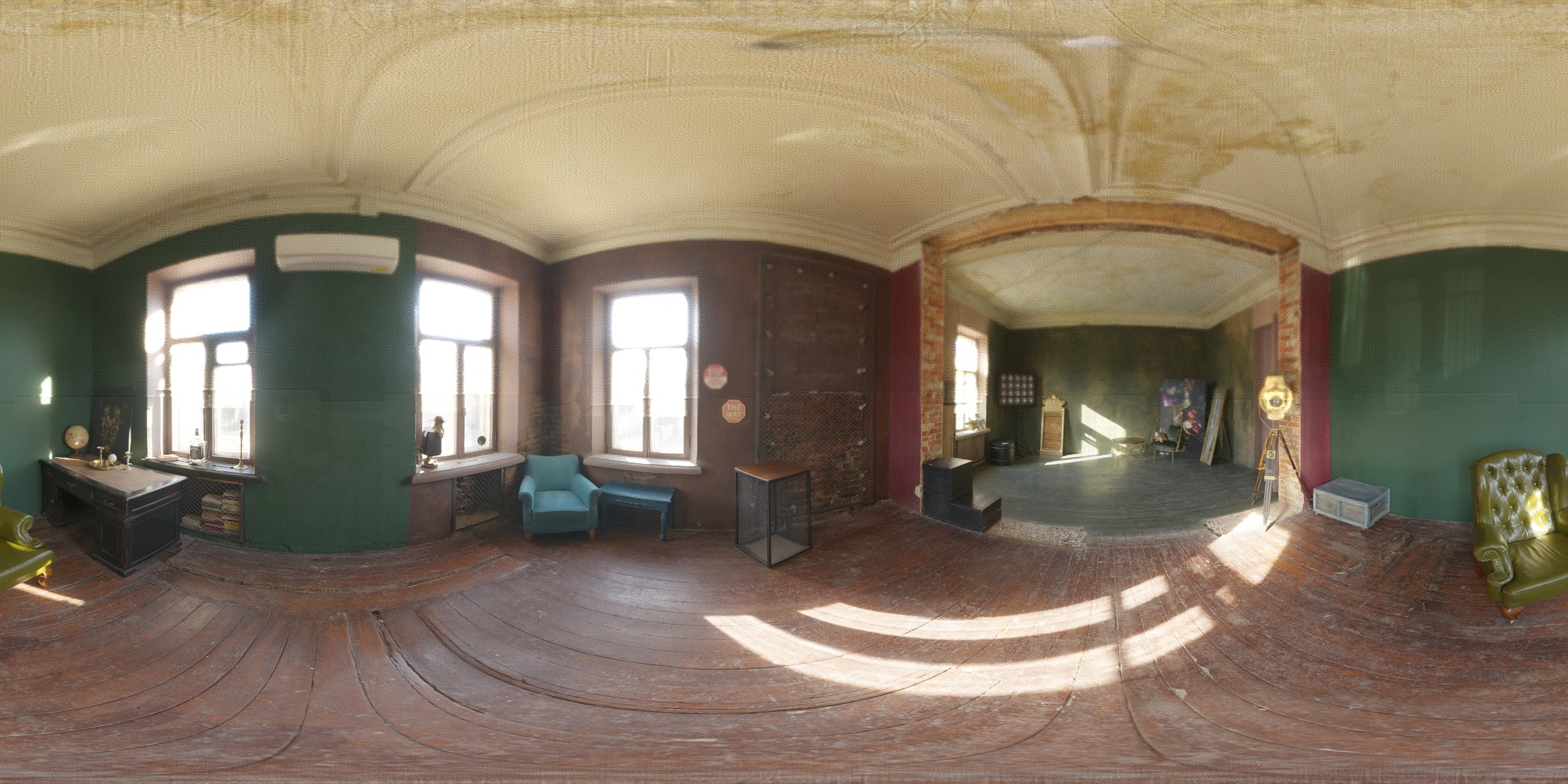}  } &
    \includegraphics[trim={0 0 0 0}, clip, height=0.133\textwidth]{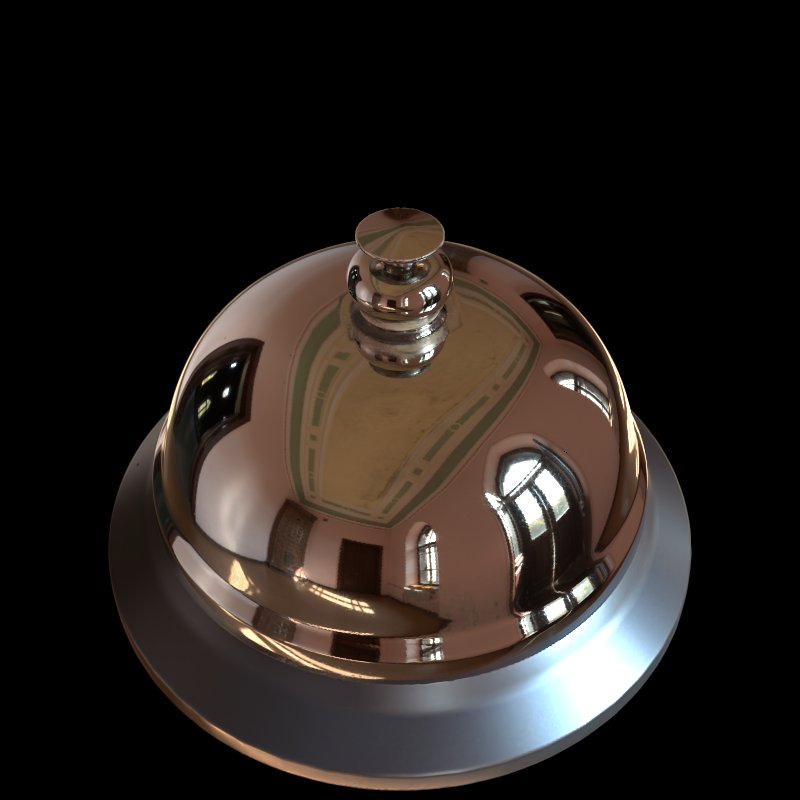}  &
    \includegraphics[trim={0 0 0 0}, clip, height=0.133\textwidth]{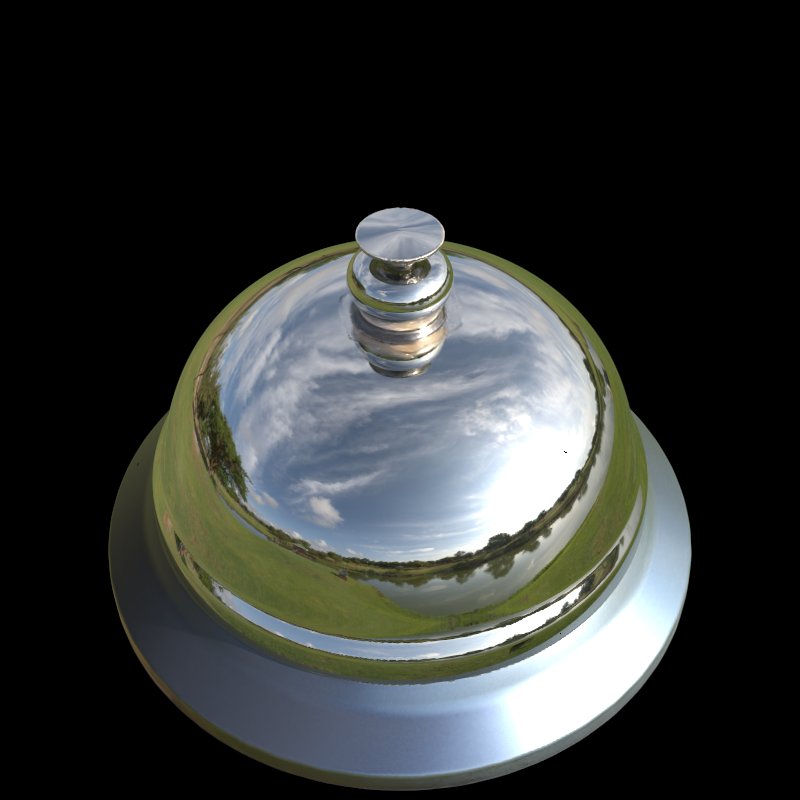}  &
    \includegraphics[trim={0 0 0 0}, clip, height=0.133\textwidth]{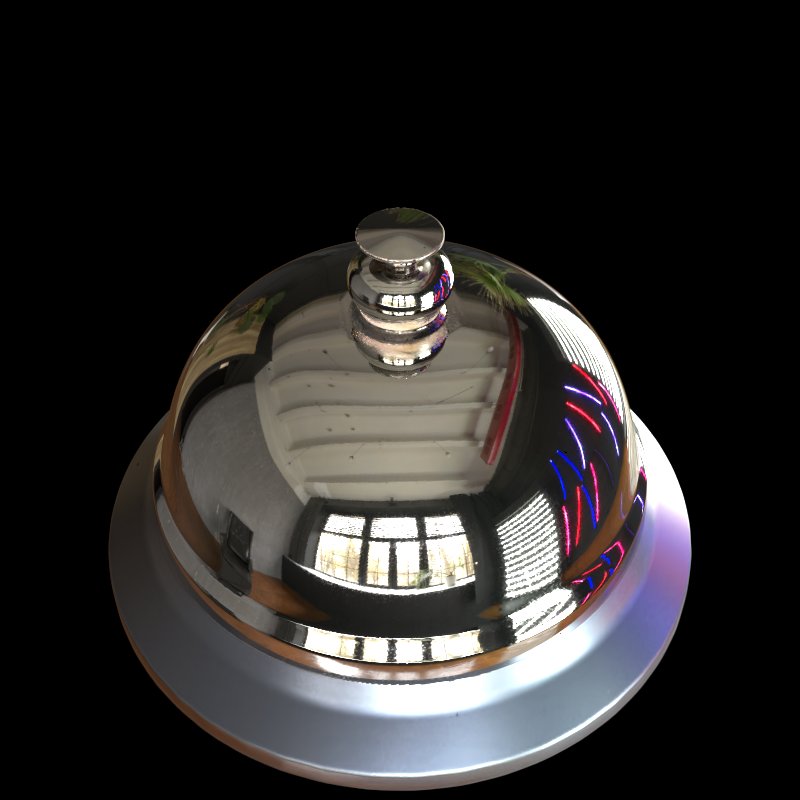} \\
    \rotatebox{90}{\parbox{0.133\textwidth}{\centering NeRO }} &
    \includegraphics[trim={0 0 0 0}, clip, height=0.133\textwidth]{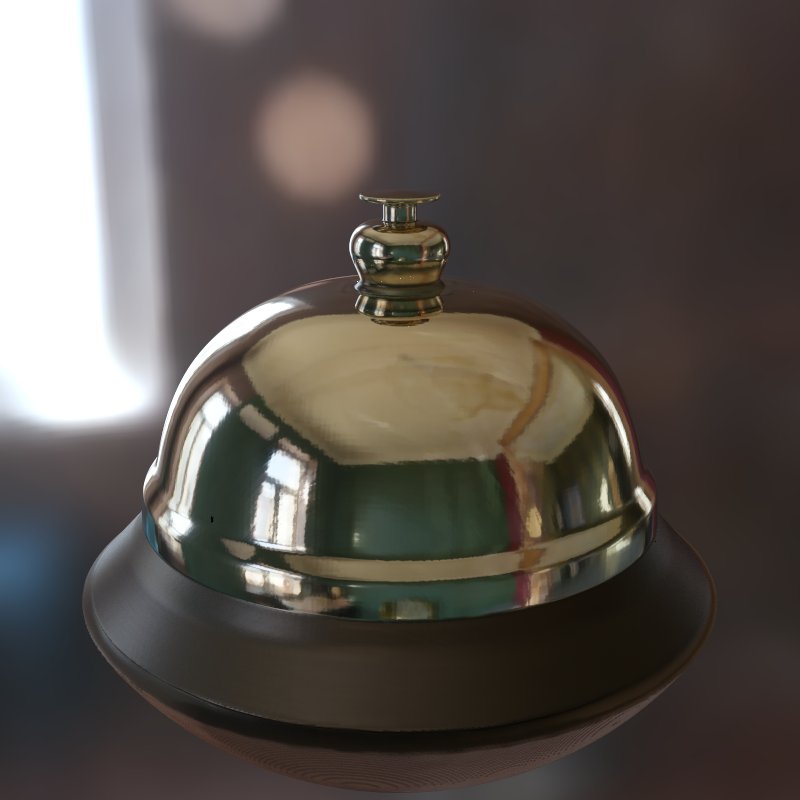} &
    \includegraphics[trim={0 0 0 0}, clip, height=0.133\textwidth]{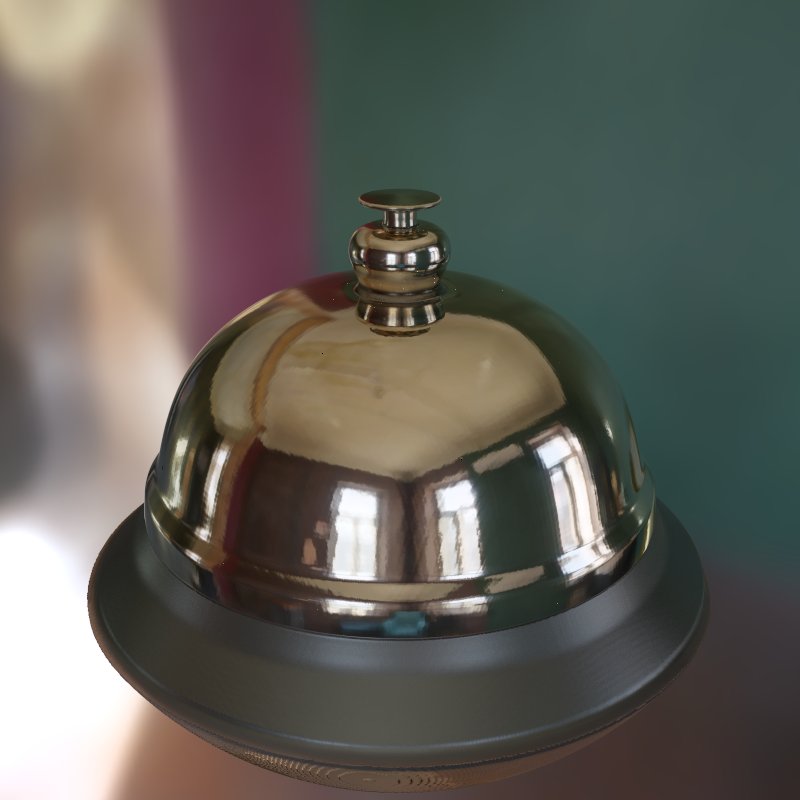} &
    \multicolumn{2}{c}{\includegraphics[trim={0 0 0 0}, clip, height=0.133\textwidth]{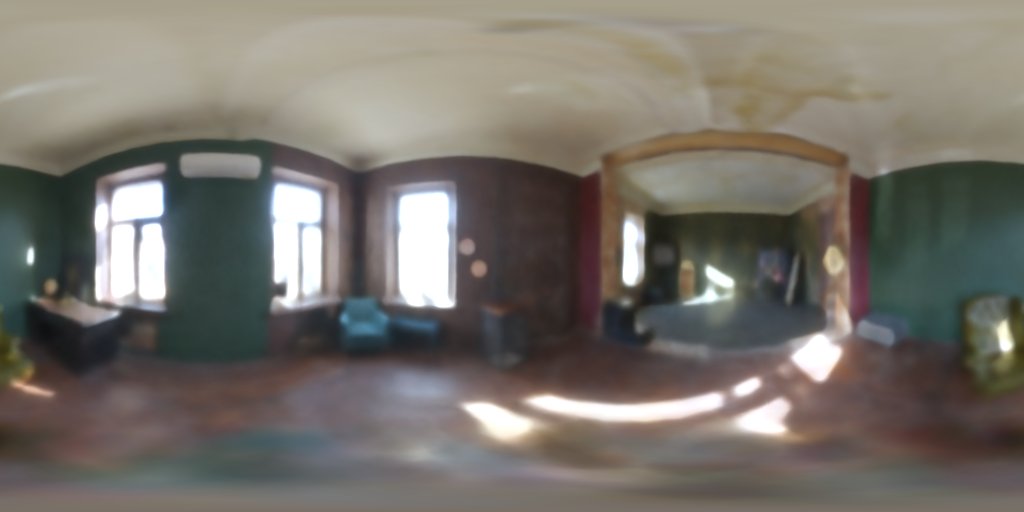}  } &
    \includegraphics[trim={0 0 0 0}, clip, height=0.133\textwidth]{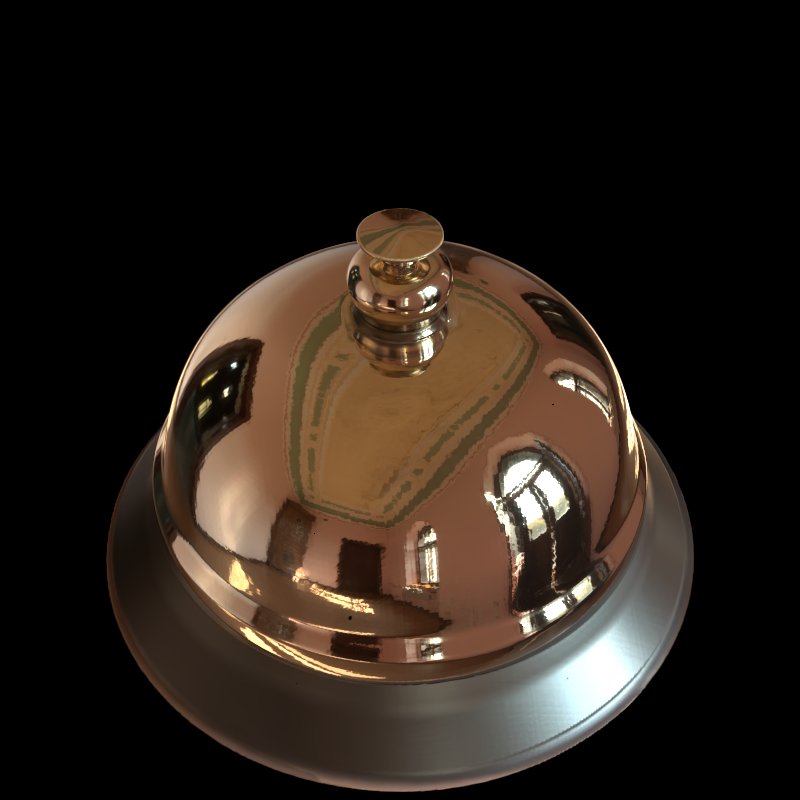}  &
    \includegraphics[trim={0 0 0 0}, clip, height=0.133\textwidth]{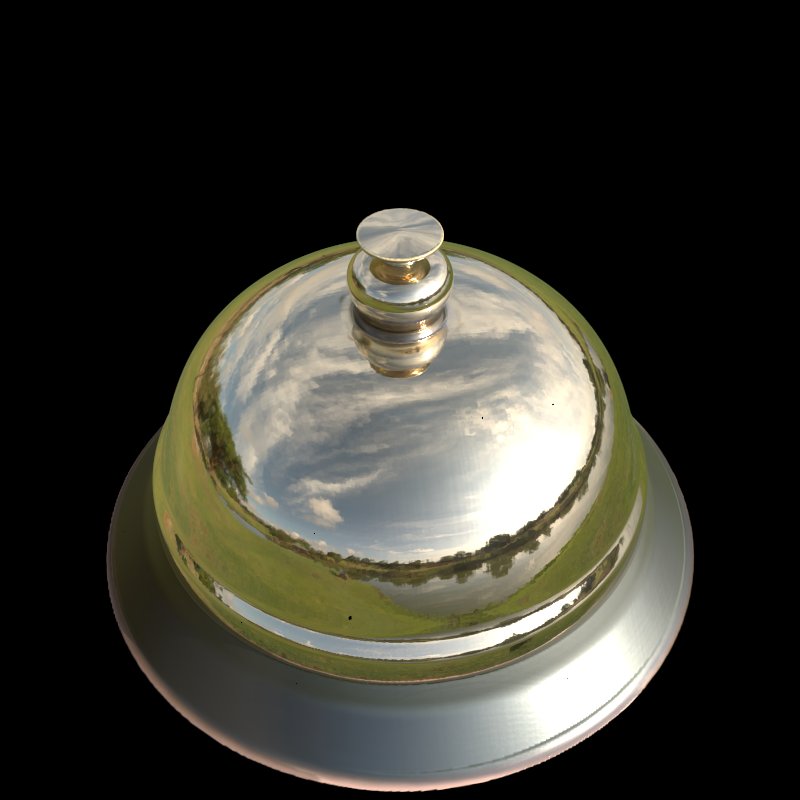}  &
    \includegraphics[trim={0 0 0 0}, clip, height=0.133\textwidth]{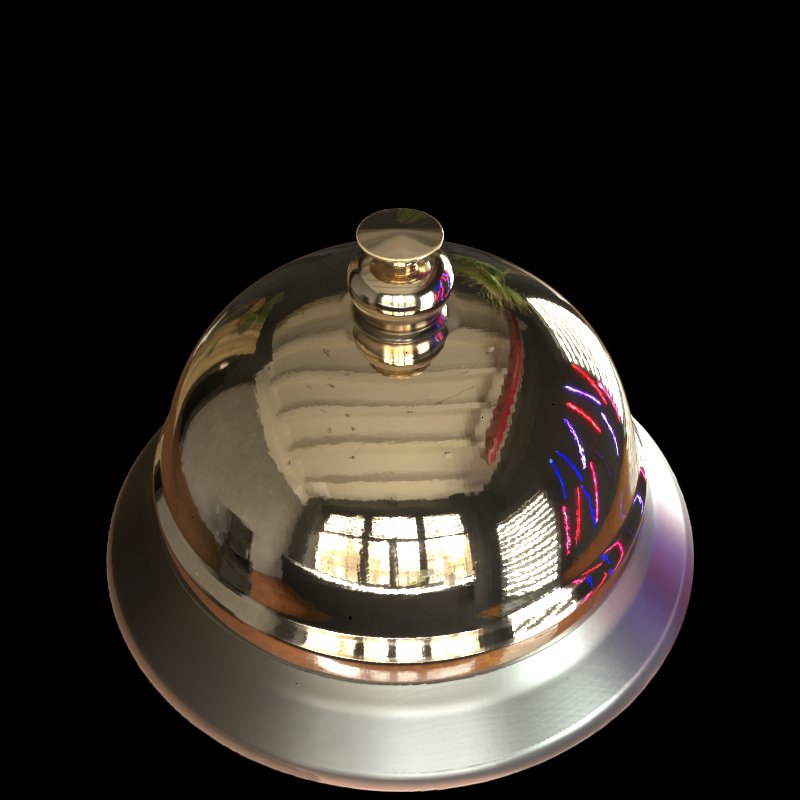} \\ \hline
        \rotatebox{90}{\parbox{0.133\textwidth}{\centering Ours }} &
    \includegraphics[trim={0 0 0 0}, clip, height=0.133\textwidth]{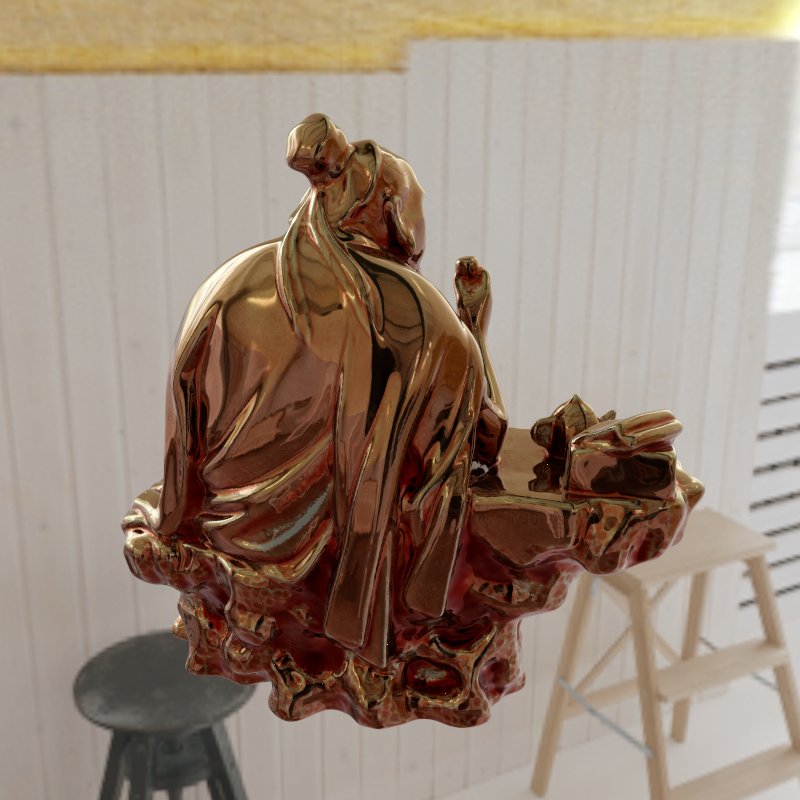} &
    \includegraphics[trim={0 0 0 0}, clip, height=0.133\textwidth]{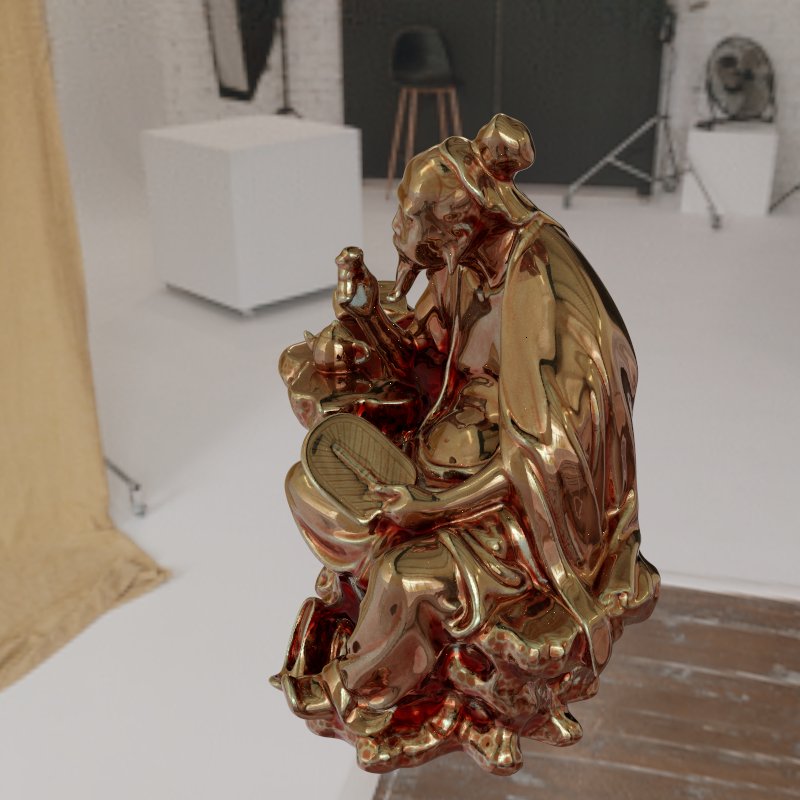} &
    \multicolumn{2}{c}{\includegraphics[trim={0 0 0 0}, clip, height=0.133\textwidth]{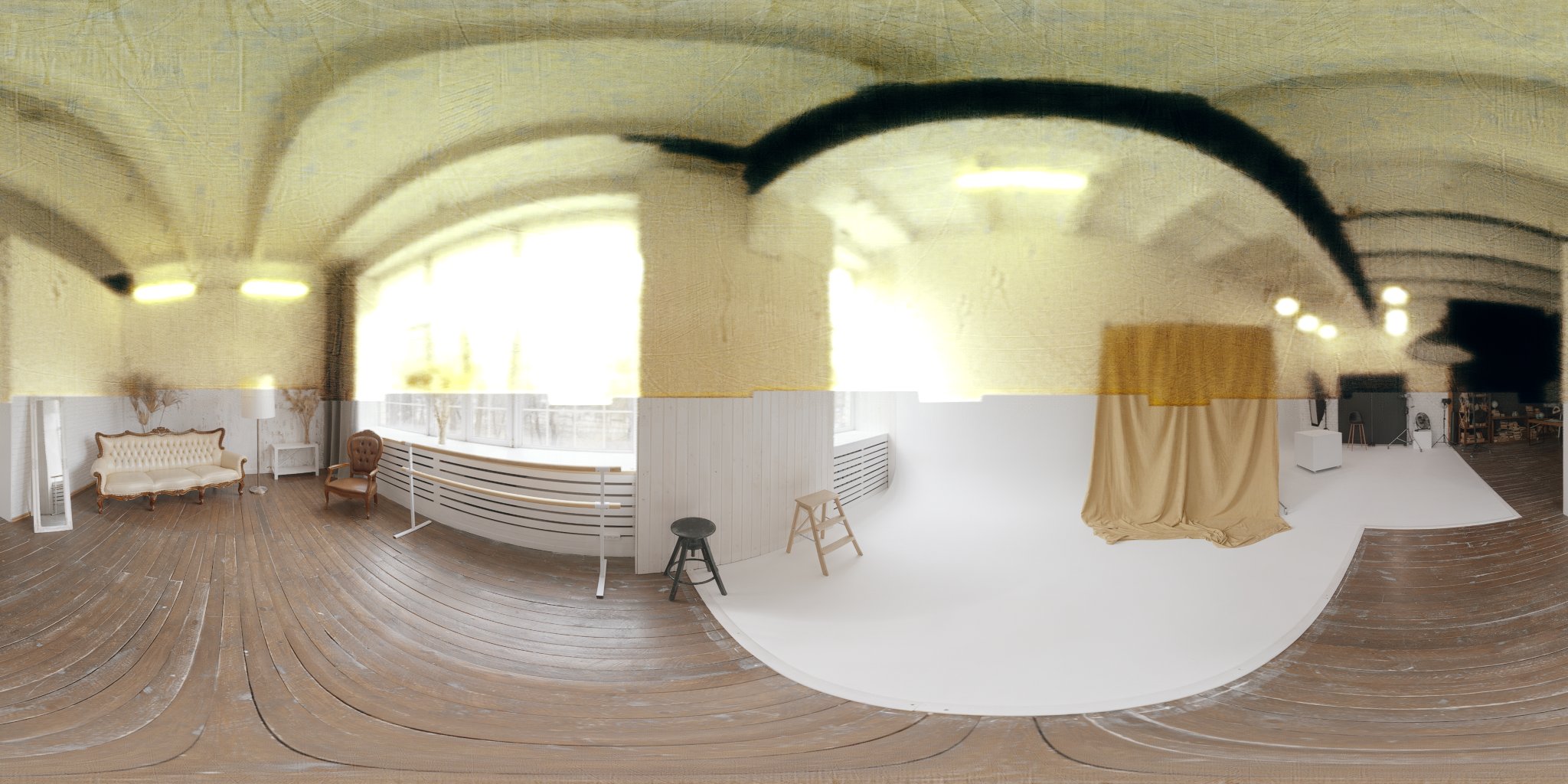}  } &
    \includegraphics[trim={0 0 0 0}, clip, height=0.133\textwidth]{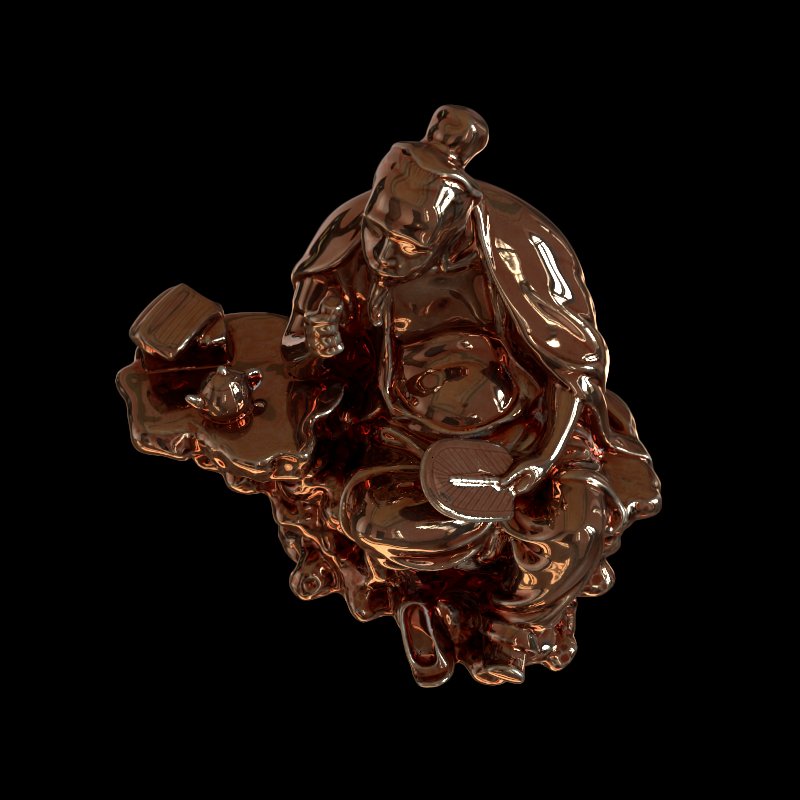}  &
    \includegraphics[trim={0 0 0 0}, clip, height=0.133\textwidth]{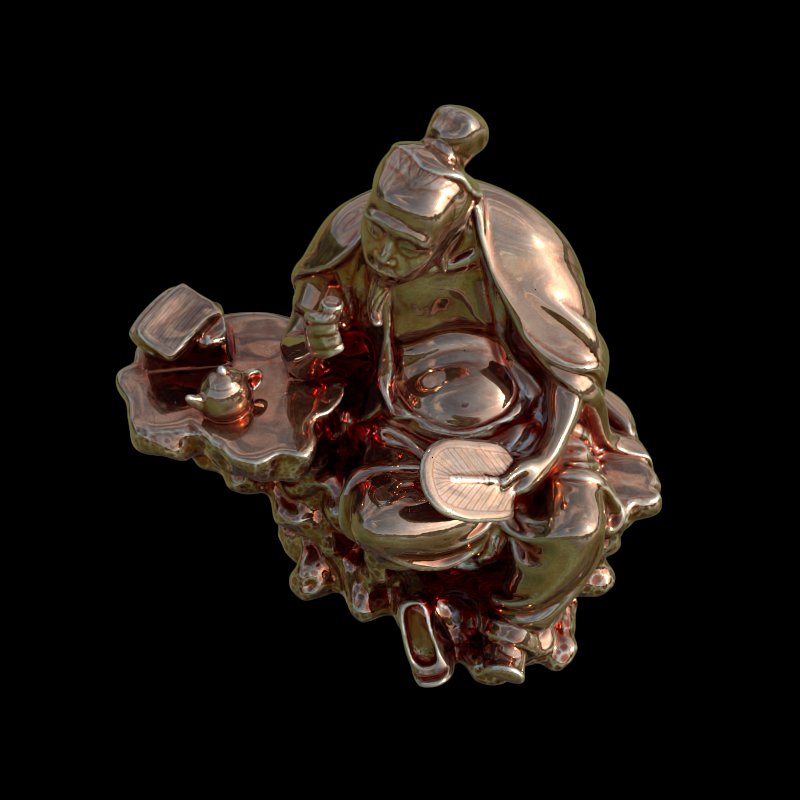}  &
    \includegraphics[trim={0 0 0 0}, clip, height=0.133\textwidth]{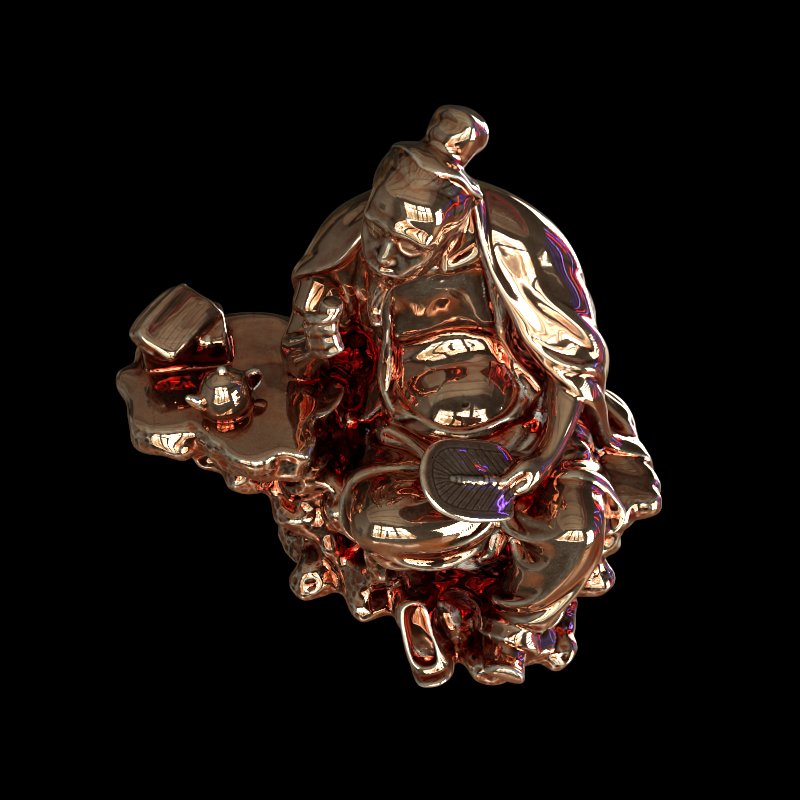} \\
    \rotatebox{90}{\parbox{0.133\textwidth}{\centering NeRO }} &
    \includegraphics[trim={0 0 0 0}, clip, height=0.133\textwidth]{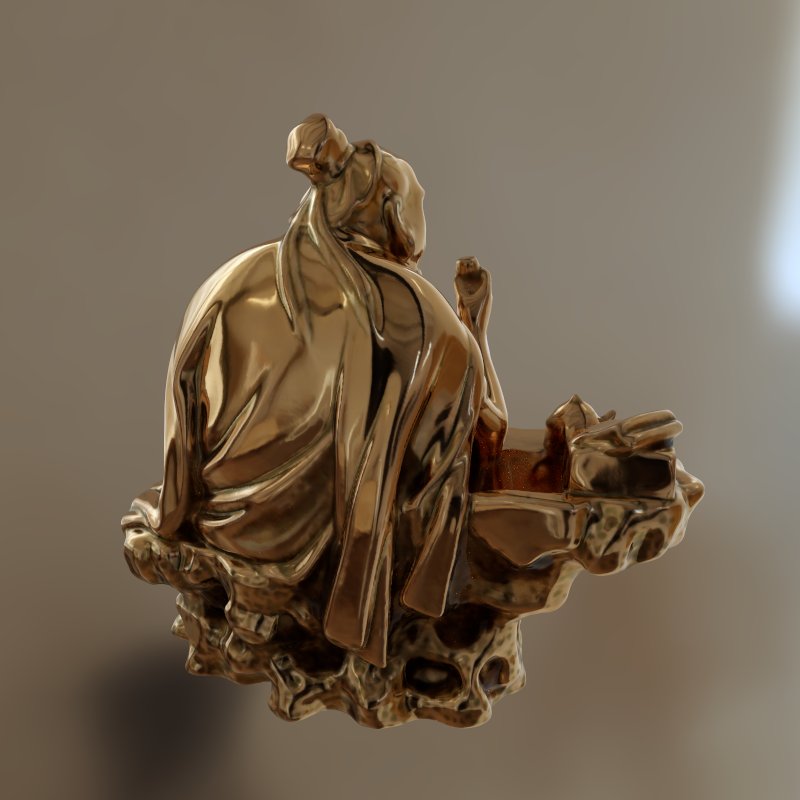} &
    \includegraphics[trim={0 0 0 0}, clip, height=0.133\textwidth]{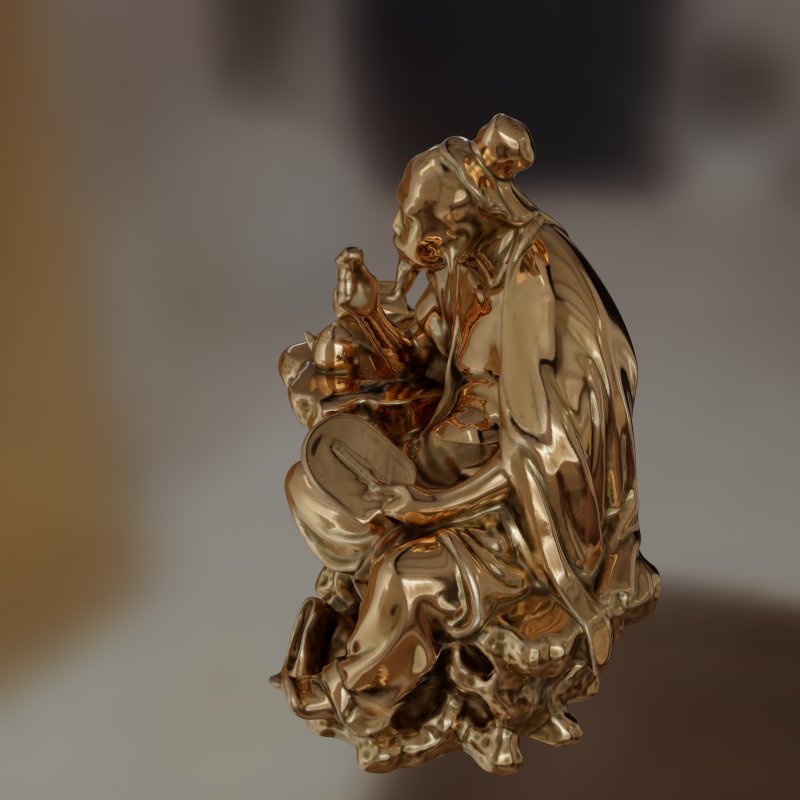} &
    \multicolumn{2}{c}{\includegraphics[trim={0 0 0 0}, clip, height=0.133\textwidth]{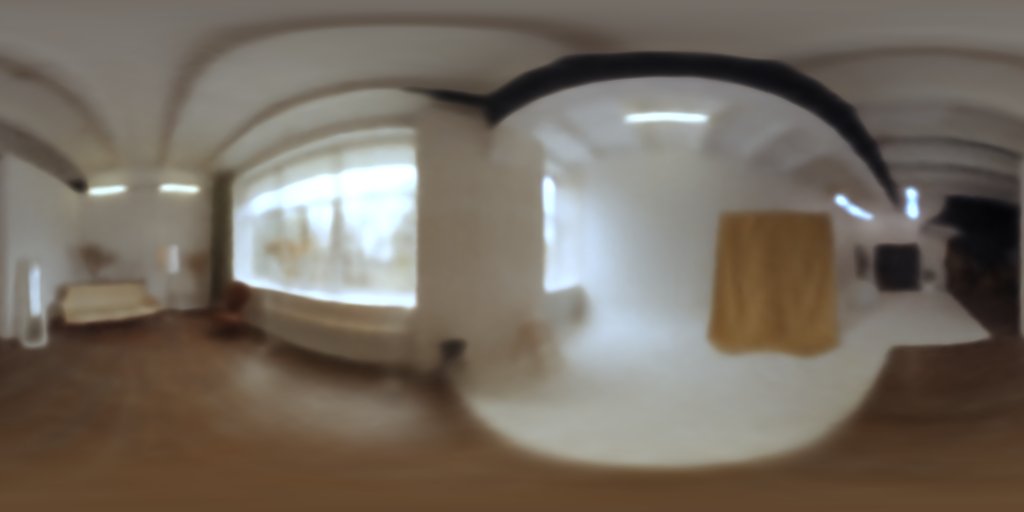}  } &
    \includegraphics[trim={0 0 0 0}, clip, height=0.133\textwidth]{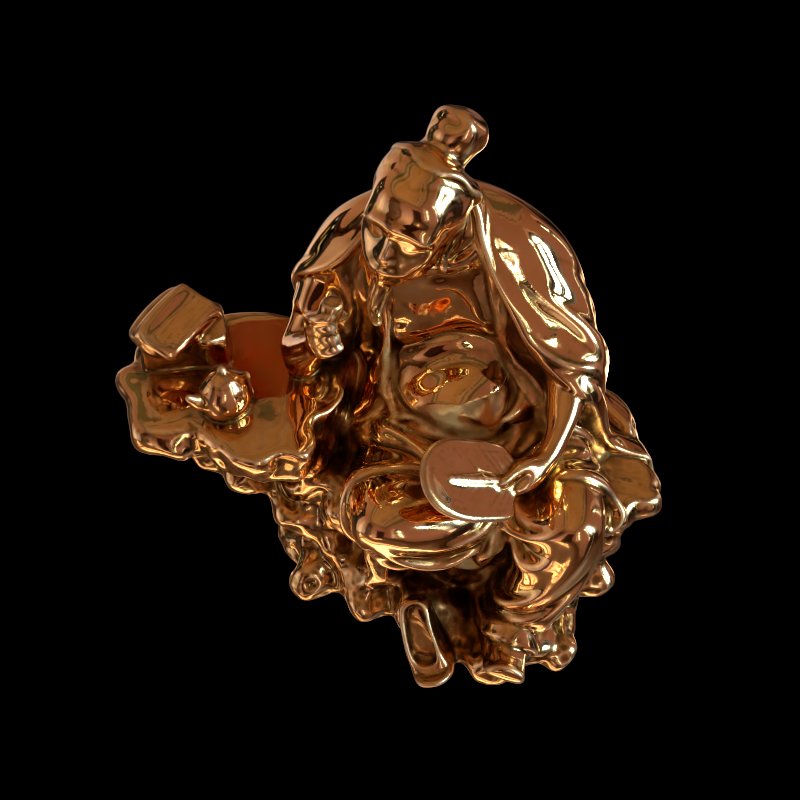}  &
    \includegraphics[trim={0 0 0 0}, clip, height=0.133\textwidth]{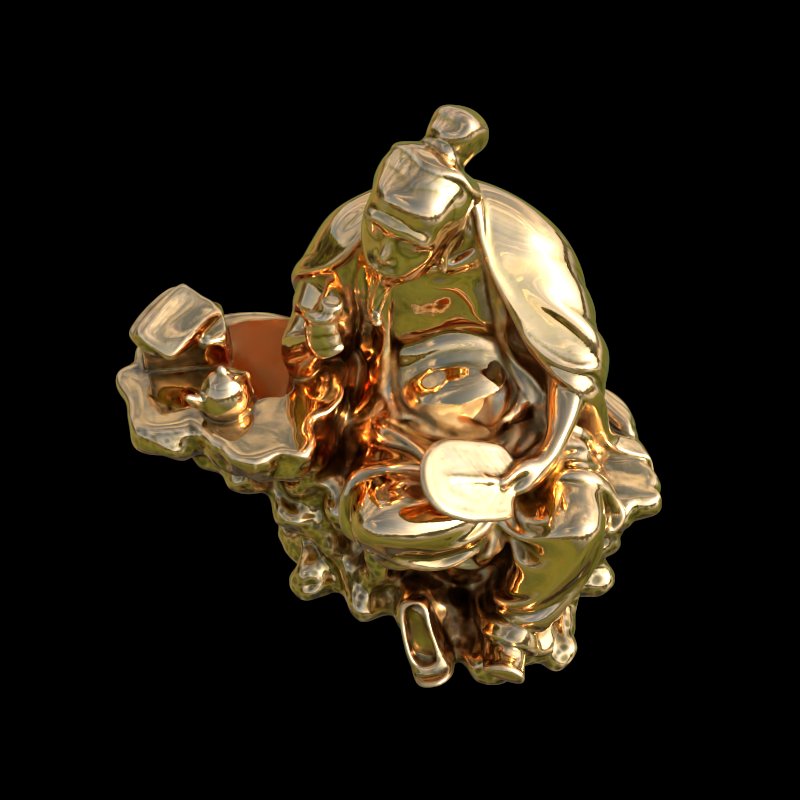}  &
    \includegraphics[trim={0 0 0 0}, clip, height=0.133\textwidth]{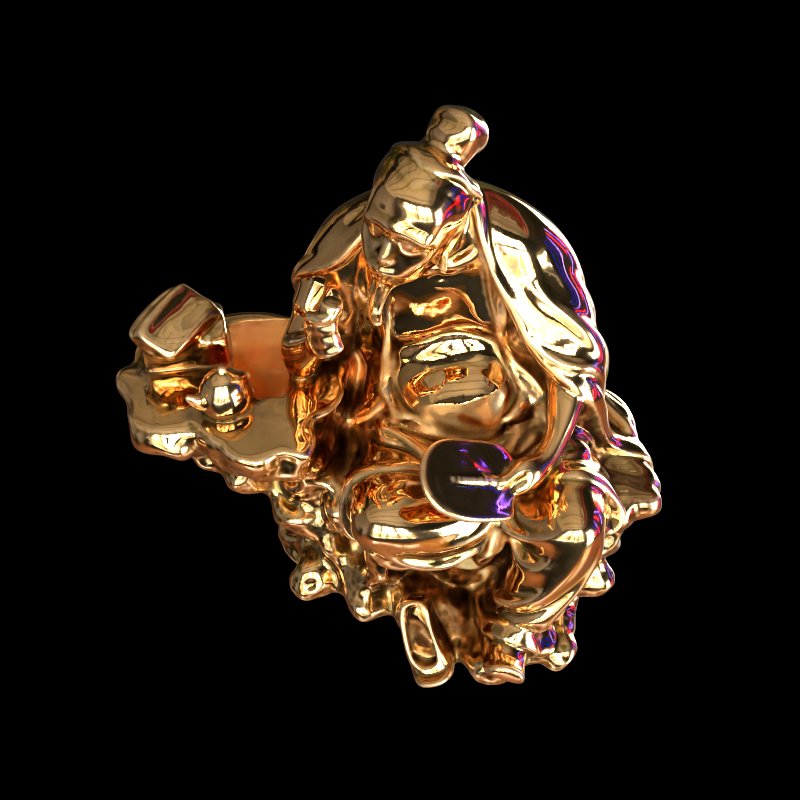} \\ \hline
    \rotatebox{90}{\parbox{0.133\textwidth}{\centering Ours }} &
    \includegraphics[trim={0 0 0 0}, clip, height=0.133\textwidth]{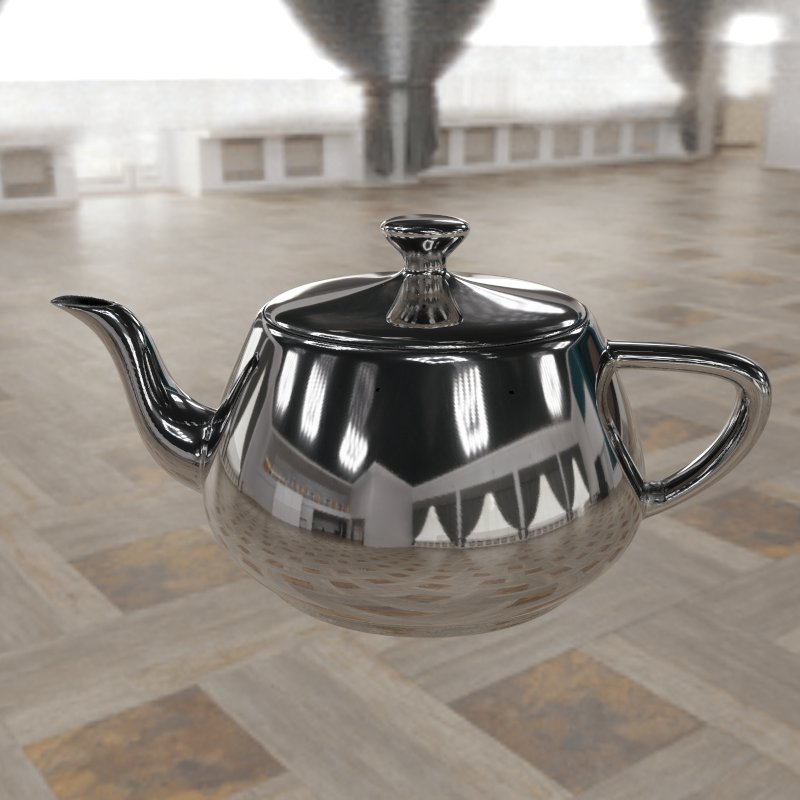} &
    \includegraphics[trim={0 0 0 0}, clip, height=0.133\textwidth]{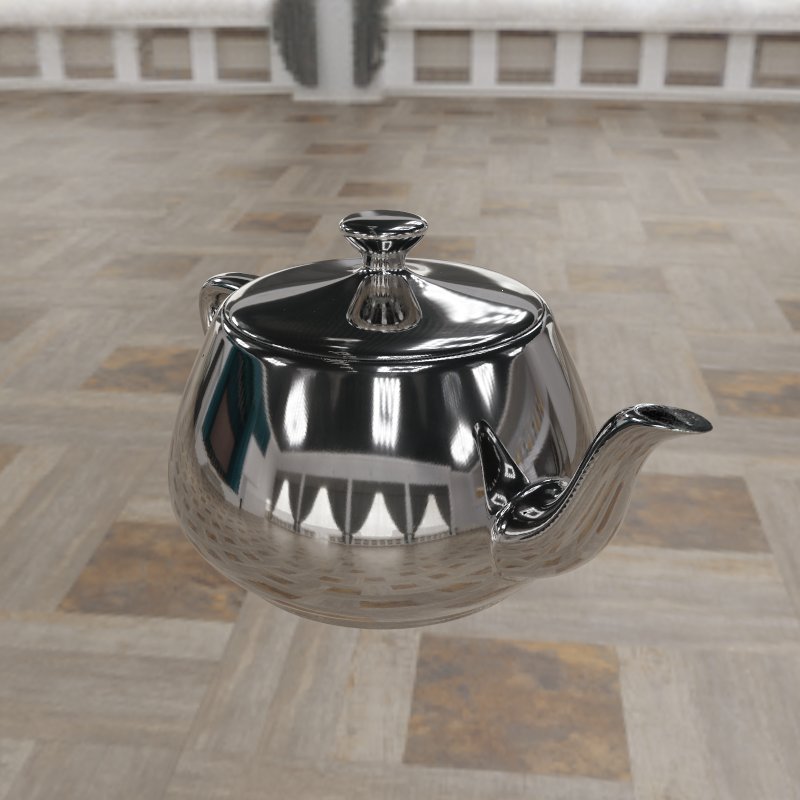} &
    \multicolumn{2}{c}{\includegraphics[trim={0 0 0 0}, clip, height=0.133\textwidth]{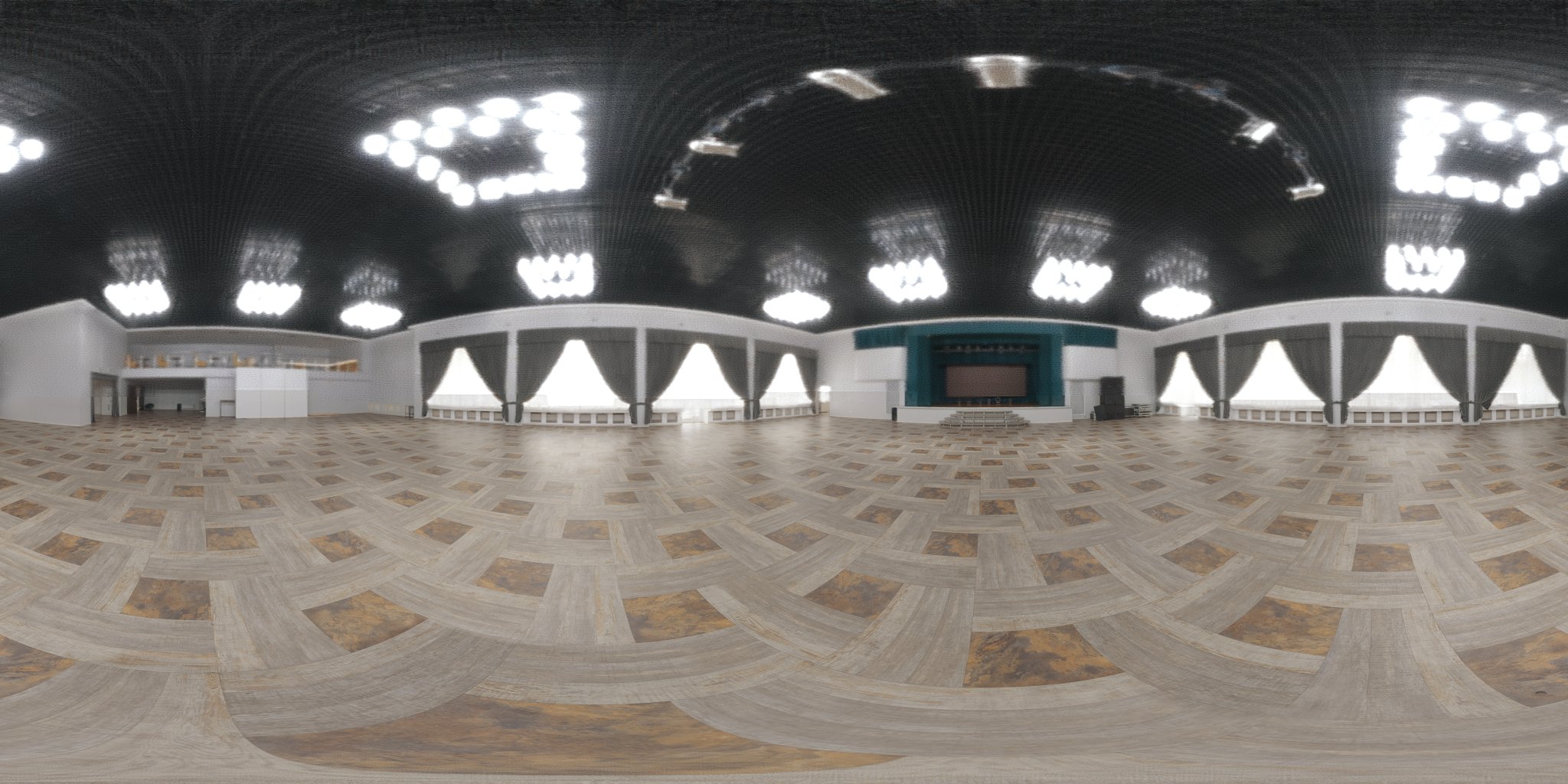}  } &
    \includegraphics[trim={0 0 0 0}, clip, height=0.133\textwidth]{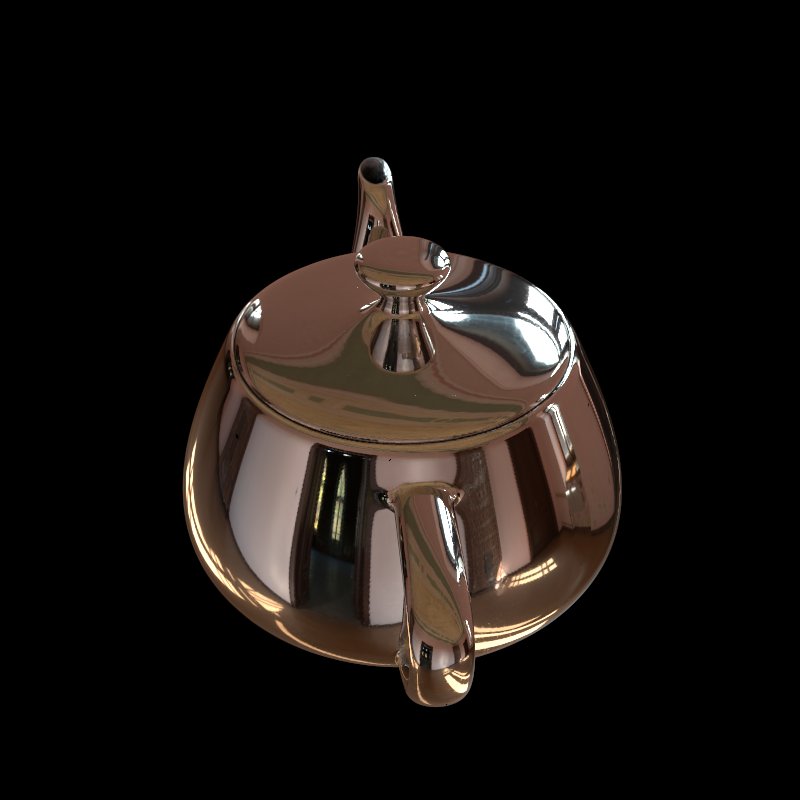}  &
    \includegraphics[trim={0 0 0 0}, clip, height=0.133\textwidth]{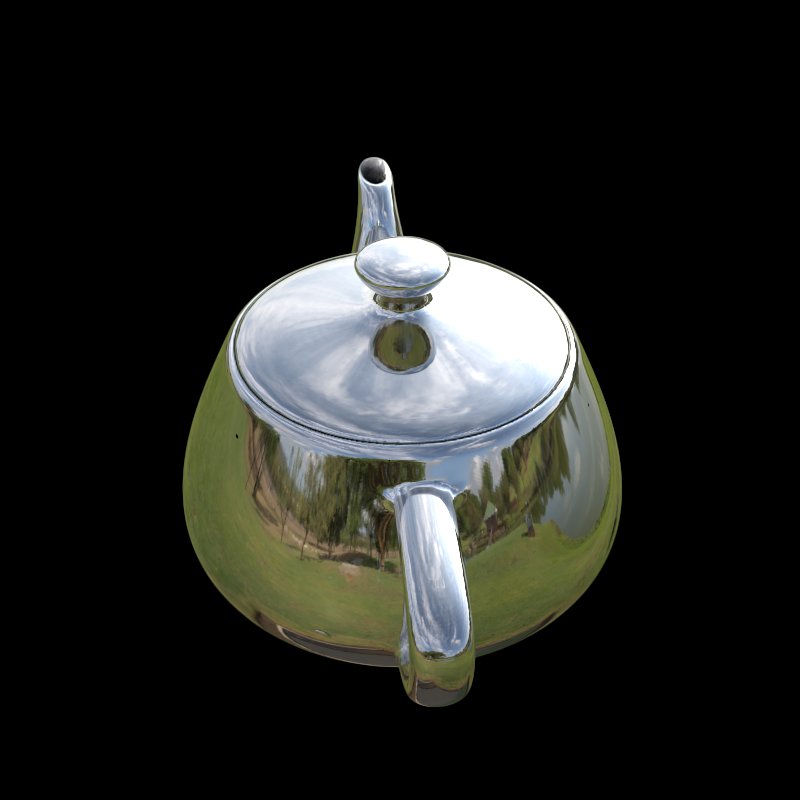}  &
    \includegraphics[trim={0 0 0 0}, clip, height=0.133\textwidth]{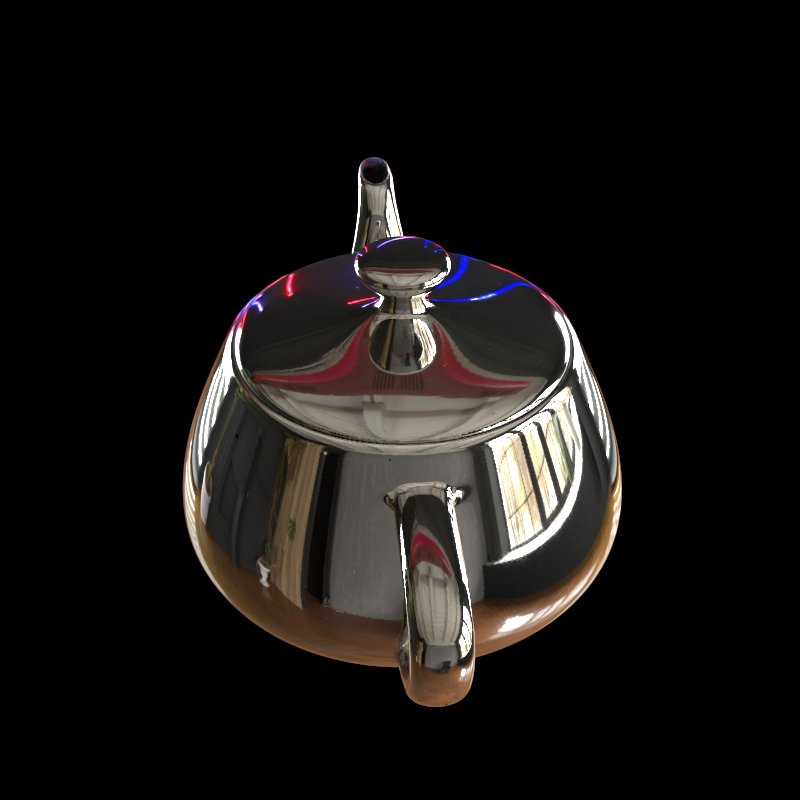} \\
    \rotatebox{90}{\parbox{0.133\textwidth}{\centering NeRO }} &
    \includegraphics[trim={0 0 0 0}, clip, height=0.133\textwidth]{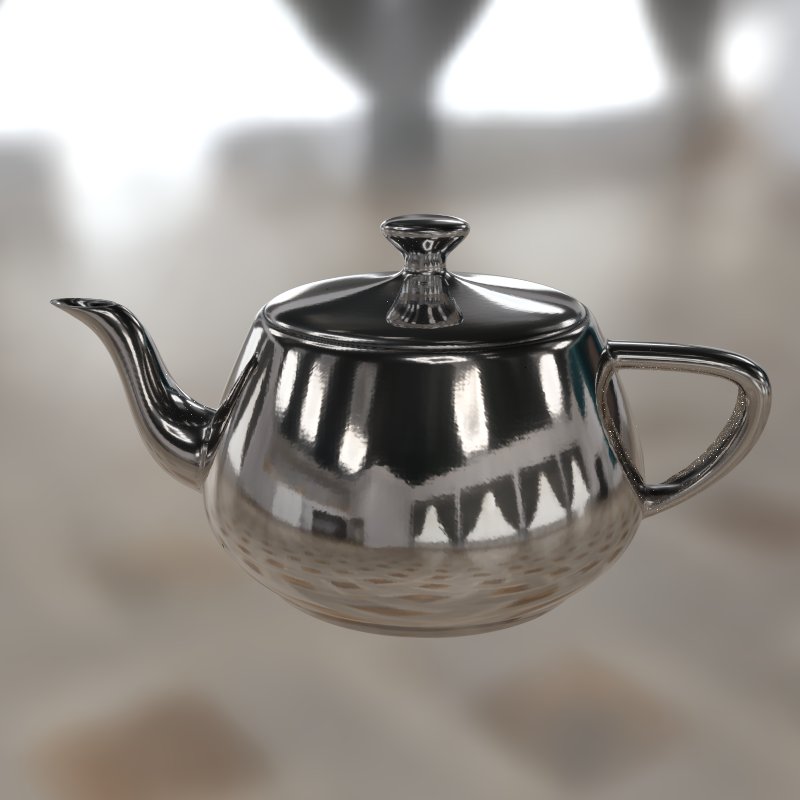} &
    \includegraphics[trim={0 0 0 0}, clip, height=0.133\textwidth]{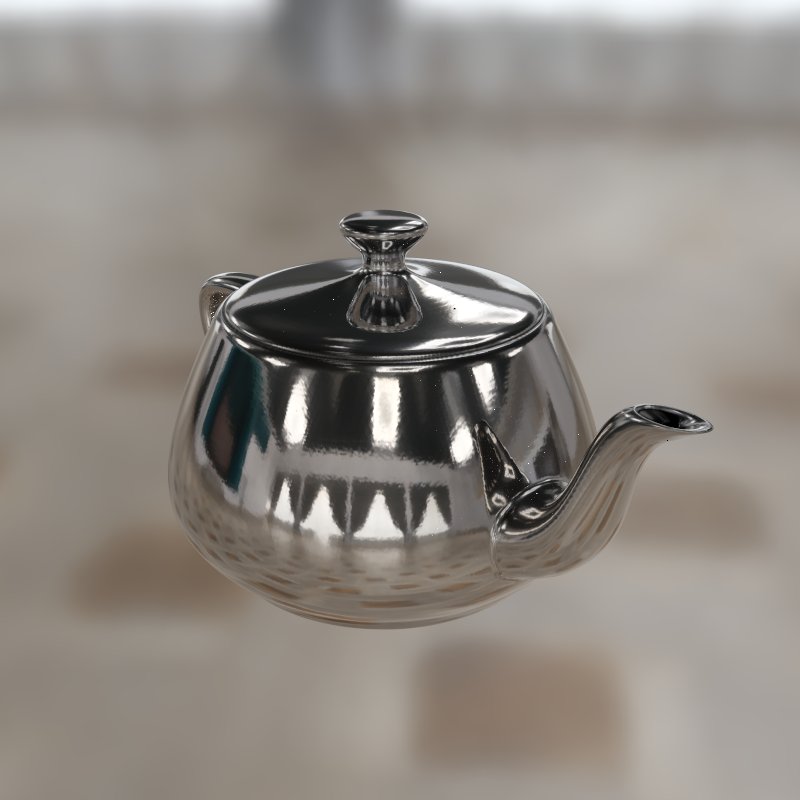} &
    \multicolumn{2}{c}{\includegraphics[trim={0 0 0 0}, clip, height=0.133\textwidth]{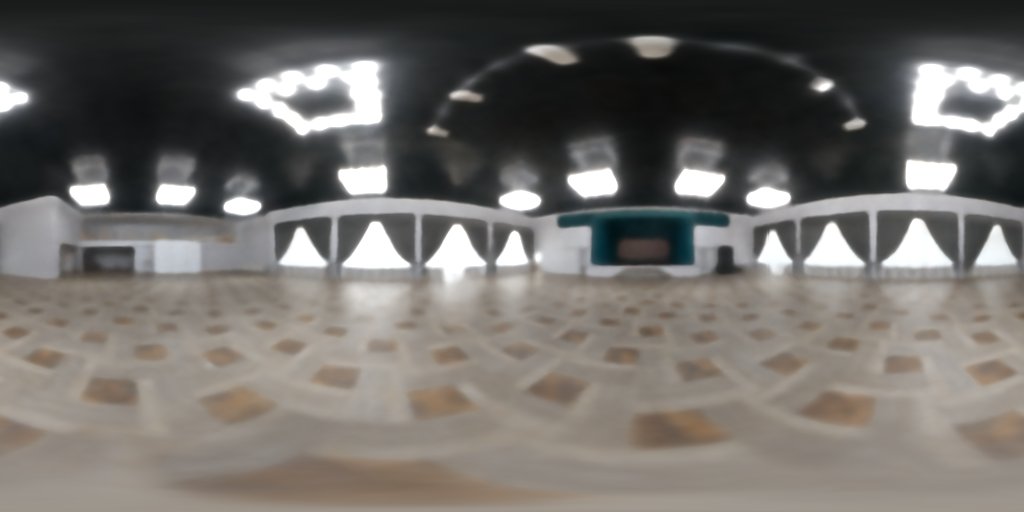}  } &
    \includegraphics[trim={0 0 0 0}, clip, height=0.133\textwidth]{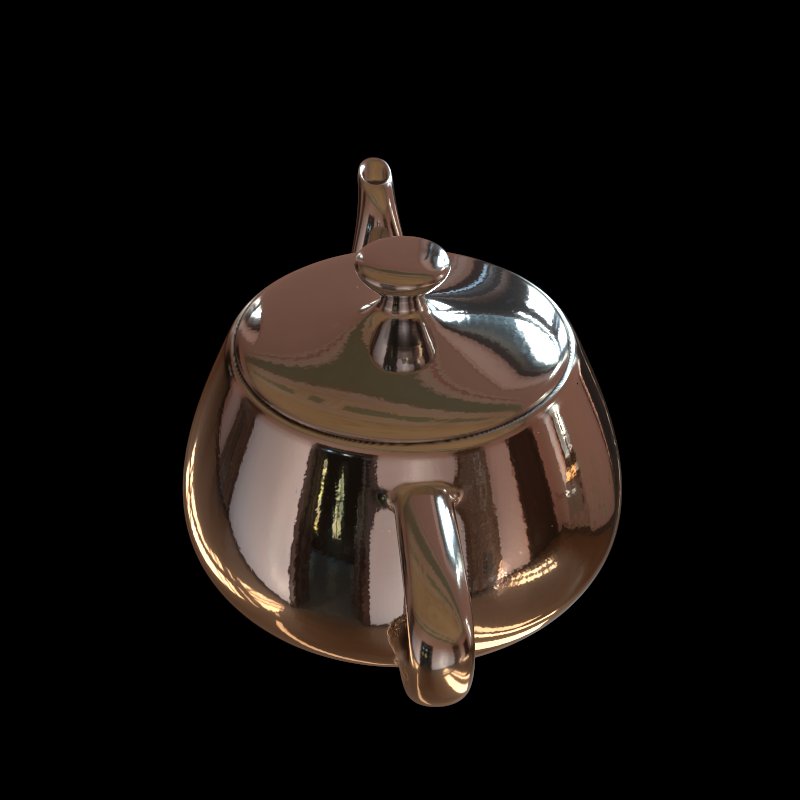}  &
    \includegraphics[trim={0 0 0 0}, clip, height=0.133\textwidth]{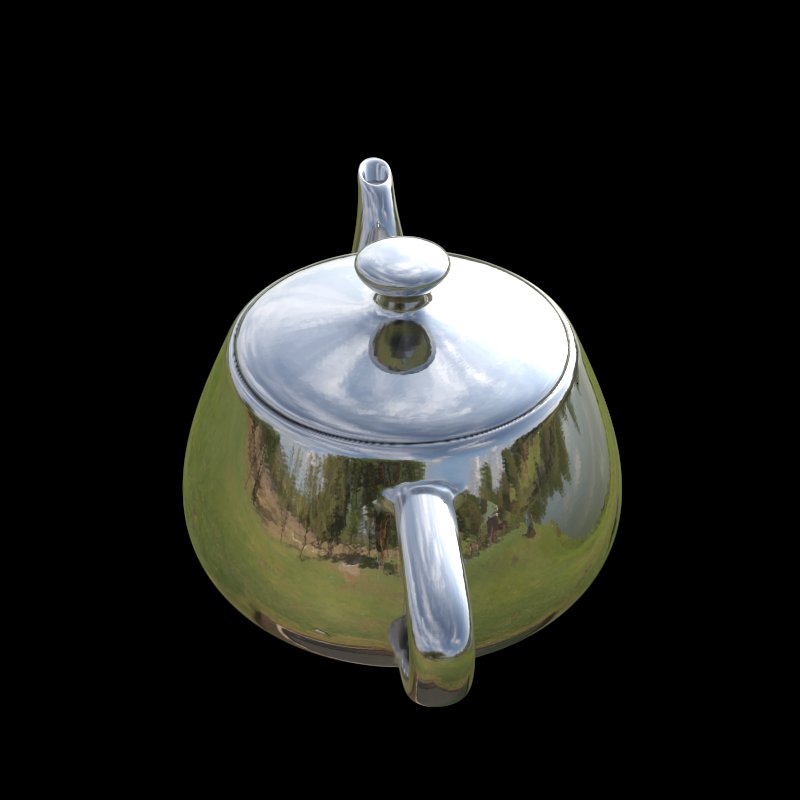}  &
    \includegraphics[trim={0 0 0 0}, clip, height=0.133\textwidth]{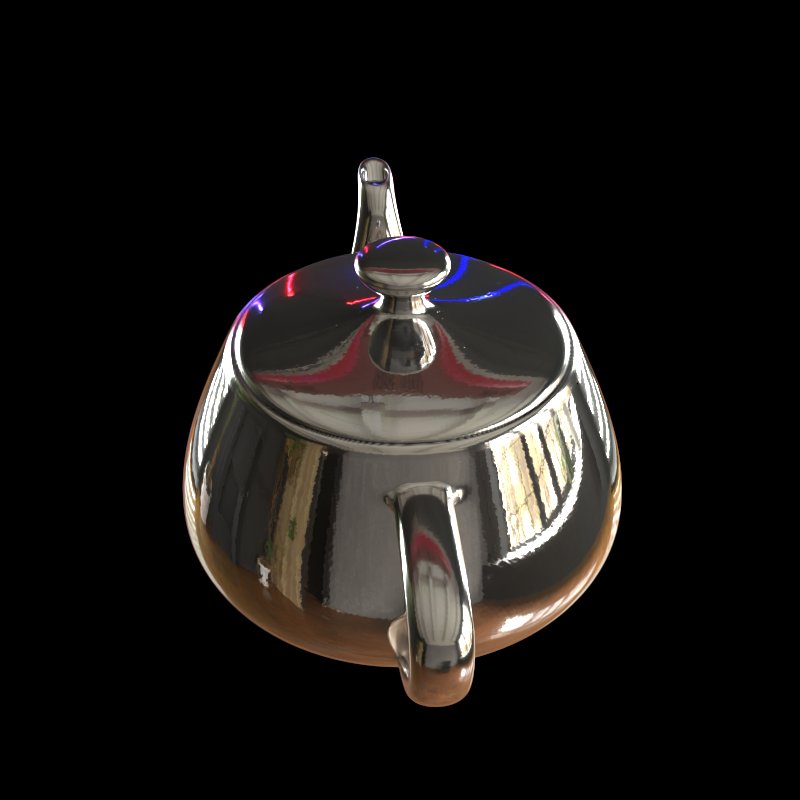} \\ \hline
    \end{tabular}
    }
    \caption{\textbf{Additional comparisons of our method against NeRO~\cite{liuNeRONeuralGeometry2023} on glossy synthetic data. } Please zoom in to better compare the results. }
    \label{fig:supp_nero2}
\end{figure*}

The figures presented below serve as supplementary material to the comparisons discussed in Sec. \ref{sec:comparison_with_baselines}
. Fig. \ref{fig:supp_sorb} provides various visualizations of a subset of our reconstructions of the Stanford-ORB dataset \cite{kuang2024stanford}. Figures \ref{fig:supp_nero1} and \ref{fig:supp_nero2} provide additional comparisons with NeRO on their glossy synthetic dataset.

\clearpage

\end{document}